\documentclass{llncs}
\usepackage{amsmath}
\usepackage{makeidx}  
\usepackage{times}
\usepackage{graphicx}
\usepackage{amssymb}
\usepackage{bm}
\usepackage[boxed]{algorithm2e}
\usepackage{todonotes}
\usepackage{dsfont}
\usepackage{array}
\usepackage{threeparttable}
\usepackage{booktabs}
\usepackage{multirow}

\SetAlCapSkip{1em}

\newtheorem{prop}{Proposition}

\usepackage{xspace}

\newcommand*{\ie}{\textit{i.e.}\@\xspace}
\newcommand*{\etal}{\textit{et al.}\@\xspace}

\newcommand{\keywords}[1]{\par\addvspace\baselineskip
\noindent\keywordname\enspace\ignorespaces#1}

\newcolumntype{"}{@{\hskip\tabcolsep\vrule width 1pt\hskip\tabcolsep}}
\makeatother

\pdfoutput=1

\begin{document}

\mainmatter              
%
\title{A Projected Gradient Descent Method for CRF Inference allowing End-To-End Training of Arbitrary Pairwise Potentials}

%
\titlerunning{Deep Structured Models}  
%

\author{M\aa ns Larsson \inst{1} \and Anurag Arnab \inst{2} \and Fredrik Kahl \inst{1,3} \and Shuai Zheng \inst{2} \and Philip Torr \inst{2}}
\authorrunning{M\aa ns Larsson et al.} 
%
%
\institute{Chalmers University of Technology, Gothenburg, Sweden\\
\and
University of Oxford, Oxford, England \\
\and
Centre for Mathematical Sciences, Lund University, Lund, Sweden}

\maketitle              

\begin{abstract}
Are we using the right potential functions in the Conditional Random Field models that are popular in the Vision community? Semantic segmentation and other pixel-level labelling tasks have made significant progress recently due to the deep learning paradigm. However, most state-of-the-art structured prediction methods also include a random field model with a hand-crafted Gaussian potential to model spatial priors, label consistencies and feature-based image conditioning. 

In this paper, we challenge this view by developing a new inference and learning framework which can learn pairwise CRF potentials restricted only by their dependence on the image pixel values and the size of the support. Both standard spatial and high-dimensional bilateral kernels are considered. Our framework is based on the observation that CRF inference can be achieved via projected gradient descent and consequently, can easily be integrated in deep neural networks to allow for end-to-end training. It is empirically demonstrated that such learned potentials can improve segmentation accuracy and that certain label class interactions are indeed better modelled by a non-Gaussian potential. In addition, we compare our inference method to the commonly used mean-field algorithm. Our framework is evaluated on several public benchmarks for semantic segmentation with improved performance compared to previous state-of-the-art CNN+CRF models.

\end{abstract} 

\section{Introduction}
\vspace{-0.05cm}
Markov Random Fields (MRFs), Conditional Random Fields (CRFs) and more generally, probabilistic graphical models are a ubiquitous tool used in a variety of domains spanning Computer Vision, Computer Graphics and Image Processing \cite{koller09,blake-etal-book-2011}. 
In this paper, we focus on the application of CRFs for Computer Vision problems involving per-pixel labelling such as image segmentation. 
There are many successful approaches in this line of research, such as the interactive segmentation of \cite{rother-etal-sg-2004} using graph cuts and the semantic segmentation works of \cite{krahenbuhl-koltun-nips-2011,vineet-etal-eccv-2012} where the parallel mean-field inference algorithm 
was applied for fast inference.
Recently, Convolutional Neural Networks (CNNs) have dominated the field in a variety of recognition tasks \cite{he2016deep,simonyan_2015,ren_2015}.
However, we observe that leading segmentation approaches still include CRFs, either as a post-processing step \cite{chen14semantic,ghiasi-fowlkes-eccv-2016,chandra-kokkinos-eccv-2016,chen_arxiv_2016}, or as part of the deep neural network itself \cite{crfasrnn_iccv2015,lin-etal-cvpr-2016,arnab_eccv_2016,liu_2015,Kirillov/accv2016}. 

\begin{figure}[t]
\begin{center}
   \includegraphics[width=0.30\linewidth]{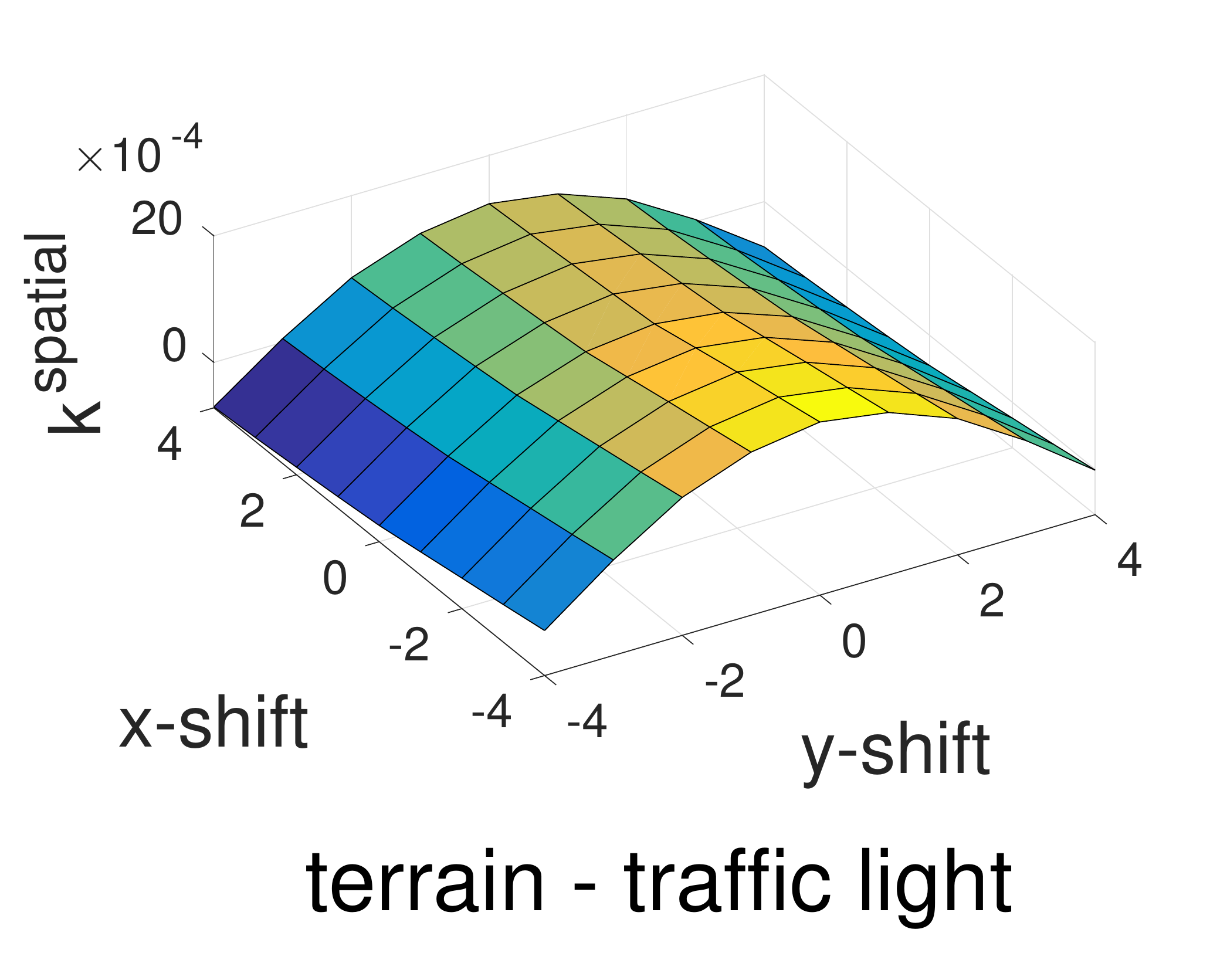}
   \includegraphics[width=0.30\linewidth]{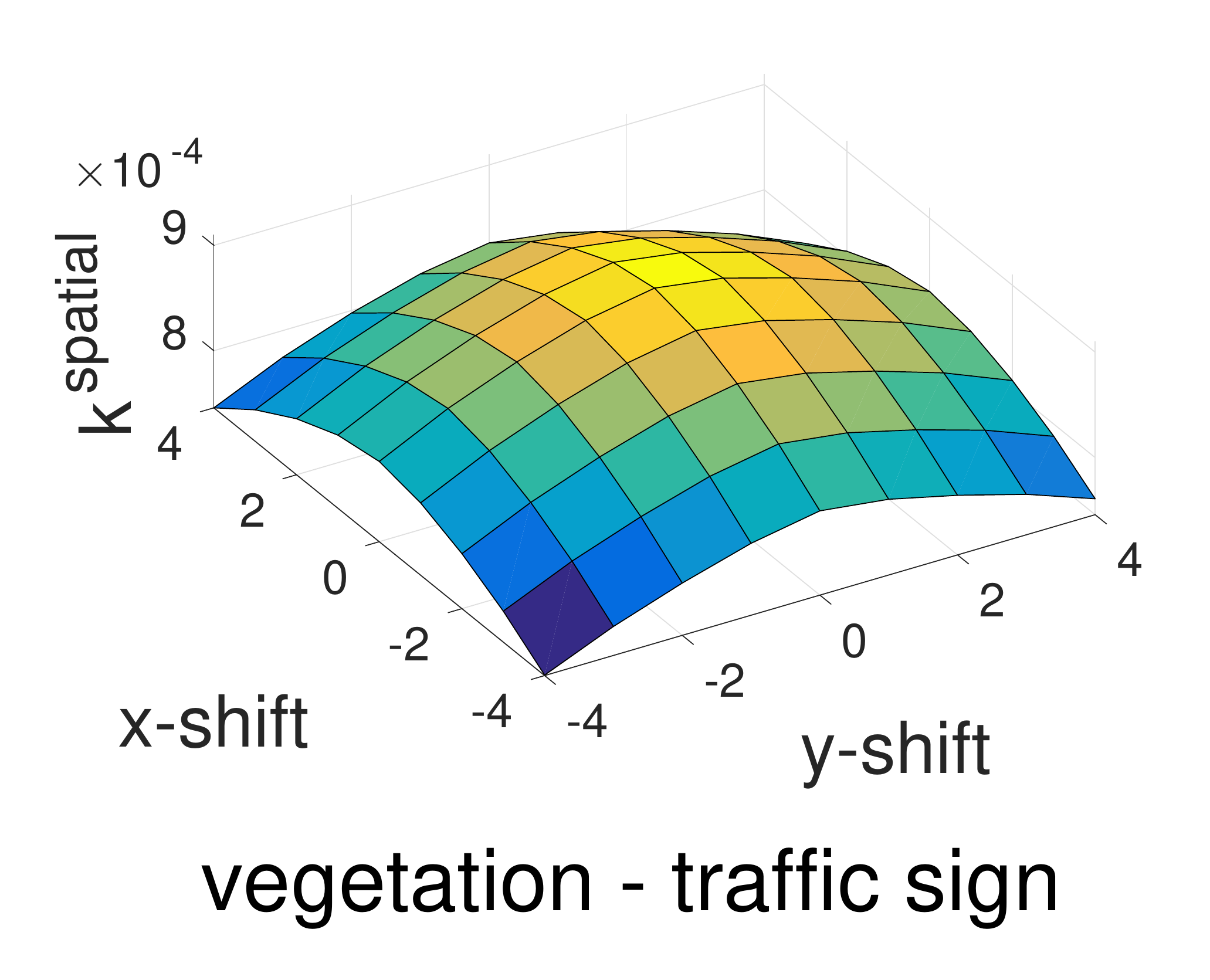}
   \includegraphics[width=0.30\linewidth]{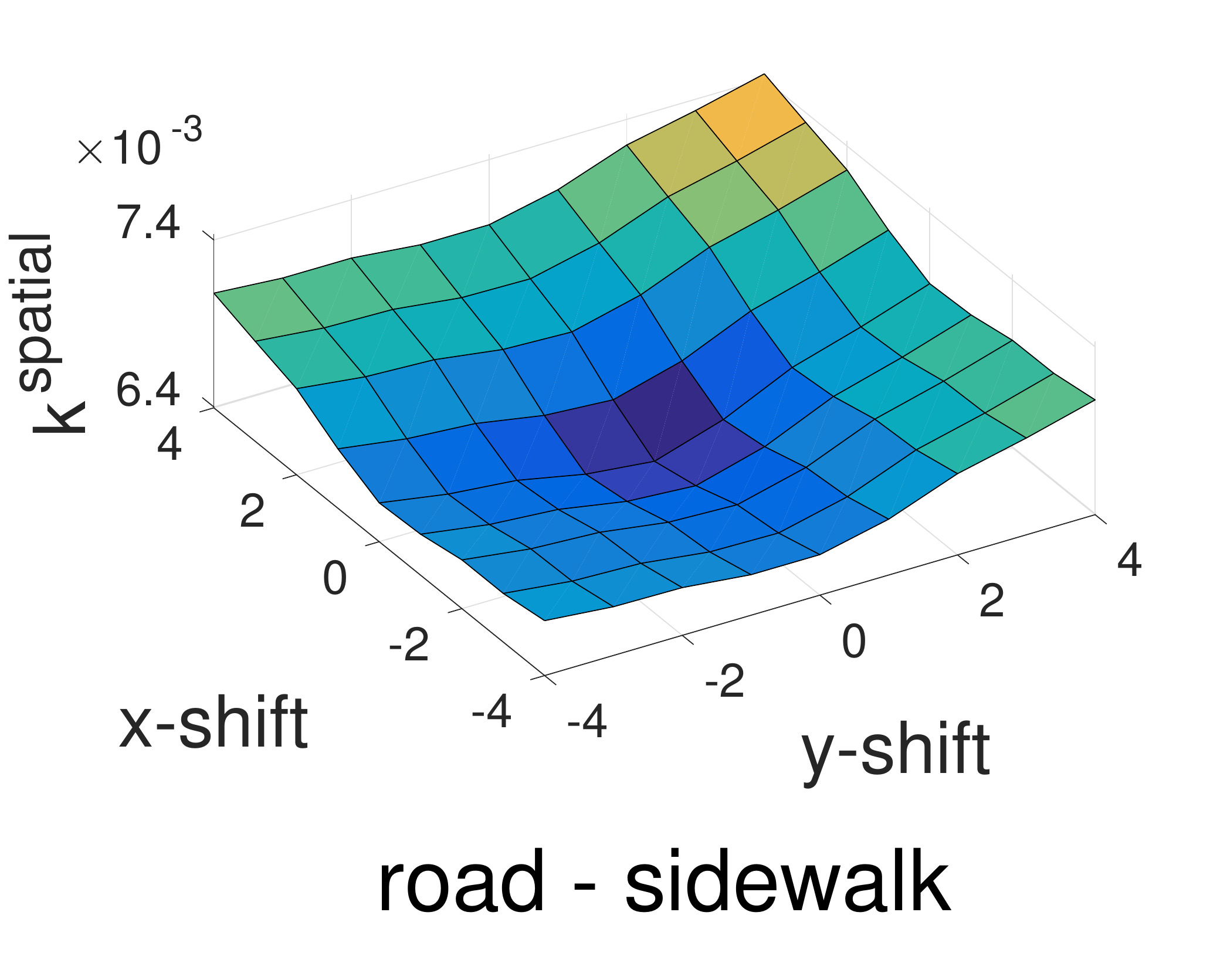}
\end{center}
    \vspace{-0.8cm}
   \caption{Learned spatially invariant CRF filters for {\sc Cityscapes}. These filters model contextual relationships between classes and their values can be understood as the energy added when setting one pixel to the first class (e.g., vegetation) and the other pixel with relative position (x-shift,y-shift) to the second class (e.g., traffic sign). Note how the terrain-traffic light filter favours vertical edges. In addition, the model can learn filters of different shapes which is shown by the road-sidewalk filter. \vspace{-0.15cm} }
\label{fig:cs:filters}
\end{figure}

We leverage on the idea of embedding inference of graphical models into a neural network. 
An early example of this idea was presented in \cite{bottou-etal-cvpr-1997} where the authors back propagated through the Viterbi algorithm when designing a document recognition system.
Similar to \cite{crfasrnn_iccv2015,arnab_eccv_2016,belanger-mccallum-icml-2016,wang-etal-nips-2016}, we use a recurrent neural network to unroll the iterative inference steps of a CRF. This was first used in \cite{crfasrnn_iccv2015} and \cite{schwing-urtasun-arxiv-2015} to imitate mean-field inference and to train a fully convolutional network \cite{Long/cvpr2015,chen14semantic} along with a CRF end-to-end via back propagation. In contrast to mean-field, we do not optimize the KL-divergence. Instead, we use a gradient descent approach for the inference that directly minimises the Gibbs energy of the random field and hence avoids the approximations of mean-field. A similar framework was recently suggested in \cite{belanger-mccallum-icml-2016} for multi-label classification problems in machine learning with impressive results. However, \cite{belanger-mccallum-icml-2016} uses a Structured SVM approach for training whereas we do back propagation through the actual steps of the gradient descent method.
Moreover, \cite{desmaison-etal-eccv-2016} have recently shown that one can obtain lower energies compared to mean-field inference using gradient descent based optimization schemes. Still, we lack formal algorithmic guarantees of the solution quality compared to, e.g., graph cuts~\cite{boykov-etal-pami-2001}.

In many works, the pairwise potentials consist of parametrized Gaussians \cite{kraehenbuehl-koltun-icml-2013,crfasrnn_iccv2015,arnab_eccv_2016} and it is only the parameters of this Gaussian which are learned. Our framework can learn arbitrary pairwise potentials which need not be Gaussian, cf.\ Fig.~\ref{fig:cs:filters}. 
An early work which learned potentials of a linear chain CRF for sequence modelling is \cite{peng-etal-nips-2009}.
In \cite{chen-etal-icml-2015}, a general framework for learning arbitrary potentials in deep structured models was proposed based on approximate Maximum Likelihood learning. One of the advantages with that framework is that data likelihood is maximized in the learning process. However, this involves approximating the partition function which is otherwise intractable. This hinders the handling of large structured output spaces like in our case.
The experiments of \cite{chen-etal-icml-2015} and \cite{belanger-mccallum-icml-2016} are limited to multi-label classification where there is no spatial relationship between different labels as in pixel-labelling tasks.
Morever, \cite{Kirillov/accv2016} have also observed that the memory requirements of the method of \cite{chen-etal-icml-2015} renders it infeasible for the large datasets common in Computer Vision.
Another approach to learning arbitrary pairwise potentials was presented in \cite{Kirillov/accv2016} which uses Gibbs sampling. Again they struggle with the difficulty of computing the partition function.
In the end, only experiments on synthetic data restricted to learned 2D potentials are presented.

The authors of \cite{lin-etal-cvpr-2016} and \cite{chandra-kokkinos-eccv-2016} also learn arbitrary pairwise potentials to model contextual relations between parts of the image. 
However, their approaches still perform post-processing with a CRF model with parametric Gaussian potentials.
In \cite{jampani:cvpr:2016}, a pairwise potential is learned based on sparse bilateral filtering. Applying such a filter can be regarded as one iteration of mean field CRF inference. Contrasting \cite{jampani:cvpr:2016}, we not only learn sparse high-dimensional bilateral filters, but also learn arbitrary spatial filters. Such spatial 2D potentials are computationally much more efficient and easier to analyze and interpret compared to their high-dimensional counterparts.
In summary, our contributions are as follows.
\begin{itemize}
\setlength\itemsep{0.05cm}

\item Our main contribution is a new framework for non-parametric CRF inference and learning which is integrated with standard CNNs. During inference, we directly minimize the CRF energy using gradient descent and during training, we back propagate through the gradient descent steps for end-to-end learning.

\item We analyze the learned filter kernels empirically and demonstrate that in many cases it is advantageous with non-Gaussian potentials.

    \item We experimentally compare our approach to several leading methodologies, e.g., \cite{crfasrnn_iccv2015,ghiasi-fowlkes-eccv-2016,Long/cvpr2015} and improve on state of the art on two public benchmarks: {\sc NYU~V2}~\cite{Wang/cvpr2014} and {\sc Cityscapes}~\cite{cityscapes}.
    
\end{itemize}
Our framework has been implemented in both {\sc Caffe}~\cite{jia2014caffe} and {\sc MatConvNet}~\cite{vedaldi15matconvnet}, and all source code will be made publicly available to facilitate further research.





\section{CRF Formulation}
Consider a Conditional Random Field over $N$ discrete random variables $\mathcal{X} = \{X_1, ... , X_N\}$ conditioned on an observation $\bm{I}$ and let $\mathcal{G} = \{\mathcal{V}, \mathcal{E}\}$ be an undirected graph whose vertices are the random variables $\{X_1, ... , X_N\}$. Each random variable corresponds to a pixel in the image and takes values from a predefined set of $L$ labels $\mathcal{L} = \{\, 0, ... , L-1 \,\}$. The pair $(\mathcal{X},\bm{I})$ is modelled as a CRF characterized by the Gibbs distribution
\begin{equation}
P(\mathcal{X} = \bm{x} | \bm{I}) = \frac{1}{Z(\bm{I})} \exp(-E(\bm{x}|\bm{I})),
\label{eq:gibbs_distribution}
\end{equation}
where $E(\bm{x}|\bm{I})$ denotes the Gibbs energy function with respect to the labeling $\bm{x} \in \mathcal{L}^N$ and $Z(\bm{I})$ is the partition function. To simplify notation the conditioning on $\bm{I}$ will from now on be dropped. The MAP inference problem for the CRF model is equivalent to the problem of minimizing the energy $E(\bm{x})$. In this paper, we only consider energies containing unary and pairwise terms. The energy function can hence be written as
\begin{equation}
 E(\bm{x}) = \sum_{i\in\mathcal{V}} \psi_i(x_i) + \sum_{(i,j)\in\mathcal{E}} \psi_{ij}(x_i,x_j)
\end{equation}
where $\psi_i : \mathcal{L} \rightarrow \mathbb{R}$ and $\psi_{ij} : \mathcal{L} \times \mathcal{L} \rightarrow \mathbb{R}$ are the unary and pairwise potentials, respectively.

\subsection{Potentials}
The unary potential $\psi_i(x_i)$ specifies the energy cost of assigning label $x_i$ to pixel $i$. In this work we obtain our unary potentials from a CNN. Roughly speaking, the CNN outputs a probability estimate of each pixel containing each class. Denoting the output of the CNN
for pixel $i$ and class $x_i$ as $z_{i:x_i}$, the unary potential is
\begin{equation}
\psi_i(x_i) = - w_u \log (z_{i:x_i} + \epsilon) \label{eq:unary}
\end{equation}
where $w_u$ is a parameter controlling the impact of the unary potentials, and $\epsilon$ is introduced to avoid numerical problems.

The pairwise potential $\psi_{ij}(x_i,x_j)$ specifies the energy cost of assigning label $x_i$ to pixel $i$ while pixel $j$ is assigned label $x_j$. Introducing pairwise terms in our model enables us to take dependencies between output data into account. We consider the following set of pairwise potentials
\begin{equation} \label{eq:pairwise}
\psi_{ij}(x_i,x_j) = k^{spatial}_{x_i,x_j}(\bm{p}_i-\bm{p}_j) + k^{bilateral}_{x_i,x_j}(\bm{f}_i - \bm{f}_j)
\end{equation}
Here $k^{spatial}_{x_i,x_j}$ denotes a spatial kernel with compact support. 
Its value depends on the relative position coordinates $\bm{p}_i - \bm{p}_j$ between pixels $i$ and $j$. We do not restrict these spatial terms to any specific shape, cf.\ Fig.~\ref{fig:cs:filters}. However, we restrict the support of the potential meaning that if pixels $i$ and $j$ are far apart, then the value of $k^{spatial}_{x_i,x_j}(\bm{p}_i-\bm{p}_j)$ will be zero. CRFs with Gaussian potentials do not in theory have compact support, and therefore, they are often referred to as dense. However, in practice, the exponential function in the kernel drops off quickly and effectively. The interactions between pixels far apart are negligible and are commonly truncated to 0 after two standard deviations.

The term $k^{bilateral}_{x_i,x_j}$ is a bilateral kernel which depends on the feature vectors $\bm{f}_i$ and $\bm{f}_j$ for pixels $i$ and $j$, respectively. Following several previous works \cite{kraehenbuehl-koltun-icml-2013,crfasrnn_iccv2015,chen_arxiv_2016}, we let the vector depend on pixel coordinates $\bm{p}_i$ and RGB values associated to the pixel, hence $\bm{f}_i$ is a $5$-dimensional vector.
Note that for both the spatial and the bilateral kernels, there is one kernel for each label-to-label ($x_i$ and $x_j$) interaction to enable the model to learn differently shaped kernels for each of these interactions.


\subsection{Multi-label Graph Expansion and Relaxation}
To facilitate a continuous relaxation of the energy minimisation problem we start off by expanding our original graph in the following manner. Each vertex in the original graph $\mathcal{G}$ will now be represented by $L$ vertices $X_{i:\lambda}$, $\lambda \in \mathcal{L}$. In this way, an assignment of labels in $\mathcal{L}$ to each variable $X_i$ is equivalent to an assignment of boolean labels $0$ or $1$ to each node $X_{i:\lambda}$, whereby an assignment of label 1 to $X_{i:\lambda}$ means that in the multi-label assignment, $X_i$ receives label $\lambda$. 
As a next step, we relax the integer program by allowing real values on the unit interval $[0,1]$ instead of booleans only. We denote the relaxed variables $q_{i:\lambda}\in[0,1]$. We can now write our problem as a quadratic program

\begin{equation}
\label{realvalprog}
\begin{array}{lll}
\displaystyle \min & \displaystyle \sum_{i\in\mathcal{V},\lambda \in \mathcal{L}} \psi_{i}(\lambda)q_{i:\lambda} +  \sum_{\substack{(i,j)\in\mathcal{E} \\ \lambda,\mu \in \mathcal{L}}} & \displaystyle \psi_{ij}(\lambda,\mu)q_{i:\lambda}q_{j:\mu}  \\
\textrm{s.t.} & q_{i:\lambda} \geq 0 & \forall i\in \mathcal{V},\lambda\in\mathcal{L}  \\
&\displaystyle \sum_{\lambda \in \mathcal{L}} q_{i:\lambda} = 1 &\forall i\in \mathcal{V}.   \vspace{-0.3cm}
\end{array}
\end{equation}

Note that the added constraints ensure that our solution lies on the probability simplex.
A natural question is what happens when the domain is enlarged, allowing real values instead of booleans. Somewhat surprisingly, the relaxation is tight \cite{boros-hammer-dam-2002}.
\begin{prop} \label{prop:tight}
Let $E(\bm{x}^*)$ and $E(\bm{q}^*)$ denote the optimal values of \eqref{realvalprog}, where $\bm{x}^*$ is restricted to boolean values. Then,
\[
        E(\bm{x}^*) = E(\bm{q}^*).
\]
\end{prop}
In the supplementary material, we show that for {\em any} real $\bm{q}$, one can obtain a binary $\bm{x}$ such that $E(\bm{x}) \le E(\bm{q})$. In particular, it will be true for $\bm{x}^*$ and $\bm{q}^*$, which implies $E(\bm{x}^*) = E(\bm{q}^*)$. Note that the proof is constructive.


%
%


In summary, it has been shown that to minimize
the energy function $E(\bm{x})$ over $\bm{x}\in \mathcal{L}^N$,
one may work in the continuous domain,
minimize over $\bm{q}$, and then
replace any solution $\bm{q}$ by a discrete solution $\bm{x}$
which has lower or equal energy. It will only
be possible to find a local solution $\bm{q}$, but
still the discrete solution $\bm{x}$ 
will be no worse than $\bm{q}$.



\section{Minimization with Gradient Descent} \label{sec:gd}
To solve the program stated in \eqref{realvalprog} we propose an optimization scheme based on projected gradient descent, see Algorithm~\ref{alg:gradient}.
It was designed with an extra condition in mind, that all operations should be differentiable to enable back propagation during training.


\begin{algorithm}
\textbf{Initialize} $\bm{q}^0$ \\
\For{$t$ from $0$ to $T-1$}{
  Compute the gradient $\nabla_{\bm{q}}E(\bm{q}^t)$. \\
  Take a step in the negative direction, $\bm{\tilde{q}^{t+1}} = \mathbf{q^{t}} - \gamma \; \nabla_{\mathbf{q}}\bm{E}$. \\ 
 
  
  Project $\tilde{q}^{t+1}_{i:\lambda}$ to the simplex $\triangle^L$ satisfying $\sum_{\lambda \in \mathcal{L}} \tilde{q}_{i:\lambda} = 1$ and $0 \leq \tilde{q}_{i:\lambda} \leq 1$, $\bm{q}^{t+1} = \text{Proj}_{\triangle^L}(\tilde{\bm{q}})$. 
}
\KwOut{$\bm{q}^{T-1}$}
\caption{\label{alg:gradient}Algorithm 1. Projected gradient descent algorithm.}
\end{algorithm}

The gradient $\nabla_{\bm{q}}E$ of the objective function $E(\bm{q})$ in \eqref{realvalprog} has the following elements
\begin{equation}
\frac{\partial E}{\partial q_{i:\lambda}} = \psi_{i}(\lambda) + \sum_{\substack{j:(i,j)\in\mathcal{E} \\ \mu \in \mathcal{L}}} \psi_{ij}(\lambda,\mu)q_{j:\mu}.
\end{equation}
The contribution from the spatial kernel in $\psi_{ij}$, cf.\ \eqref{eq:pairwise}, can be written as
\begin{equation}
v^{spatial}_{i:\lambda} = \sum_{\substack{j:(i,j)\in\mathcal{E} \\ \mu \in \mathcal{L}}} k^{spatial}_{\lambda,\mu}(\bm{p}_i-\bm{p}_j) q_{j:\mu}.
\end{equation}
Since the value of the kernel $v^{spatial}_{i:\lambda}$ only depends on the relative position of pixels $i$ and $j$, the contribution for all pixels and classes can be calculated by passing $q_{j:\mu}$ through a standard convolution layer consisting of $L \times L$ filters of size $(2s+1) \times (2s+1)$ where $L$ is the number of labels and $s$ the number of neighbours each pixel interacts with in each dimension.

The contribution from the bilateral term is
\begin{equation}
v^{bilateral}_{i:\lambda} = \sum_{\substack{j:(i,j)\in\mathcal{E} \\ \mu \in \mathcal{L}}} k^{bilateral}_{\lambda,\mu}(\bm{f}_i - \bm{f}_j) q_{j:\mu}.
\end{equation}
For this computation we utilize the method presented by Jampani \etal \cite{jampani:cvpr:2016} which is based on the permutohedral lattice introduced by Adams \etal \cite{adams-etal-2010}. Efficient computations are obtained by using the fact that the feature space is generally sparsely populated. Similar to the spatial filter we get $L \times L$ filters, each having size of $(s+1)^{d+1}-s^{d+1}$ where $s$ is the number of neighbours each pixel interacts with in each dimension in the sparse feature space. 


Next, we simply take a step in the negative direction of the gradient according to 
\begin{equation}
\bm{\tilde{q}^{t+1}} = \mathbf{q^{t}} - \gamma \; \nabla_{\mathbf{q}}\bm{E},    
\end{equation}
where $\gamma$ is the the step size. 

Finally, we want to project our values onto the simplex $\triangle^L$ satisfying $\sum_{\lambda \in \mathcal{L}} q_{i:\lambda} = 1$ and $0 \leq q_{i:\lambda} \leq 1$. This is done following the method by Chen \etal \cite{chen2011} that efficiently calculates the euclidean projection on the probability simplex $\triangle^L$, for details see the supplementary materials.
Note that this projection is done individually for each pixel~$i$.

\paragraph{Comparison to Mean-Field. }
In recent years, a popular choice for CRF inference is to apply the mean-field algorithm. One reason is that the kernel evaluations can be computed with fast bilateral filtering \cite{krahenbuhl-koltun-nips-2011}. As we have seen in this section, it can be accomplished with our framework as well. The main difference is that our framework directly optimizes the Gibbs energy which corresponds to the MAP solution while mean-field optimizes KL-divergence which does not.

\section{Integration in a Deep Learning Framework} \label{sec:deepgrad}

In this section we will describe how the gradient descent steps of Algorithm~\ref{alg:gradient} can be formulated as layers in a neural network. We need to be able to calculate error derivatives with respect to the input given error derivatives with respect to the output. In addition we need to be able to calculate the error derivatives with respect to the network parameters.
This will enable us to unroll the entire gradient descent process as a Recurrent Neural Network (RNN).
A schematic of the data flow for one step is shown in Fig.~\ref{fig:gradstep}. In the supplementary material, all derivative formulae are given. 


\begin{figure}[t]
\begin{center}
   \includegraphics[width=0.5\linewidth]{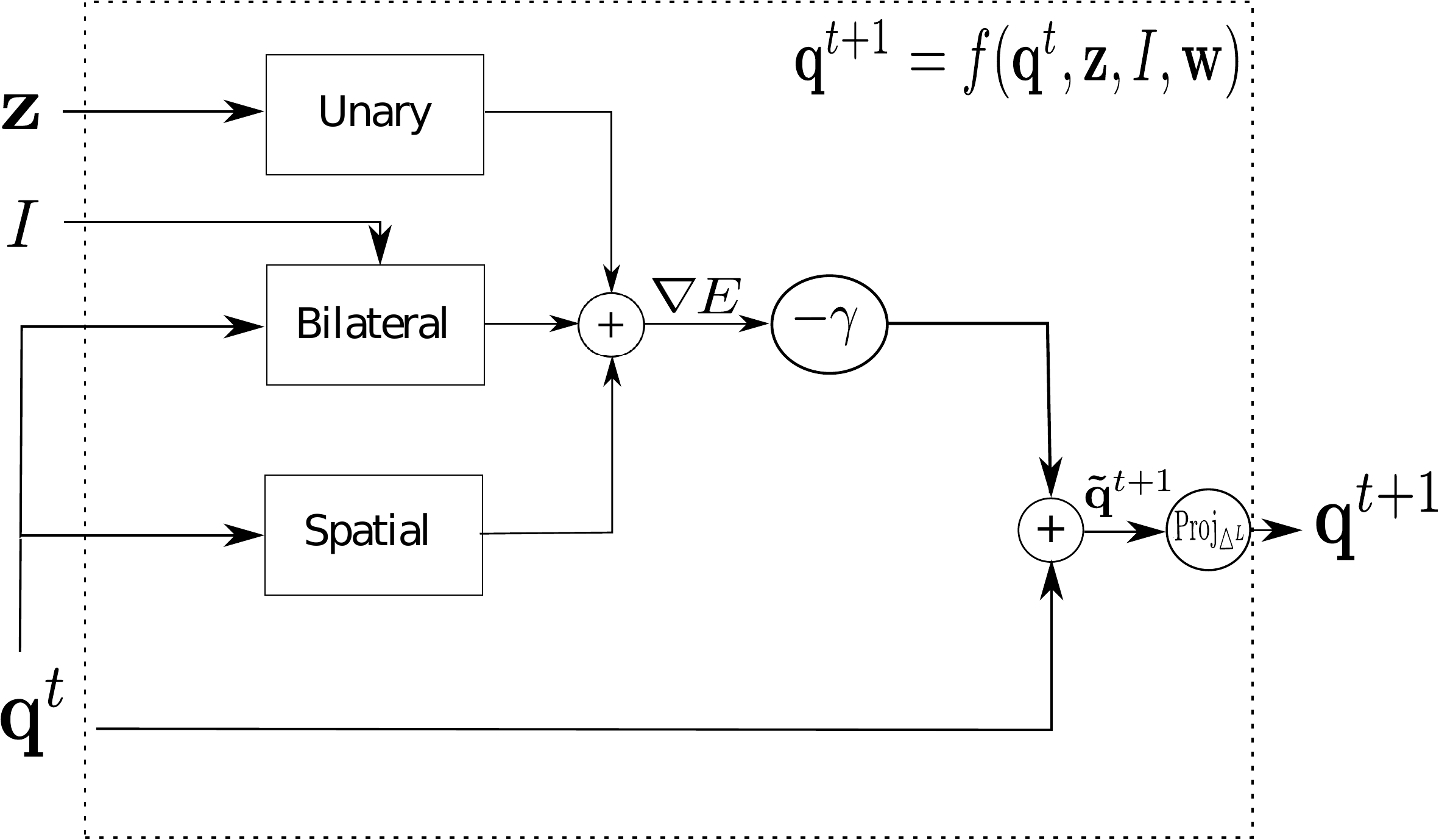}
\end{center}
    \vspace{-0.5cm}
   \caption{The data flow of one iteration of the projected gradient descent algorithm. Each rectangle or circle represent an operation that can be performed within a deep learning framework.}
\label{fig:gradstep}
\end{figure}

\paragraph{Initialization. }
The variables $\bm{q}^0$ are set as the output of the CNN, which has been pre-trained to estimate the probability of each pixel containing each class and has a softmax layers at the end to ensure that the variables lies within zero and one as well as sum to one for each pixel.

\paragraph{Gradient Computations. }
We have previously explained the gradient computations in Section~\ref{sec:gd} for the forward pass. To describe the calculation of the error derivatives we first notice that the gradient is calculated by summing three terms, the unary, spatial and bilateral pairwise terms. We can hence treat these three terms separately and combine them using element-wise summing.

The unary term in \eqref{eq:unary} is an element-wise operation with the CNN output as input and the unary weight $w_u$ as parameter. The operation is obviously differentiable with respect to both the layer input as well as its parameter.
Note that for $w_u$ we get a summation over all class and pixel indexes for the error derivatives while for the input the error derivatives are calculated element-wise.
The spatial pairwise term of the gradient can be calculated efficiently using standard 2D convolution. In addition to giving us an efficient way of performing the forward pass we can also utilize the 2D convolution layer to perform the backward pass, calculating the error derivatives with respect to the input and parameters.
Similar to the spatial term, the bilateral term is also calculated utilizing a bilateral filtering technique. Jampani \etal \cite{jampani:cvpr:2016} also presented a way to calculate the error derivatives with respect to the parameters for an arbitrary shaped bilateral filter. 


\paragraph{Gradient Step. }
Taking a step in the negative direction of the gradient is easily incorporated in a deep learning framework by using an element-wise summing layer. The layer takes the variables $\bm{q}^t$ as the first input and the gradient (scaled by $-\gamma$) as the second input.

\paragraph{Simplex Projection}
As a final step, the variables from the gradient step $\tilde{\bm{q}}^t$ are projected onto the simplex $\triangle^L$. In reality we use a leaky version of the last step of the projection algorithm to avoid error derivatives becoming zero during back propagation. Since the projection is done individually for each pixel it can be described as a function $\bm{f}(\tilde{\bm{q}}): \mathbb{R}^L \rightarrow \mathbb{R}^L$ of which we can calculate the Jacobian, see supplementary materials. Knowing the Jacobian, the error derivatives with respect to the input can be computed during back propagation.

\begin{figure}[t]
\begin{center}
   \includegraphics[width=0.5\linewidth]{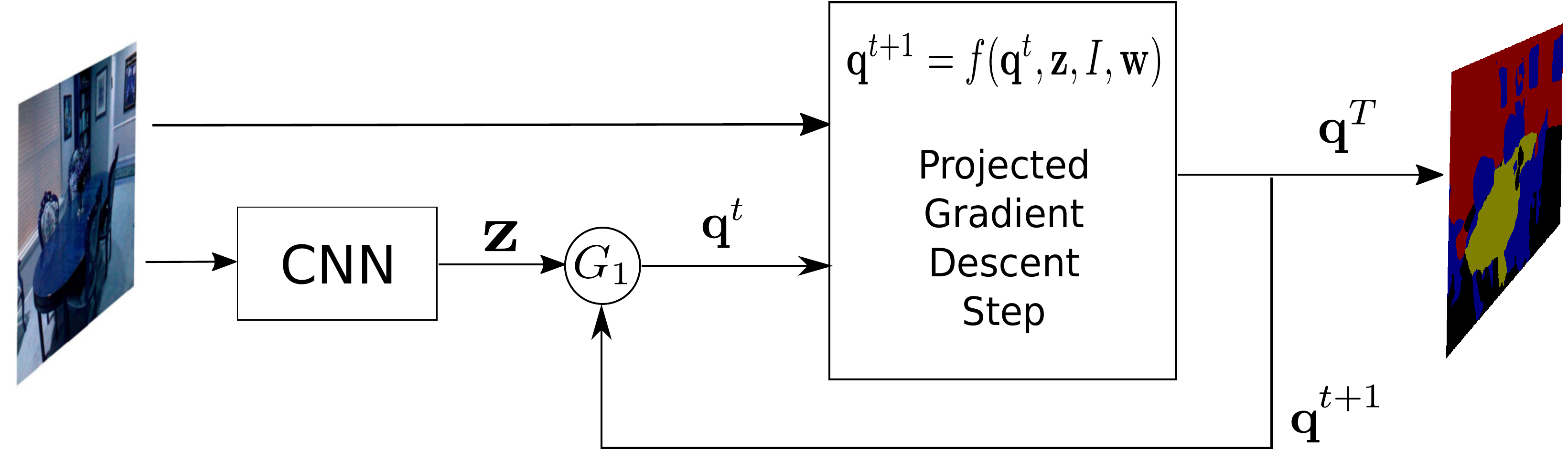}
\end{center}
\vspace{-0.6cm}
   \caption{The data flow of the deep structured model. Each rectangle or circle represent an operation that can be performed within a deep learning framework.}
\label{fig:rnn}
\end{figure}

\section{Recurrent Formulation as a Deep Structured Model}

Our iterative solution to the CRF energy minimisation problem by projected gradient descent, as described in the previous sections, is formulated as a Recurrent Neural Network (RNN).
The input to the RNN is the image, and the outputs of a CNN, as shown in Fig. \ref{fig:rnn}.
The CNN's output, $\mathbf{z}$, are the unary potentials and obtained after the final softmax layer (since the CNN is initially trained for classification).

Each iteration of the RNN performs one projected gradient descent step to approximately solve (\ref{realvalprog}). Thus, one update step can be represented by:
\begin{equation}
\bm{q}^{t+1} = f(\bm{q}^{t},\bm{z},I,\bm{w}).
\end{equation}
As illustrated in Fig. \ref{fig:rnn}, the gating function $G_1$ sets $\bm{q}^t$ to $\mathbf{z}$ at the first time step, and to $\bm{q}^{t-1}$ at all other time steps. 
In our iterative energy minimisation formulated as an RNN, the output of one step is the input to the next step.
We initialise at $t=0$ with the output of the unary CNN.

The output of the RNN can be read off $\bm{q}^{T}$ where $T$ is the total number of steps taken. 
In practice, we perform a set number of $T$ steps where $T$ is a hyperparameter.
It is possible to run the RNN until convergence for each image (thus a variable number of iterations per image), but we observed minimal benefit in the final Intersection over Union (IoU) from doing so, as opposed to $T=5$ iterations.

The parameters of the RNN are the filter weights for the spatial and bilateral kernels, and also the weight for the unary terms.
Since we are able to compute error derivatives with respect to the parameters, and input of the RNN, we can back propagate error derivates through our RNN to the preceding CNN and train our entire network end-to-end. 
Furthermore, since the operations of the RNN are formulated as filtering, training and inference can be performed efficiently in a fully-convolutional manner.


\paragraph{Implementation Details.}
Our proposed CRF model has been implemented in the {\sc Caffe} \cite{jia2014caffe} library, and also has a Matlab wrapper allowing it to be used in {\sc MatConvNet}~\cite{vedaldi15matconvnet}.
The Unary CNN part of our model is initialised from a pre-trained segmentation network. 
As we show in our experiments, different pre-trained networks for different applications can be used. 

The CRF model has several tunable parameters. The step size $\gamma$ and the number of iterations $T$ specify the properties of the gradient decent algorithm. Too high a step size $\gamma$ might make the algorithm not end up in a minimum while setting a low step size and a low number of iterations might not give the algorithm a chance to converge. The kernel sizes for the spatial and bilateral kernels also need to be set. Choosing the value of these parameters gives a trade-off between model expression ability and number of parameters, which may cause (or hinder) over-fitting.   

The spatial weights of the CRF model are all initialized as zero with the motivation that we did not want to impose a shape for these filters, but instead see what was learned during training. The bilateral filters were initialized as Gaussians with the common Potts class interaction (the filters corresponding to interactions between the same class were set to zero) as done in \cite{krahenbuhl-koltun-nips-2011,chen14semantic,crfasrnn_iccv2015}. Note that unlike \cite{krahenbuhl-koltun-nips-2011,chen14semantic} we are not limited to only Potts class interactions. 

\section{Experiments} \label{sec:exp}

We evaluate the proposed approach on three datasets: {\sc Weizmann Horse}~\cite{weizmann},
{\sc NYU V2}~\cite{Wang/cvpr2014} and
{\sc Cityscapes}~\cite{cityscapes}.
In these experiments, we show that the proposed approach, denoted CRF-Grad, has advantages over baseline approaches such as CRFasRNN~\cite{crfasrnn_iccv2015} and complement other networks such as FCN-8s~\cite{Long/cvpr2015} and LRR~\cite{ghiasi-fowlkes-eccv-2016}. In addition we compare our inference method to mean-field inference. Hyperparameter values not specified in this section can be found in the supplementary materials.


\subsection{Weizmann Horse}
The {\sc Weizmann Horse} dataset contains 328 images of horses in different environments. We divide these images into a training set of 150 images, a validation set of 50 images and a test set of 128 images. Our purpose is to verify our ability to learn reasonable kernels and study the effects of different settings on a relatively small dataset.

The CNN part of our model was initialized as an FCN-8s network~\cite{Long/cvpr2015} pre-trained without the CRF layer. We then compare several variants of our model. We start off by training a variant of our CRF model only using the 2D spatial kernel. We compare these results to using a Gaussian spatial filter, where the parameters for the Gaussian kernel were evaluated using cross-validation. In addition we train the full model with both the spatial and bilateral kernels, once keeping the filters fixed as Gaussians and once allowing arbitrarily shaped filters. 

\begin{figure*}[t]
\begin{center}
   \setlength\tabcolsep{1pt} 
   \begin{tabular}{cccccccc}
   
   Input & $\bm{z}_{\text{horse}}$ &  $\bm{q}^1_{\text{horse}}$ & $\bm{q}^2_{\text{horse}}$ & $\bm{q}^3_{\text{horse}}$ & $\bm{q}^4_{\text{horse}}$ & $\bm{q}^5_{\text{horse}}$ & Ground truth \\
   
   \includegraphics[width=0.12\linewidth]{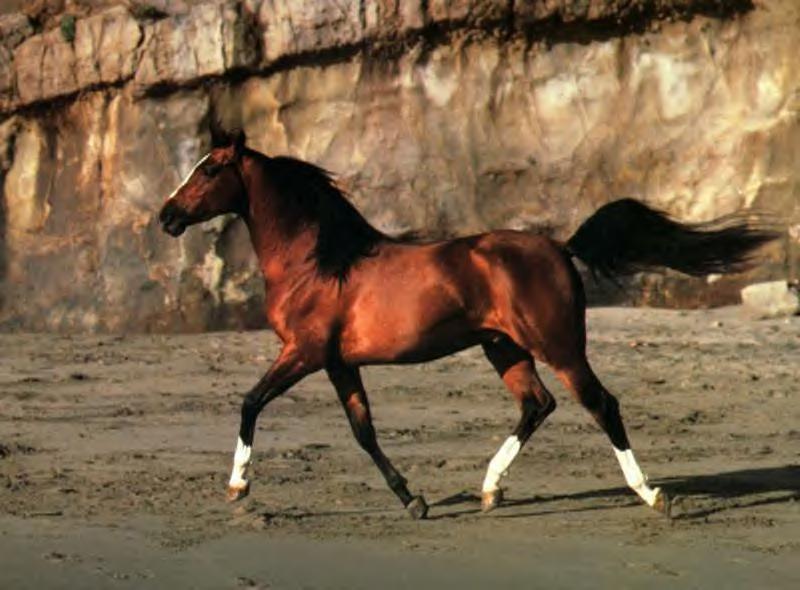} &
   \includegraphics[width=0.12\linewidth]{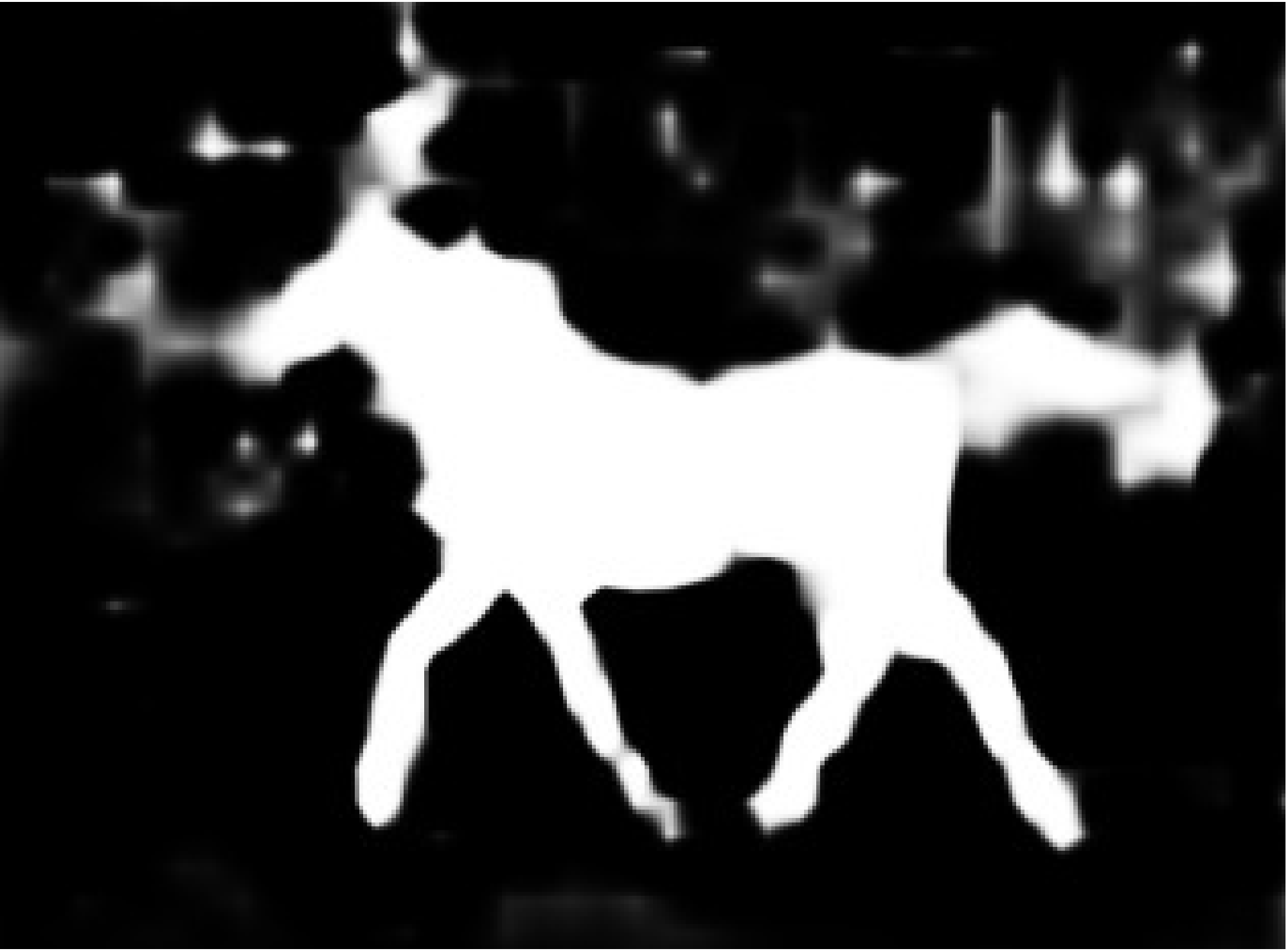} &
   \includegraphics[width=0.12\linewidth]{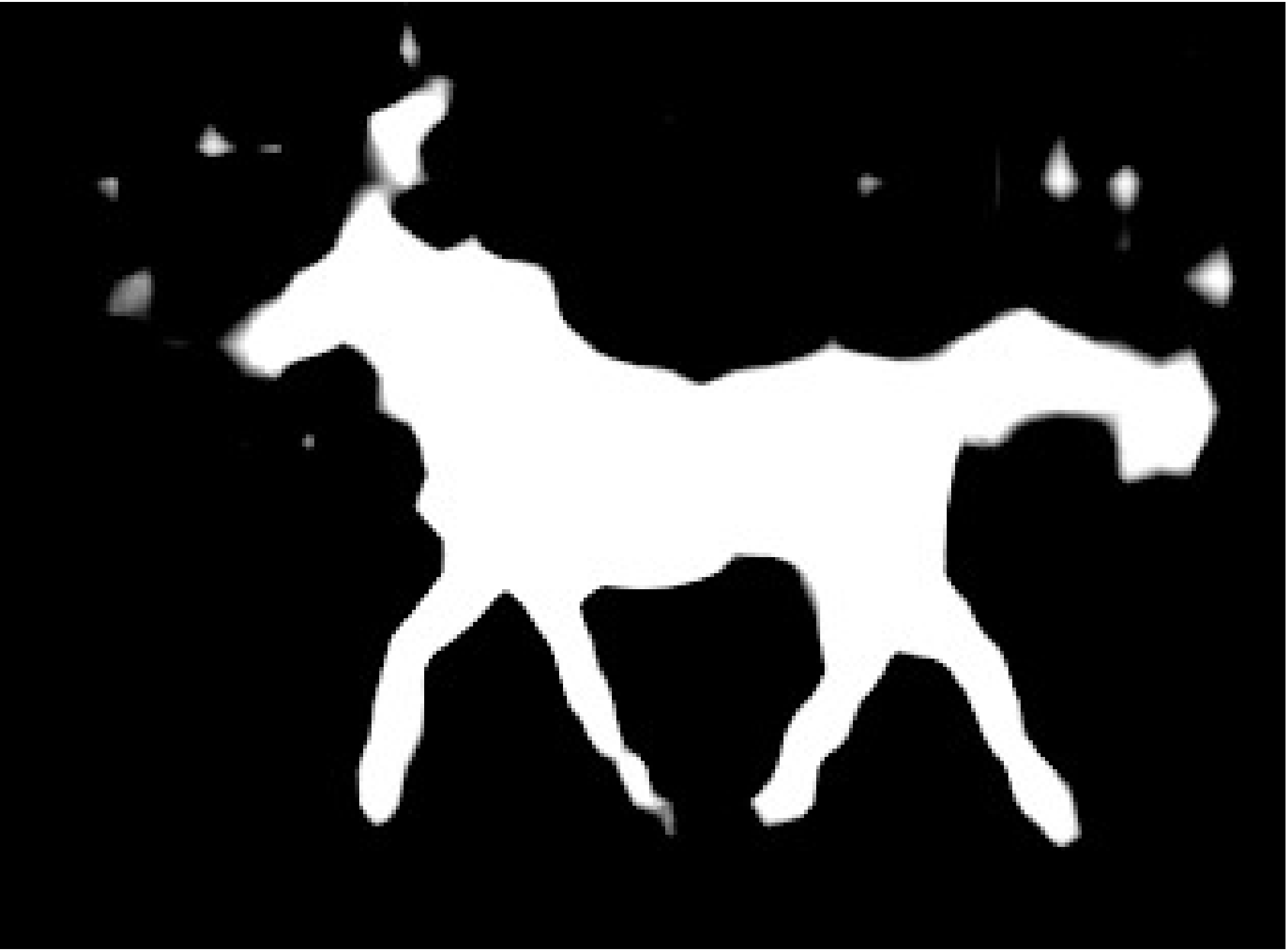} &
   \includegraphics[width=0.12\linewidth]{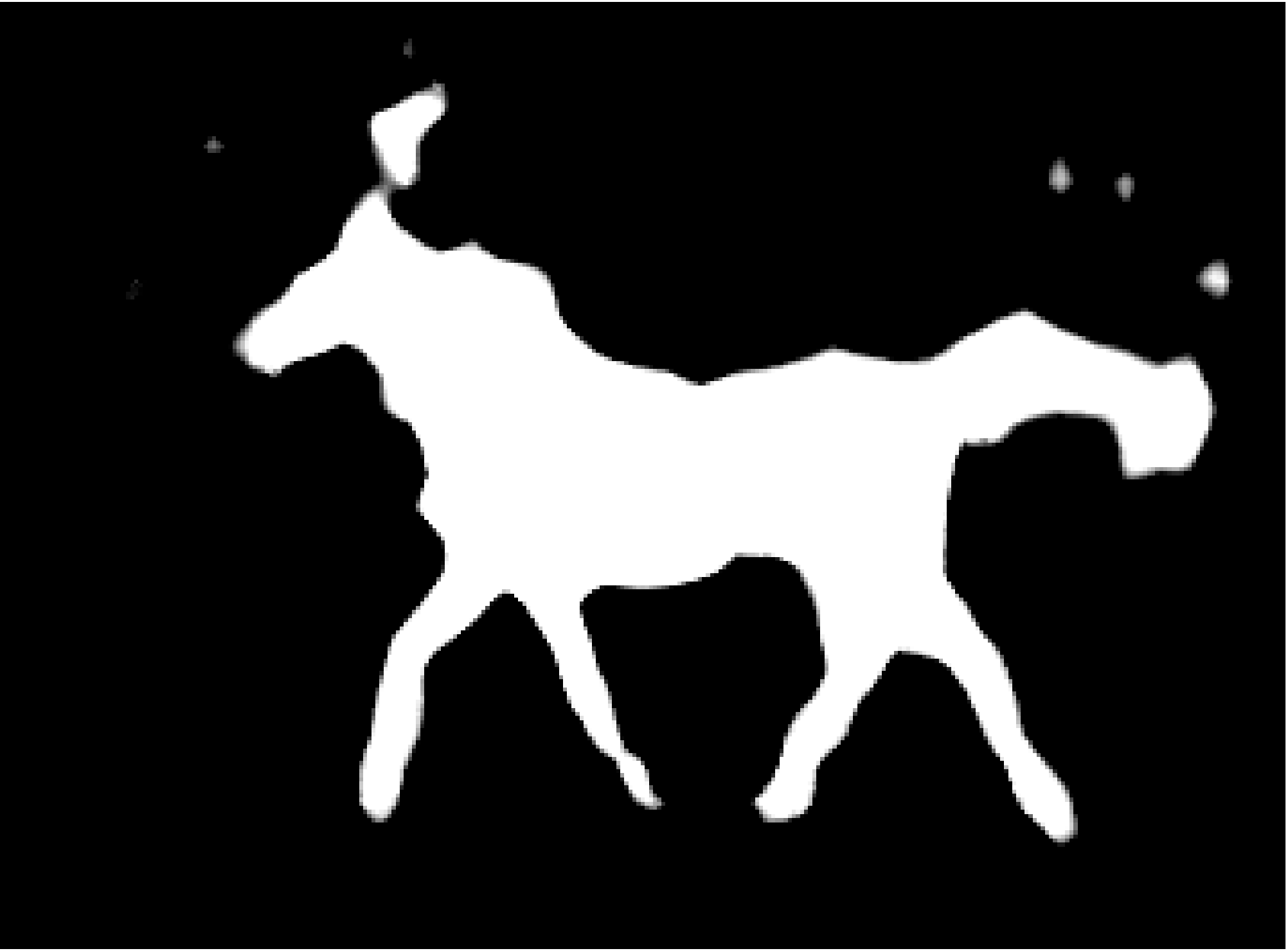} &
   \includegraphics[width=0.12\linewidth]{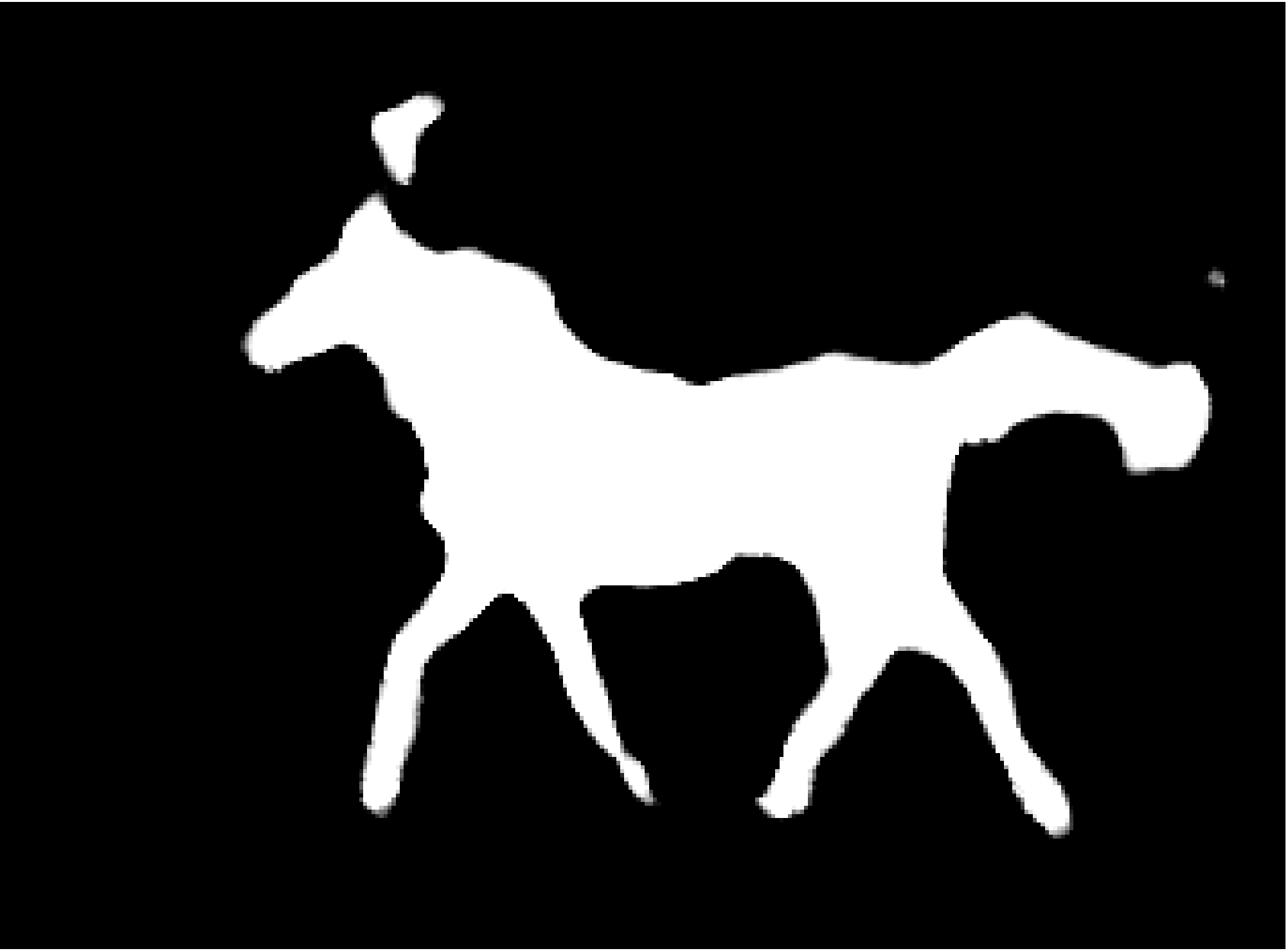} &
   \includegraphics[width=0.12\linewidth]{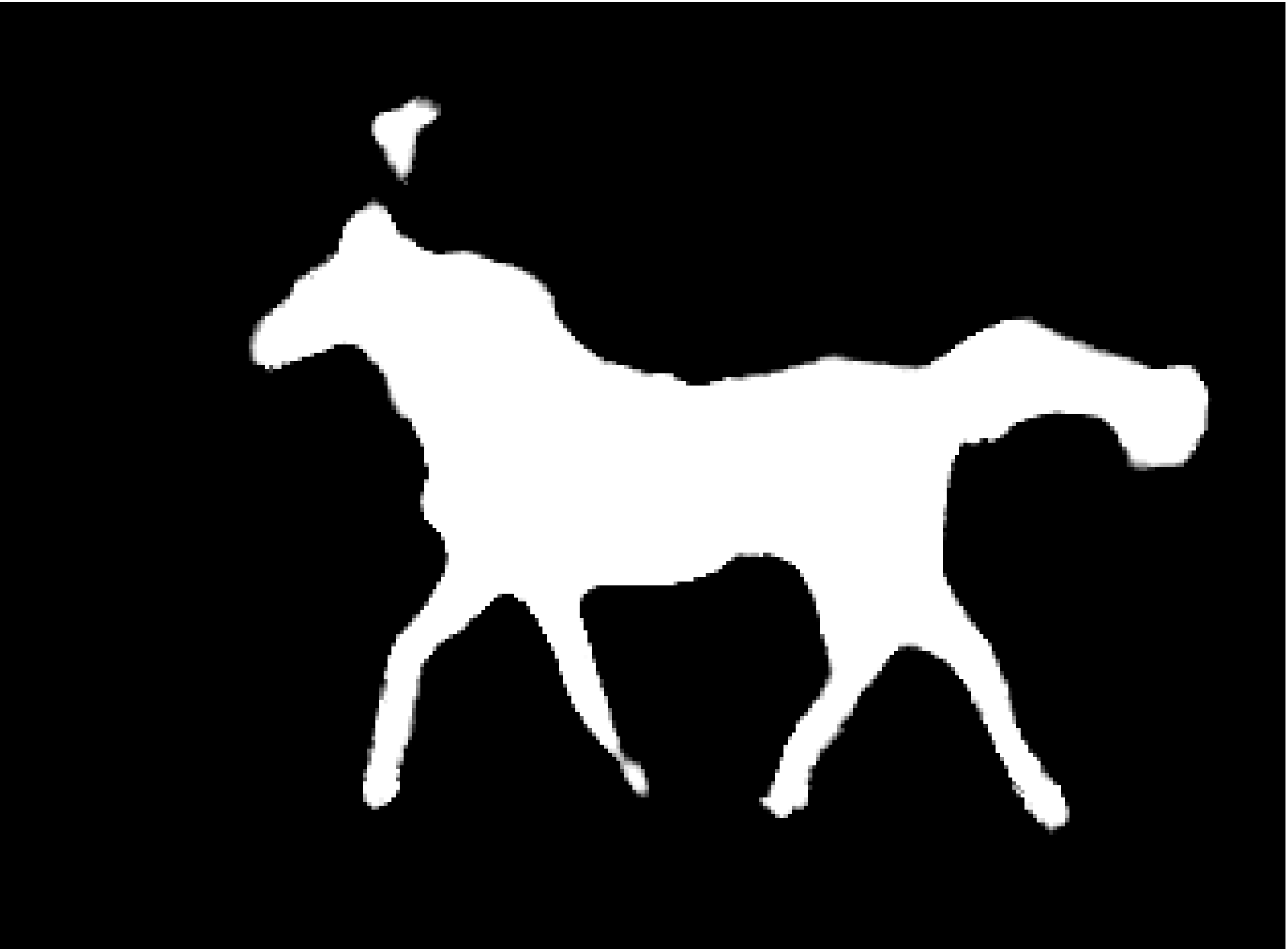} &
   \includegraphics[width=0.12\linewidth]{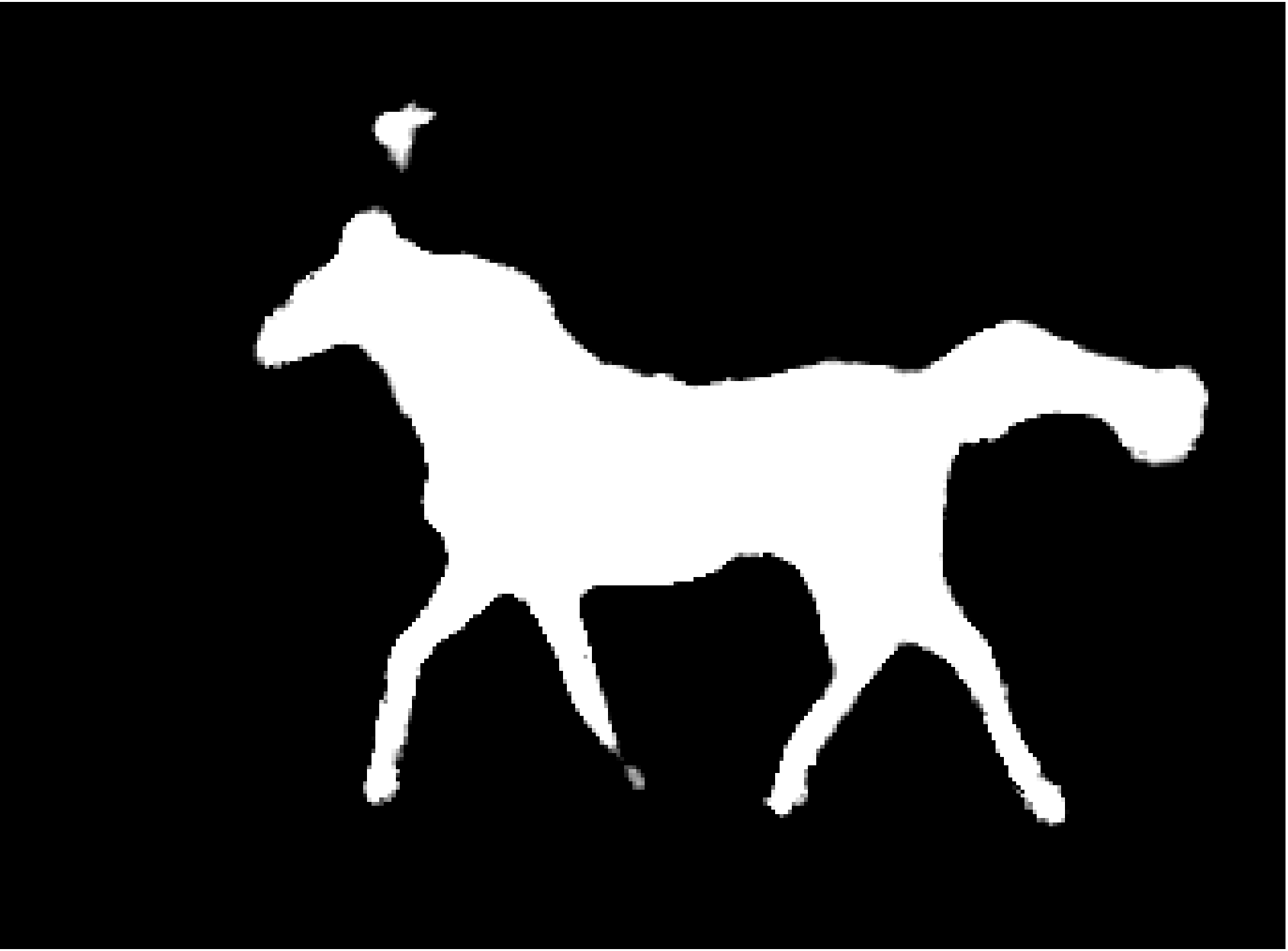} &
   \includegraphics[width=0.12\linewidth]{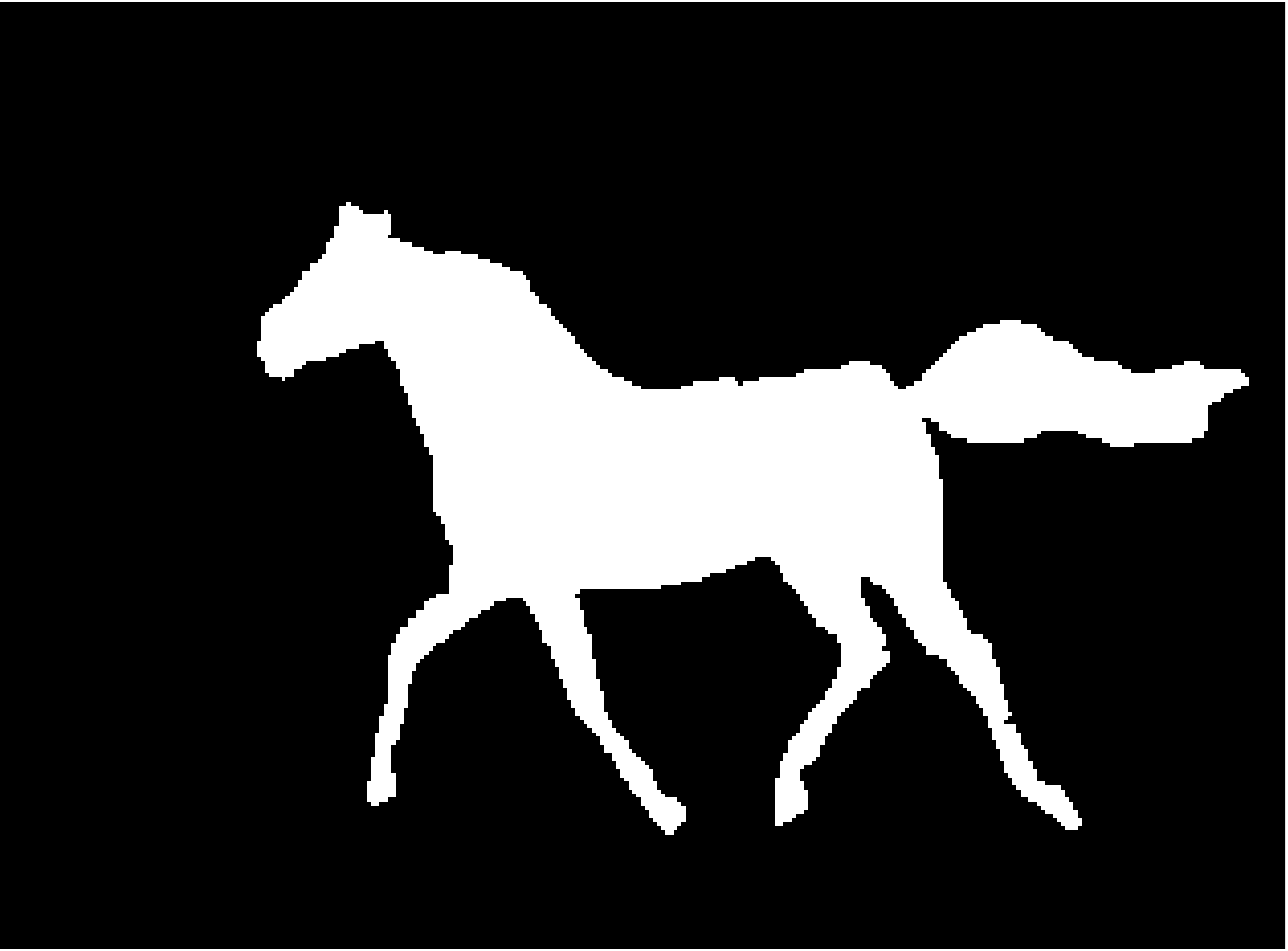} \\

   \includegraphics[width=0.12\linewidth]{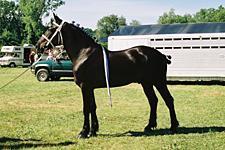} &
   \includegraphics[width=0.12\linewidth]{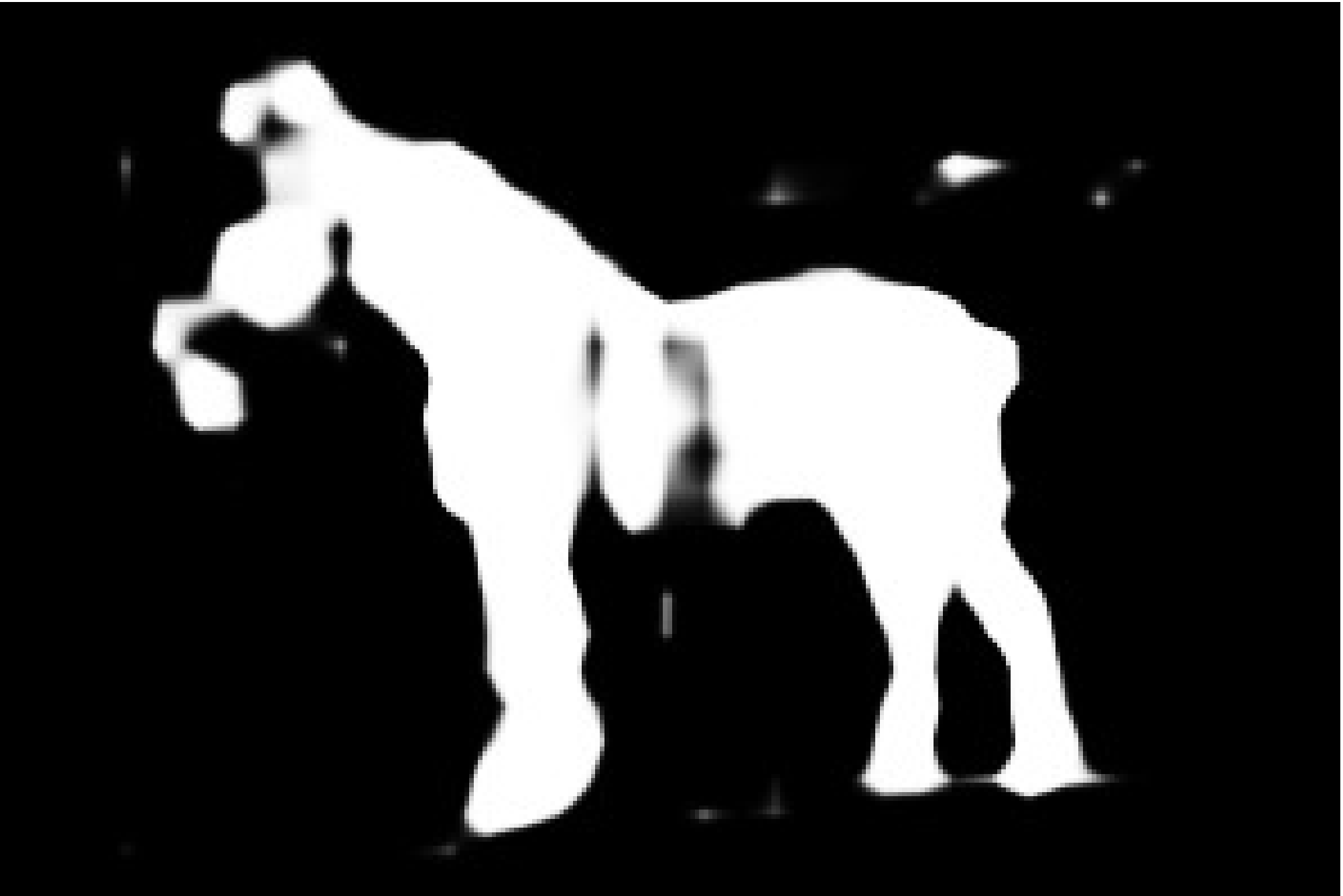} &
   \includegraphics[width=0.12\linewidth]{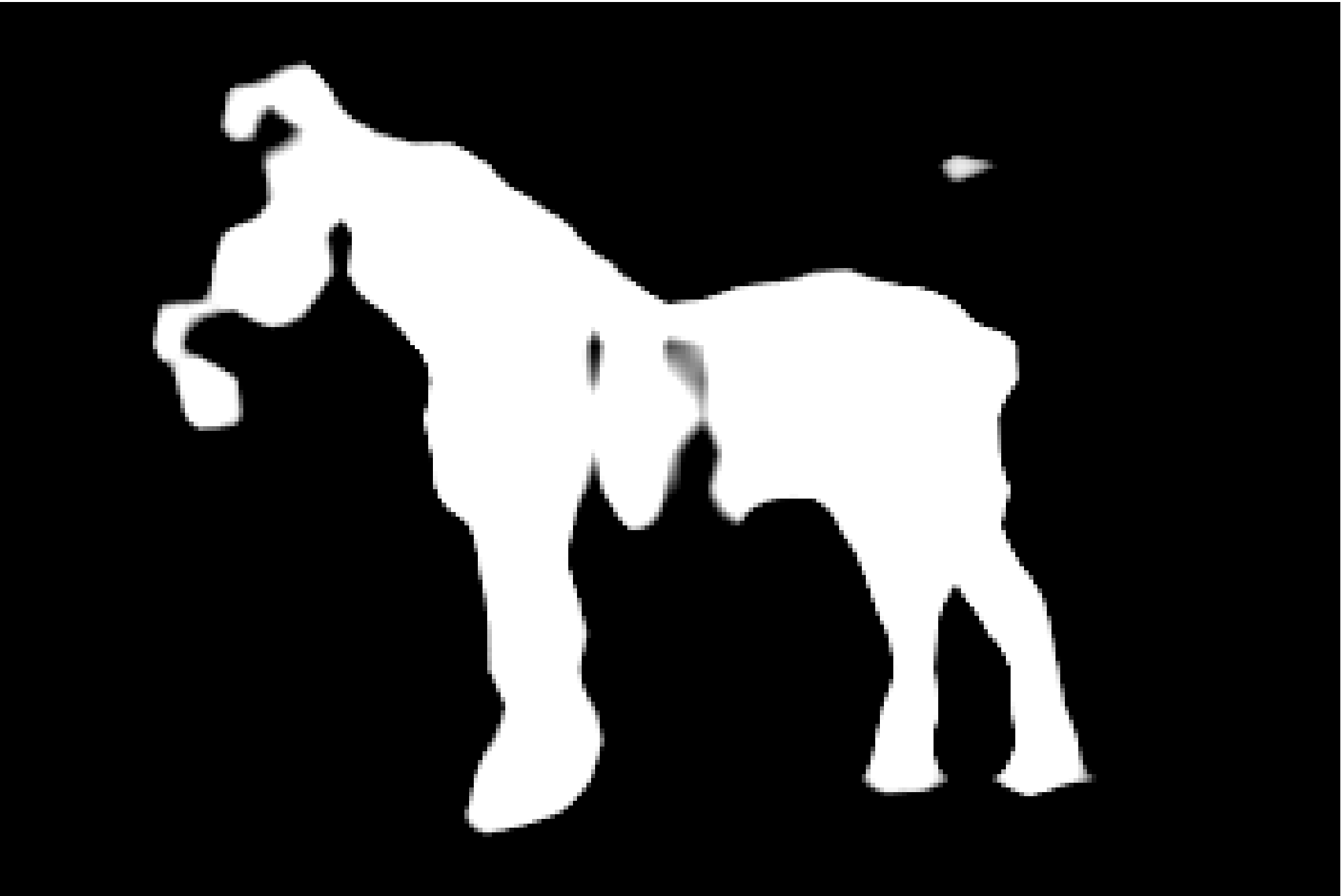} &
   \includegraphics[width=0.12\linewidth]{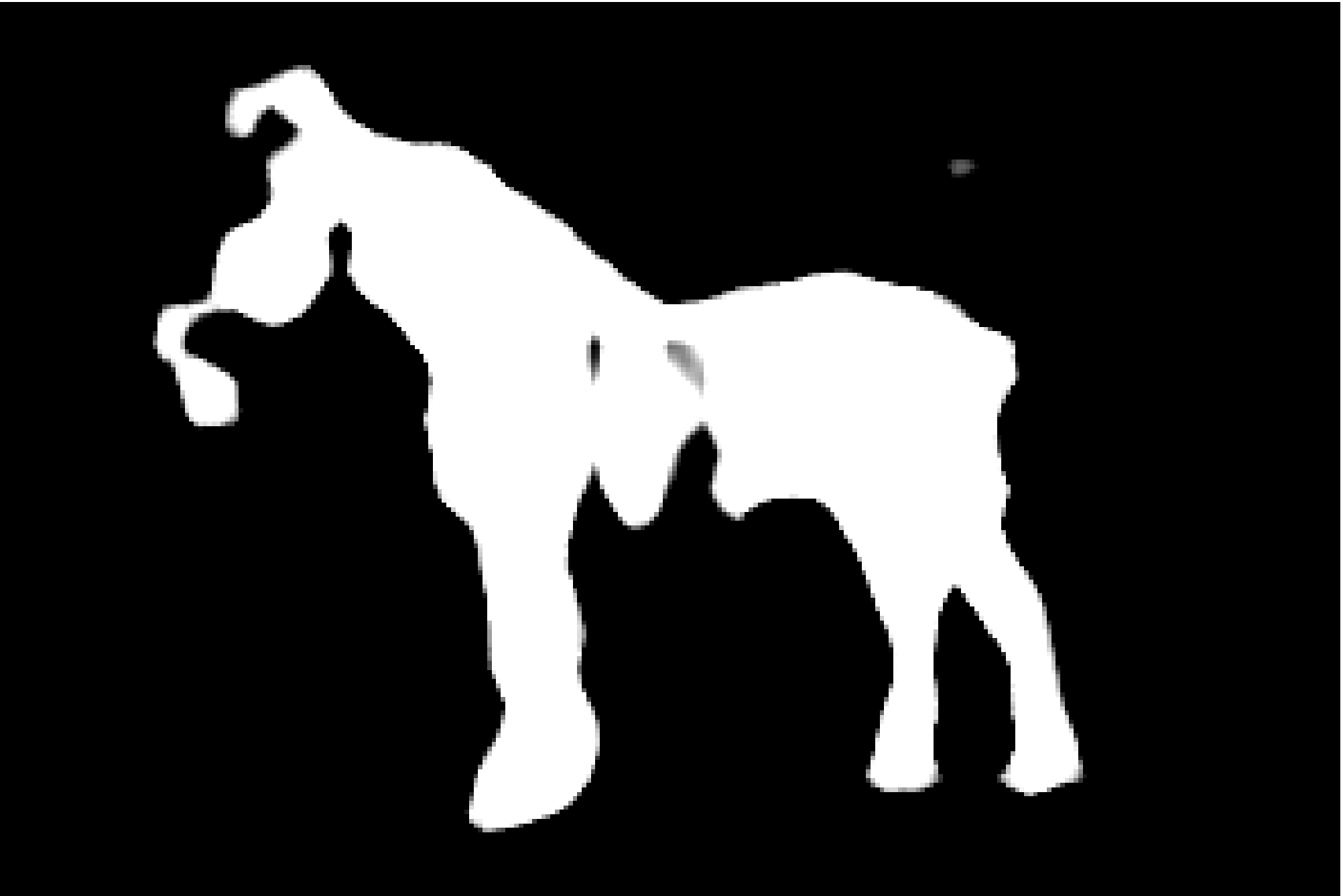} &
   \includegraphics[width=0.12\linewidth]{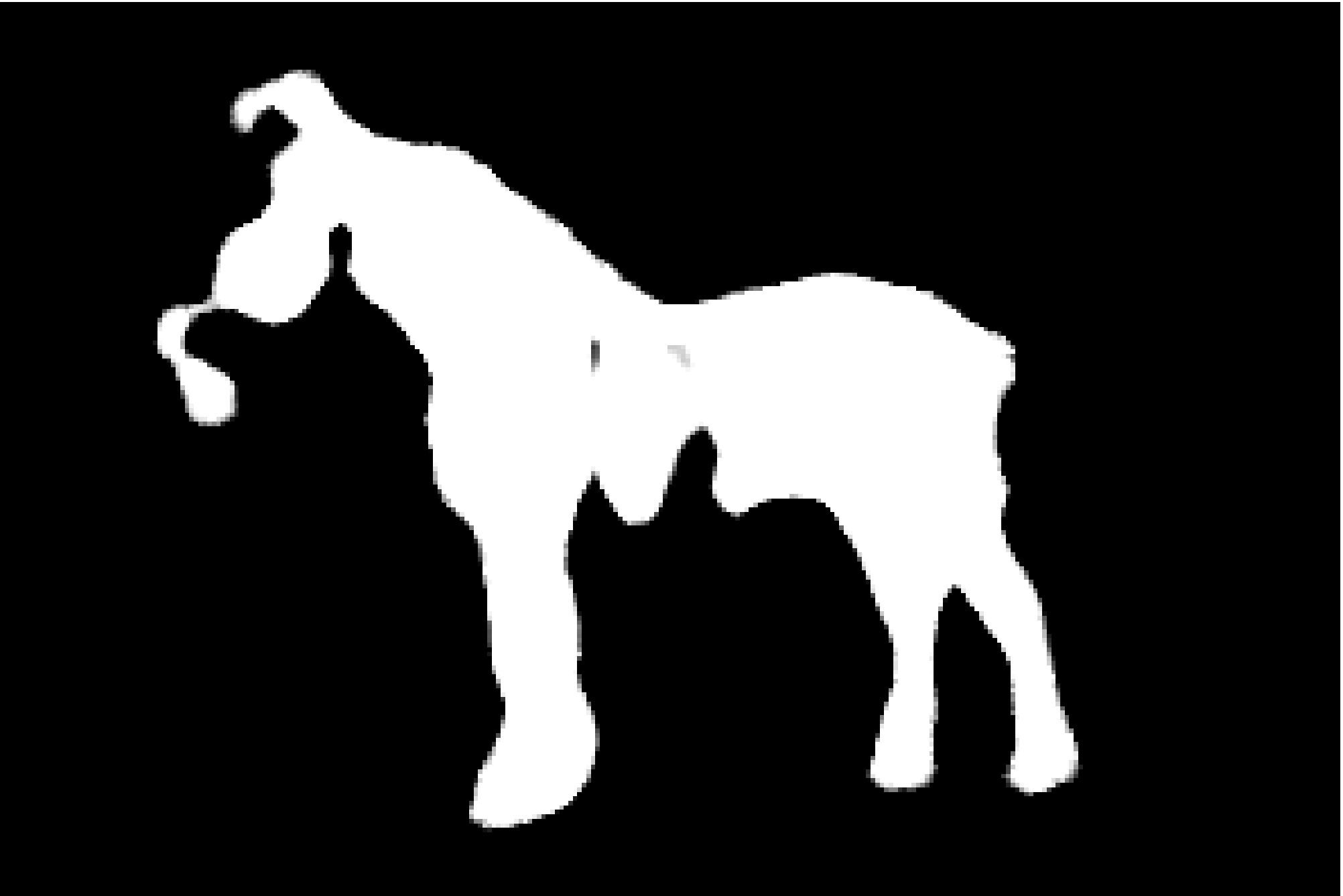} &
   \includegraphics[width=0.12\linewidth]{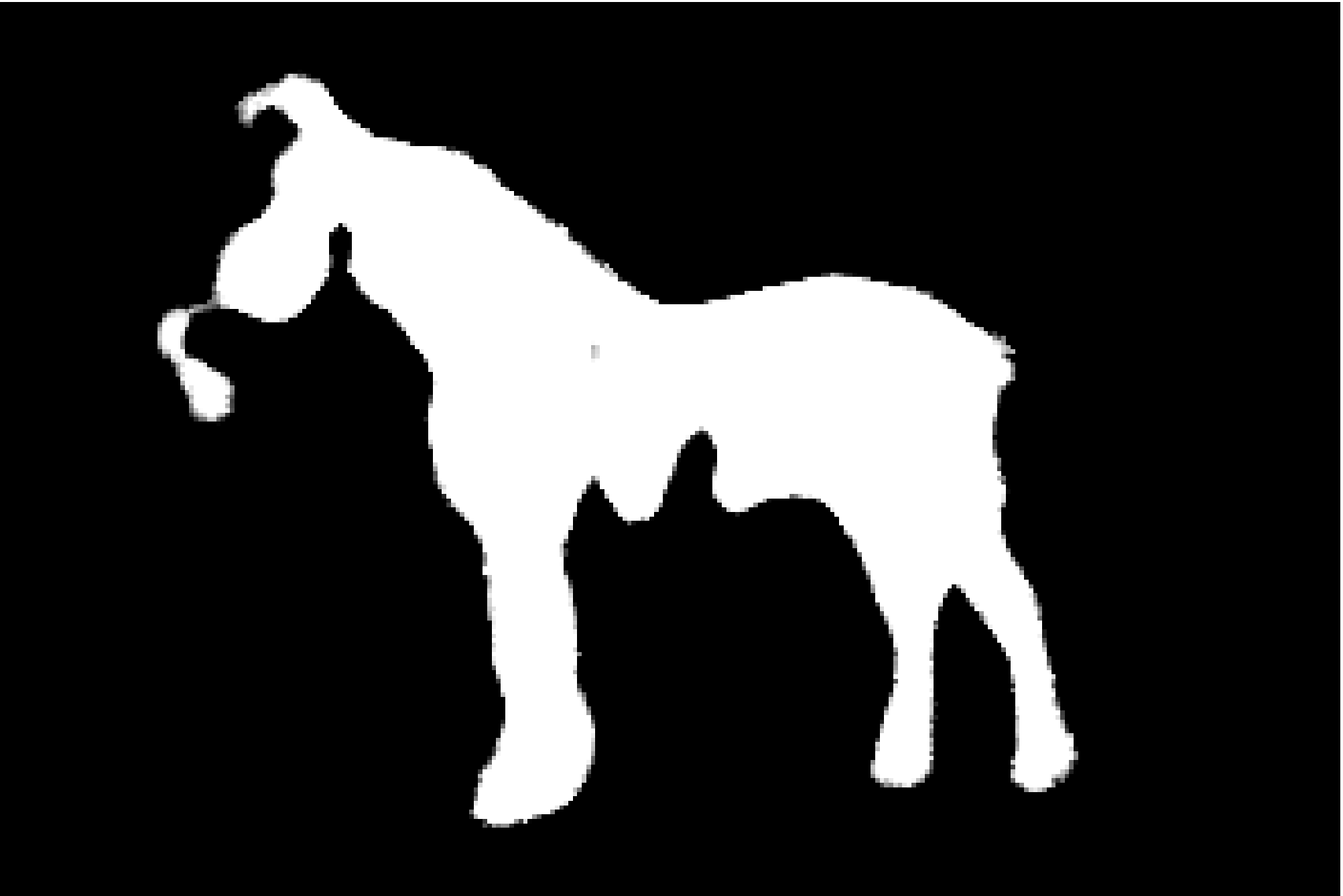} &
   \includegraphics[width=0.12\linewidth]{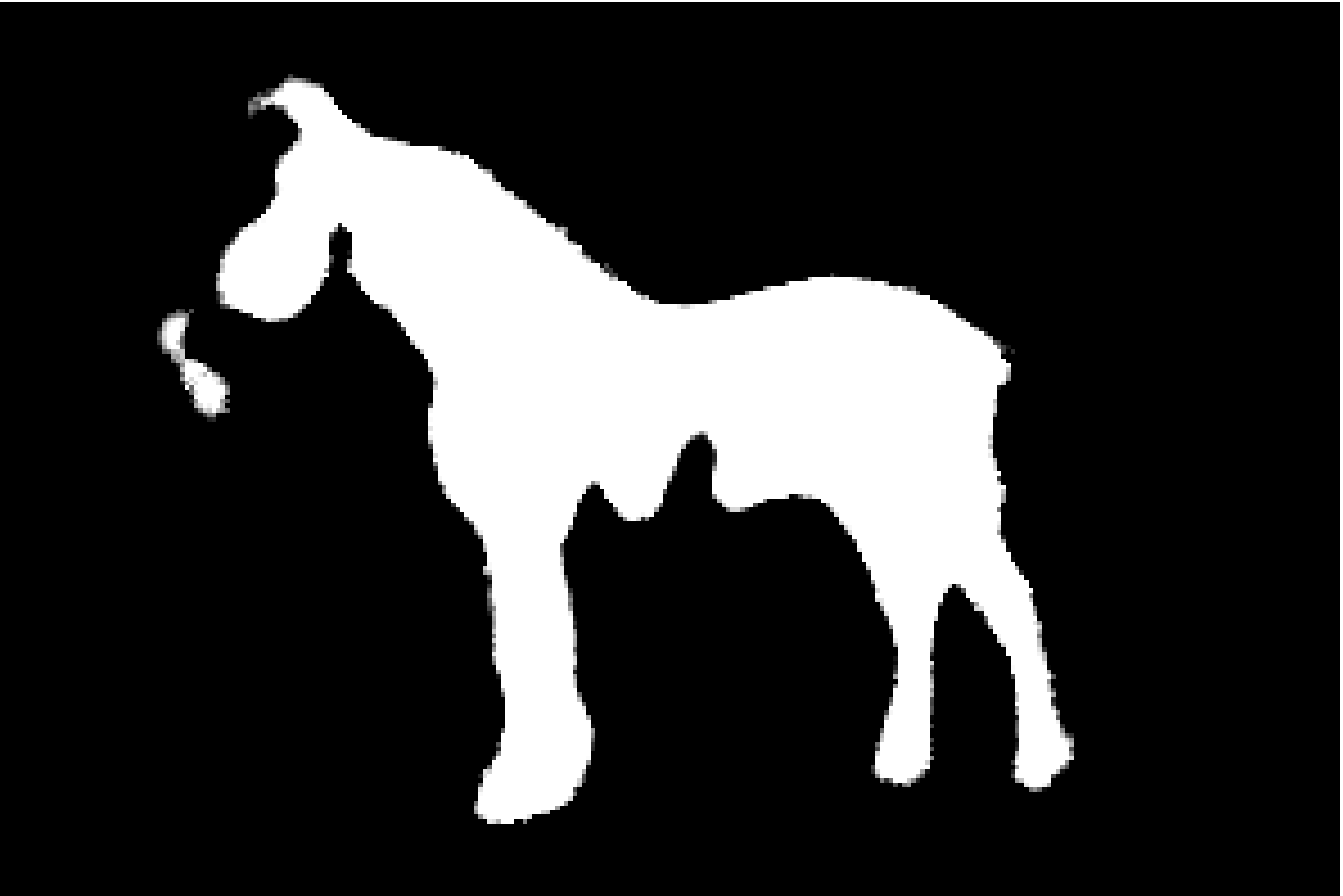} &
   \includegraphics[width=0.12\linewidth]{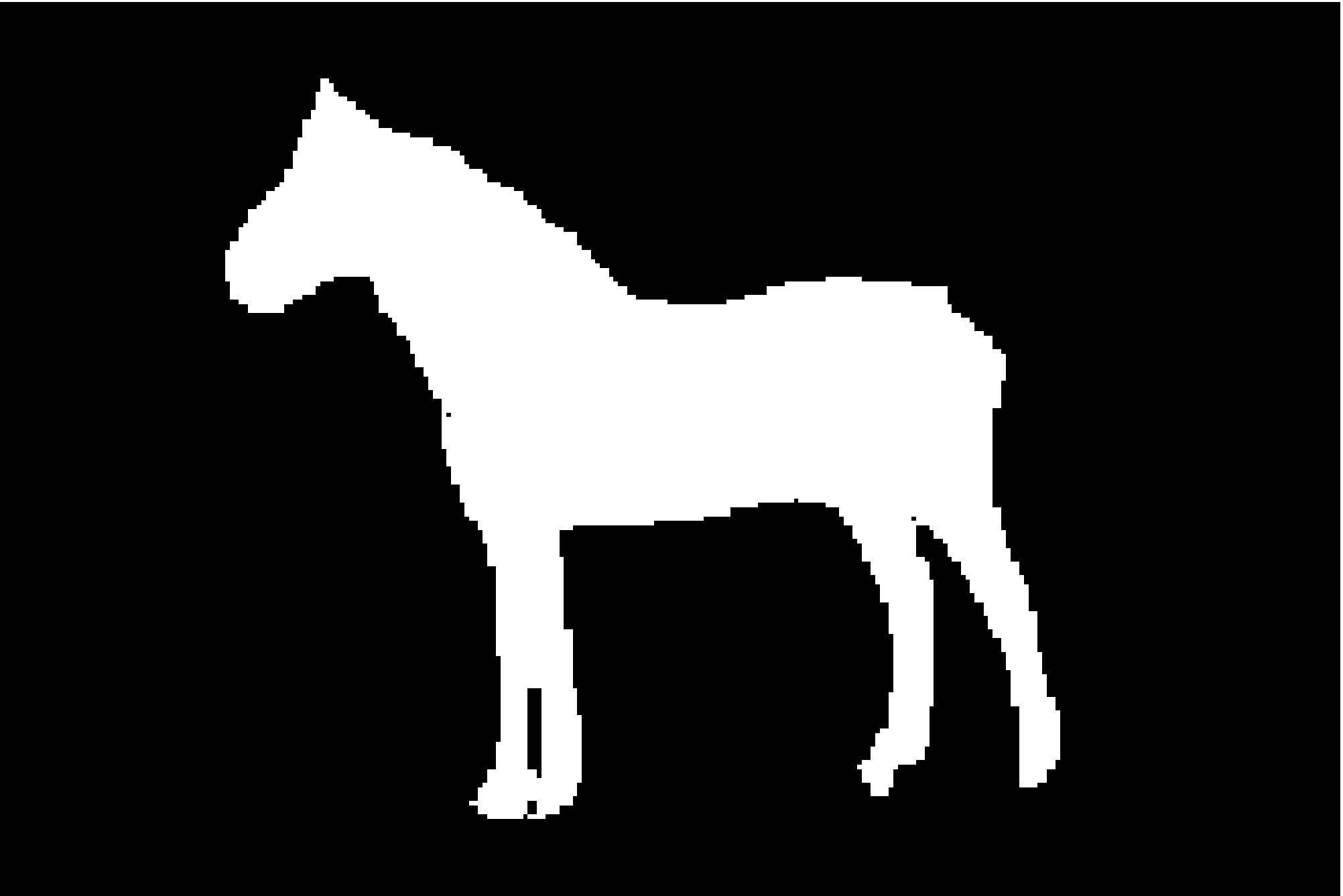} \\

   \includegraphics[width=0.12\linewidth]{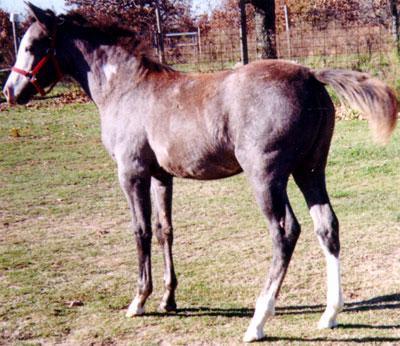} &
   \includegraphics[width=0.12\linewidth]{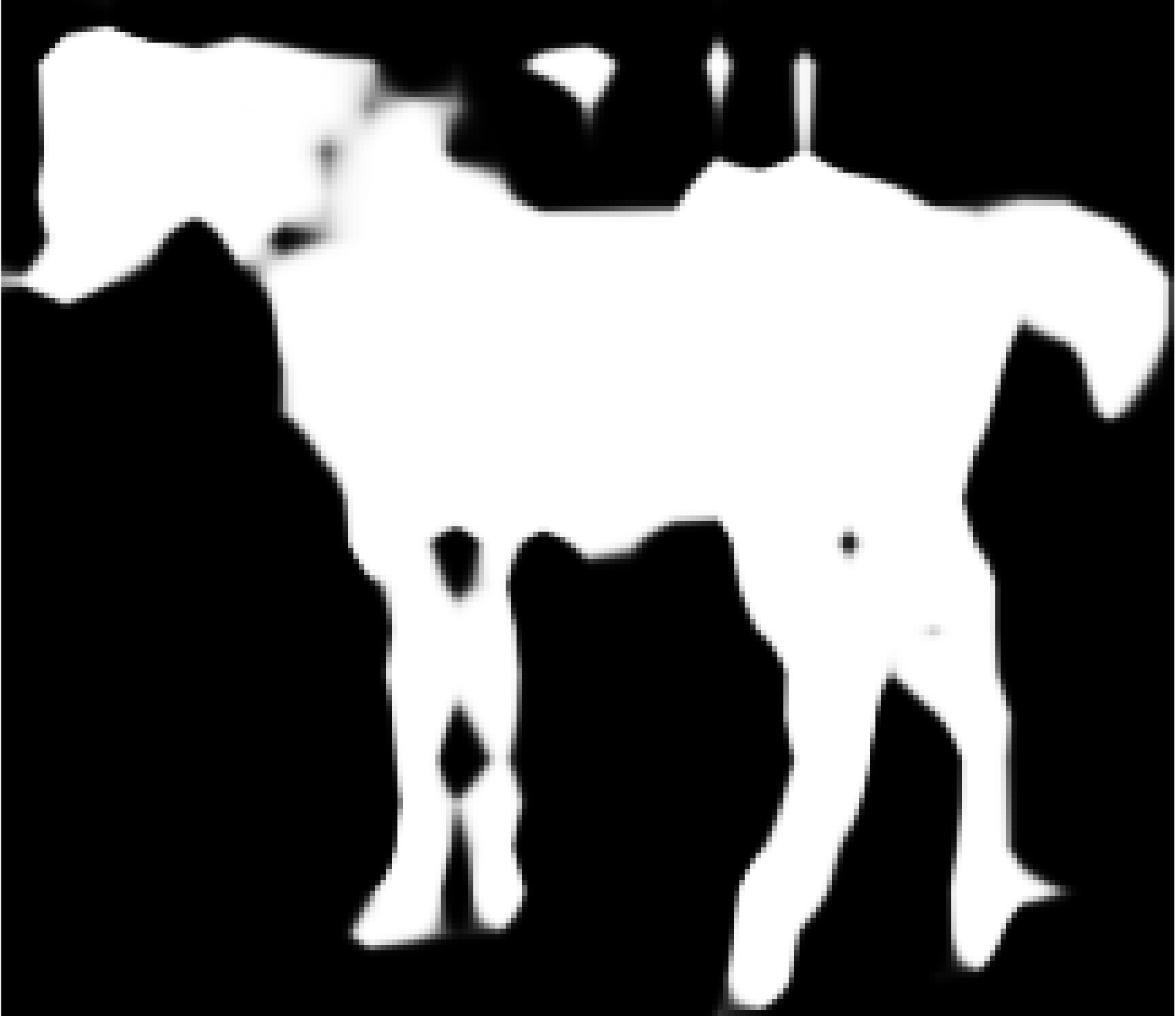} &
   \includegraphics[width=0.12\linewidth]{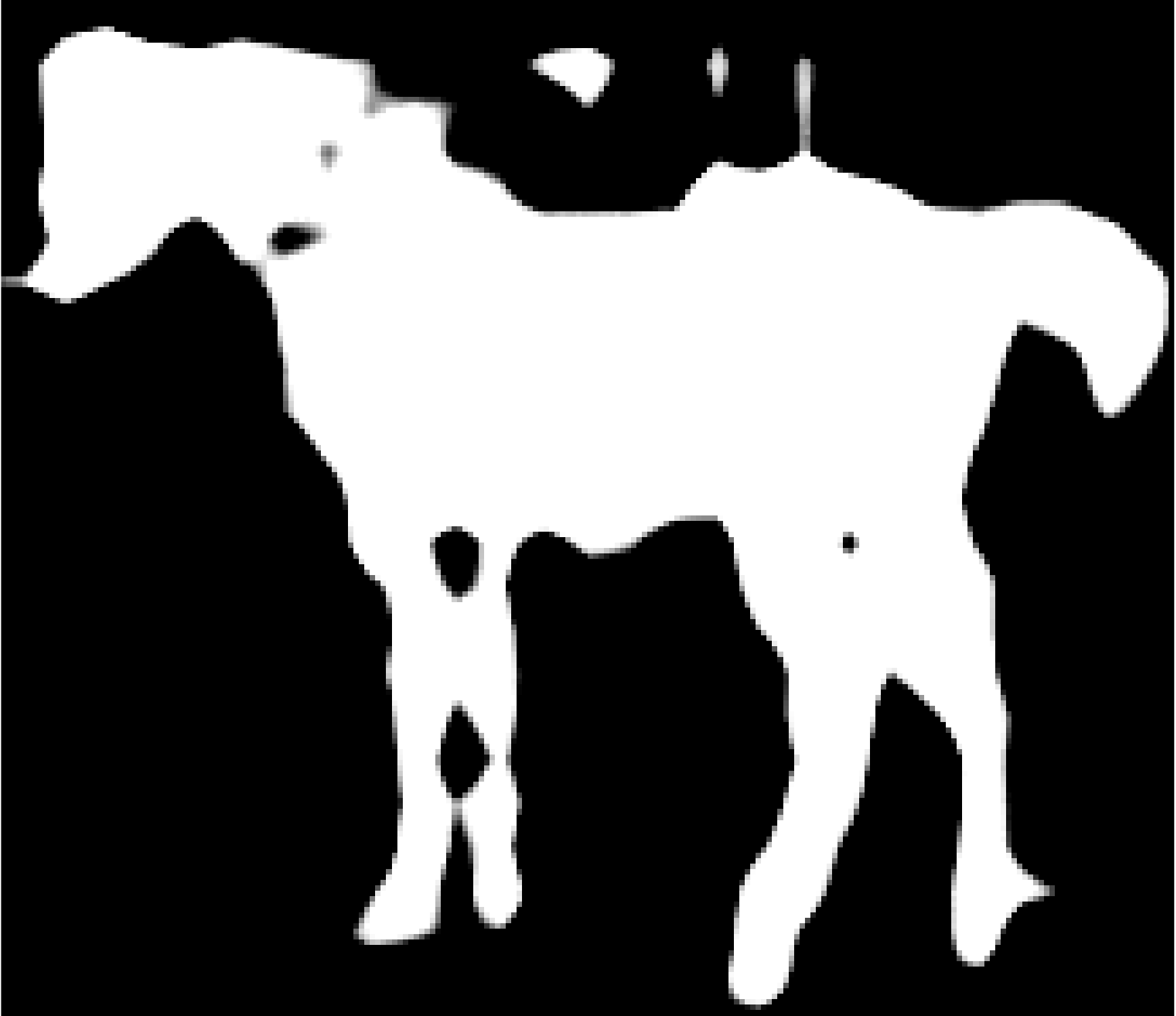} &
   \includegraphics[width=0.12\linewidth]{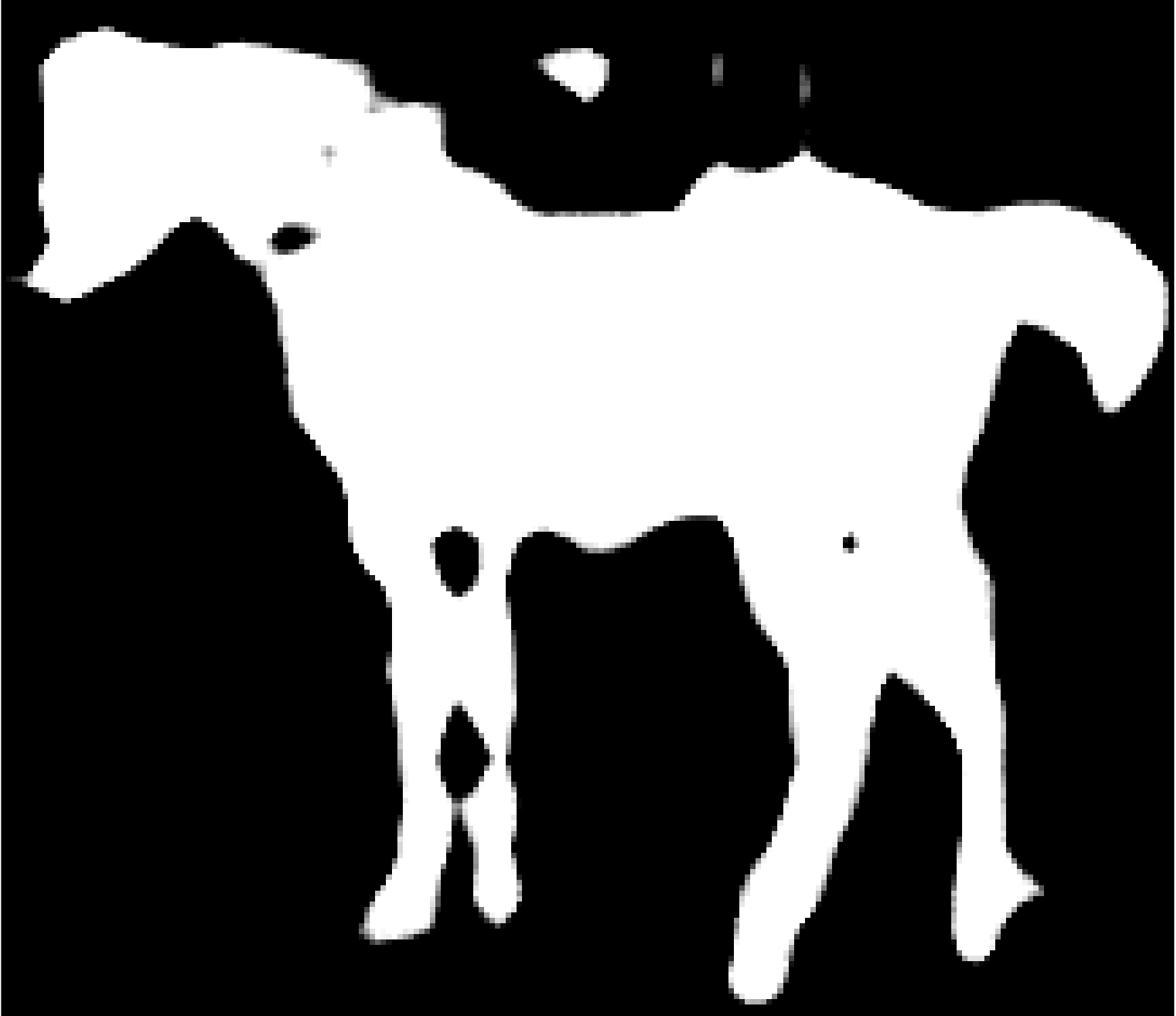} &
   \includegraphics[width=0.12\linewidth]{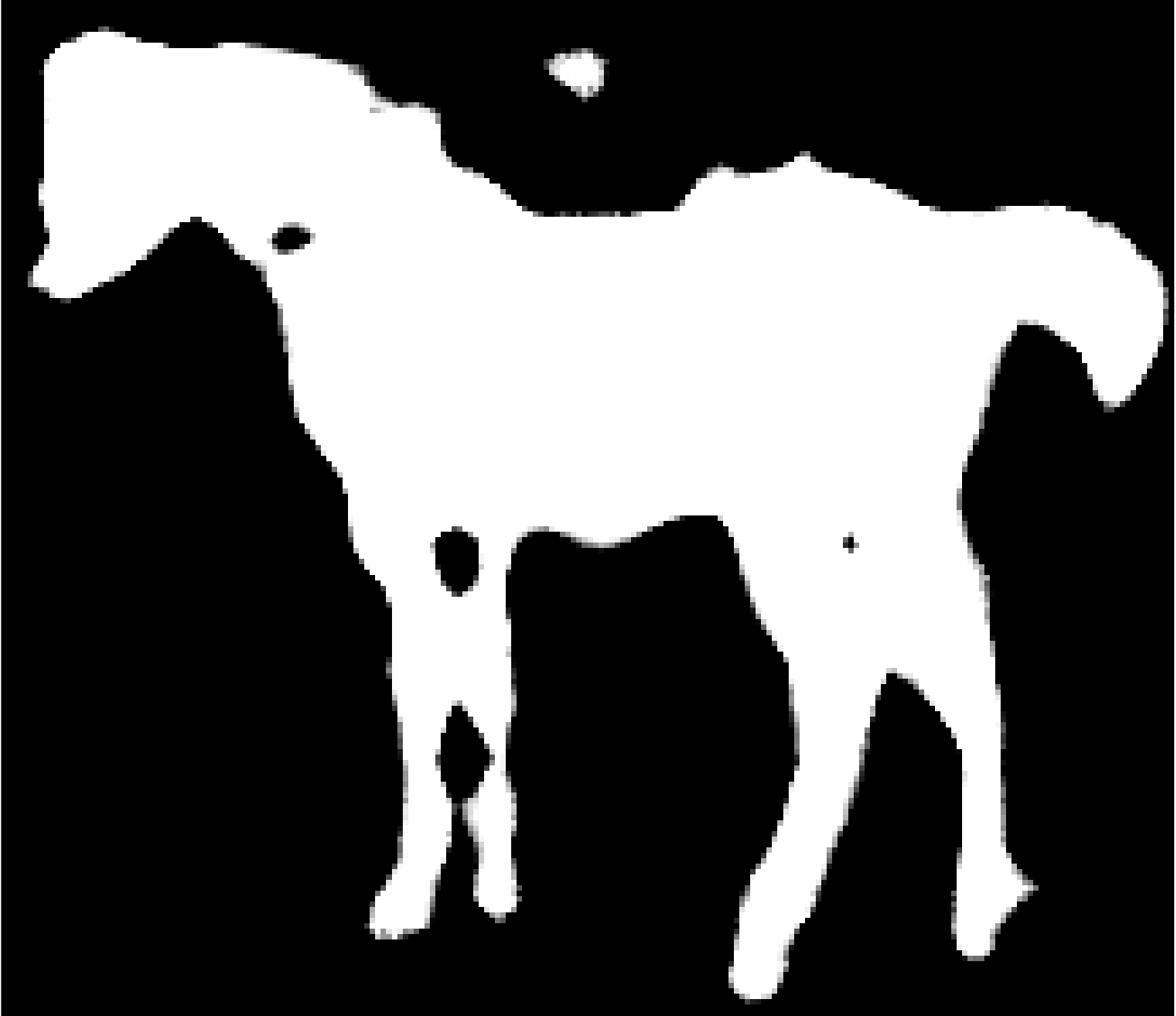} &
   \includegraphics[width=0.12\linewidth]{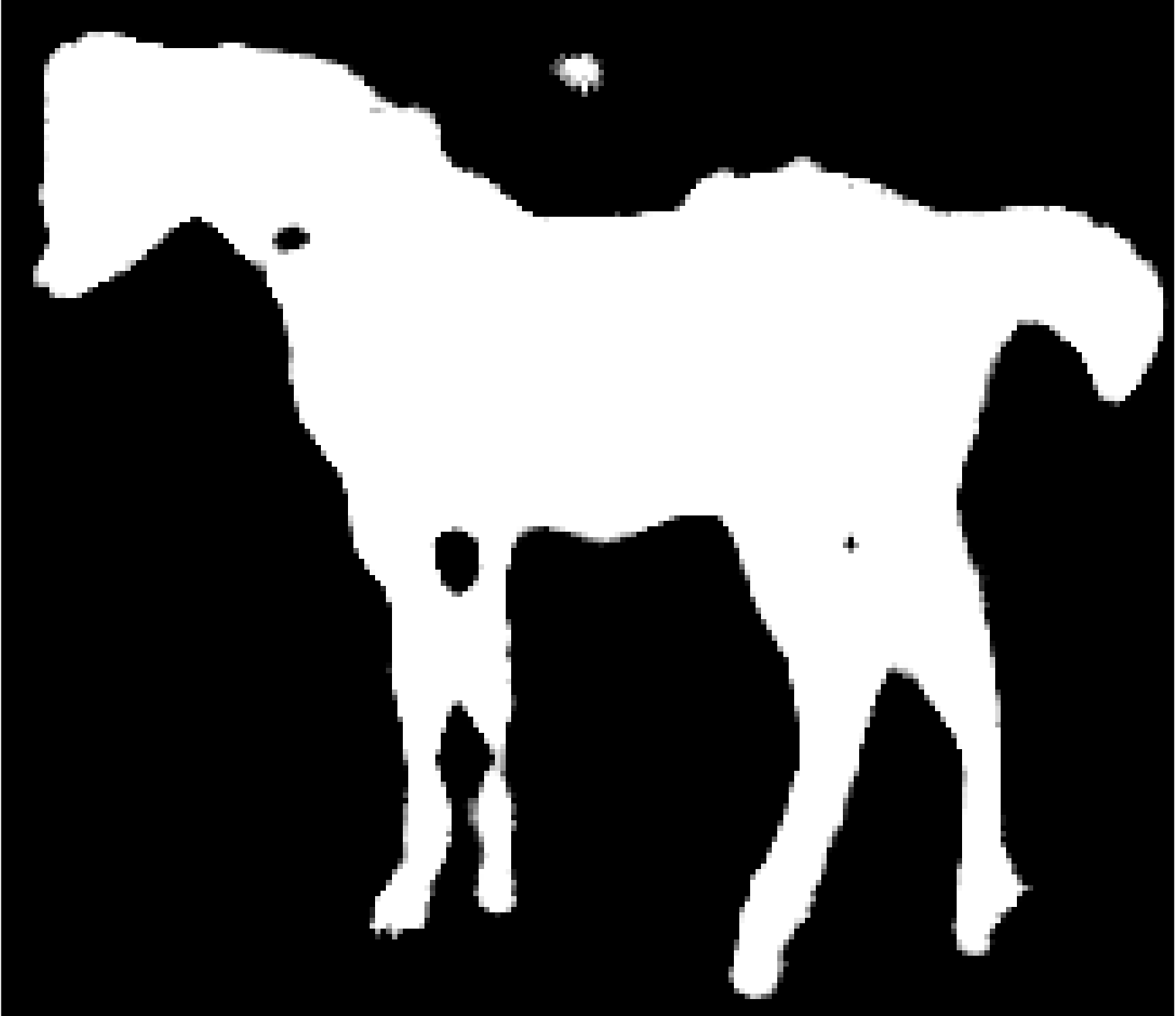} &
   \includegraphics[width=0.12\linewidth]{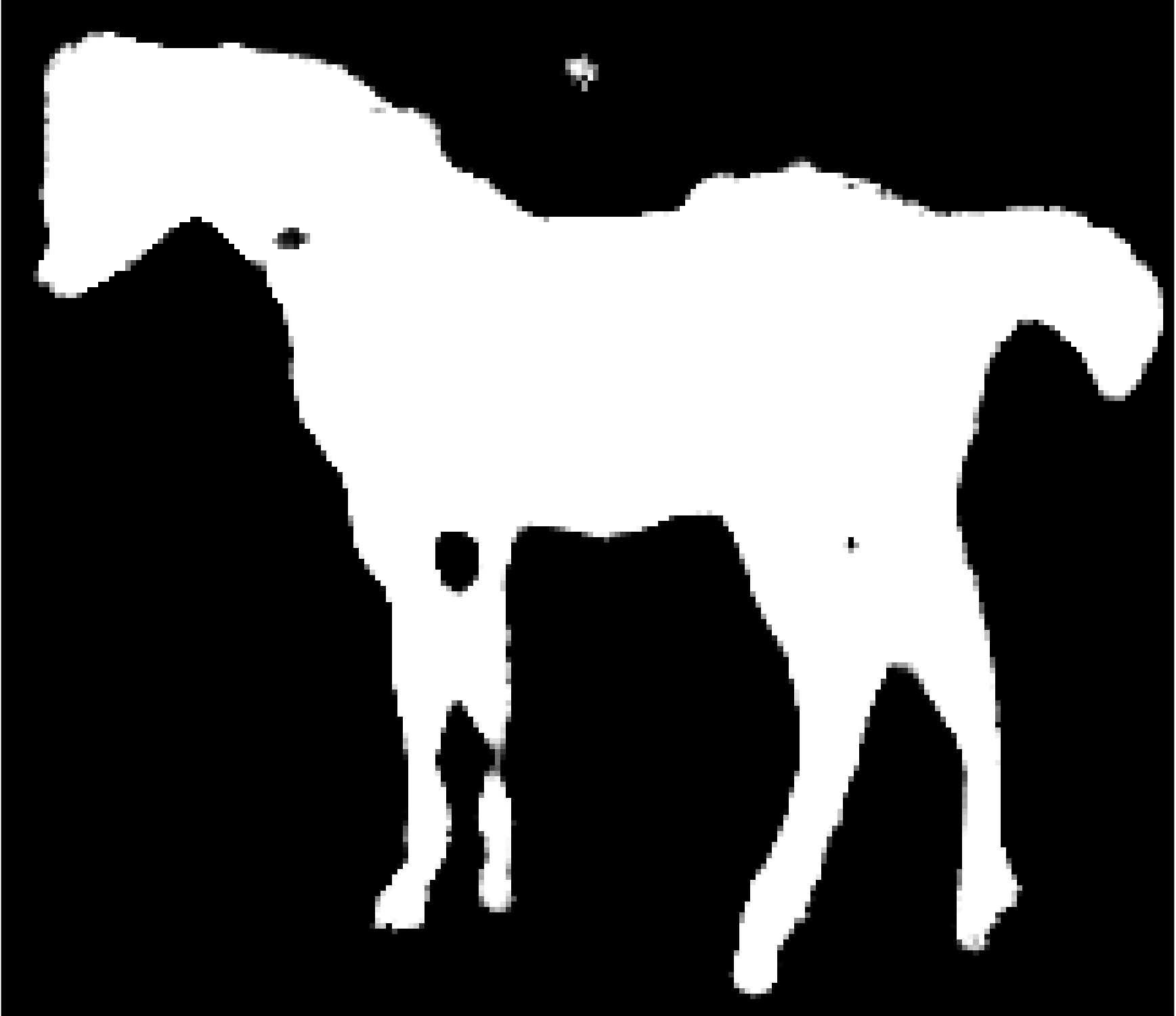} &
   \includegraphics[width=0.12\linewidth]{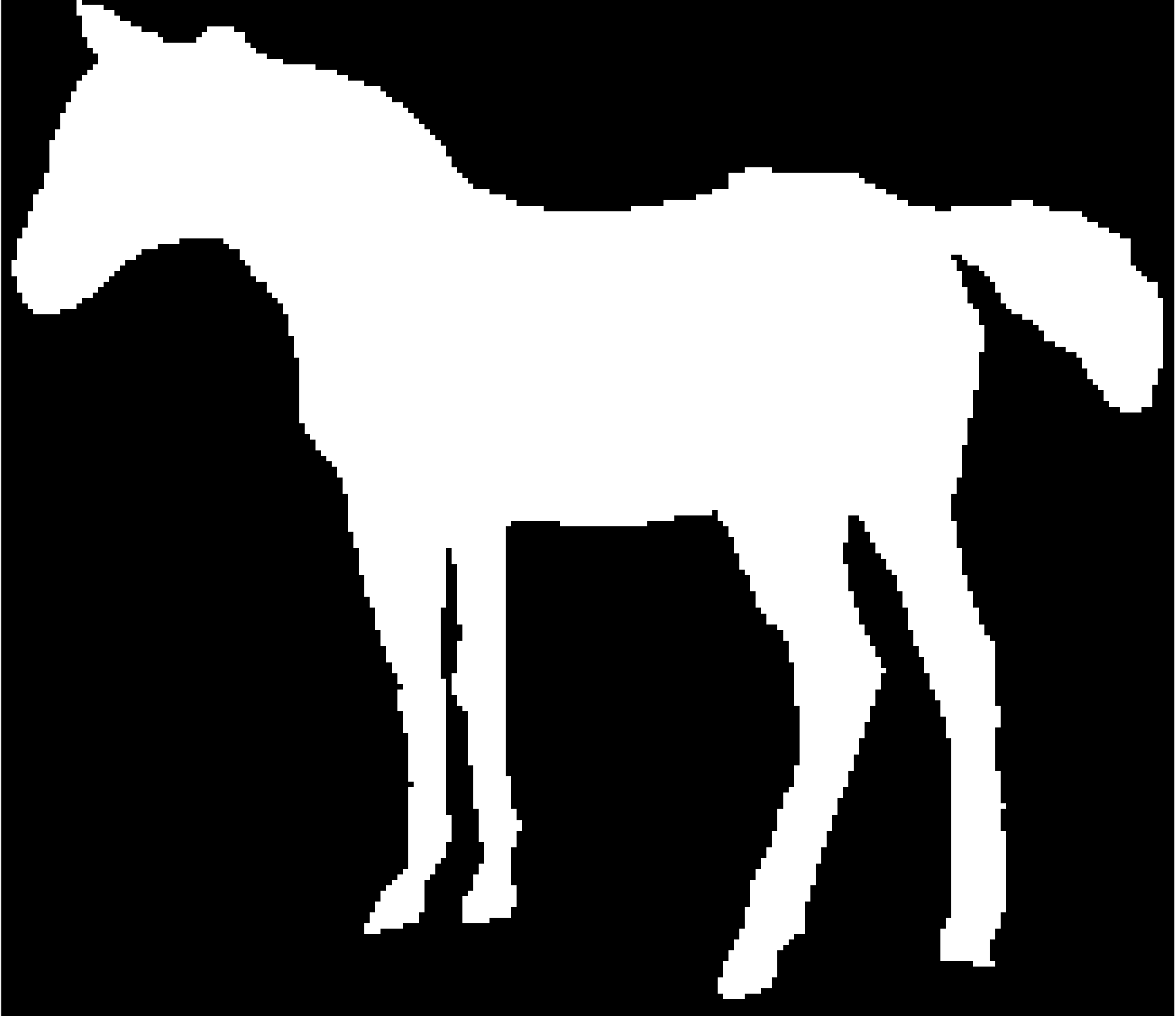} \\
   \end{tabular}
   \setlength\tabcolsep{6pt} 

\end{center}
   \vspace{-0.8cm}
   \caption{Visualization of intermediate states of the CRF-Grad layer for the {\sc Weizmann Horse} dataset. 
   Note how each step of the gradient descent algorithm refines the segmentation slightly, removing spurious outlier pixels classified as horse.
   $\bm{z}_{\text{horse}}$ is the CNN output.
   }
\label{fig:intmediate}
\vspace{-0.5\baselineskip}

\end{figure*}

The mean Intersection over Union (IoU) on the test set, for different configurations of our model, is shown in Table~\ref{tab:quant_res}. Our results show that allowing arbitrarily shaped filters gives better performance than keeping the filters fixed as Gaussian. This is the case for both the spatial and bilateral kernels. Also, adding bilateral filters improves the results compared to using spatial filters only. In addition the model trained with our inference method performs better than the model (with the same unary and pairwise potentials) trained with mean-field inference.

\begin{table}
\begin{center}
\begin{tabular}{|l|c " " l|c|l|c|}
\hline
\multicolumn{2}{|c " "}{{\sc Weizmann Horse}} & \multicolumn{2}{c|}{{\sc NYU V2}} & \multicolumn{2}{c|}{{\sc Cityscapes}} \\
\hline\hline
Method & IoU (\%)& Method  &IoU (\%) & Method &IoU (\%)\\
\hline
FCN-8s (only)& $80.0$ & R-CNN~\cite{girshick2014rich}          & $40.3$  & CRFasRNN\textsuperscript{*}~\cite{crfasrnn_iccv2015} & $62.5$\\
FCN-8s + spatial (G) &  $81.3$ & Joint HCRF~\cite{Wang/cvpr2014}        & $44.2$ & Deeplabv2-CRF~\cite{chen_arxiv_2016} & $70.4$\\
FCN-8s + spatial &  $82.0$ & Modular CNN \cite{jafari2017analyzing} & $54.3$  & Adelaide\_context~\cite{lin-etal-cvpr-2016} &$71.6$ \\
FCN-8s + full (G) &  $82.9$ &  CRFasRNN~\cite{crfasrnn_iccv2015}      & $54.4$  & LRR-4x~\cite{ghiasi-fowlkes-eccv-2016} & $71.8$ \\
FCN-8s + full &  $\bm{84.0}$ &  CRF-Grad (Ours)                        &$\mathbf{55.0}$   & CRF-Grad (Ours) &$\mathbf{71.9}$\\
FCN-8s + full (MF) &  83.3 & & & & \\
\hline
\multicolumn{6}{l}{\textsuperscript{*}\footnotesize{Note that CRFasRNN uses a different CNN model than ours on the {\sc Cityscapes} dataset.}}
\end{tabular}
\end{center}
\caption{Quantitative results comparing our method to baselines as well as state-of-the-art methods. Mean intersection over union for the test set is shown for {\sc Weizmann Horse} and {\sc Cityscapes}, for {\sc NYU V2} there is no test set and validation score is presented.
In the entry denoted ``full'' the complete CRF-Grad layer was used, while in ``spatial'' no bilateral kernel was used. A (G) means that the filters were restricted to a Gaussian shape and (MF) means that the inference method was switched to mean field. For the entries denoted CRF-Grad the full model was used.}
\label{tab:quant_res}
\end{table}

\begin{figure}[t]
\begin{center}
   \includegraphics[width=0.3\linewidth]{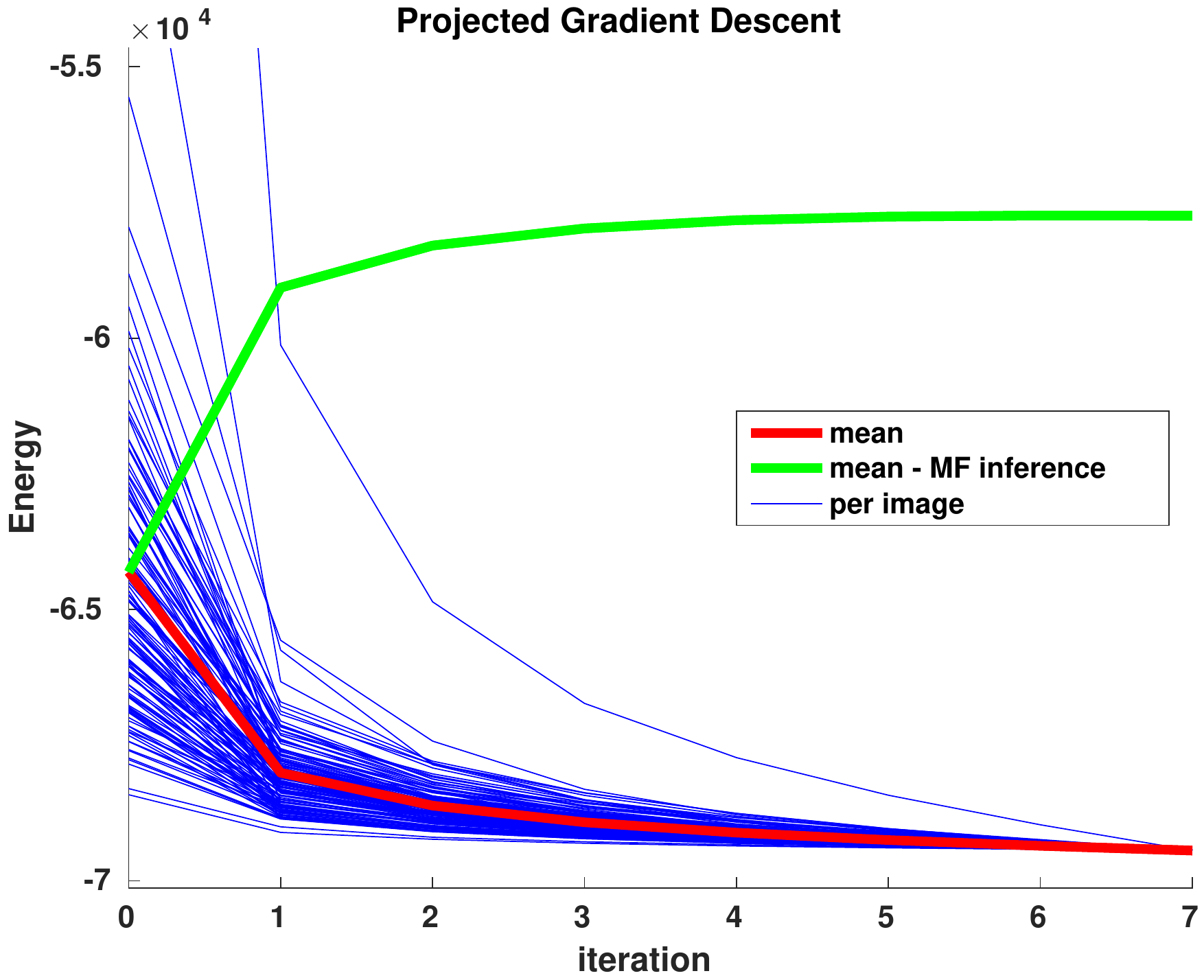}
   \includegraphics[width=0.3\linewidth]{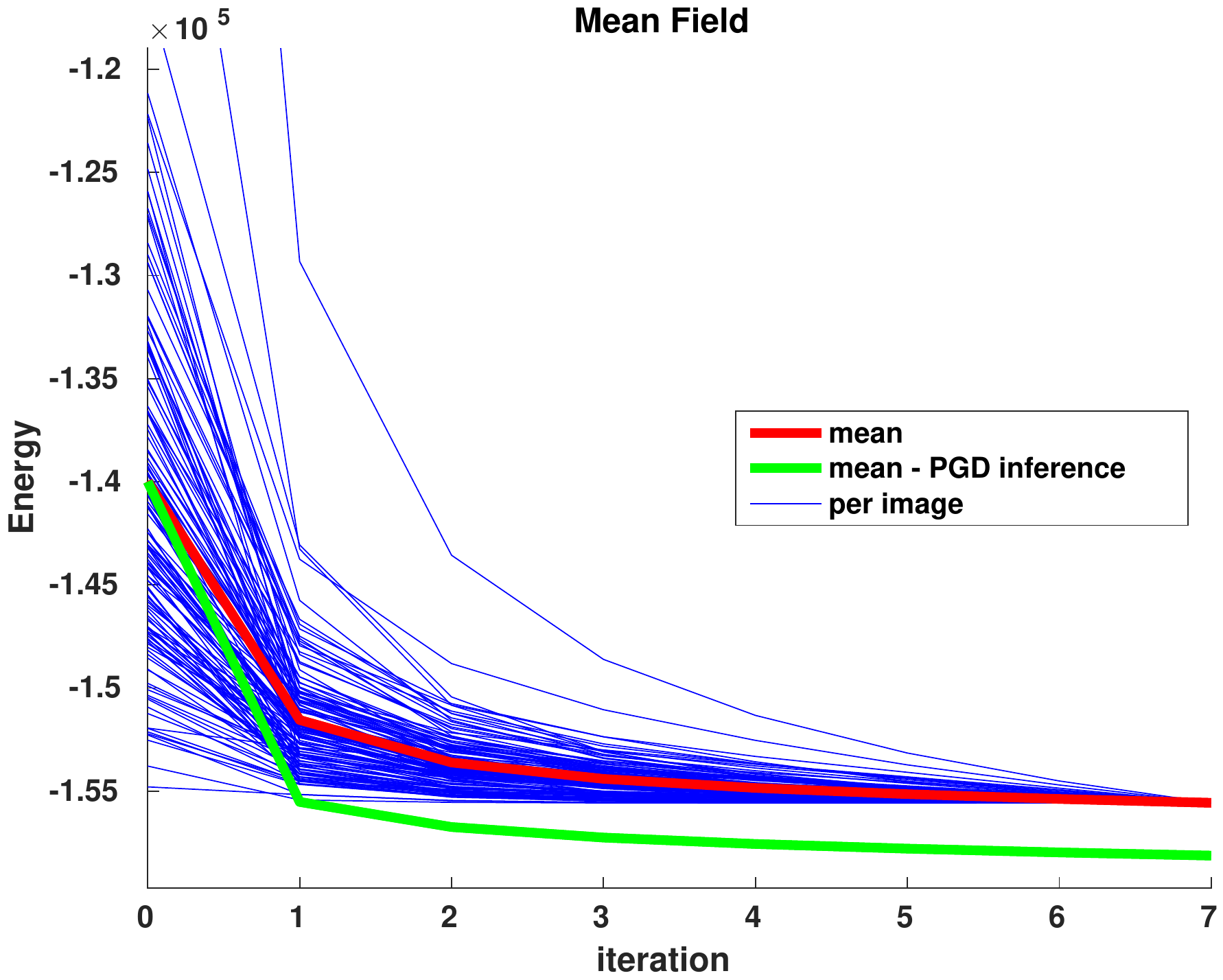}
   \includegraphics[width=0.3\linewidth]{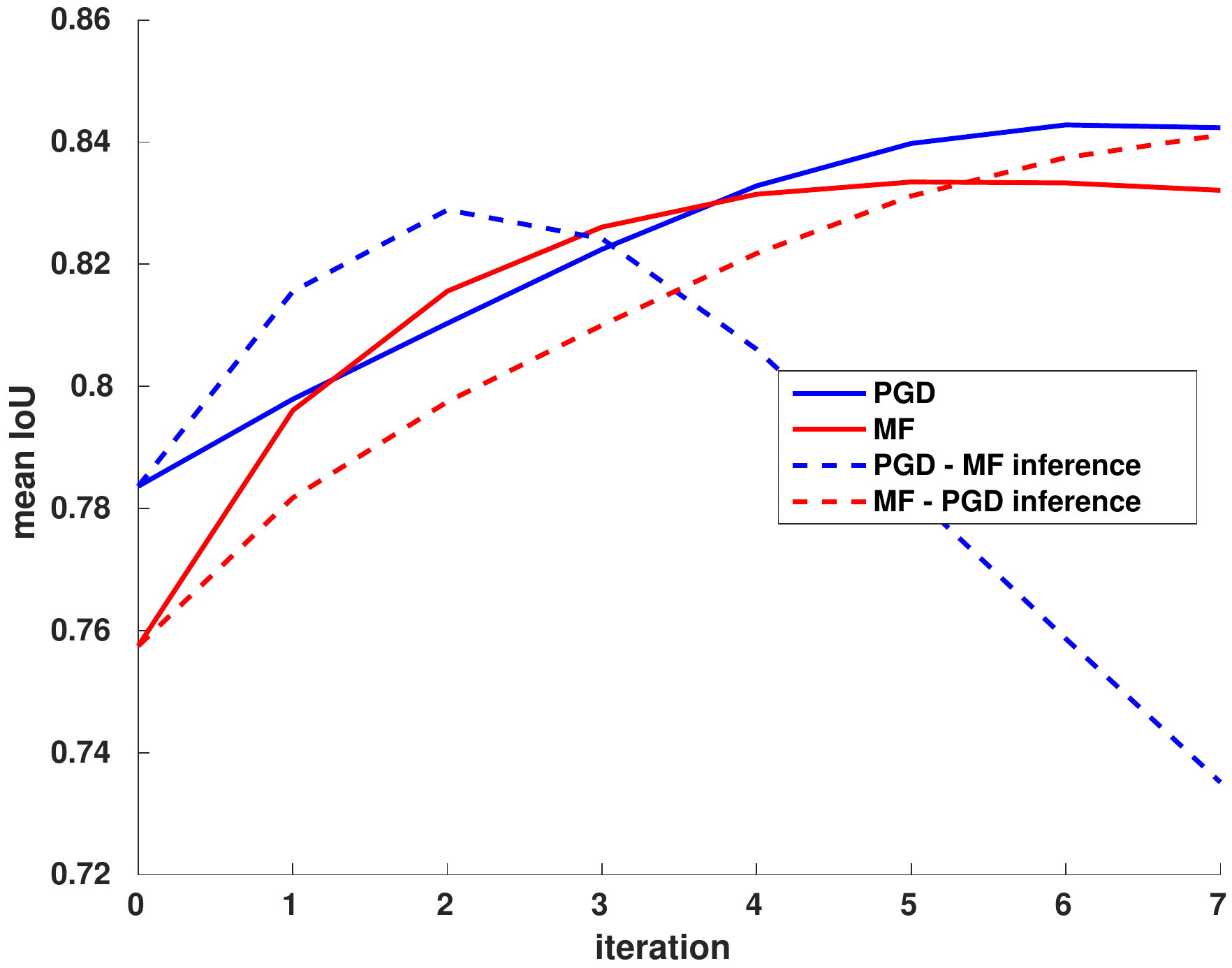}
\end{center}
\vspace{-0.6cm}
   \caption{Left and Middle: CRF energy, as defined in \eqref{realvalprog}, as a function of inference iterations. The thin blue lines show the different instances, the thicker red line show the mean while the green line shows the mean when the inference method has been switched. The model has been trained with the inference method shown in respective title. For these calculations the leak factor was set to zero, meaning that the solutions satisfy the constraints of \eqref{realvalprog}. Note that, for presentation purposes, all energies have been normalized to have the same final energy. Right: Mean IoU as a function of iterations for the two different models and inference methods. The number of iterations were set to five during training. All results are from the test set of the {\sc Weizmann Horse} data set.}
\label{fig:weiz:convergence}
\end{figure}

In Fig.~\ref{fig:intmediate} the intermediate states of the layer ($\mathbf{q}^t$ for each gradient descent step) are shown. Fig.~\ref{fig:weiz:convergence} presents a convergence analysis for the full version of our layer as well as a comparison to mean-field inference. The results show that the CRF energy converges in only a few iteration steps and that increasing the  number of iterations barely affects performance after $T=6$ iterations. It also shows that our inference method achieves lower Gibbs energy as well as higher IoU results compared to mean-field, even for models trained with mean-field inference. Looking at using mean field inference for the model trained with projected gradient descent we see that the energy increases for each iteration. The reason for this is that the mean-field method doesn't actually minimize the energy but the Kullback-Liebler divergence between the Gibbs distribution and the solution. This is, as shown in the supplementary materials, equivalent to minimizing the energy minus the entropy. The extra entropy term favors ``uncertain" solutions, i.e. solution with values not close to zero or one, which increases the energy for some models. 

\subsection{NYU V2}
The {\sc NYU V2} dataset contains images taken by Microsoft Kinect V-1 camera in 464 indoor scenes. We use the official training and validation splits consisting of 795 and 654 images, respectively. Following the setting described in Wang \etal~\cite{Wang/cvpr2014}, we also include additional images for training. These are the images from the {\sc NYU V1} dataset that do not overlap with the images in the official validation set. This gives a total of 894 images with semantic label annotations for training. As in \cite{Wang/cvpr2014} we consider 5 classes  conveying strong geometric properties: ground, vertical, ceiling, furniture and objects. The CNN part of our model was initialized as the fully convolutional network FCN-8s~\cite{Long/cvpr2015} pre-trained on the data. Afterwards we added our CRF-Grad layer and trained the model end-to-end. 

\begin{figure*}[t]
\begin{center}

\setlength\tabcolsep{1pt} 
\begin{tabular}{ccccc}

    Input & FCN-8s & CRF-RNN & CRF-Grad & Ground truth  \\

   \includegraphics[width=0.19\linewidth]{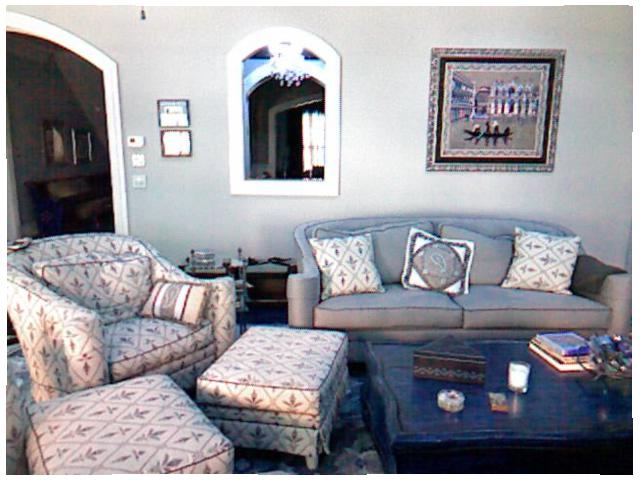} &
   \includegraphics[width=0.19\linewidth]{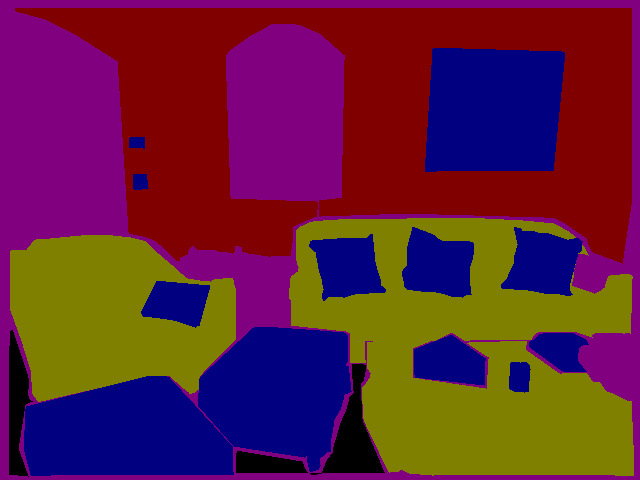} &
   \includegraphics[width=0.19\linewidth]{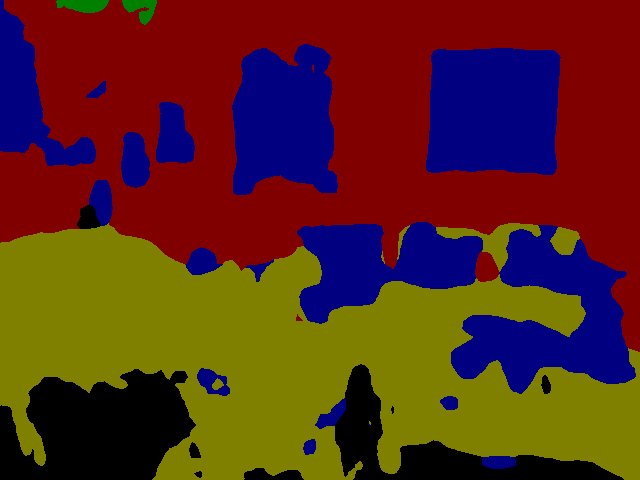} &
   \includegraphics[width=0.19\linewidth]{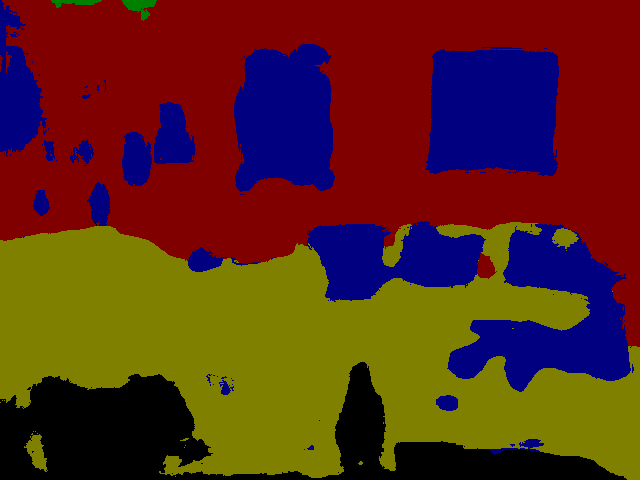} &
   \includegraphics[width=0.19\linewidth]{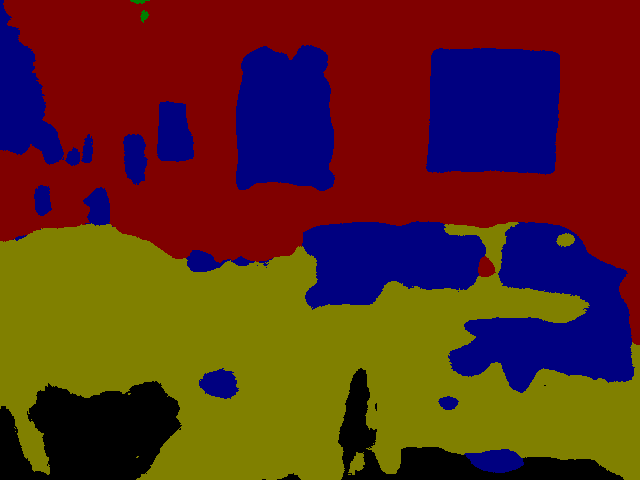} \\

   \includegraphics[width=0.18\linewidth]{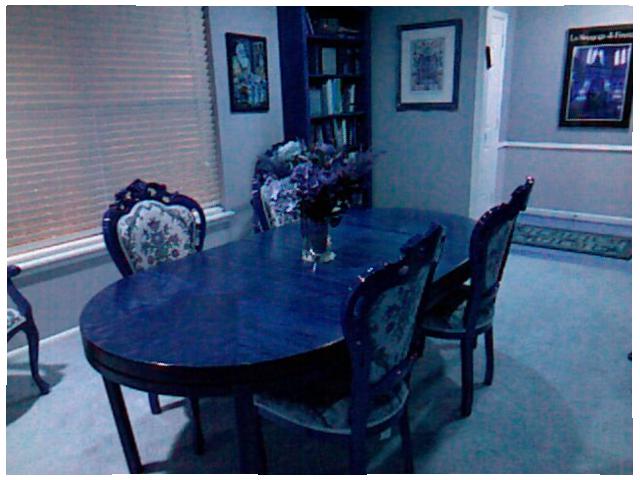} &
   \includegraphics[width=0.18\linewidth]{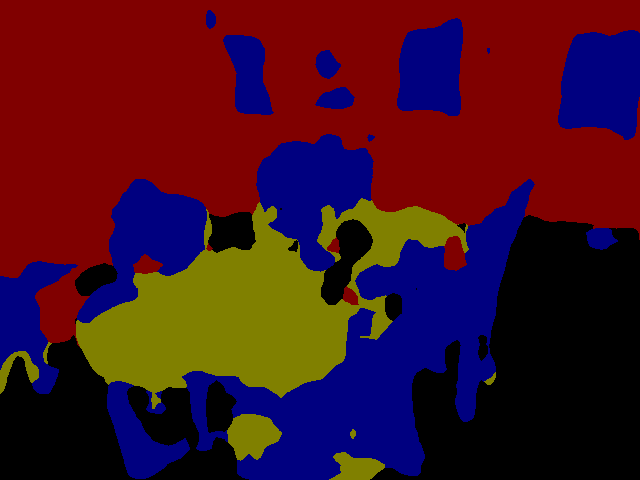} &
   \includegraphics[width=0.18\linewidth]{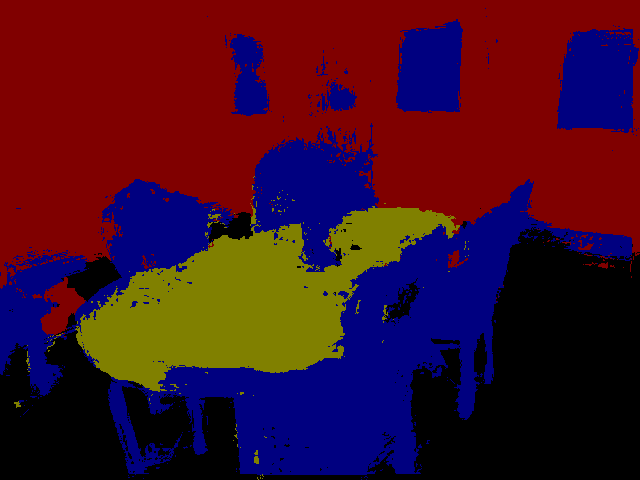} &
   \includegraphics[width=0.18\linewidth]{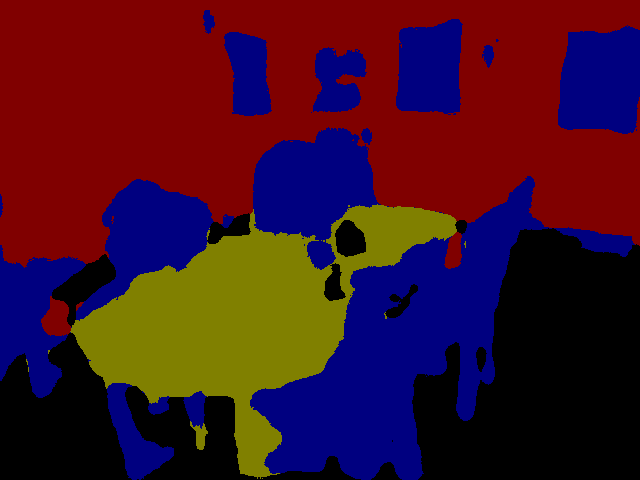} &
   \includegraphics[width=0.18\linewidth]{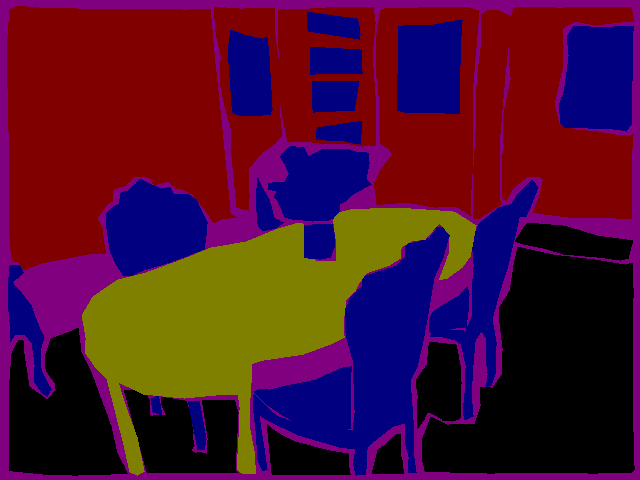} \\

      
   
   \includegraphics[width=0.19\linewidth]{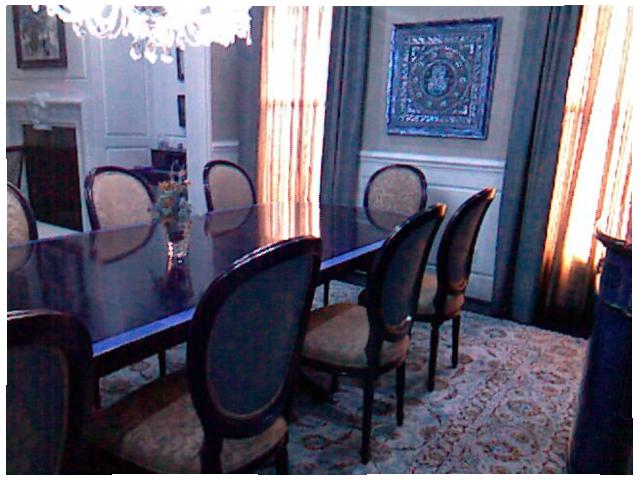}&
   \includegraphics[width=0.19\linewidth]{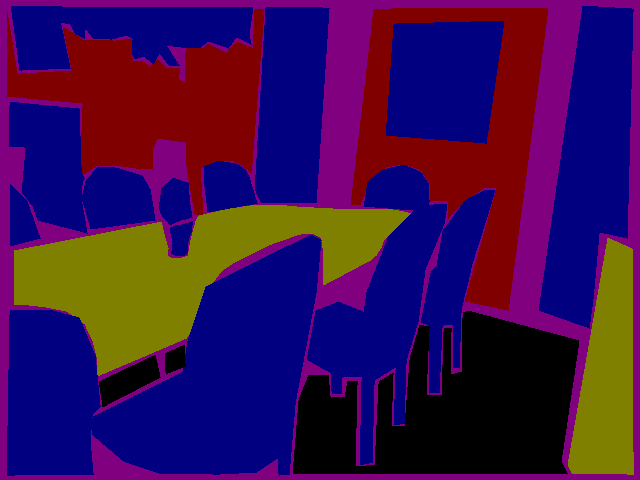}&
   \includegraphics[width=0.19\linewidth]{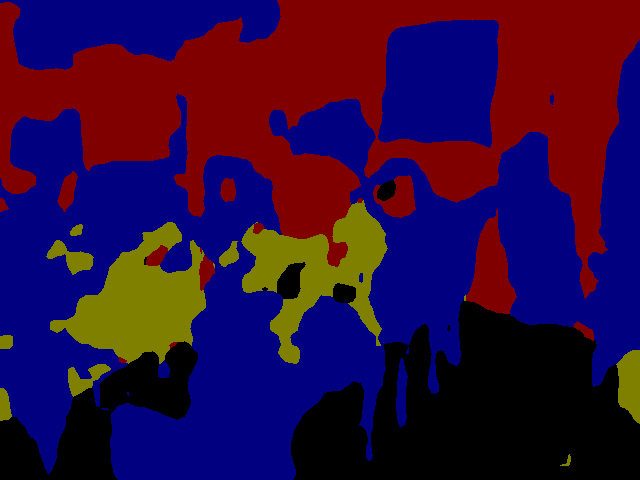}&
   \includegraphics[width=0.19\linewidth]{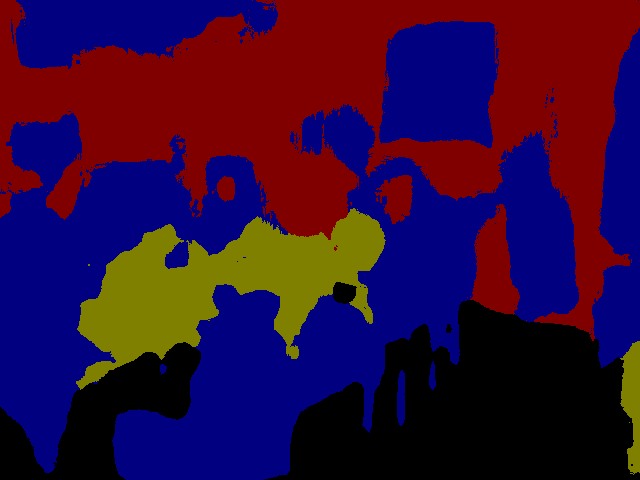}&
   \includegraphics[width=0.19\linewidth]{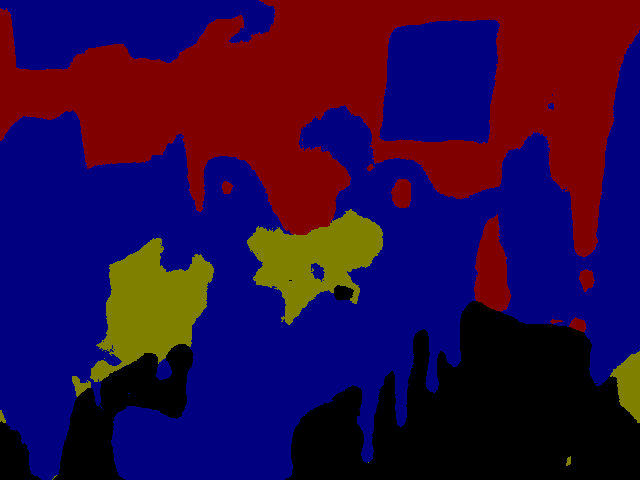}    \\ 
\end{tabular}
\setlength\tabcolsep{6pt} 
 
\end{center}
\vspace{-0.6cm}
   \caption{Qualitative results on the {\sc NYU V2} dataset. Note that the CRF-Grad captures the shape of the object instances better compared to the baselines. This effect is perhaps most pronounced for the paintings hanging on the walls.}
\label{fig:nyu_res_main}
\end{figure*}

As shown in Table~\ref{tab:quant_res}, we achieved superior results for semantic image segmentation on the {\sc NYU V2} dataset. Some example segmentations are shown in Fig.~\ref{fig:nyu_res_main}. Additional examples are included in the supplementary material.

\subsection{Cityscapes}
The {\sc Cityscapes} dataset~\cite{cityscapes} consists of a set of images of street scenes collected from 50 different cities. The images are high resolution (1024 $\times$ 2048) and are paired with pixel-level annotations of 19 classes including road, sidewalk, traffic sign, pole, building, vegetation and sky. 
The training, validation and test sets consist of 2975, 500 and 1525 images, respectively. In addition there are 20000 coarsely annotated images that can be used for training. The CNN part of our model was initialized as an LRR network~\cite{ghiasi-fowlkes-eccv-2016} pre-trained on both the fine and the coarse annotations. We then added our CRF-Grad layer and trained the model end-to-end on the finely annotated images only. 

\begin{figure*}[t]
\begin{center}

\setlength\tabcolsep{1pt} 
\begin{tabular}{cccc}

    Input & LRR & LRR + CRF-Grad & Ground truth \\

   \includegraphics[width=0.23\linewidth]{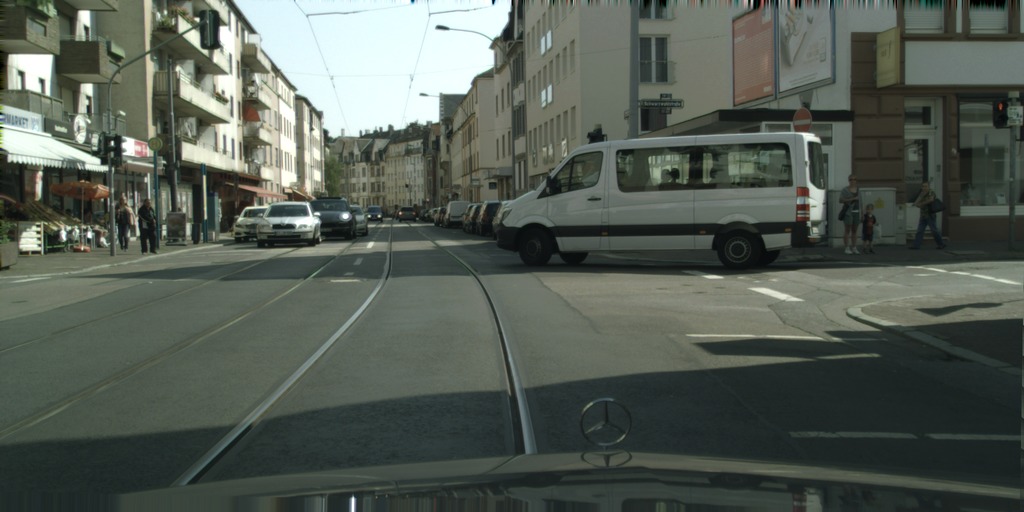} &
   \includegraphics[width=0.23\linewidth]{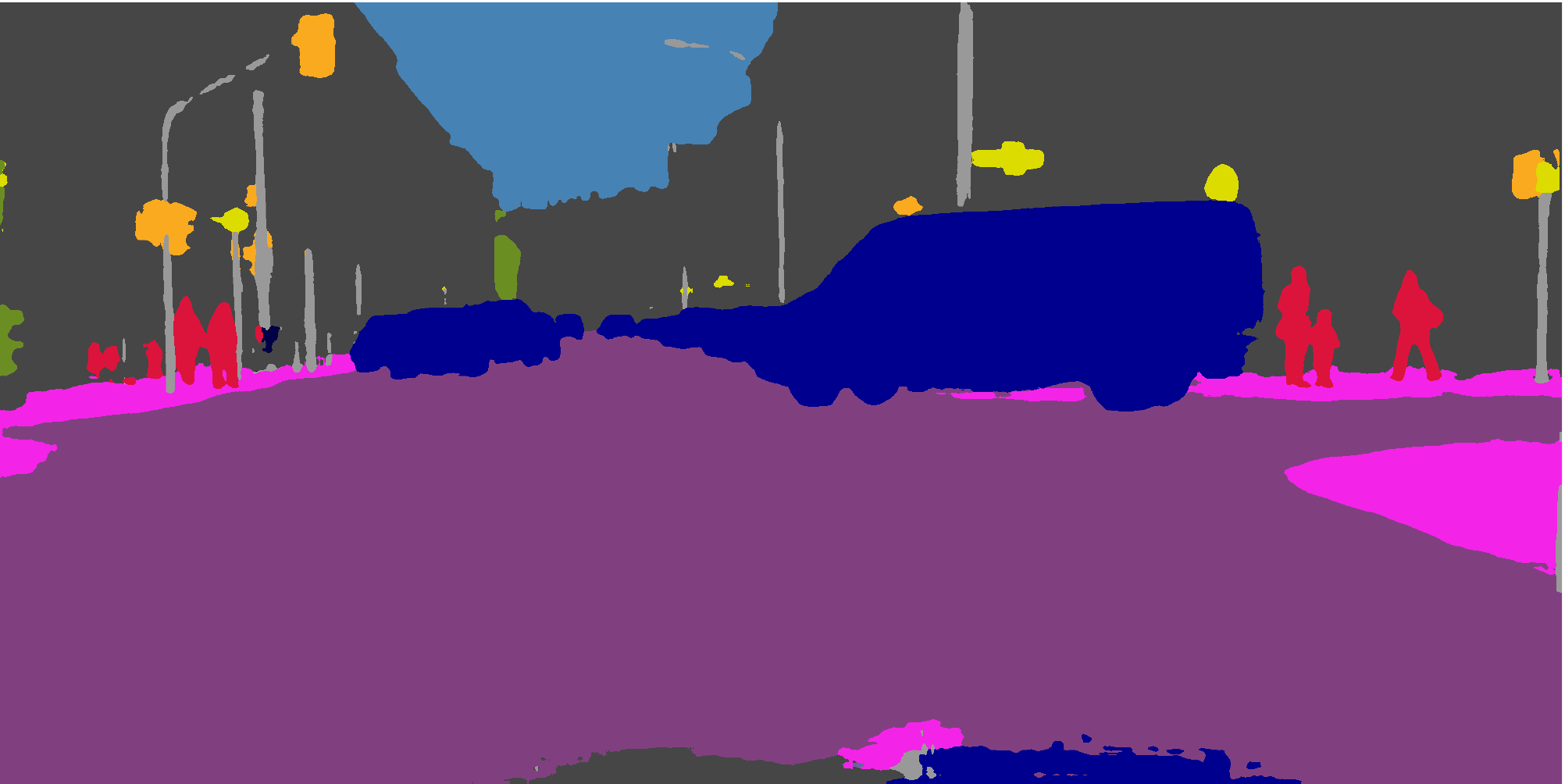} &
   \includegraphics[width=0.23\linewidth]{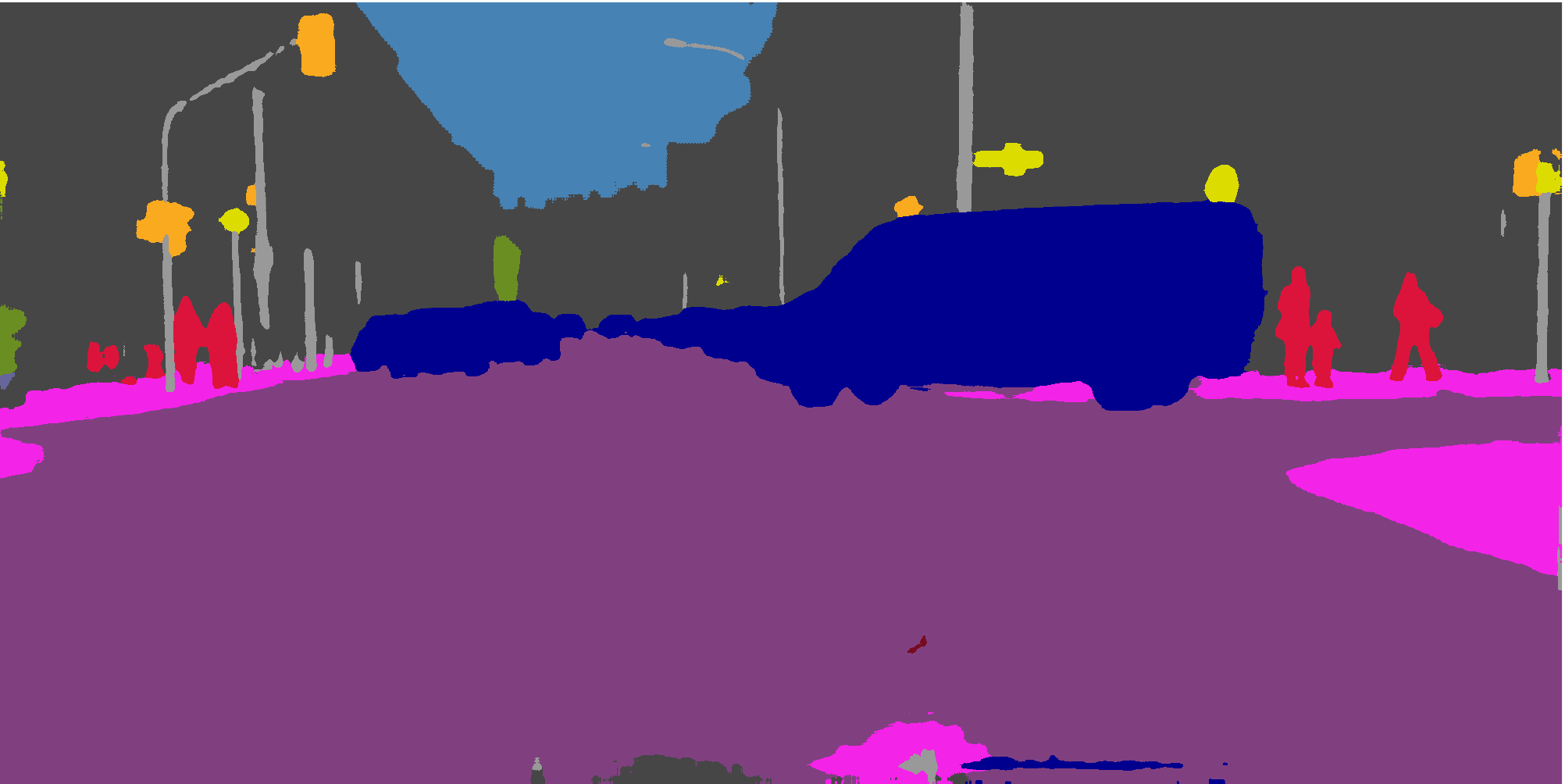} &
   \includegraphics[width=0.23\linewidth]{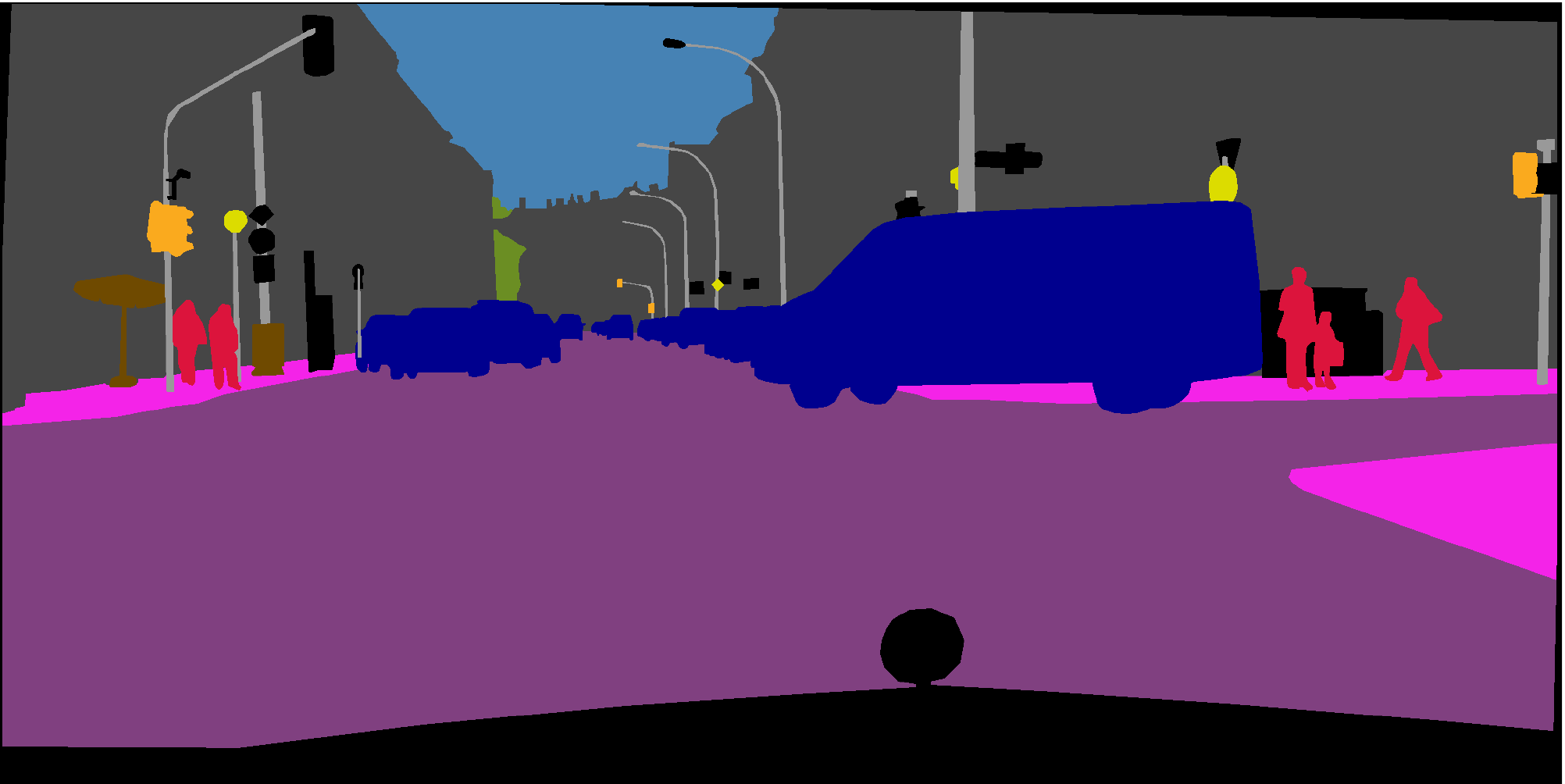} \\

   \includegraphics[width=0.23\linewidth]{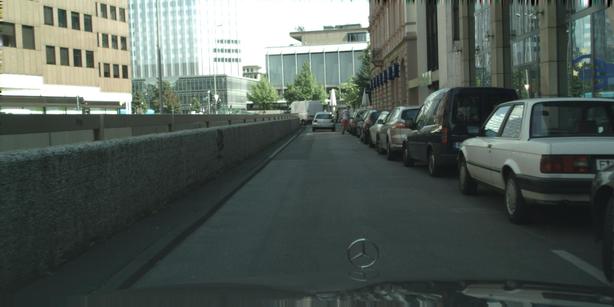} &
   \includegraphics[width=0.23\linewidth]{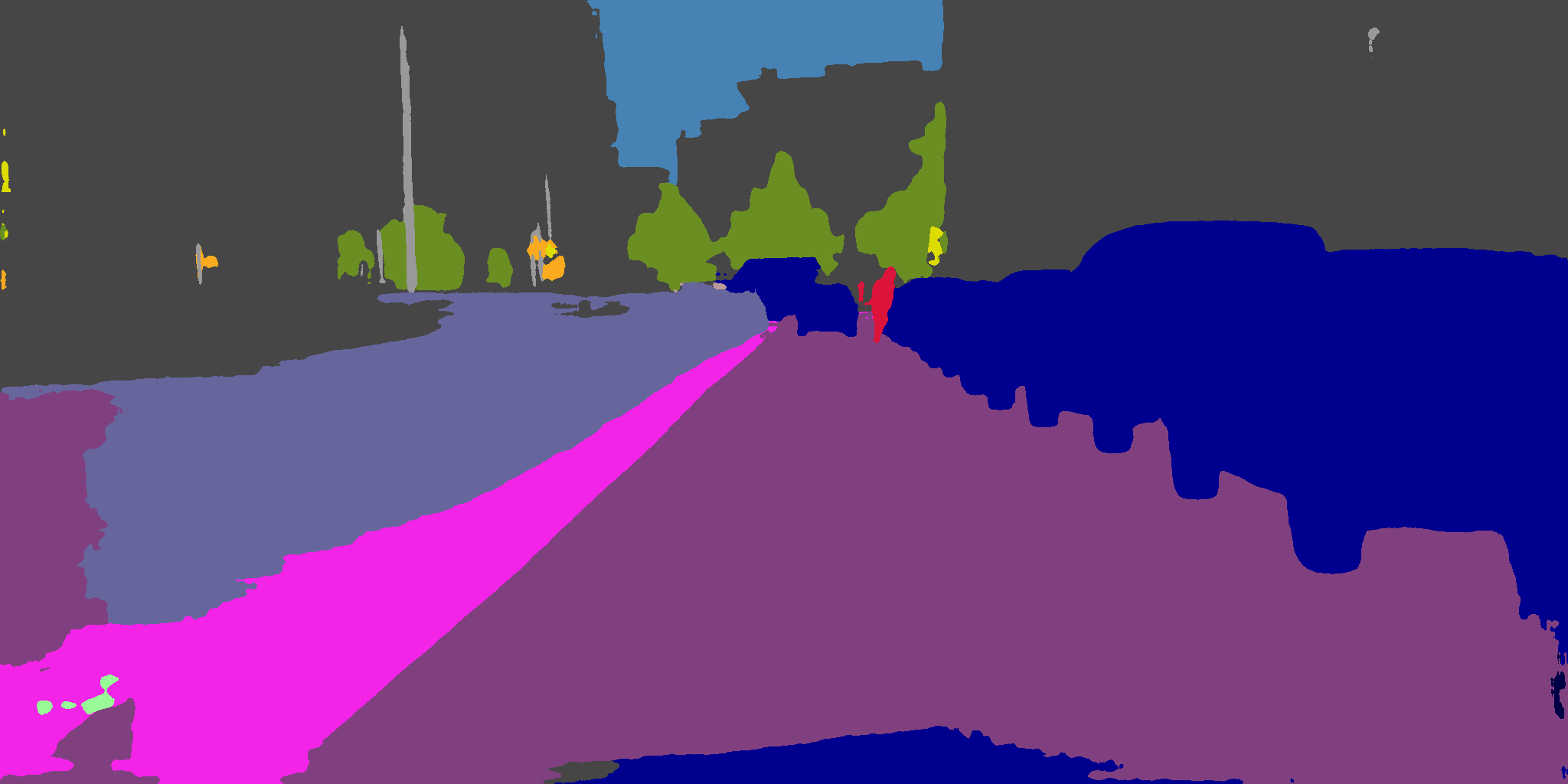} &
   \includegraphics[width=0.23\linewidth]{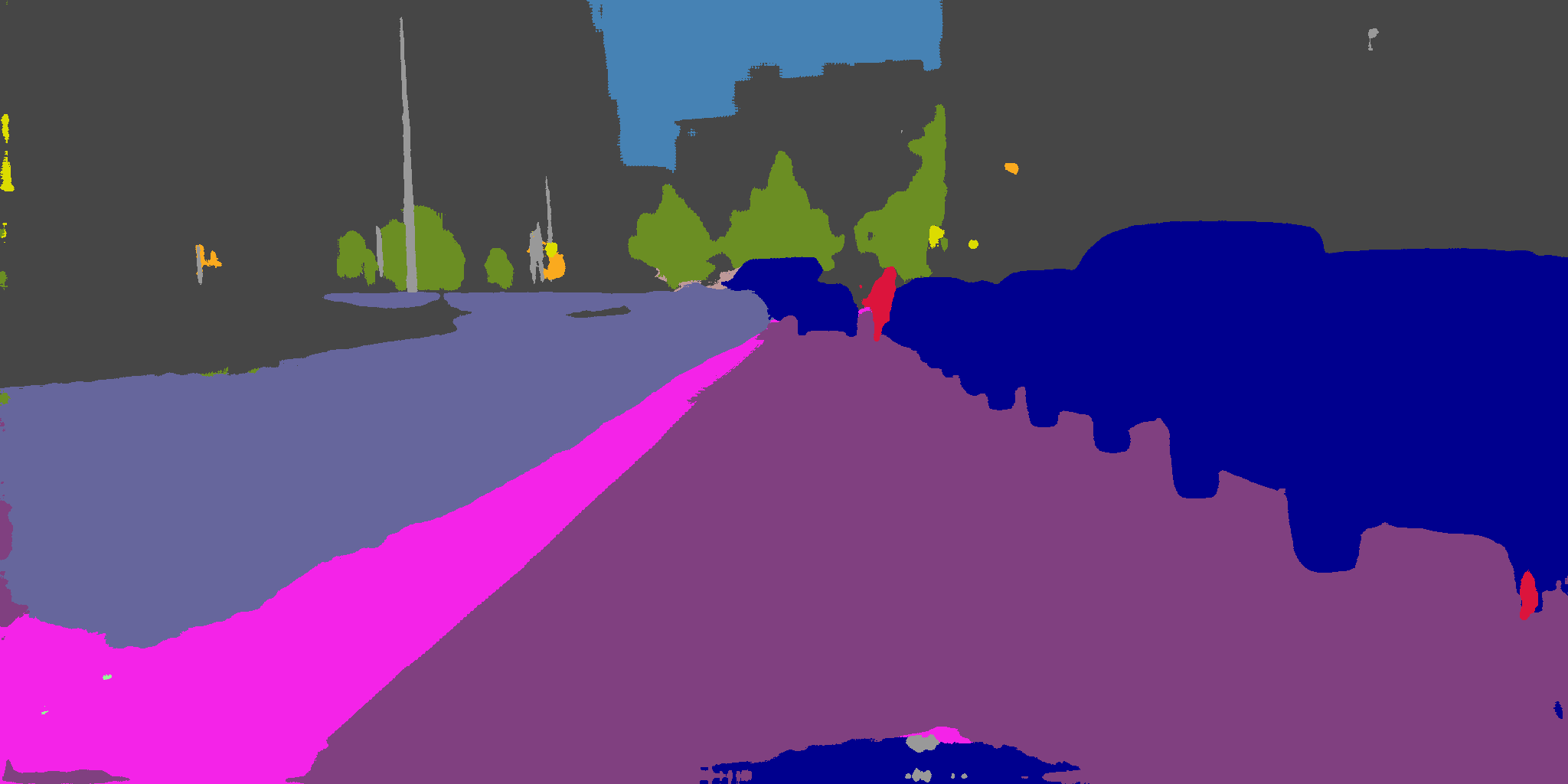} &
   \includegraphics[width=0.23\linewidth]{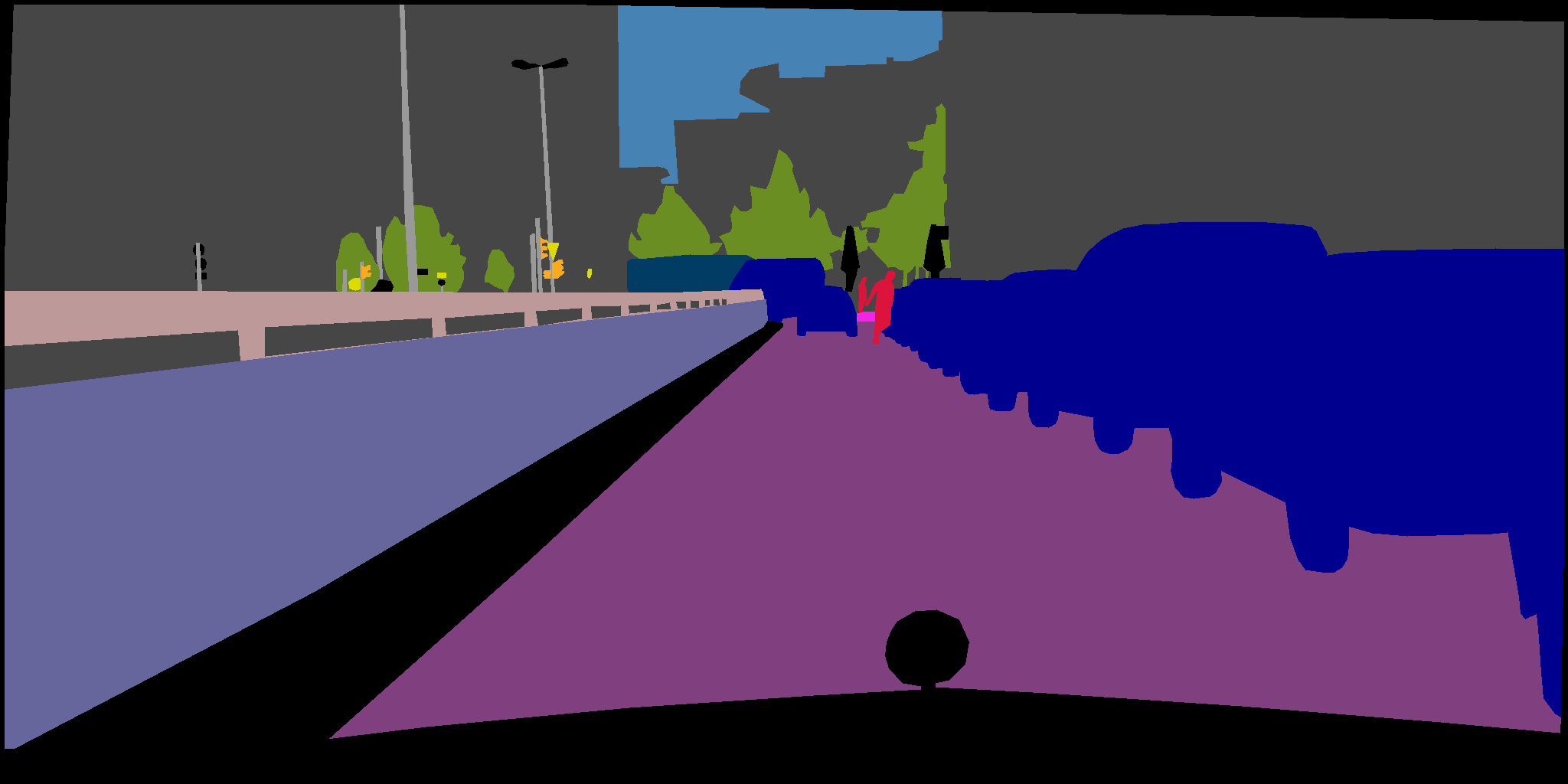} \\

   \includegraphics[width=0.23\linewidth]{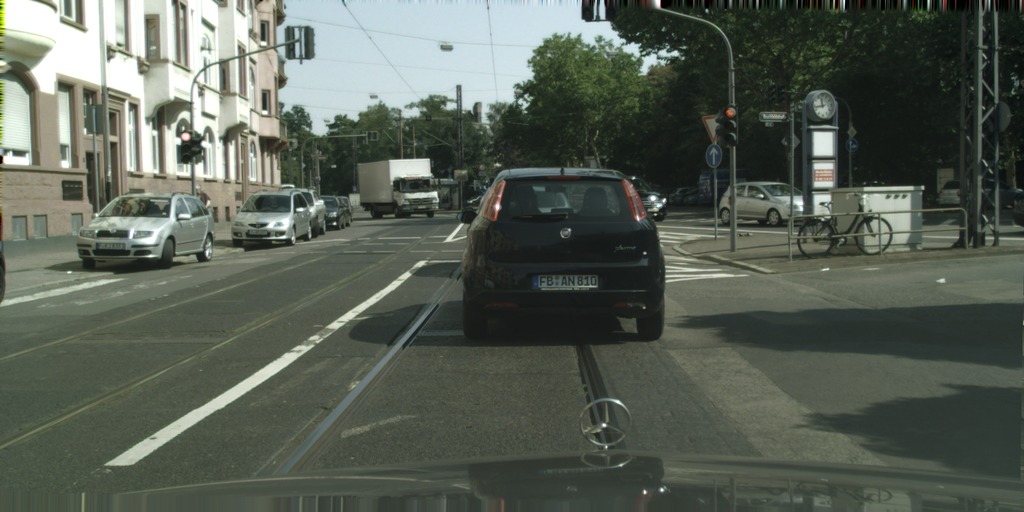} &
   \includegraphics[width=0.23\linewidth]{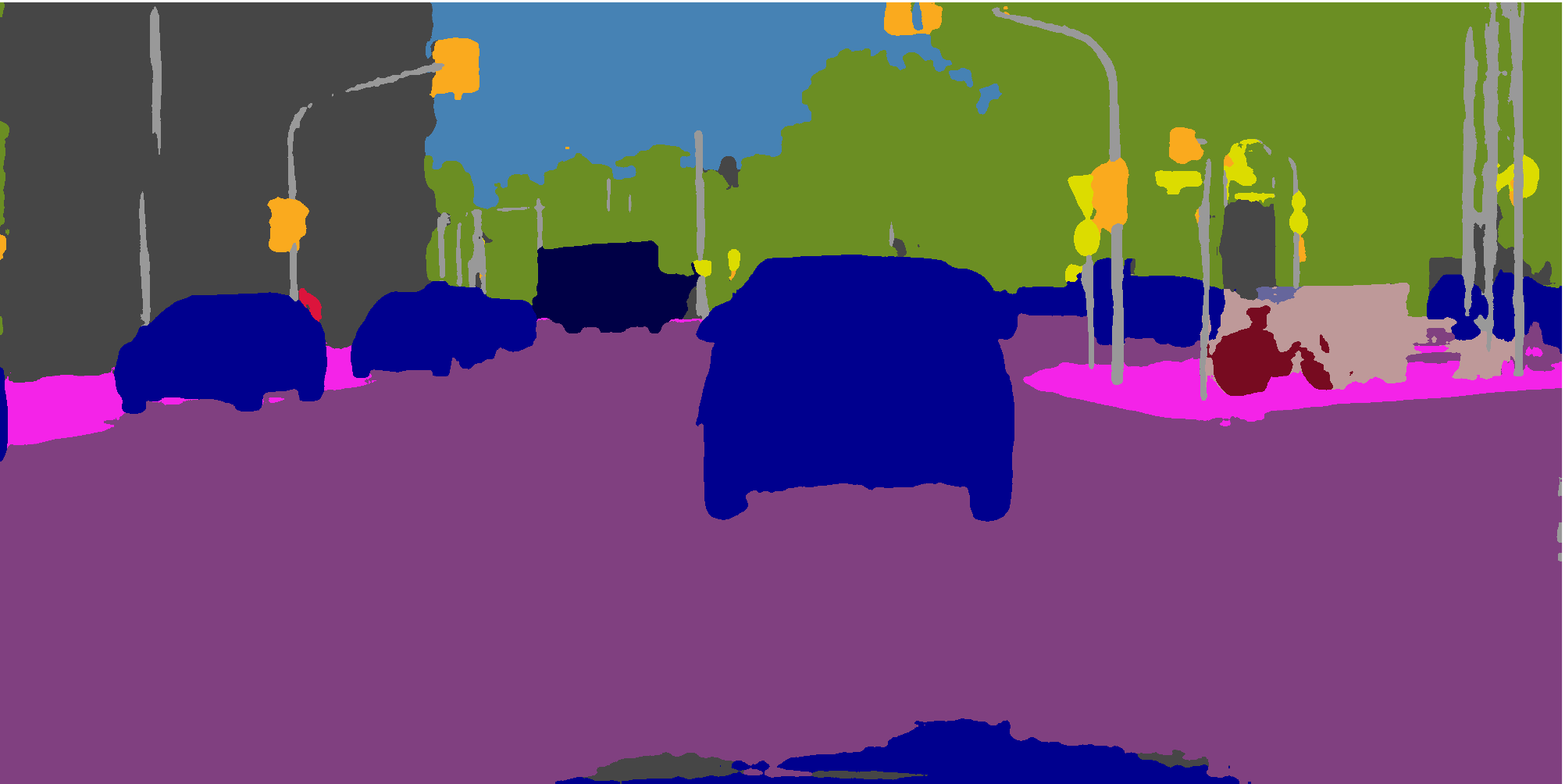} &
   \includegraphics[width=0.23\linewidth]{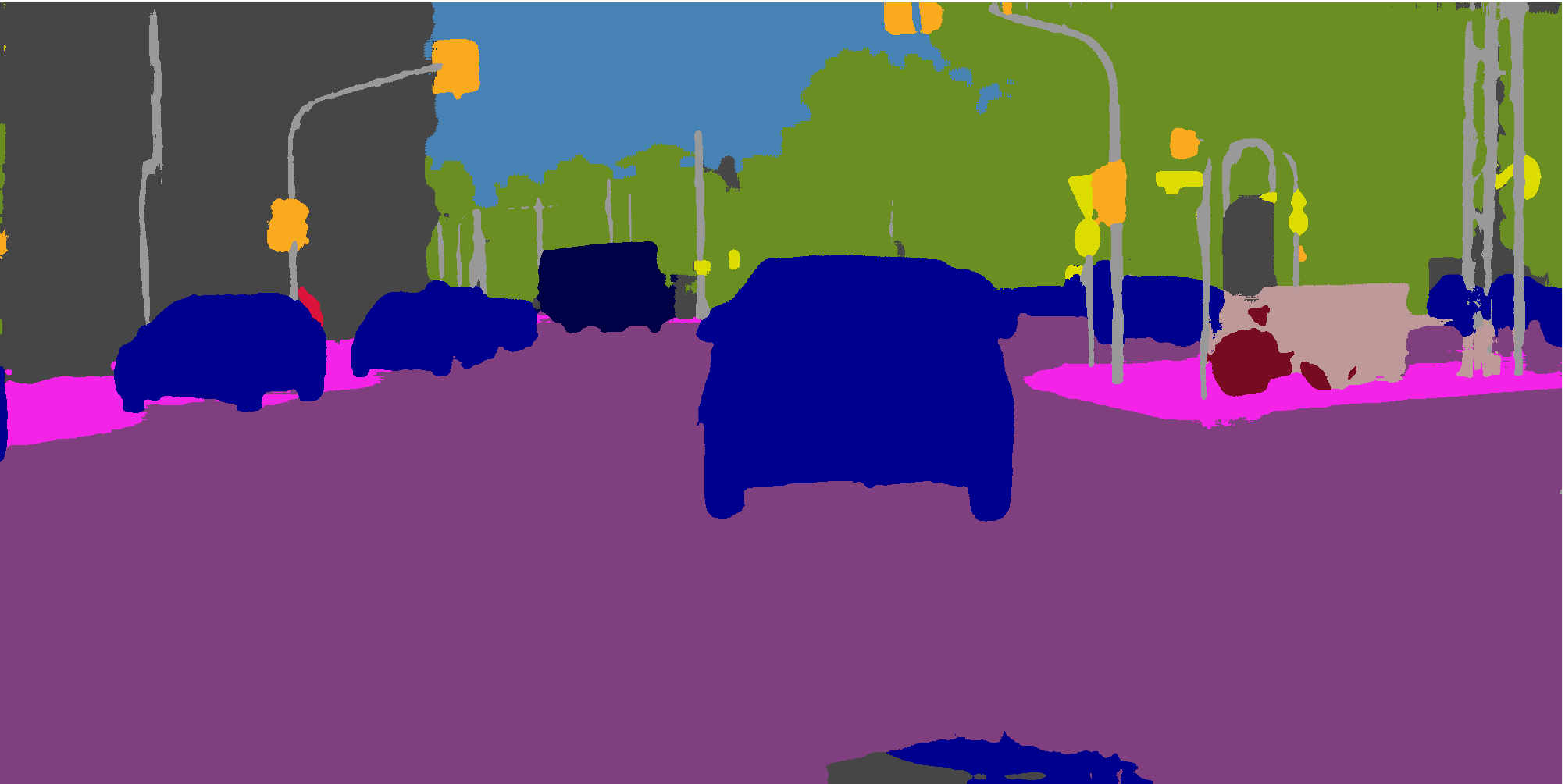} &
   \includegraphics[width=0.23\linewidth]{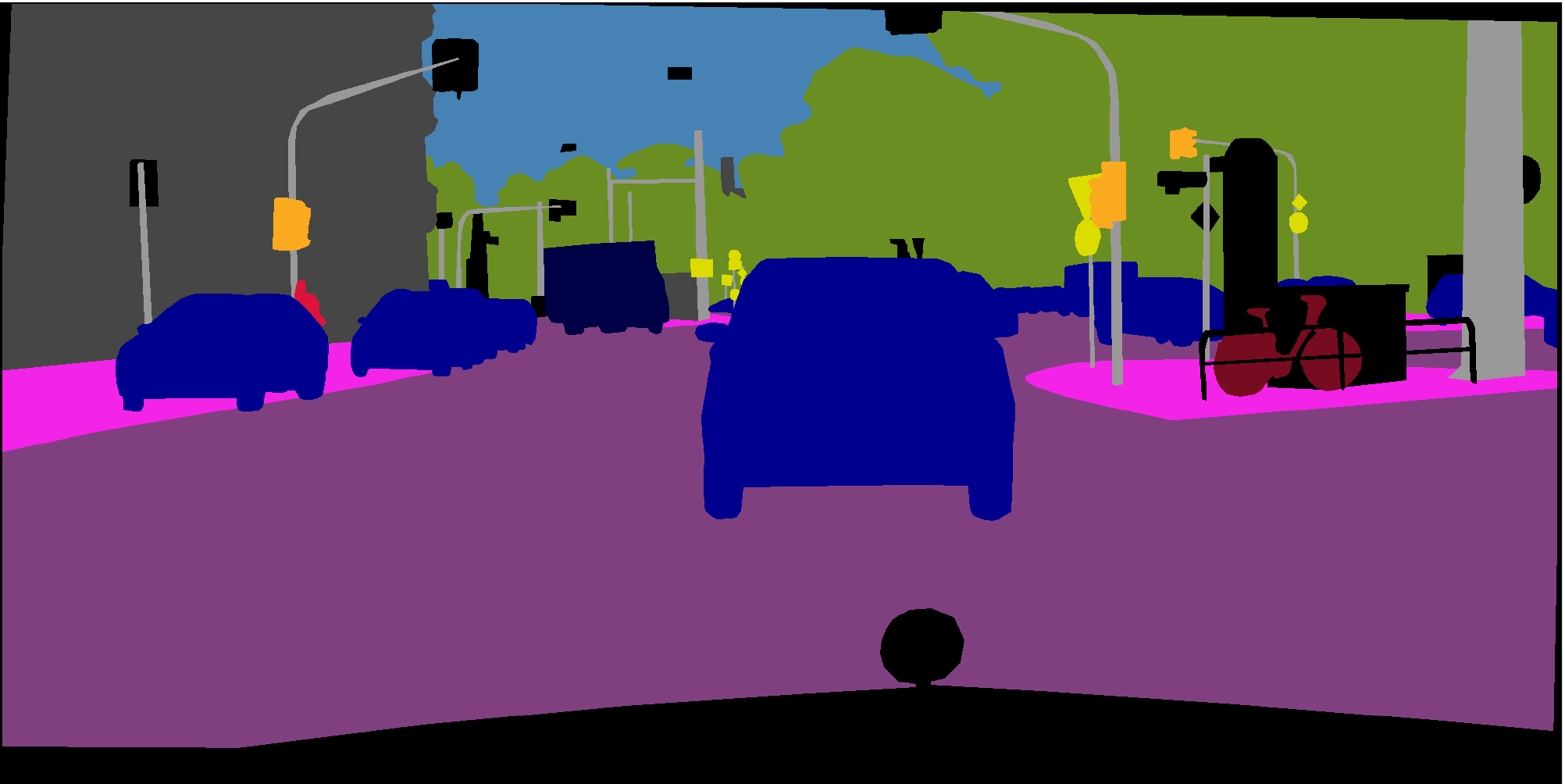} \\

   \includegraphics[width=0.23\linewidth]{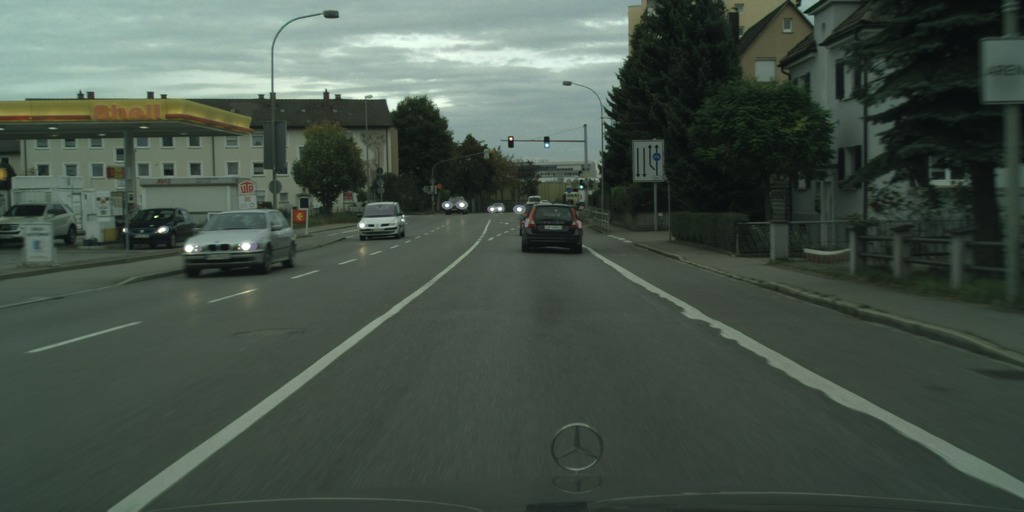} &
   \includegraphics[width=0.23\linewidth]{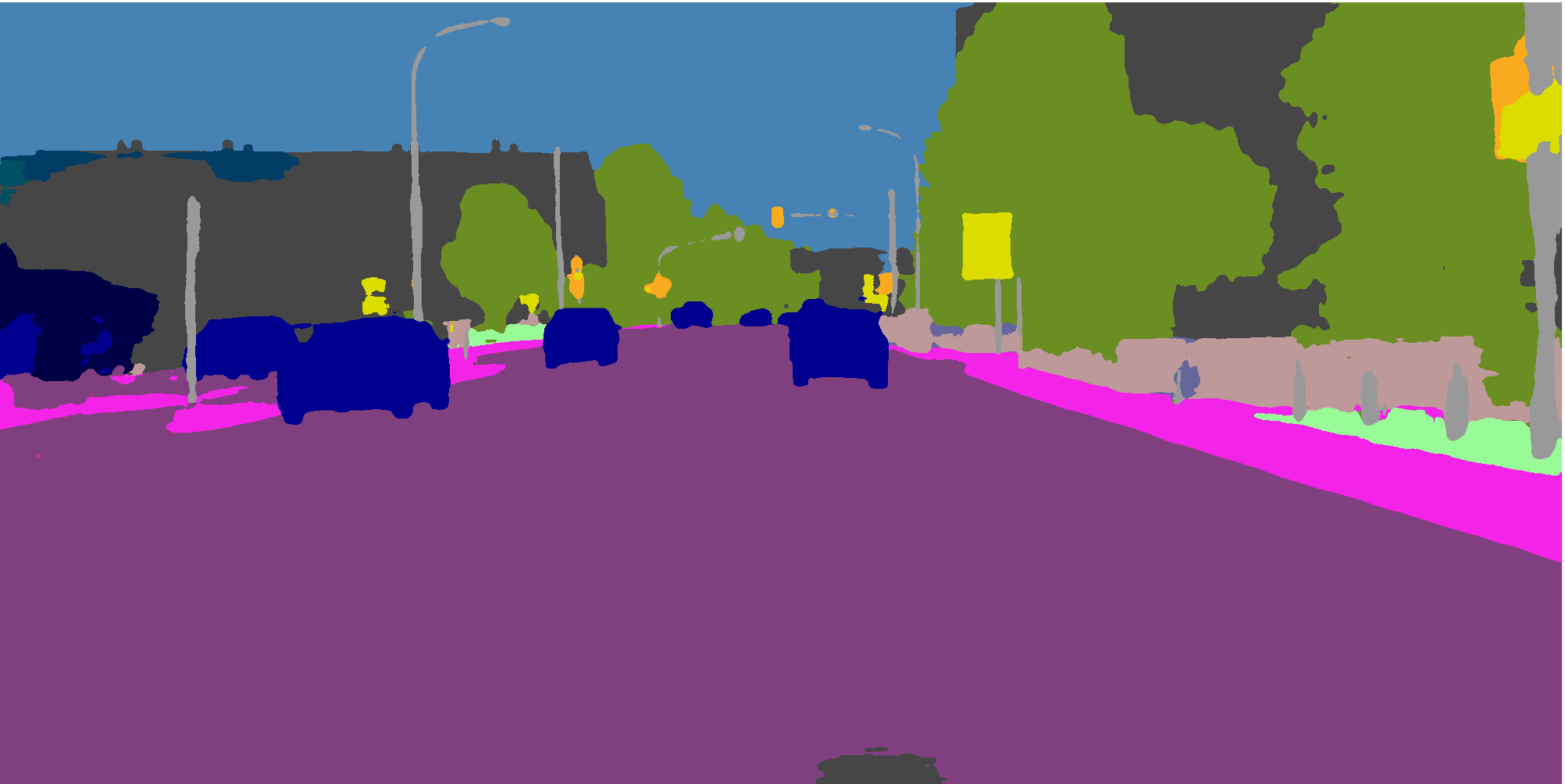} &
   \includegraphics[width=0.23\linewidth]{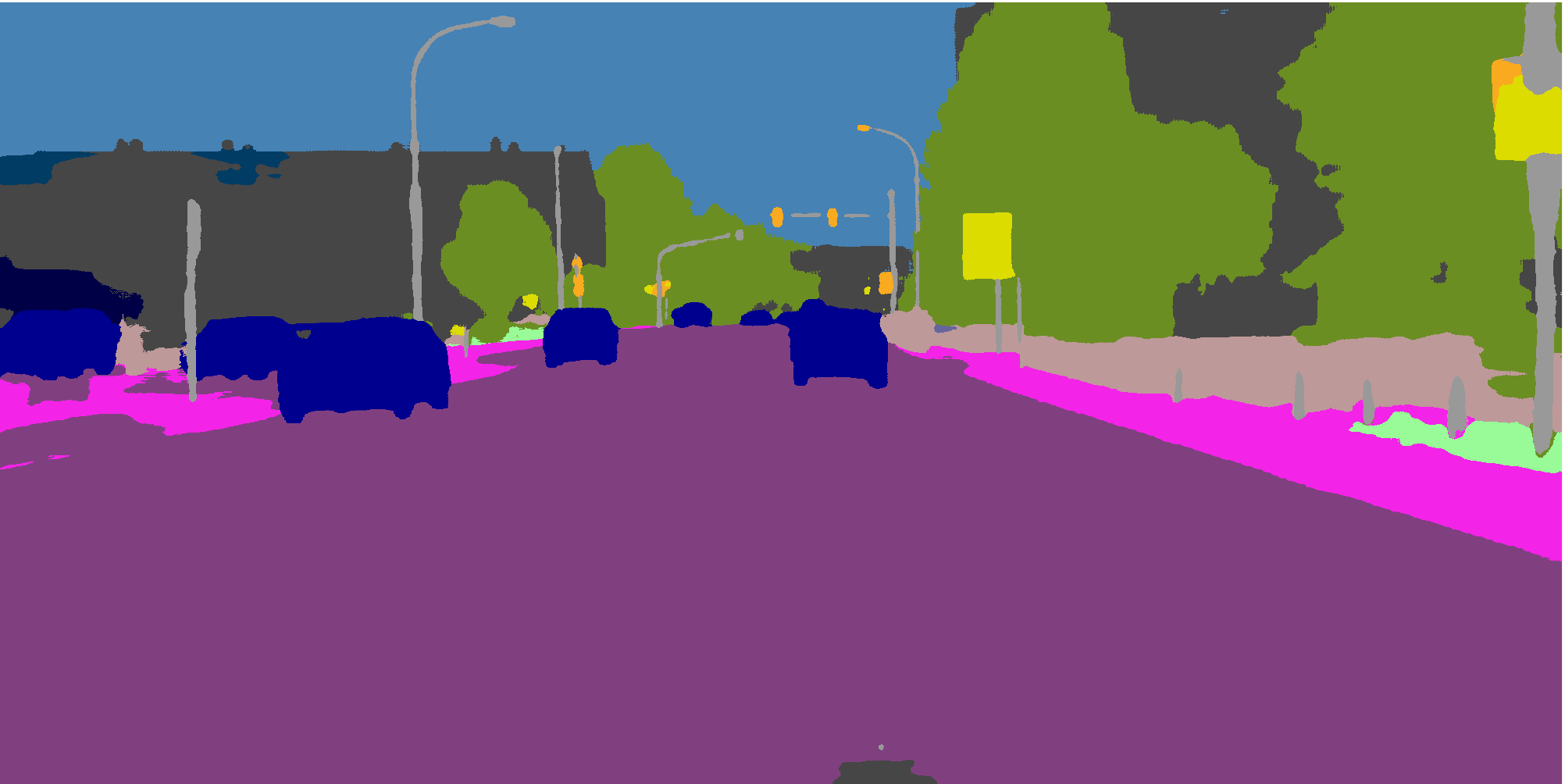} &
   \includegraphics[width=0.23\linewidth]{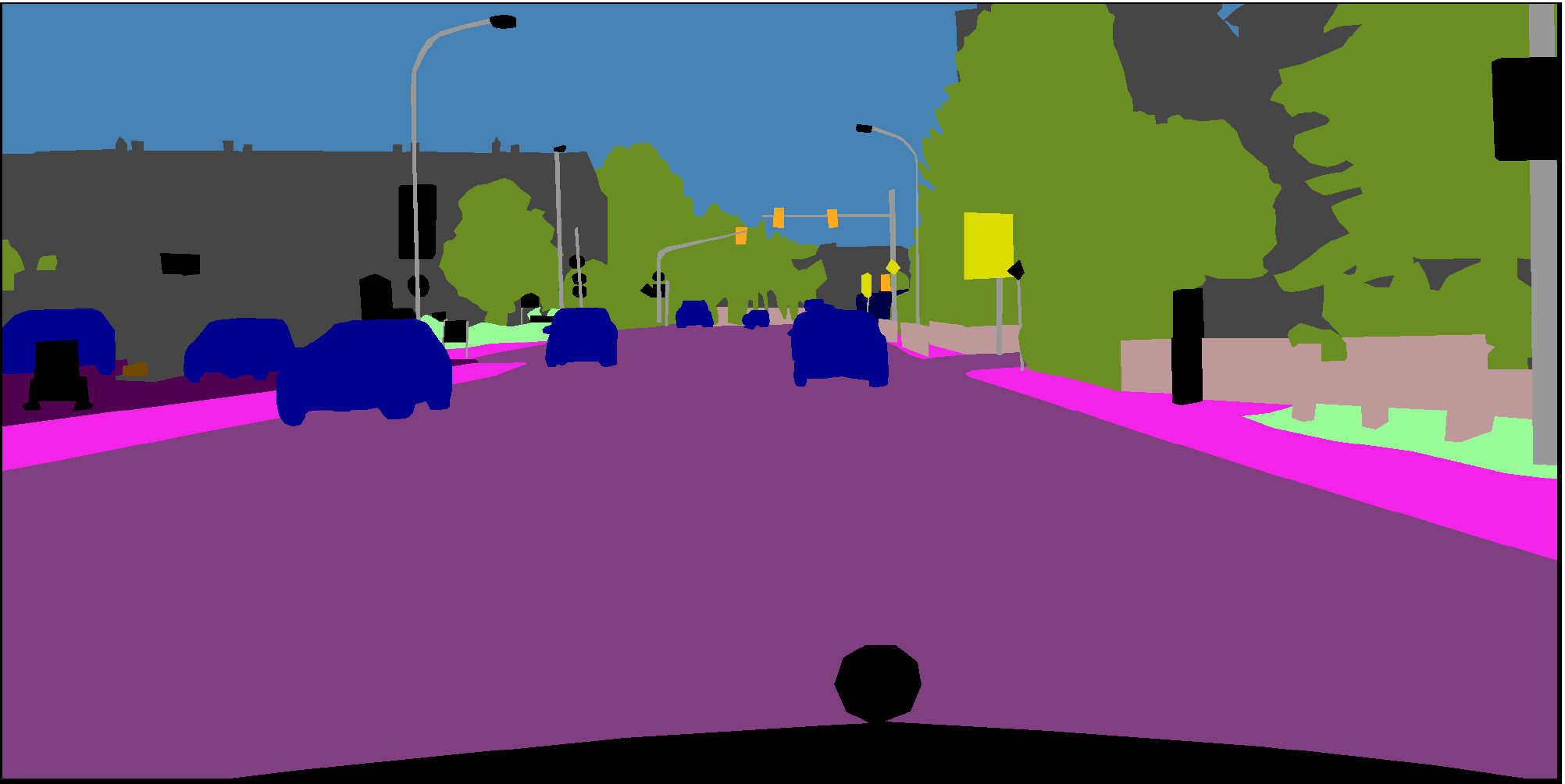} \\

\end{tabular}   
\setlength\tabcolsep{6pt} 

\end{center}
\vspace{-0.6cm}   
   \caption{Qualitative results on the {\sc Cityscapes} validation set. Black regions in the ground truth are ignored during evaluation. Our CRF models contextual relationships between classes, hence unlike LRR, it does not label ``road'' as being on top of ``sidewalk'' (Row 2). Note that the traffic lights are better segmented with the additional CRF-Grad layer. Adding the CRF-Grad layer increased the IoU of the class traffic lights from $66.8$ to $68.1$. }
\label{fig:cs_res}
\end{figure*}
In Table~\ref{tab:quant_res} the results of evaluating our model on the test set are compared to current state of the art. As can be seen, our model is on par although the improvement upon LRR is minor. An interesting aspect of {\sc Cityscapes} is that it contains classes of thin and vertical objects, e.g., traffic light and pole. What we noticed is that the spatial filters for these classes usually get a more oblong shape. This type of pairwise filters does not add as much energy for switching classes going in the horizontal direction, favoring vertically elongated segmentations. This can be seen in the spatial filter for the class interaction between ``terrain'' and ``traffic light'' in Fig.~\ref{fig:cs:filters}. Some example segmentations are shown in Fig.~\ref{fig:cs_res}. Additional examples as well as class-wise results are included in the supplementary material. Our method got better, or equal, results for 14 of the 19 classes compared to the baseline.

\section{Concluding Remarks}
In this paper, we have introduced a new framework capable of learning arbitrarily shaped pairwise potentials in random fields models. In a number of experiments, we have empirically demonstrated that our developed framework can improve state-of-the-art CNNs by adding a CRF layer. We have also seen that the learned filters are not necessarily Gaussian, and may capture other kinds of interactions between labels. In addition, we have shown advantages of our inference method compared to the commonly used mean-field method.

A key factor for the success of deep learning and by now a well-established paradigm is that the power of convolutions should be used, especially for the first layers in a CNN. Our work supports that repeated usage of convolutions in the final layers is also beneficial. We also note that our gradient descent steps resemble the highly successful {\sc ResNet}~\cite{he2016deep}, as one step in gradient descent is, in principle, an identity transformation plus a correction term.

There are several future research avenues that we intend to explore. In our model, many free variables are introduced and this may lead to over-fitting.
One way to compensate would be to collect larger datasets and consider data augmentation. An alternative approach would be to directly encode geometric shape priors into the random fields and thereby reducing the required amount of data.

\paragraph{Acknowledgements.} This work has been funded by the Swedish Research Council (grant no. 2016-04445), the Swedish Foundation for Strategic Research (Semantic Mapping and Visual Navigation for Smart Robots), Vinnova / FFI (Perceptron, grant no. 2017-01942), ERC (grant ERC-2012-AdG 321162-HELIOS) and EPSRC (grant Seebibyte EP/M013774/1 and EP/N019474/1).

{\small
\bibliographystyle{plain}
\bibliography{sample}
}

\clearpage

\newpage
\newpage



\section*{Supplementary Material}
\subsection*{Proof of Proposition~\ref{prop:tight}}

Let $E(\bm{x}^*)$ and $E(\bm{q}^*)$ denote the optimal values of \eqref{realvalprog}, where $\bm{x}^*$ is restricted to boolean values. Then,
\[
        E(\bm{x}^*) = E(\bm{q}^*).
\]
\begin{proof}
We will show that for {\em any} real $\bm{q}$, one can obtain a binary $\bm{x}$ such that $E(\bm{x}) \le E(\bm{q})$. In particular, it will be true for $\bm{x}^*$ and $\bm{q}^*$, which implies $E(\bm{x}^*) = E(\bm{q}^*)$.

Let $\bm{q}$ be given, and let $\bm{x} \in \mathcal{L}^N$.
One may define
$E^m(\bm{x}, \bm{q}) = E(x_1, \ldots, x_m, q_{m+1}, \ldots, q_N)$
such that each $x_i$ or $q_i$ is a vector with entries such as $q_{i:\lambda}$
or 
$x_{i:\lambda}$, but for each $i$  only one value 
$x_{i:\lambda}$ is non-zero (and equal to $1$).
Since $E^0 = E(\bm{q})$ and 
$E^N = E(\bm{x})$ it will be sufficient
to find a $\bm{x}$ such 
that $E^{m}(\bm{x}, \bm{q}) \le E^{m-1}(\bm{x}, \bm{q})$ for all
$m$.
The required $\bm{x}$ will be constructed one element at a time.

The key observation is that $E^m$ is multilinear in
the $q_i$. Then, it follows that
%
\begin{align*}
\begin{split}
E^{m-1}(\bm{x}, \bm{q}) &= 
E(x_1, \ldots, x_{m-1}, q_{m}, \ldots q_N) \\
&= \sum_{x_m\in\mathcal{L}} q_{m:x_m} 
E(x_1, \ldots, x_m, q_{m+1}, \ldots q_N).
\end{split}
\end{align*}
Here, $x_m$ is treated as a variable and $x_1, \ldots, x_{m-1}$
are fixed.  Since $\sum_{x_m\in\mathcal{L}} q_{m:x_m} = 1$ there must be at
least one choice of $x_m$ such that 
$E^{m-1}(\bm{x}, \bm{q}) \ge
E(x_1, \ldots, x_m, q_{m+1}, \ldots q_m)  = 
E^{m}(\bm{x}, \bm{q})$.
\end{proof}

\subsection*{Mean-Field Objective}
In this section, we give a brief derivation of what is minimized with the mean-field method. The key idea behind the method is to approximate 
a complex probability distribution $P$ by a simpler one $Q$
that one can solve
(find its mode), using the simpler distribution as a stand-in for the
actual probability distribution (MRF) of interest.
As previously, we consider a probability distribution $P$
for a random field, given by the Gibbs distribution.
The probability of an assignment $\bm{x}\in\mathcal{L}^N$ is
\begin{equation}
\label{probability-Gibbs-form}
P(\bm{x}) 
= \frac{1}{Z} \exp \left( -E(\bm{x}) \right).
\end{equation}
A natural measure commonly used as a measure of
distance between two probability distributions $P$ and $Q$
is the KL divergence $D(Q\,\|\,P)$ defined by
\begin{align}
\begin{split}
\label{DQP}
D(Q\,\|\,P)  & =  \sum_{\bm{x}\in\mathcal{L}^N} Q(\bm{x}) \log  \left(
\frac{Q(\bm{x})}{P(\bm{x})} \right) \\
& = - \hspace{-0.1cm} \sum_{\bm{x}\in\mathcal{L}^N} Q(\bm{x}) \log P(\bm{x}) \hspace{-0.1cm}+\hspace{-0.1cm} \sum_{\bm{x}\in\mathcal{L}^N} 
Q(\bm{x}) \log Q(\bm{x}) .
\end{split}
\end{align}

Now, 
plugging in the form of the probability $P(\bm{x})$ given
in \eqref{probability-Gibbs-form} into \eqref{DQP}, we obtain $D(Q\,\|\,P) = $
\begin{align*}
 - \hspace{-0.1cm} \sum_{\bm{x}\in\mathcal{L}^N} Q(\bm{x})\log \left (\frac{1}{Z} \exp
\left( -E(\bm{x}) \right)  \right) \hspace{-0.1cm}+\hspace{-0.1cm}
\sum_{\bm{x}\in\mathcal{L}^N}  Q(\bm{x}) \log
Q(\bm{x}) \\
 = \sum_{\bm{x}\in\mathcal{L}^N} Q(\bm{x}) E(\bm{x})
+ \log Z   + \sum_{\bm{x}\in\mathcal{L}^N}  Q(\bm{x}) \log Q(\bm{x}) ,
\end{align*}
where we have used the fact that $\sum_{\bm{x}\in\mathcal{L}^N} Q(\bm{x}) = 1$.

Because $ \log Z$ is a constant, minimizing the KL divergence between
these two distributions is equivalent
to minimizing the following quantity:
\begin{equation}
\label{FPQ}
\sum_{\bm{x}\in\mathcal{L}^N} Q(\bm{x}) E(\bm{x}) +
 \sum_{\bm{x}\in\mathcal{L}^N}  Q(\bm{x}) \log Q(\bm{x}).
\end{equation}
The first term is equal to
the expected value of the energy $E(\bm{x})$ under the probability
distribution $Q$; for now we will write this as $\mathcal{E}_Q[E]$.
Now, consider the second term.
The entropy of a probability distribution $Q$ is equal to
$-\sum_{\bm{x}\in\mathcal{L}^N} Q(\bm{x}) \log Q(\bm{x})$.  Writing the entropy
of $Q$ as $H(Q)$, \eqref{FPQ} takes the form
\begin{equation}
\label{eq:FPQ-2}
\mathcal{E}_Q[E] - H(Q),
\end{equation}
so the quantity is equal to the expectation of the Gibbs energy under
distribution $Q$ minus the entropy of $Q$.

\section*{Hyperparameter Settings}
\subsection*{Weizmann Horse}
All the models presented in the experiments section were trained end-to-end with learning rate of $10^{-3}$ (normalized by the number of pixels), momentum $0.9$, weight decay $5 \cdot 10^{-3}$ and batch size 20. The size of the spatial filters for these runs were $9 \times 9$ if not specified as other. The number of iterations were set to $5$, step size to $0.5$ and unary weight was initialized as $0.5$. For the models with bilateral filter $s$, the number of neighbours each pixel interacts with in each dimension, were set to $1$.
\subsection*{NYU V2}
The model used on the {\sc NYU V2} data set was trained with learning rate $10^{-11}$ (not normalized), momentum $0.99$, weight decay $0.005$ and batch size $10$. The size of the spatial filters were set to $9 \times 9$ and the bilateral filters to $s=1$. The number of iterations were set to $5$, step size to $0.5$ and unary weight was initialized as $0.5$.
\subsection*{Cityscapes}
The model used on the {\sc Cityscapes} data set was trained with a learning rate of $10^{-3}$ (normalized by the number of pixels), momentum $0.9$ and weight decay $5 \cdot 10^{-3}$. The size of the spatial filters for these runs were $9 \times 9$ and the bilateral filters to $s=1$. The number of iterations were set to $5$, step size to $0.5$ and unary weight was initialized as $0.5$.

\section*{Simplex Projection}
As mentioned in the paper we want to project our values onto the simplex $\triangle^L$ satisfying $\sum_{\lambda \in \mathcal{L}} q_{i:\lambda} = 1$ and $0 \leq q_{i:\lambda} \leq 1$ for each step in our gradient descent algorithm. This is done following the method presented by Chen \etal \cite{chen2011} which is summarized in Algorithm~\ref{alg:projection}. Note that this projection is done individually for each pixel $i$, for derivation of this algorithm we refer to the article.

\begin{algorithm}
1. Sort $\tilde{\bm{q}}_i \in \mathbb{R}^L$ in ascending order and set k = L - 1 \\
2. Compute $t_i = \frac{\sum^L_{j=k+1} \tilde{{q}}_{i:j} - 1}{L-k}$, If $t_i \geq \tilde{{q}}_{i:k}$ set $\hat{t} = t_i$ and go to step 4, otherwise set $k \leftarrow k - 1$ and redo step 2, if $k = 0$ go to step 3 \\
3. Set $\hat{t} = \frac{\sum^L_{j=1} \tilde{{q}}_{i:j} - 1}{L}$ \\
4. Return $\bm{q}_i$, where $q_{i:\lambda} = \max \left(\tilde{q}_{i:\lambda}- \hat{t},0 \right)$, $\lambda \in \mathcal{L}$

\caption{\label{alg:projection}Algorithm 2. Projection of $\tilde{\bm{q}}_i \in \mathbb{R}^L$ onto the simplex  $\triangle^L$ satisfying $\sum_{\lambda \in L} q_{i:\lambda} = 1$ and $0 \leq q_{i:\lambda} \leq 1$.}
\end{algorithm}

In reality we use a leaky version of the last step of the projection algorithm, \ie instead of the $\max(\cdot,0)$ operator we use the following function
\begin{equation}
f_{\alpha}(\tilde{q}_{i:\lambda}) = \left\{\begin{matrix}
\tilde{q}_{i:\lambda} -\hat{t} & 0 \leq \tilde{q}_{i:\lambda} -\hat{t} \\
 \alpha \left ( \tilde{q}_{i:\lambda} -\hat{t} \right ) & \tilde{q}_{i:\lambda} -\hat{t} < 0
\end{matrix}\right. 
\end{equation}

where $\alpha$ is a parameter controlling the level of leakage. Note that for $\alpha = 0$ we get the the strict $\max(\cdot,0)$ operator. As previously mentioned, the projection is done individually for each pixel. It can be described as a function $\bm{f}(\tilde{\bm{q}}): \mathbb{R}^L \rightarrow \mathbb{R}^L$, which Jacobian has the elements 
\begin{equation}
\frac{\partial f_{\lambda}}{\partial \tilde{q}_{\mu}} = \left\{\begin{matrix} \left\{\begin{matrix}
\alpha \left ( 1 - \frac{\partial \hat{t}}{\partial \tilde{q}_{\mu}} \right ) &  \tilde{q}_{\mu} - \hat{t} < 0 \\
1 - \frac{\partial \hat{t}}{\partial \tilde{q}_{\mu}} &   \tilde{q}_{\mu} - \hat{t} \geq 0 \end{matrix}\right.
& \mu = \lambda \\ 
\left\{\begin{matrix}
\hspace{2mm} -\alpha \frac{\partial \hat{t}}{\partial \tilde{q}_{\mu}} \hspace{6mm} &  \tilde{q}_{\mu} - \hat{t} < 0 \\
\hspace{3mm} - \frac{\partial \hat{t}}{\partial \tilde{q}_{\mu}} \hspace{5mm} &   \tilde{q}_{\mu} - \hat{t} \geq 0 \end{matrix}\right. 
&   \mu \neq \lambda \end{matrix}\right.
\end{equation}
where $\frac{\partial \hat{t}}{\partial \tilde{q}_{\mu}} = \frac{1}{L-k}$ if $\tilde{q}_{\mu} > \hat{t} $ and $0$ otherwise ($L$ and $k$ defined as in Algorithm~\ref{alg:projection}). Knowing the Jacobian, the error derivatives with respect to the input can be computed during back propagation. The reason for introducing the leaky version is to avoid error derivatives becoming zero during back propagation. A non-zero $\alpha$ was found to facilitate the training process considerably with the drawback that we do not necessarily satisfy the constraints in \eqref{realvalprog}. However, one could set $\alpha$ to zero during inference to strictly satisfy \eqref{realvalprog}.

\section*{Convergence analysis}
\begin{figure*}[t]
\begin{center}
   \includegraphics[width=0.45\linewidth]{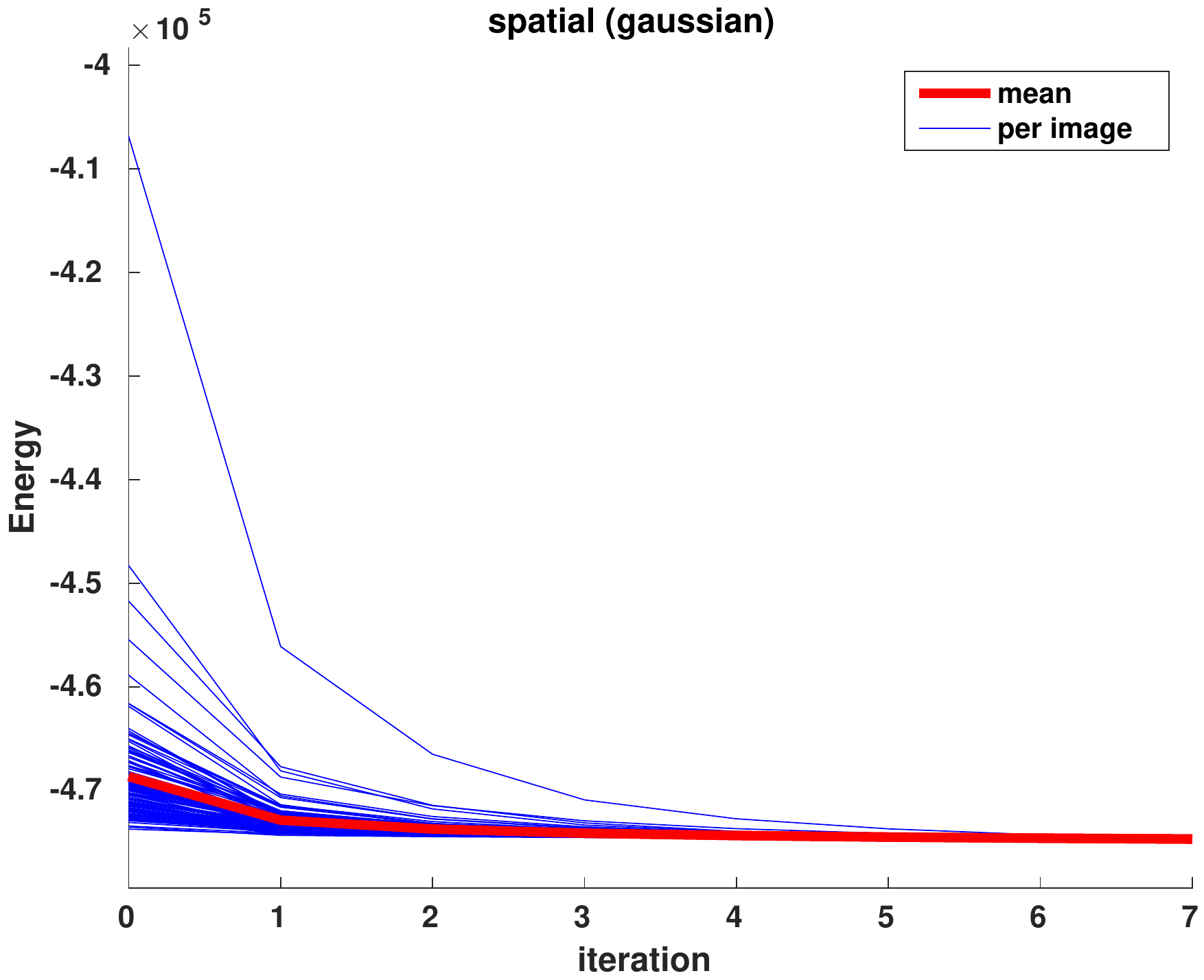}
   \includegraphics[width=0.45\linewidth]{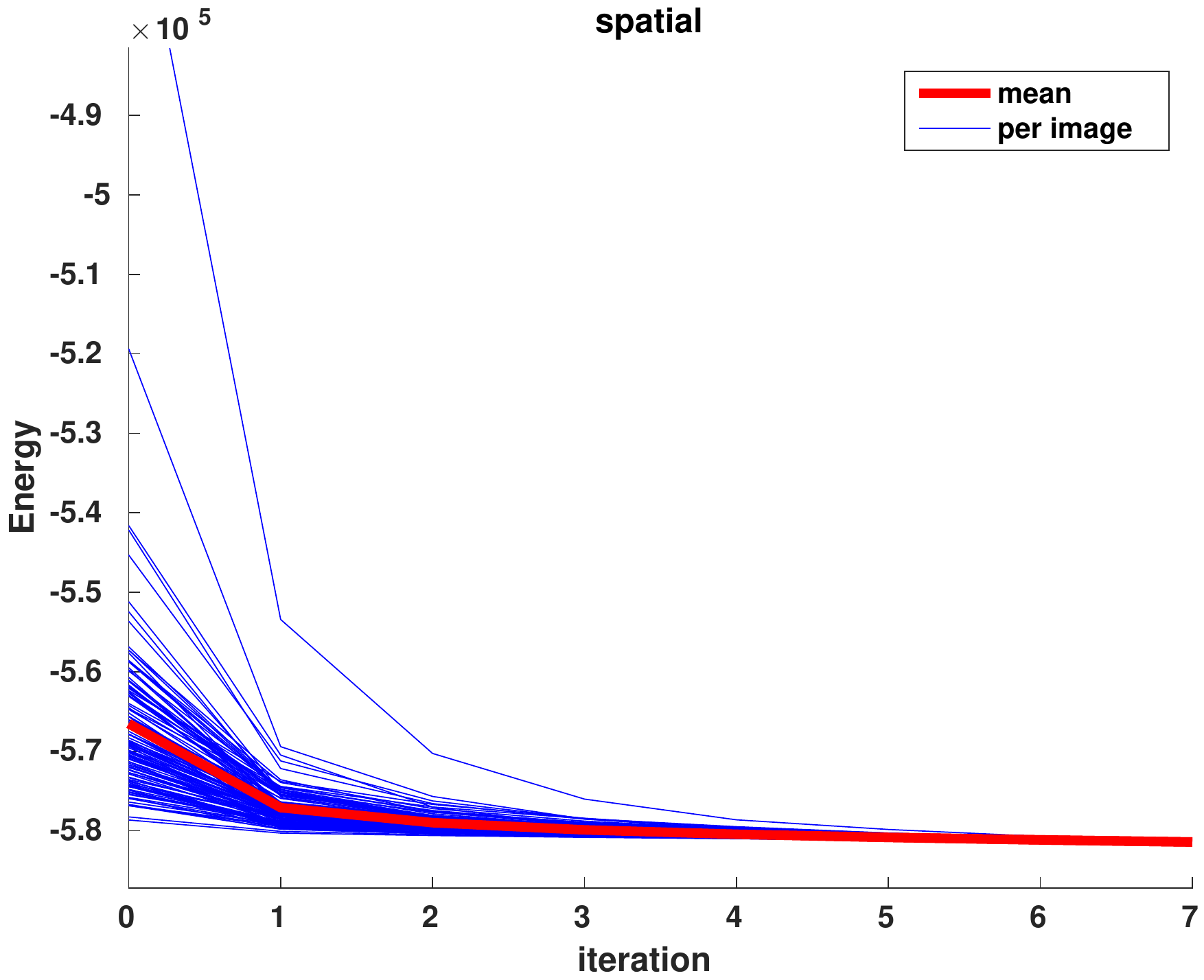}
   \includegraphics[width=0.45\linewidth]{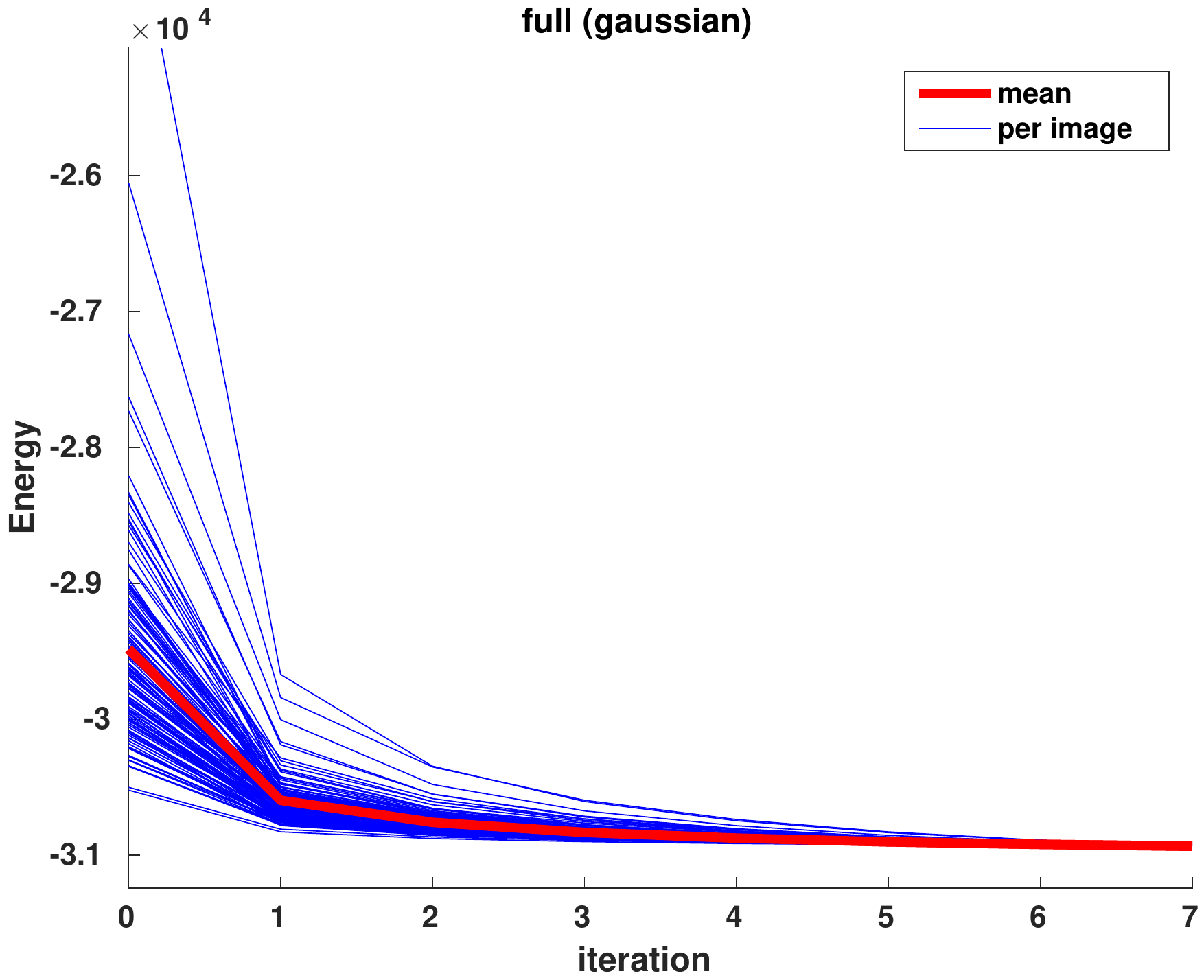}
   \includegraphics[width=0.45\linewidth]{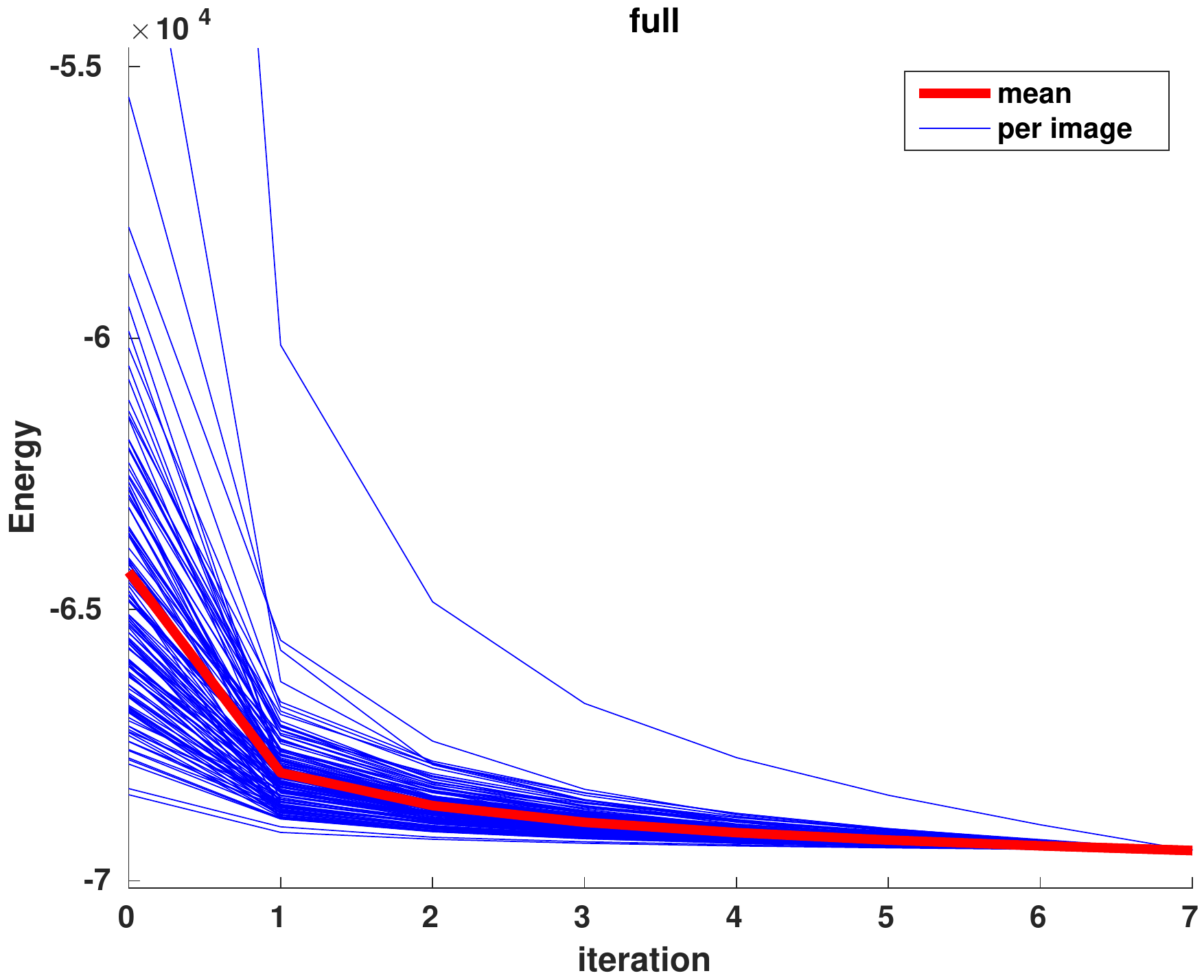}
   
\end{center}
   \caption{CRF energy, as defined in \eqref{realvalprog}, plotted as a function of iterations of the gradient descent method. These energies are calculated while running the trained models on the {\sc Weizmann Horse} dataset, The thin blue lines shows the imagewise energy change while the thicker red line show the mean. For these calculations the leak factor was set to zero, meaning that the solutions satisfy the constraints of \eqref{realvalprog}. Note that, for presentation purposes, all energies have been normalized to have the same final energy.}
\label{fig:crfenergy-supp}
\end{figure*}

\begin{figure*}[t]
\begin{center}
   \includegraphics[width=0.45\linewidth]{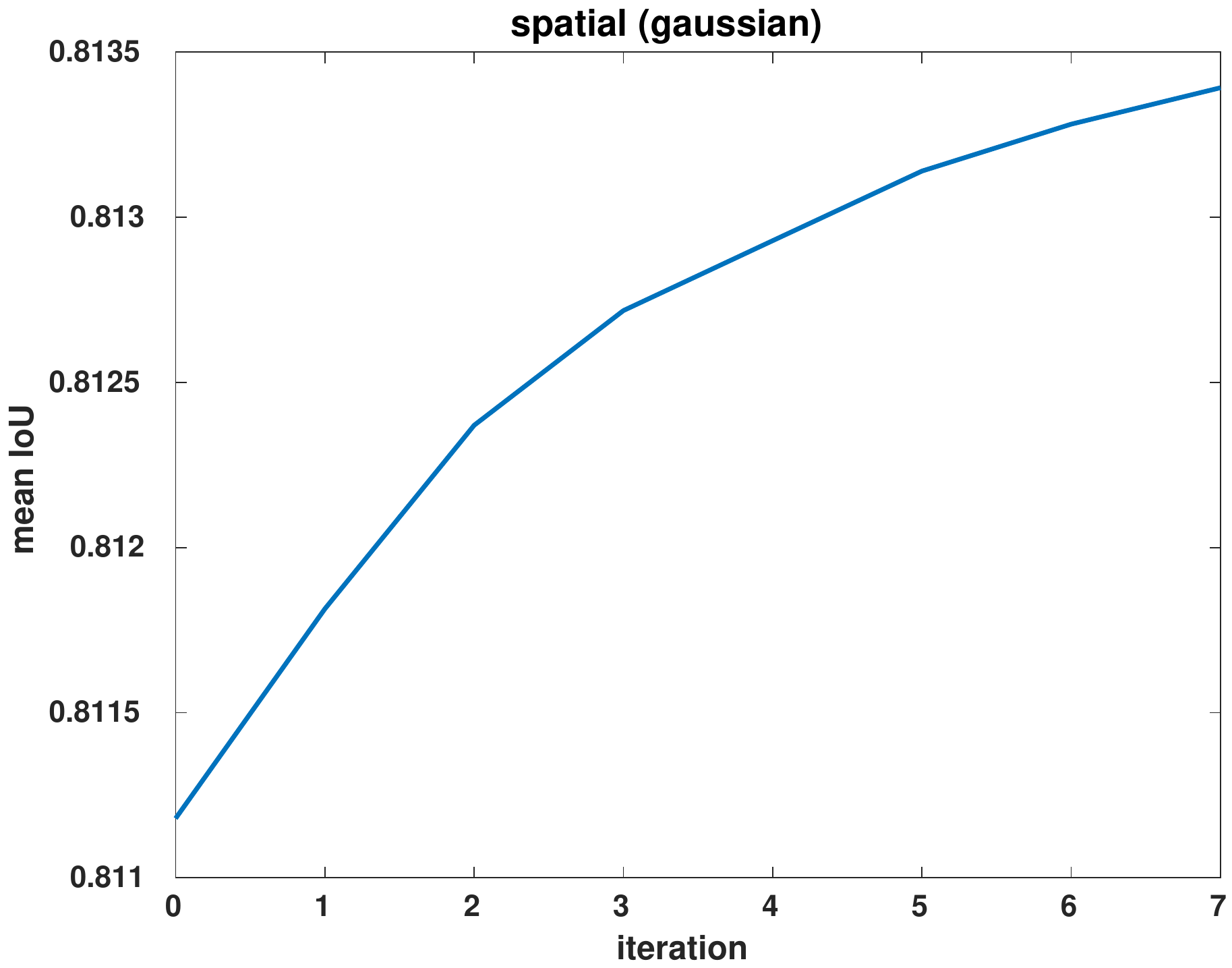}
   \includegraphics[width=0.45\linewidth]{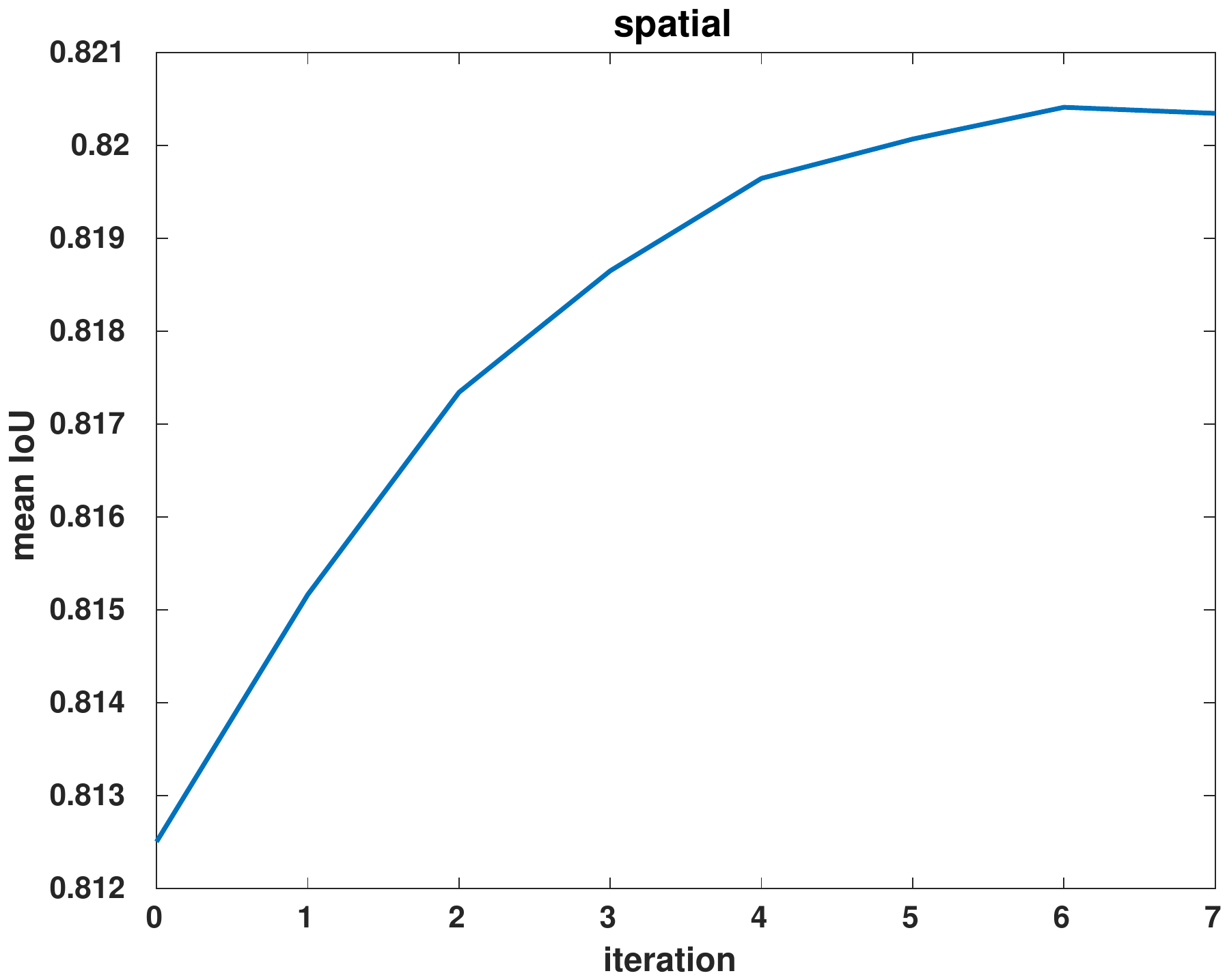}
   \includegraphics[width=0.45\linewidth]{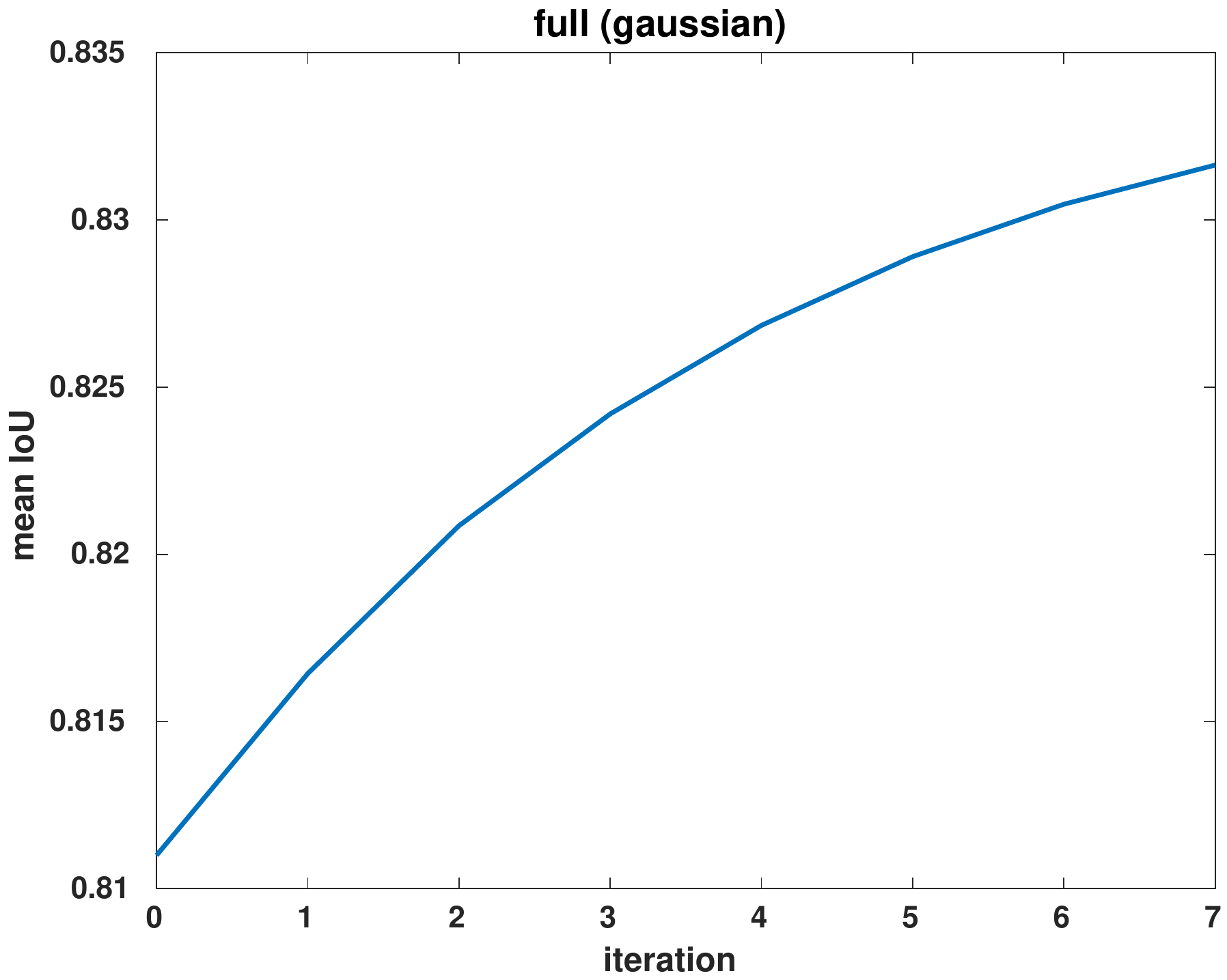}
   \includegraphics[width=0.45\linewidth]{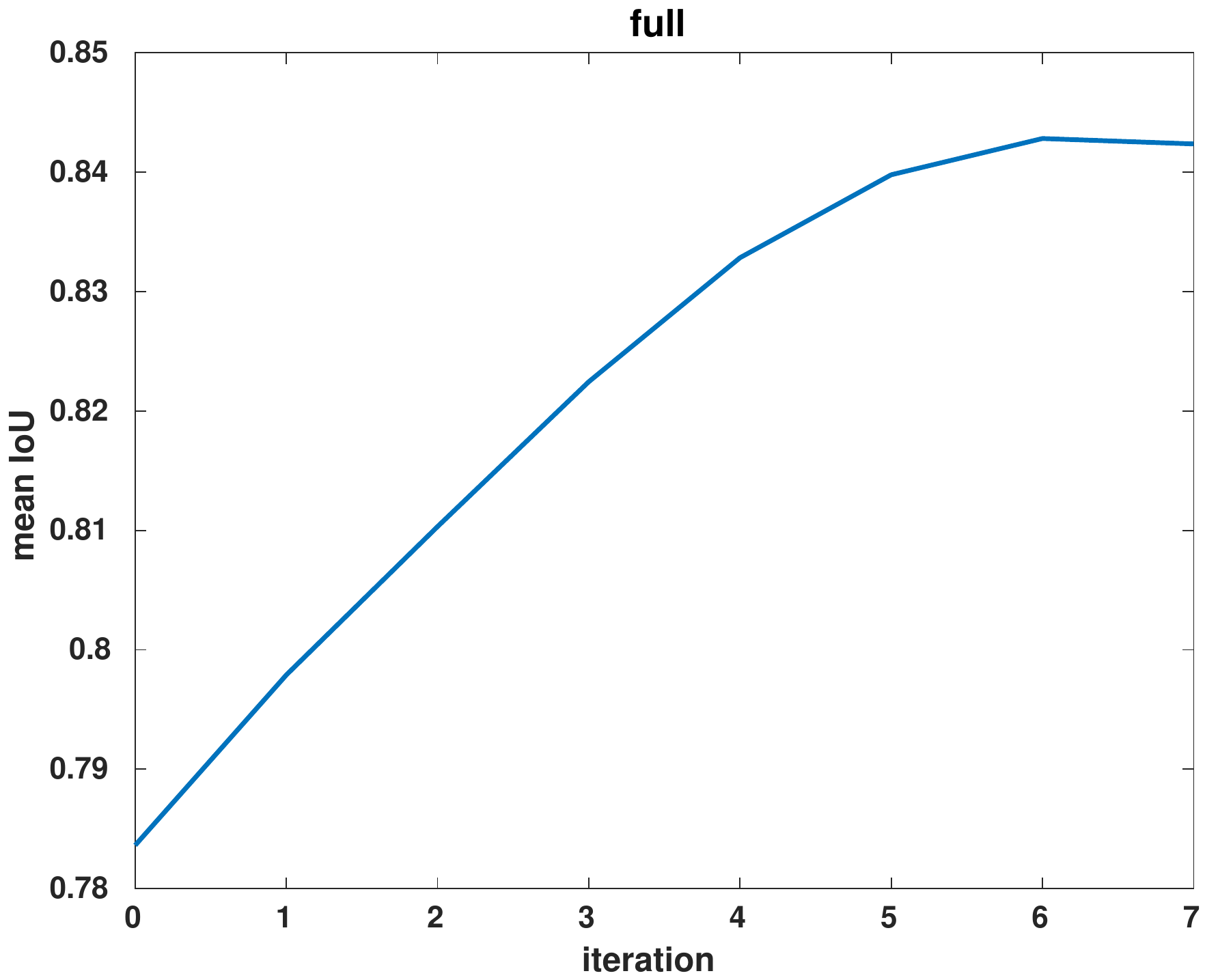}
   
\end{center}
   \caption{Mean IoU plotted as a function of iterations for the {\sc Weizmann Horse} test data. Note that, for the non-gaussian versions of the model, the performance of the layer does not increase after six iterations. The number of iterations were set to five during training.}
\label{fig:crfjac-supp}
\end{figure*}

Since the forward operation of the CRF-Grad layer performs gradient descent, we are interested in knowing how many iterations are needed to converge. In Fig.~\ref{fig:crfenergy-supp} we have plotted the CRF energy as a function of the number of iterations on the {\sc Weizmann Horse} data set for the different models. In addition Fig.~\ref{fig:crfjac-supp} shows the mean IoU as a function of the number of iterations for the {\sc Weizmann Horse} test data. Both of these results point to the fact that it is sufficient to run the model for about five iterations and that running it further would not increase the result considerably.

It also might be interesting to investigate the intermediate states $\bm{q}^t$. in Fig.~\ref{fig:intmed} the intermediate states of the CRF-Grad layers is shown. This figure gives a good indication on affect the CRF-Grad layer has. As can be seen in the figure, each step refines the segmentation slightly, removing spurious outlier pixels classified as horse, in addition to refining the boundaries slightly. 

Also included in the supplementary material is short movie, crfgd2\_mov.avi, showing how the CRF energy decreases with each step of our projected gradient descent algorithm. For this example a Potts model is used and the projected gradient descent solution is compared to the globally optimal solution obtained by graph cut.

\subsection*{Error Derivatives for the CRF-Grad layer}
In this section, we will explicitly formulate the error derivative necessary to train our deep structure model jointly. The notation used in the section is not very strict. Derivatives, gradients and jacobians are all referred to as derivatives. Denoting the output of our CRF-Grad layer $\bm{y}$ we need expressions for the derivatives $\frac{\partial \mathbf{y}}{\partial \mathbf{z}}$, where $\bm{z}$ is the output from the CNN and hence also the input to the CRF-Grad layer. In addition we need to calculate $\frac{\partial \mathbf{y}}{\partial w_u}$, $\frac{\partial \mathbf{y}}{\partial \mathbf{w}_{s}}$ and $\frac{\partial \mathbf{y}}{\partial \mathbf{w}_{b}}$ to be able to update the weights of the layer. To simplify the notation we abbreviate the update step by $\bm{q}^{t+1} = f(\bm{q}^{t},\bm{z},I,\bm{w})$. Note that the output $\bm{y} = \bm{q}^{T}$ where $T$ is the total number of iterations for the RNN. We have
\begin{align}
 \frac{\partial \bm{y}}{\partial w_{u}} &=& \frac{\partial \bm{y}}{\partial \bm{q}^T} \frac{\partial f(\mathbf{q}^{T-1})}{\partial w_{u}} + \ldots + \frac{\partial \bm{y}}{\partial \bm{q}^1} \frac{\partial f(\mathbf{q}^{0})}{\partial w_{u}} \\
 \frac{\partial \bm{y}}{\partial \bm{w}_{s}} &=& \frac{\partial \bm{y}}{\partial \bm{q}^T} \frac{\partial f(\mathbf{q}^{T-1})}{\partial \bm{w}_{s}} + \ldots + \frac{\partial \bm{y}}{\partial \bm{q}^1} \frac{\partial f(\mathbf{q}^{0})}{\partial \bm{w}_{s}} \\
 \frac{\partial \bm{y}}{\partial \bm{w}_{b}} &=& \frac{\partial \bm{y}}{\partial \bm{q}^T} \frac{\partial f(\mathbf{q}^{T-1})}{\partial \bm{w}_{b}} + \ldots + \frac{\partial \bm{y}}{\partial \bm{q}^1} \frac{\partial f(\mathbf{q}^{0})}{\partial \bm{w}_{b}} \\
 \frac{\partial \bm{y}}{\partial \bm{z}} &=& \frac{\partial \bm{y}}{\partial \bm{q}^0} \frac{\partial \bm{q}^0}{\partial \bm{z}} + \frac{\partial \bm{y}}{\partial \bm{\psi}_u} \frac{\partial \bm{\psi}_u}{\partial \bm{z}},
\end{align}
where $\bm{\psi}_u$ denote the unary part of the CRF energy function. Note that
\begin{equation}
 \frac{\partial \bm{y}}{\partial \bm{q}^{t-1}} = \frac{\partial \bm{y}}{\partial \bm{q}^{t}} \frac{\partial f(\mathbf{q}^{t-1})}{\partial \bm{q}^{t-1}})
\end{equation}

To be able to calculate these we need the derivatives of the function f with respect to $\mathbf{q}^{t}$, $w_{u}$, $\bm{w}_s$ and $\bm{w}_b$. We denote the spatial and bilateral filtering operations as $\bm{\psi_{s}} * \bm{q}^t$ and $\bm{\psi_{b}} * \bm{q}^t$ respectively. An update step can then be written as 
\begin{align*}
    \bm{q}^{t+1} &= \text{Proj}_{\triangle^L}(\mathbf{q^{t}} - \gamma (\bm{\psi}_u + \bm{\psi_{s}} * \bm{q}^t + \bm{\psi_{b}} * \bm{q}^t)) \\
    &= \text{Proj}_{\triangle^L}(\mathbf{\tilde{q}^{t+1}}).
\end{align*}
And the aforementioned derivatives become
\begin{align}
\begin{split}
\frac{\partial f}{\partial \mathbf{q}^{t}} =  &\text{Proj}'_{\triangle^L}  ( \mathbf{\tilde{q}}^{t+1}) \\
&\cdot \left(\bm{1} - \gamma  ( \bm{\psi}_u + \bm{\psi}_{s} * \mathbf{q^{t}} + \bm{\psi}_{b} * \mathbf{q^{t}}) \right ) \\
& \cdot \left ( \frac{\partial(\bm{\psi}_{s} * \mathbf{q^{t}})}{\partial \mathbf{q^{t}} } +  \frac{\partial(\bm{\psi}_{b} * \mathbf{q^{t}})}{\partial \mathbf{q^{t}} } \right ),  
\end{split}
\end{align}
for $\bm{q}^t$, and for the weights

\begin{align}
\begin{split}
&\frac{\partial f}{\partial w_u} =  \text{Proj}'_{\triangle^L}  ( \mathbf{\tilde{q}}^{t+1}) \cdot \\
&\left(- \gamma  ( \bm{\psi}_u + \bm{\psi}_{s} * \mathbf{q^{t}} + \bm{\psi}_{b} * \mathbf{q^{t}} \right ) \cdot \frac{\partial \bm{\psi}_{u}}{\partial w_u },  
\end{split}
\end{align}

\begin{align}
\begin{split}
&\frac{\partial f}{\partial \bm{w}_s} =  \text{Proj}'_{\triangle^L}  ( \mathbf{\tilde{q}}^{t+1}) \cdot \\
&\left(- \gamma  ( \bm{\psi}_u + \bm{\psi}_{s} * \mathbf{q^{t}} + \bm{\psi}_{b} * \mathbf{q^{t}} \right ) \cdot \frac{\partial(\bm{\psi}_{s} * \mathbf{q^{t}})}{\partial \bm{w}_{s} } ,  
\end{split}
\end{align}

\begin{align}
\begin{split}
&\frac{\partial f}{\partial \bm{w}_b} =  \text{Proj}'_{\triangle^L}  ( \mathbf{\tilde{q}}^{t+1}) \cdot \\
&\left(- \gamma  ( \bm{\psi}_u + \bm{\psi}_{s} * \mathbf{q^{t}} + \bm{\psi}_{b} * \mathbf{q^{t}} \right ) \cdot \frac{\partial(\bm{\psi}_{b} * \mathbf{q^{t}})}{\partial \bm{w}_{b} } ,  
\end{split}
\end{align}
Note that $\frac{\partial(\bm{\psi}_{s} * \mathbf{q^{t}})}{\partial \bm{q}^{t} } $, $\frac{\partial(\bm{\psi}_{b} * \mathbf{q^{t}})}{\partial \bm{q}^{t} }$, $\frac{\partial(\bm{\psi}_{s} * \mathbf{q^{t}})}{\partial \bm{w}_{s} } $ and $\frac{\partial(\bm{\psi}_{b} * \mathbf{q^{t}})}{\partial \bm{w}_{b} }$ can be calculated using the backward routines for a standard convolutional layer and bilateral filtering layer described in the main paper.

\begin{table*}
\begin{center}
\setlength\tabcolsep{3pt}
\begin{tabular}{c|cccccccccc}
 & \rotatebox{90}{road}&\rotatebox{90}{sidewalk}&\rotatebox{90}{building}&\rotatebox{90}{wall}&\rotatebox{90}{fence}&\rotatebox{90}{pole}&\rotatebox{90}{traffic light}&\rotatebox{90}{traffic sign}&\rotatebox{90}{vegetation}&\rotatebox{90}{terrain} \\ \hline
LRR-4x & \textbf{97.9}&\textbf{81.5}&\textbf{91.4}&50.5&52.7&59.4&66.8&72.7&92.5&\textbf{70.1} \\
CRF-Grad & \textbf{97.9}&81.3&\textbf{91.4}&\textbf{51.4}&\textbf{52.9}&\textbf{60.65}&\textbf{68.1}&\textbf{73.5}&\textbf{92.6}&69.1 \\
\hline \hline
& \rotatebox{90}{sky}&\rotatebox{90}{person}&\rotatebox{90}{rider}&\rotatebox{90}{car}&\rotatebox{90}{truck}&\rotatebox{90}{bus}&\rotatebox{90}{train}&\rotatebox{90}{motorcycle}&\rotatebox{90}{bicycle} \\

LRR-4x & \textbf{95}&81.3&\textbf{60.1}&\textbf{94.3}&\textbf{51.2}&67.7&54.6&55.6&69.6 \\
CRF-Grad &\textbf{95}&\textbf{81.6}&59.5&94.1&45.5&\textbf{68.4}&\textbf{56}&\textbf{56}&\textbf{70.2} \\
\hline
\end{tabular}
\end{center}
\caption{Class-wise results compairing our model to the baseline, LRR. The results are presented as IoU (\%).}
\label{tab:cityscapes_classwise}
\end{table*}

\subsection*{Additional Results}
\begin{figure*}[t]
\begin{center}
   \includegraphics[width=0.19\linewidth]{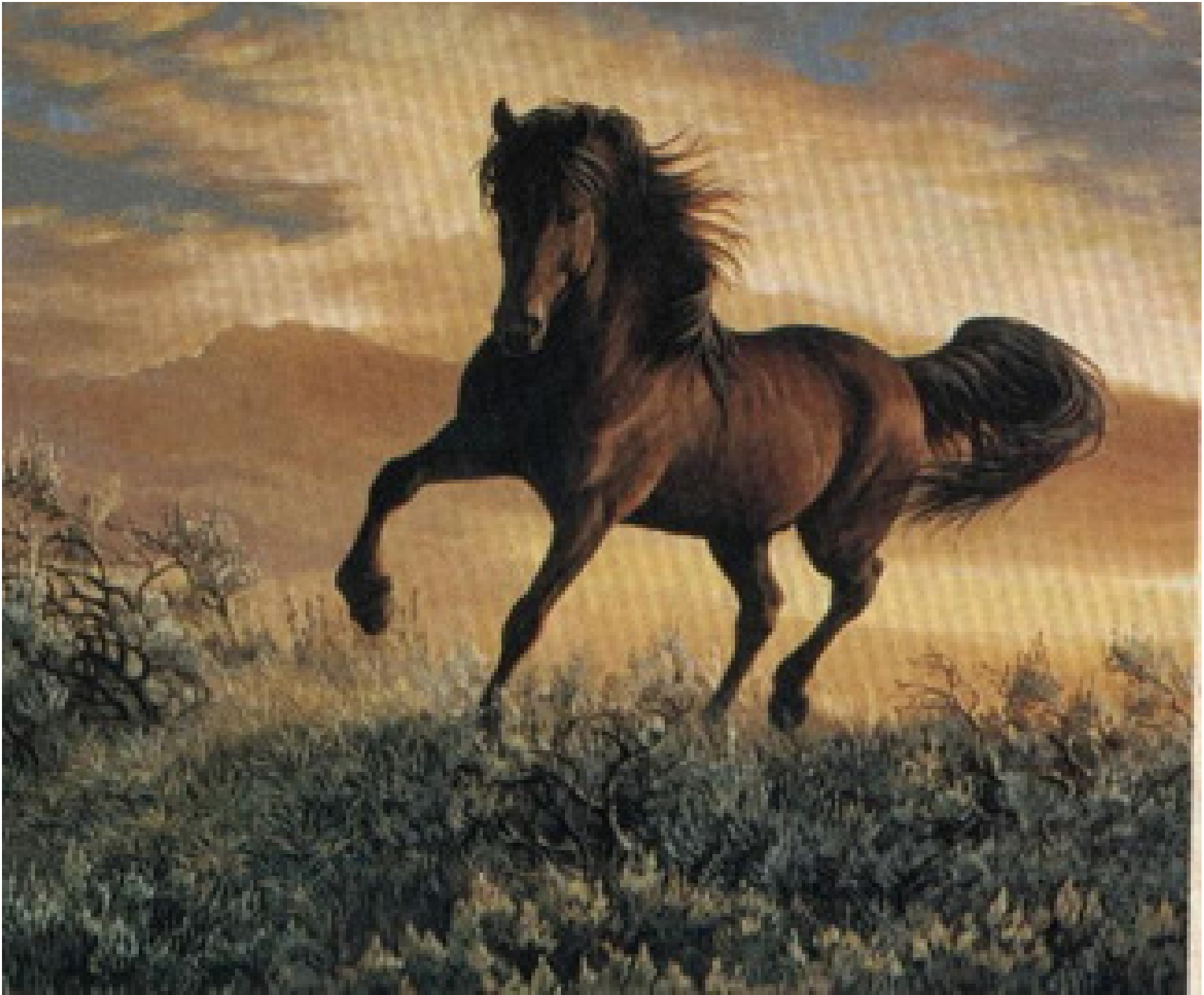}
   \includegraphics[width=0.19\linewidth]{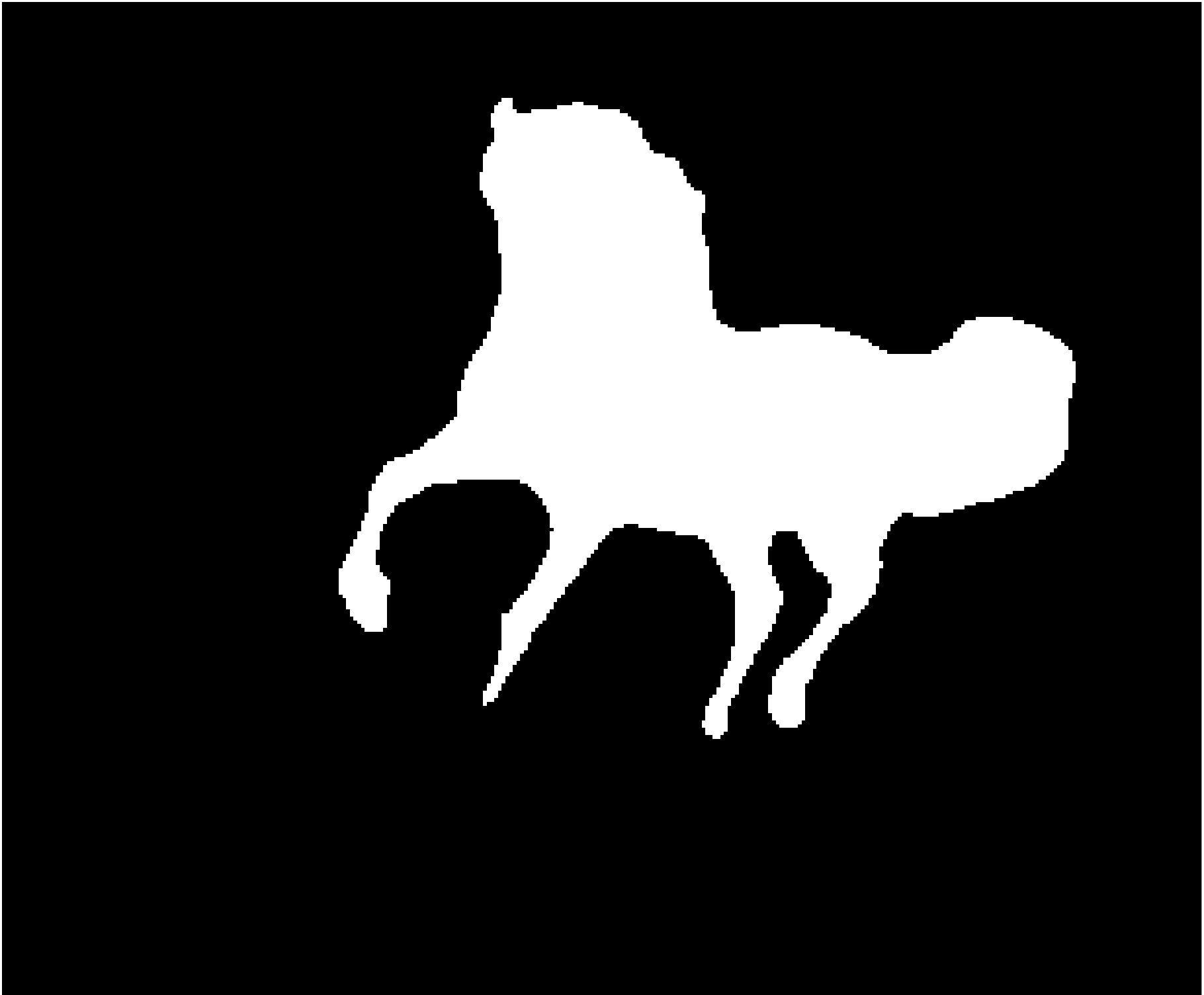}
   \includegraphics[width=0.19\linewidth]{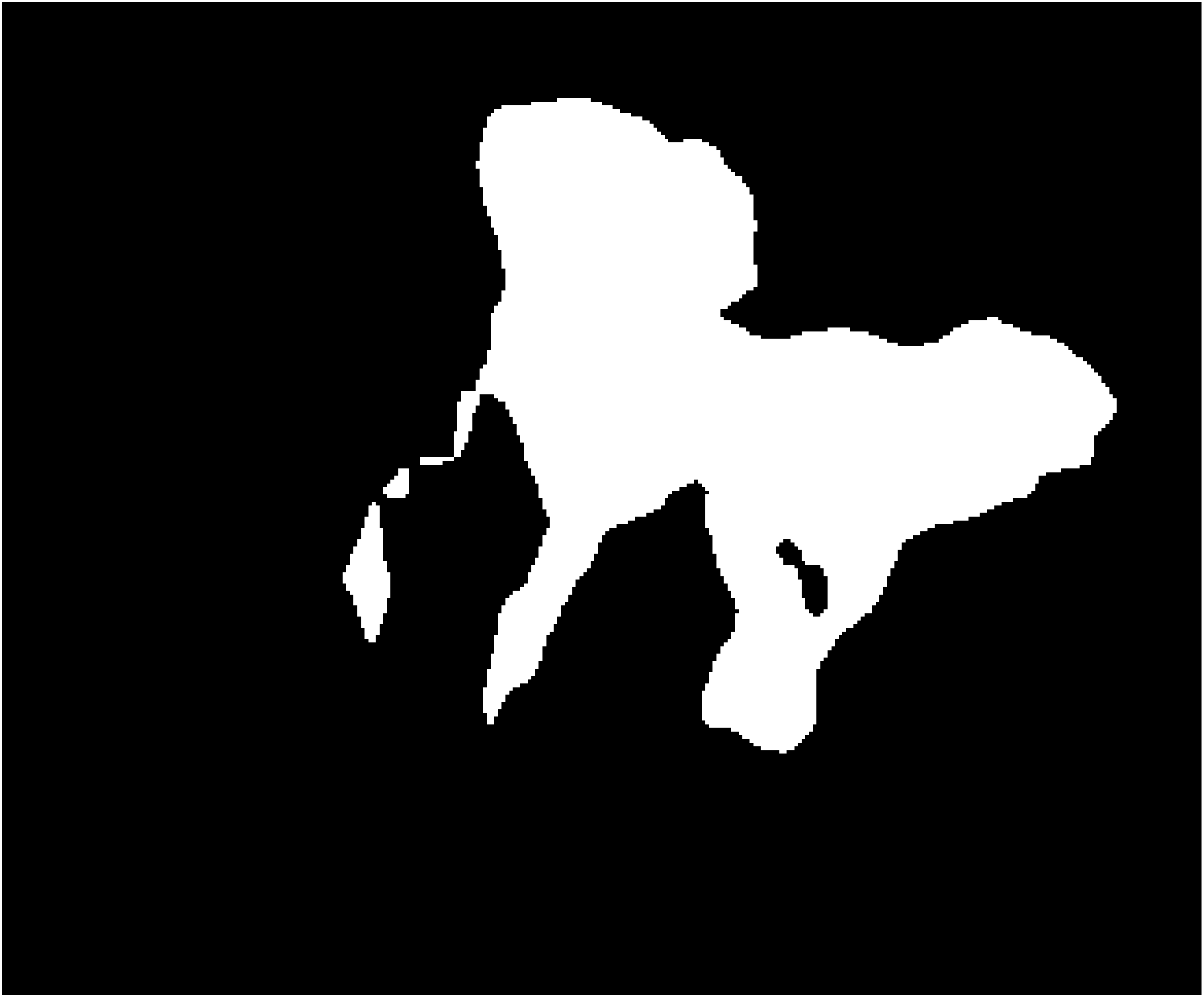}
   \includegraphics[width=0.19\linewidth]{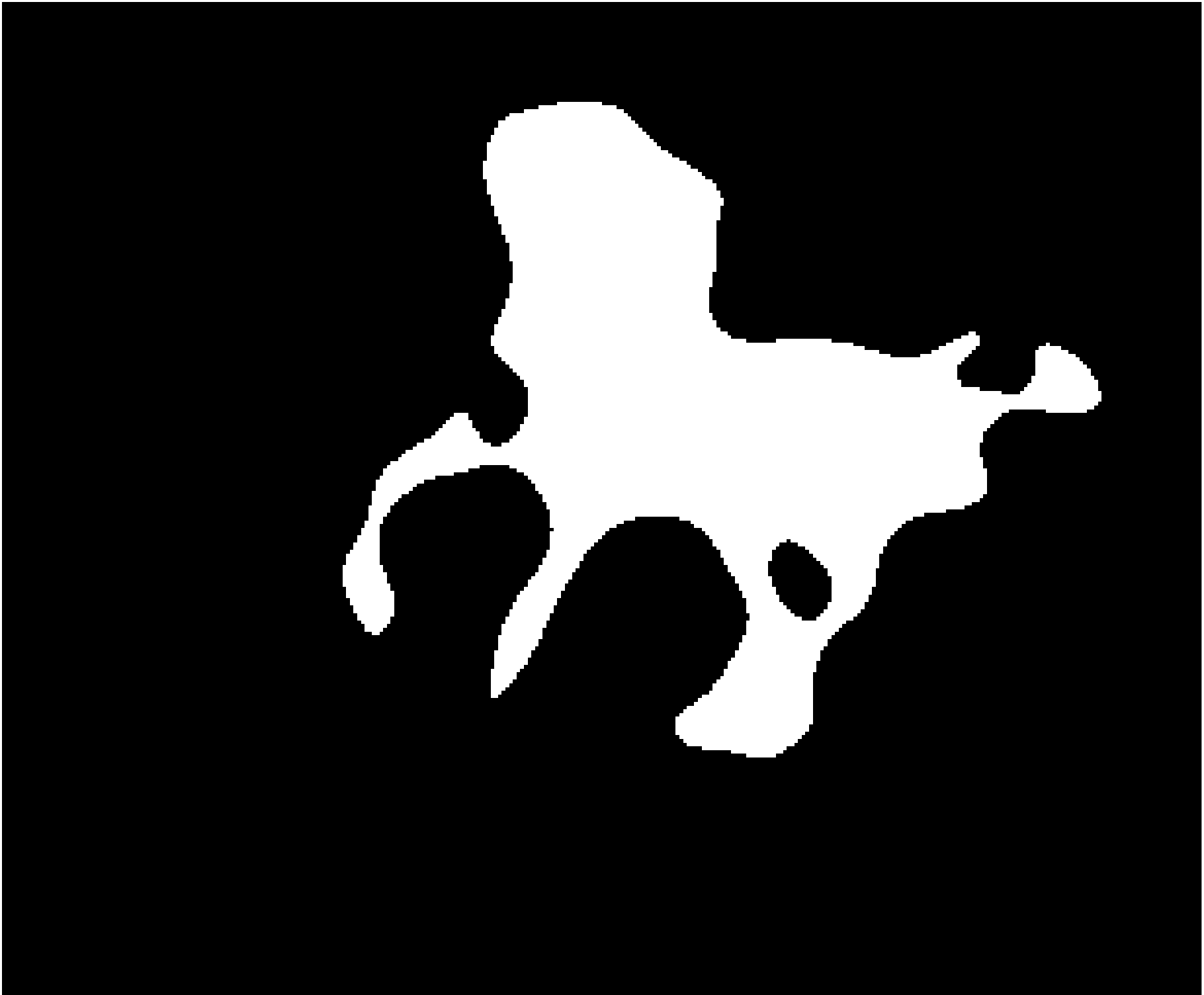}
   \includegraphics[width=0.19\linewidth]{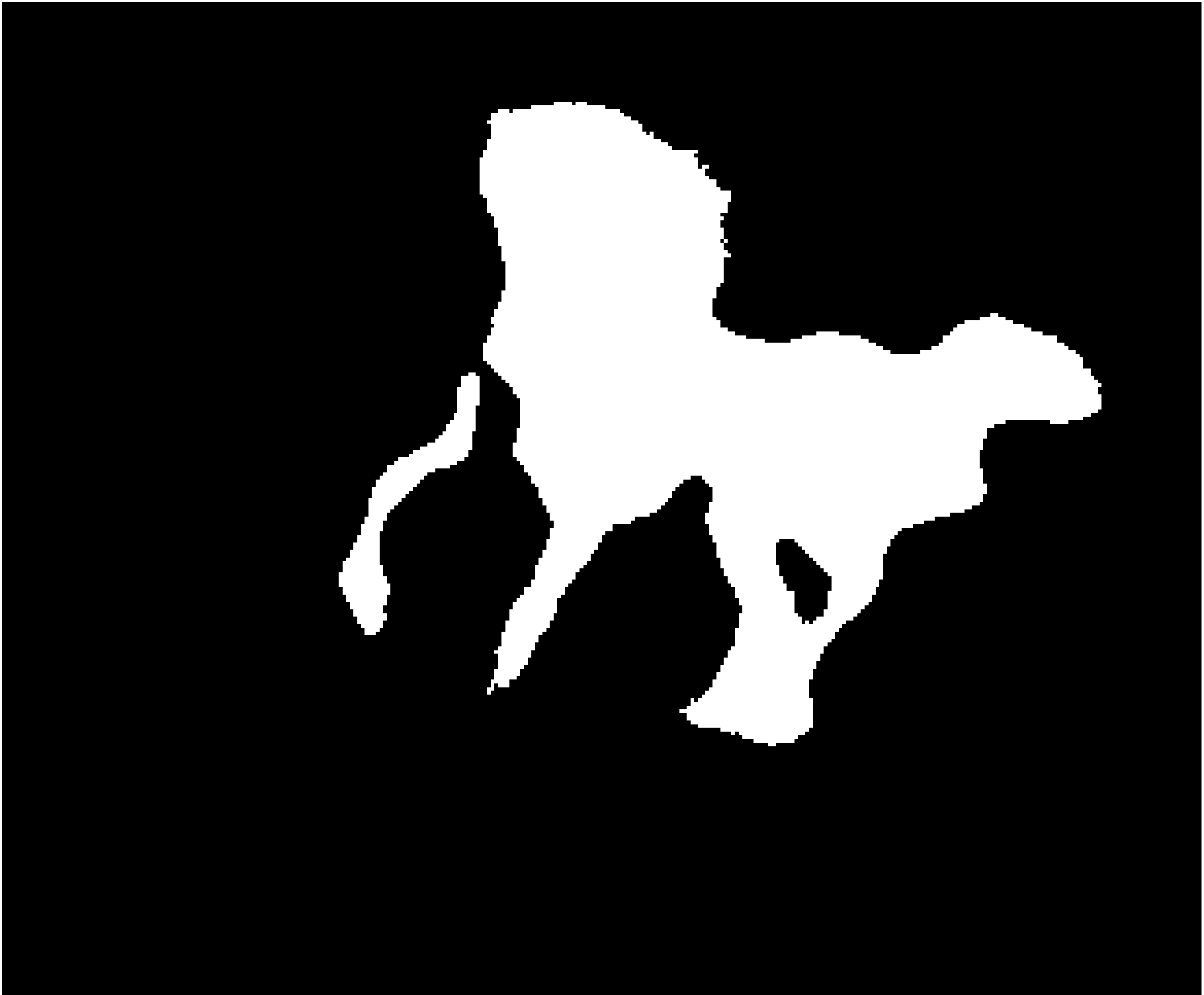}
   
   \includegraphics[width=0.19\linewidth]{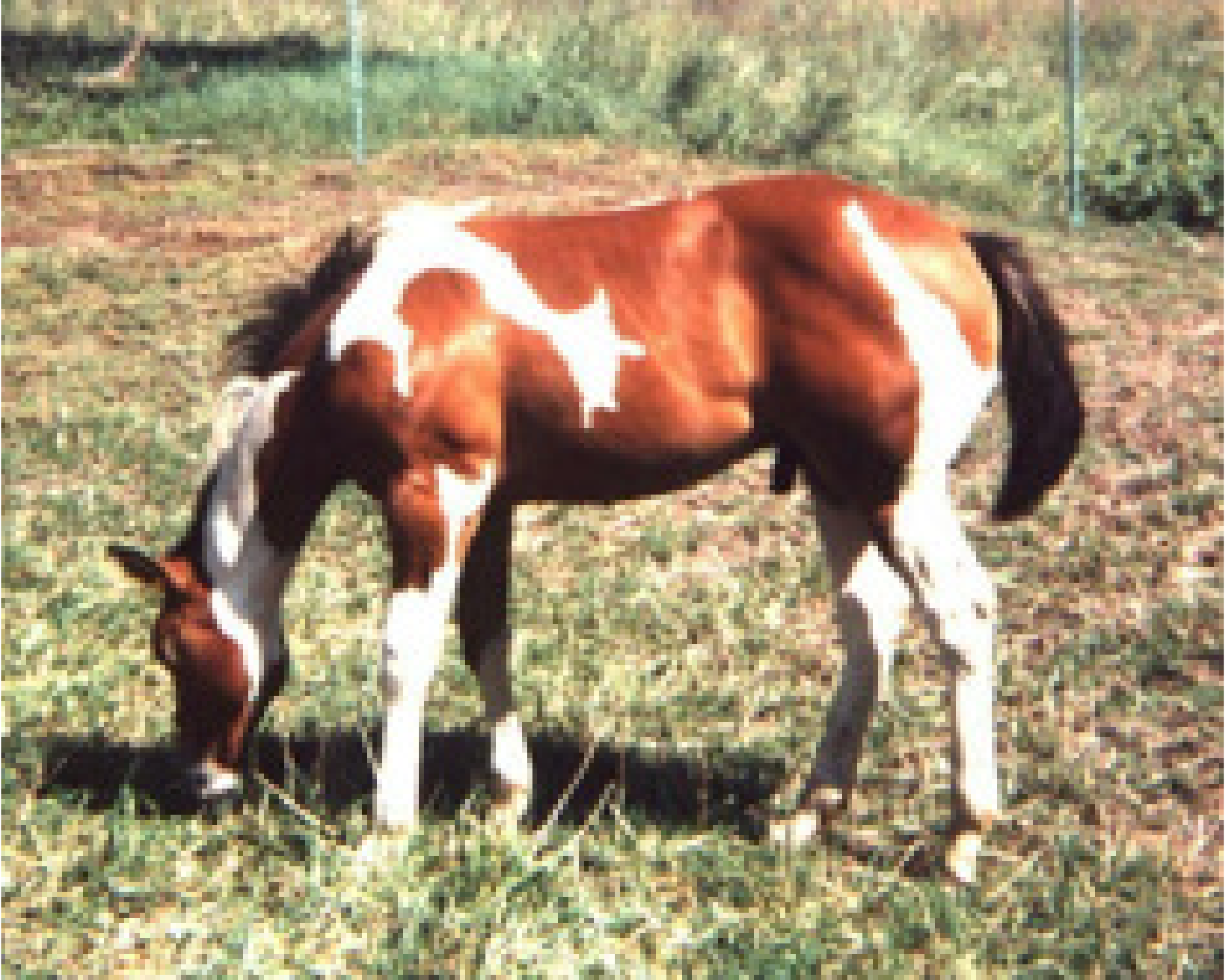}
   \includegraphics[width=0.19\linewidth]{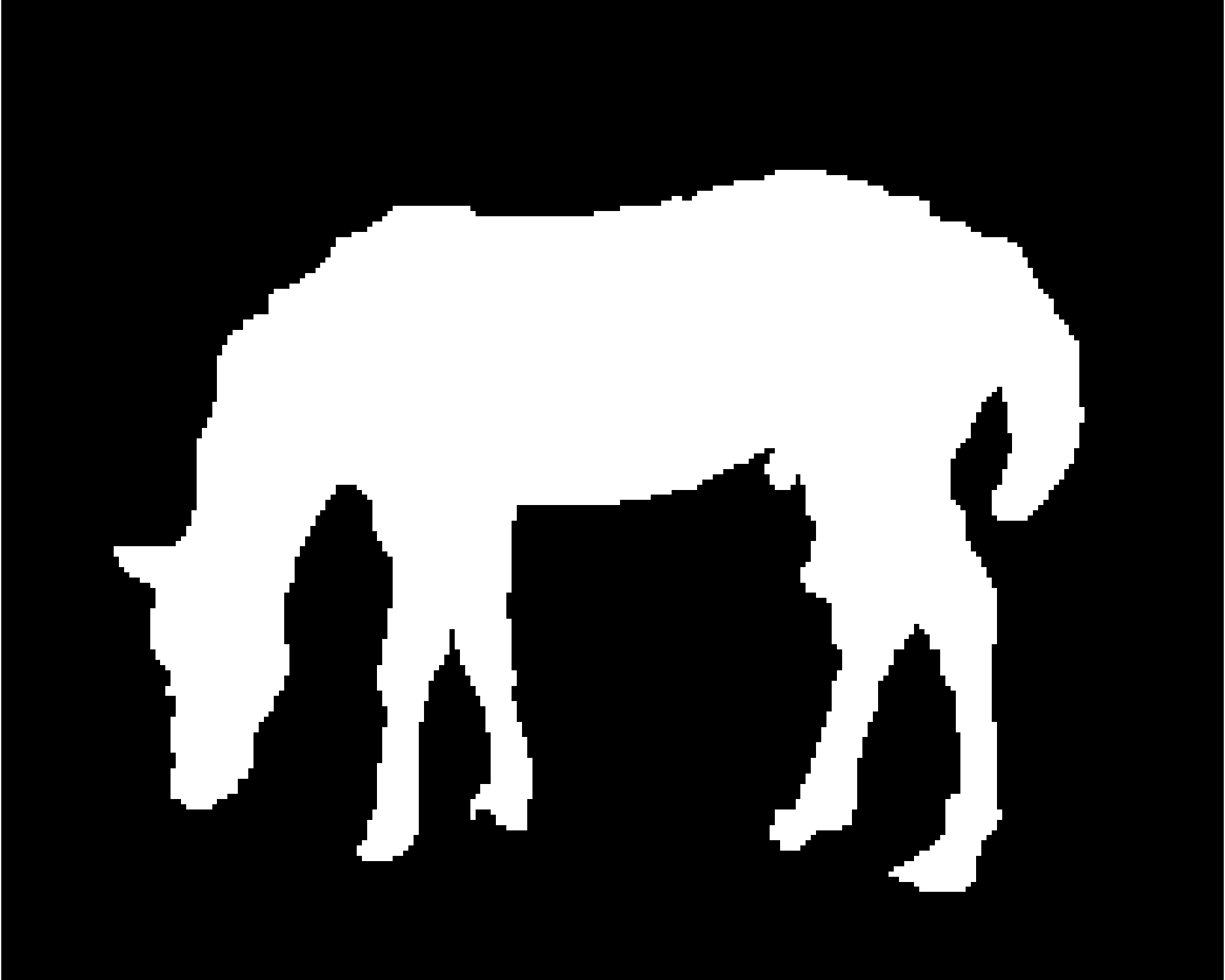}
   \includegraphics[width=0.19\linewidth]{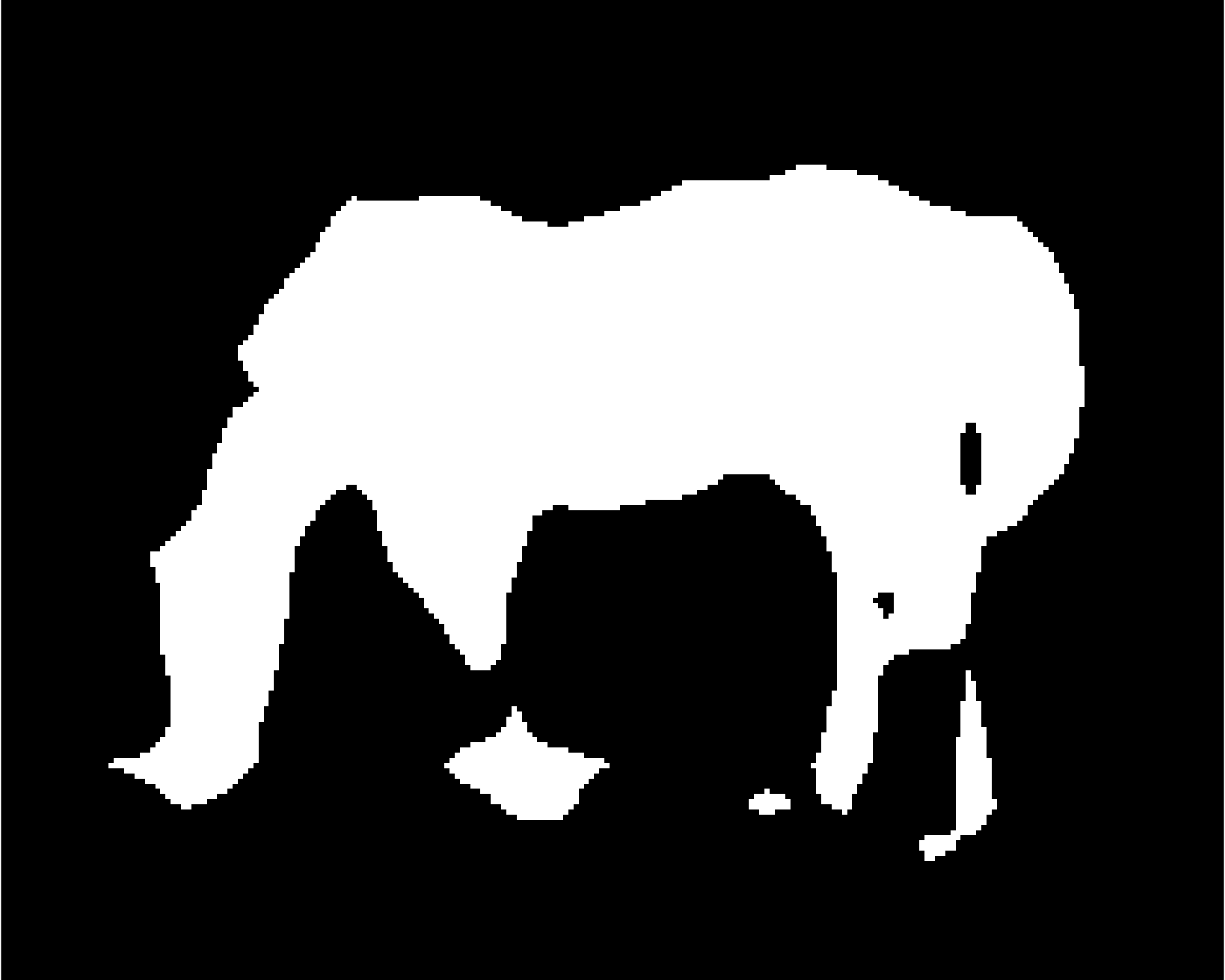}
   \includegraphics[width=0.19\linewidth]{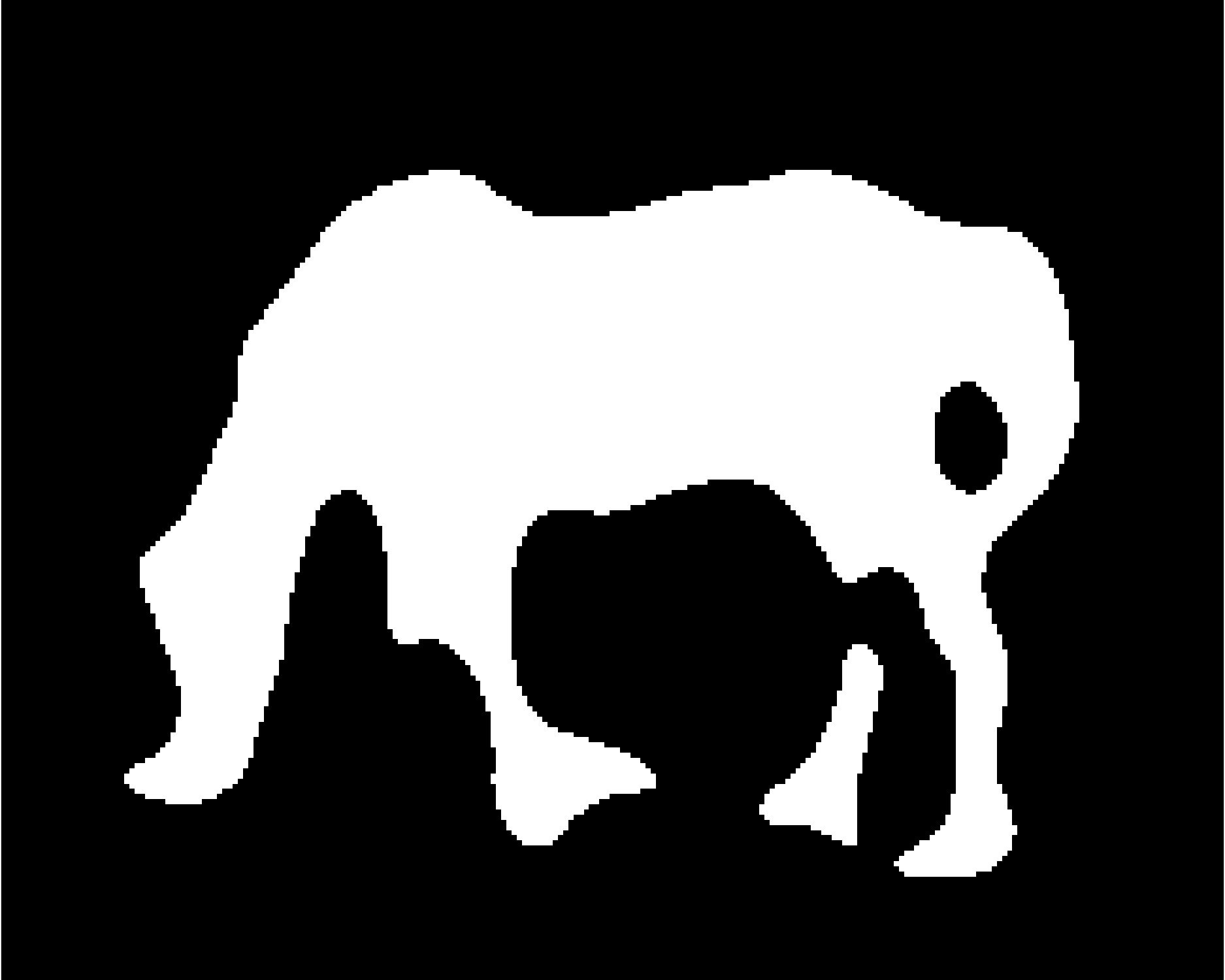}
   \includegraphics[width=0.19\linewidth]{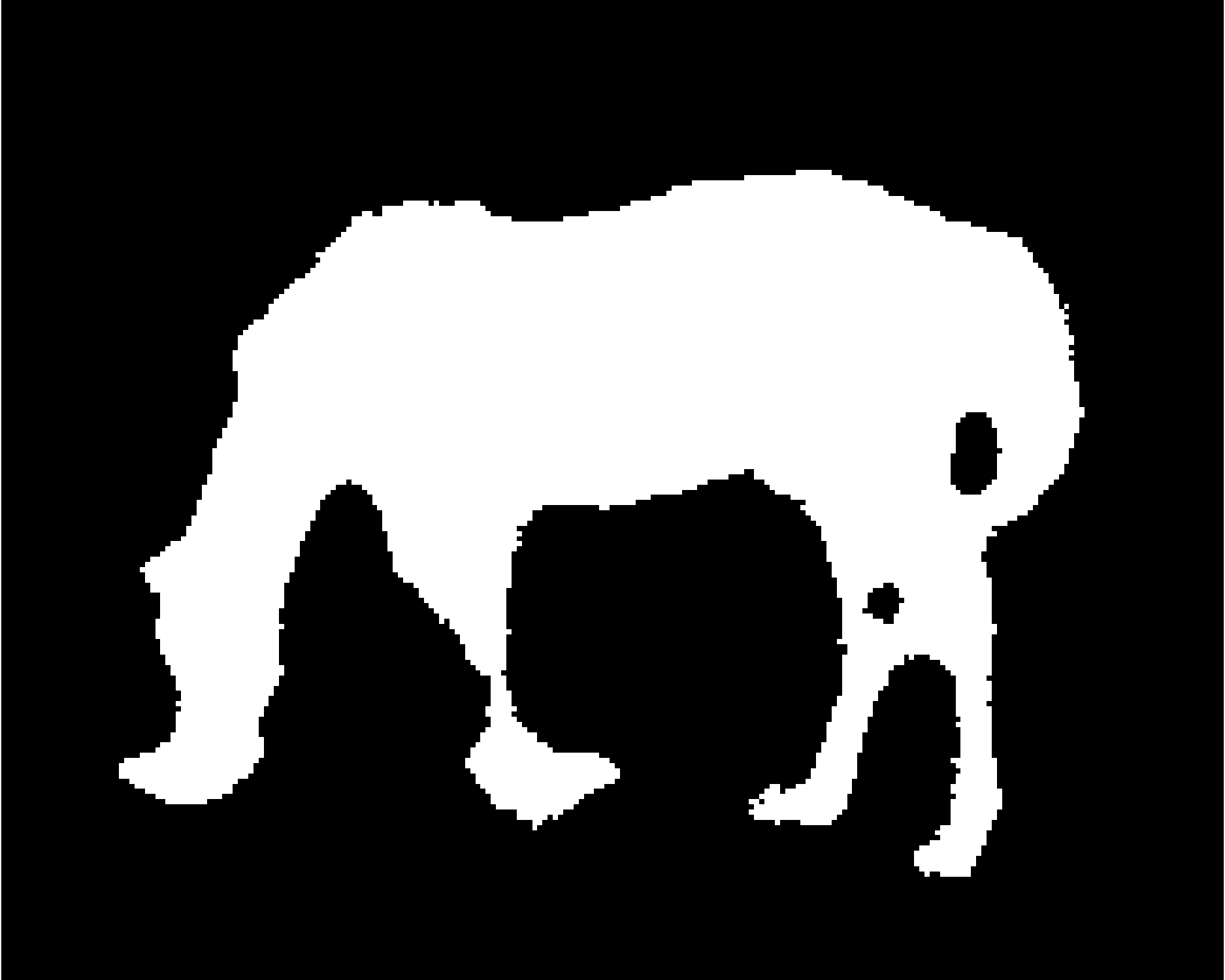}
   
   \includegraphics[width=0.19\linewidth]{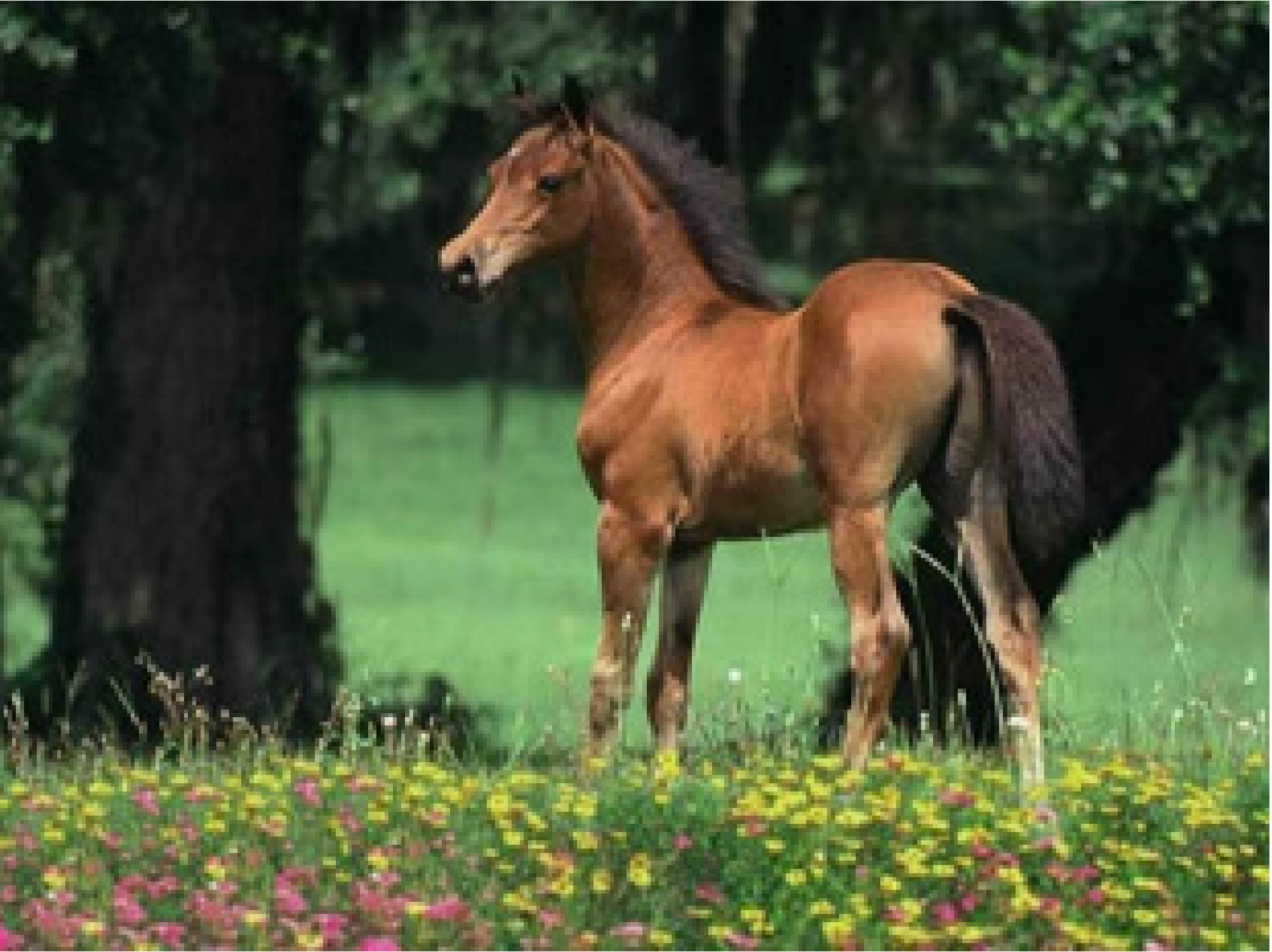}
   \includegraphics[width=0.19\linewidth]{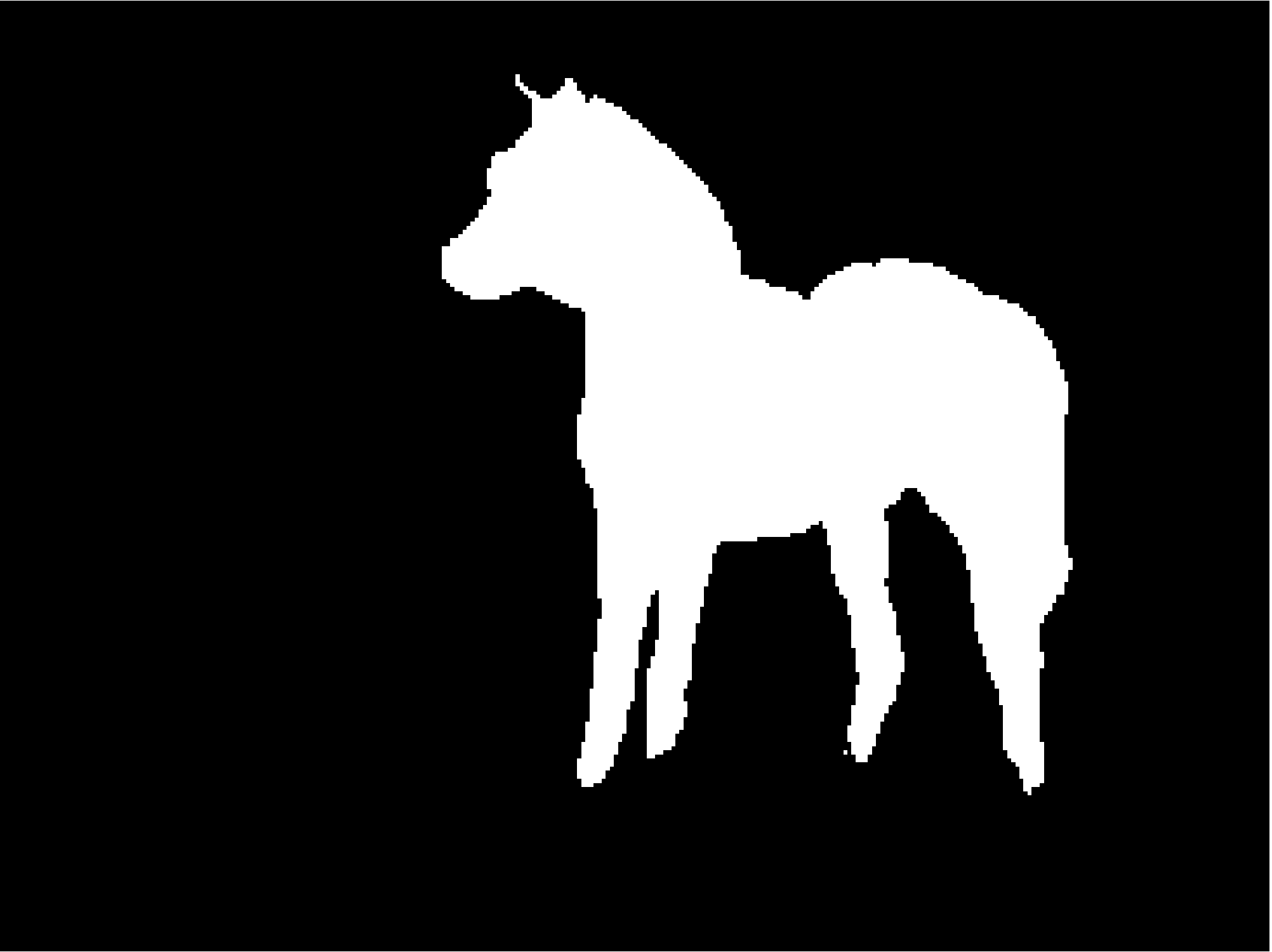}
   \includegraphics[width=0.19\linewidth]{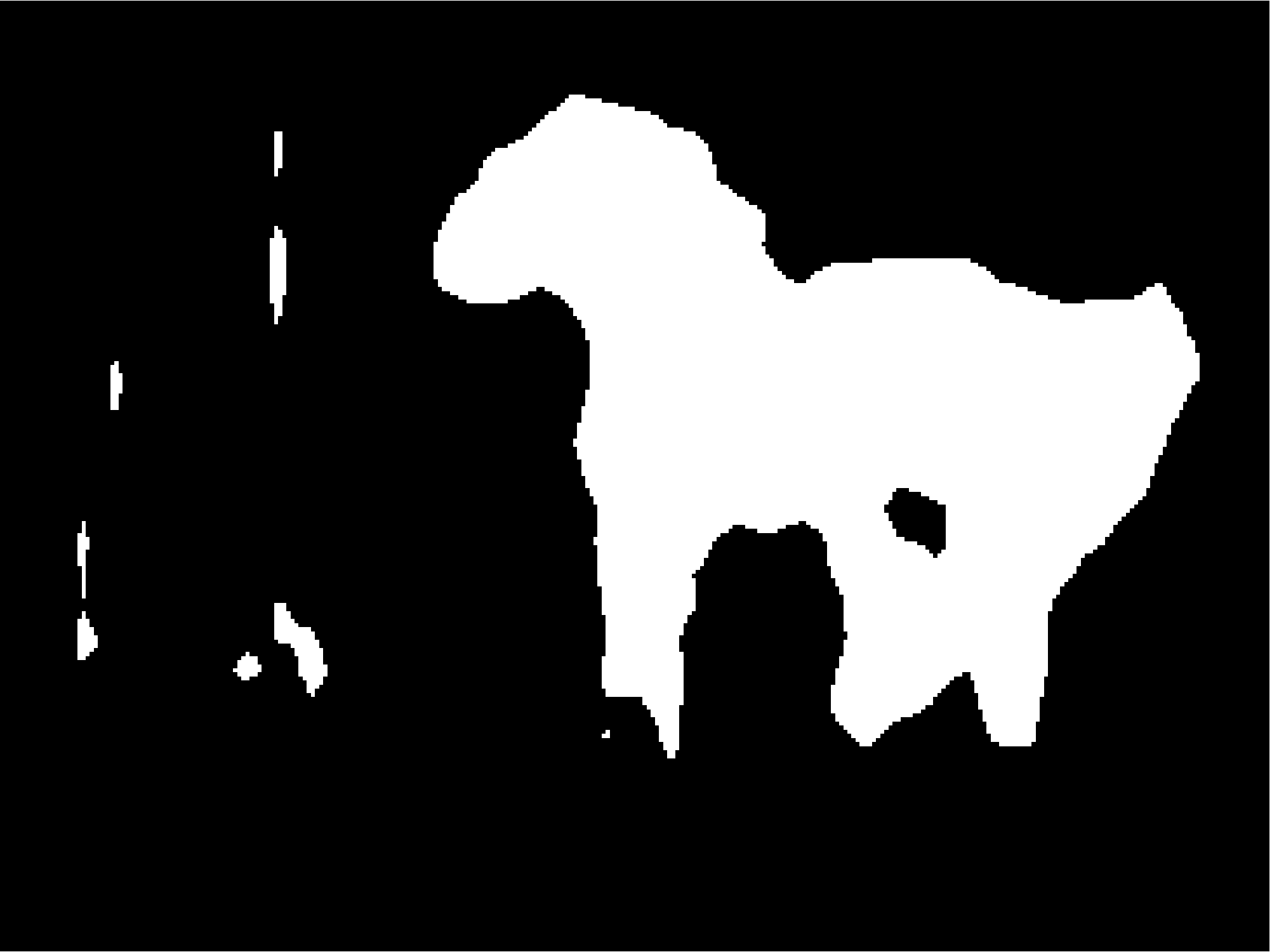}
   \includegraphics[width=0.19\linewidth]{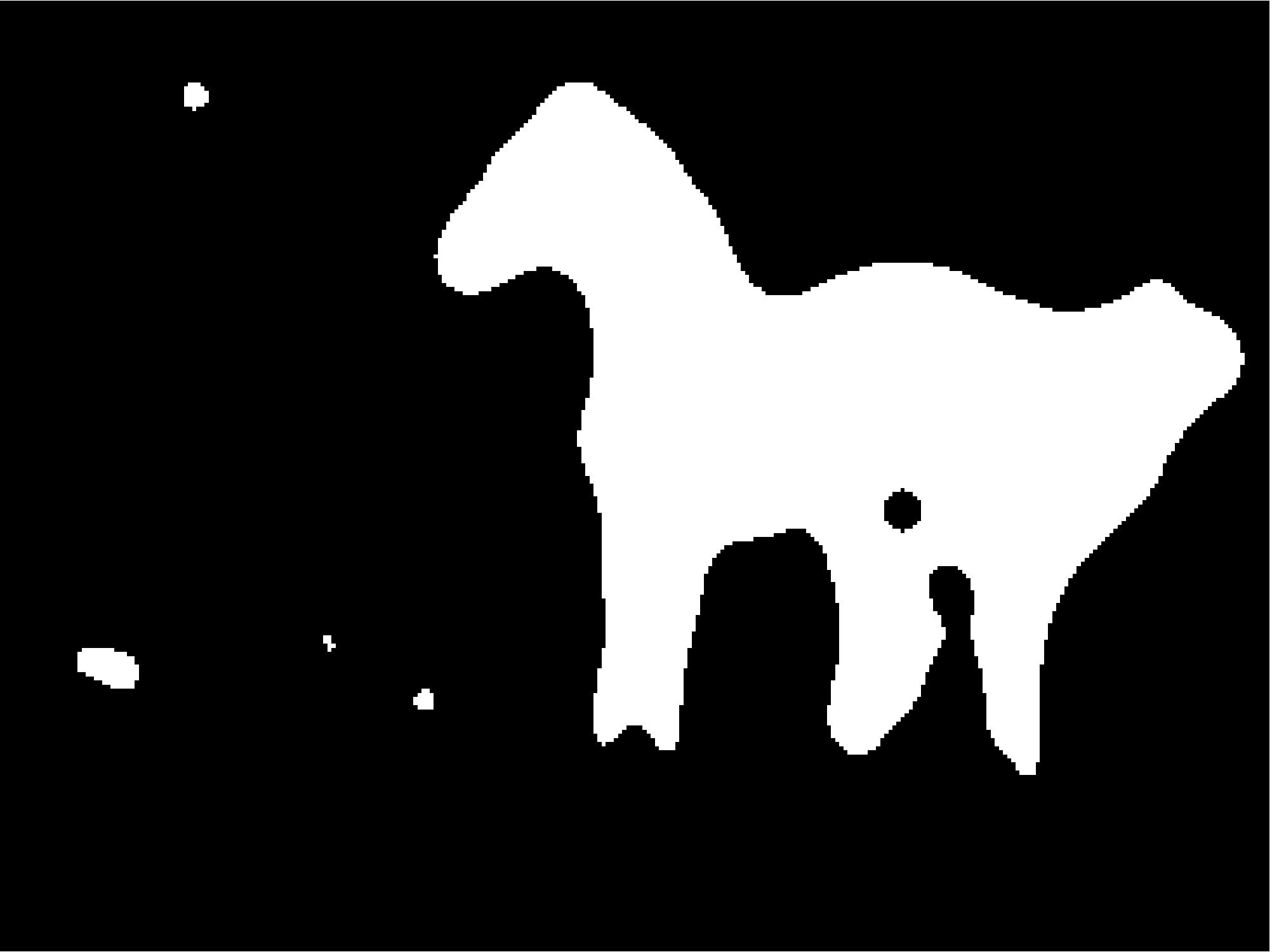}
   \includegraphics[width=0.19\linewidth]{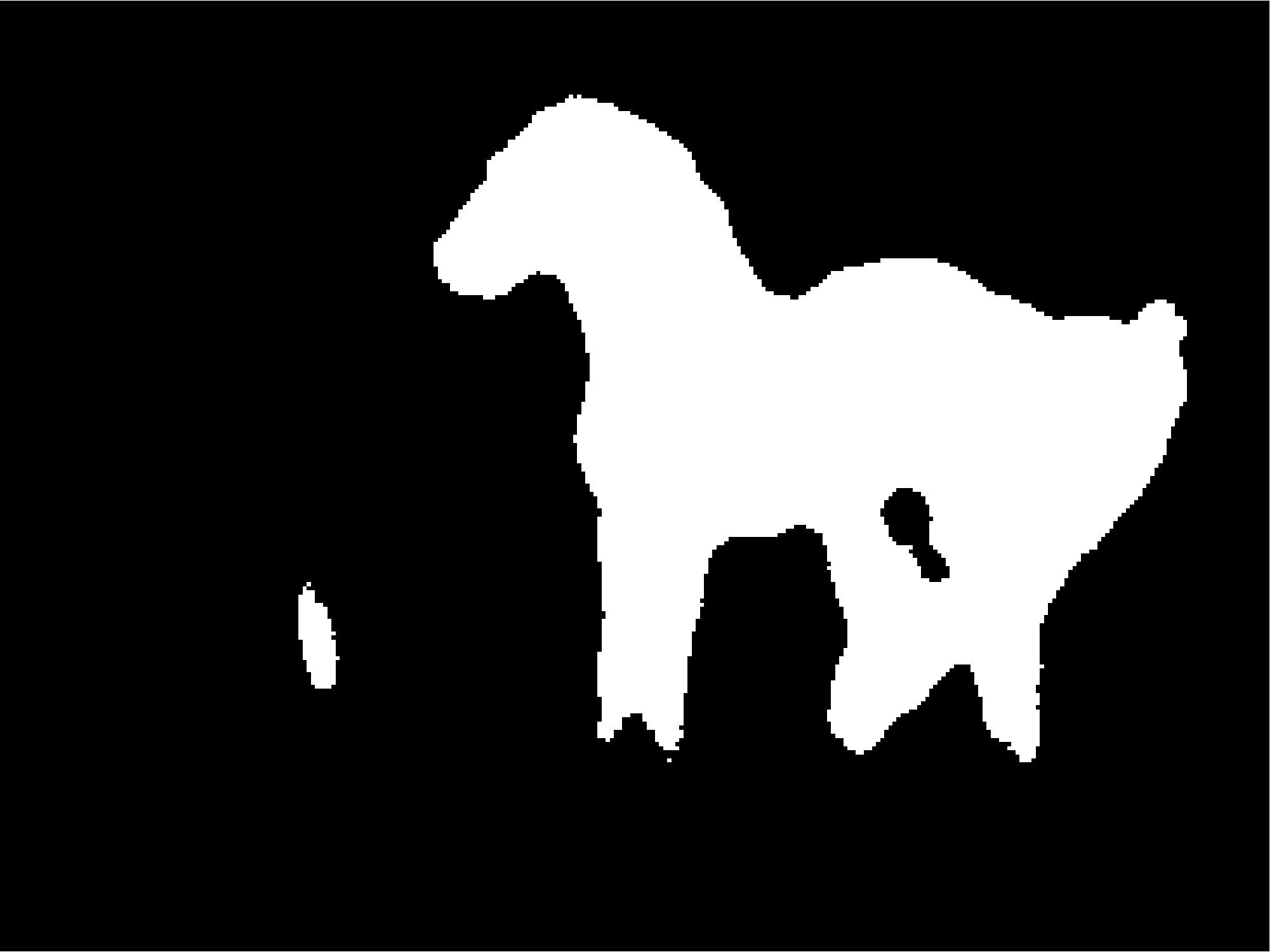}
   
   \includegraphics[width=0.19\linewidth]{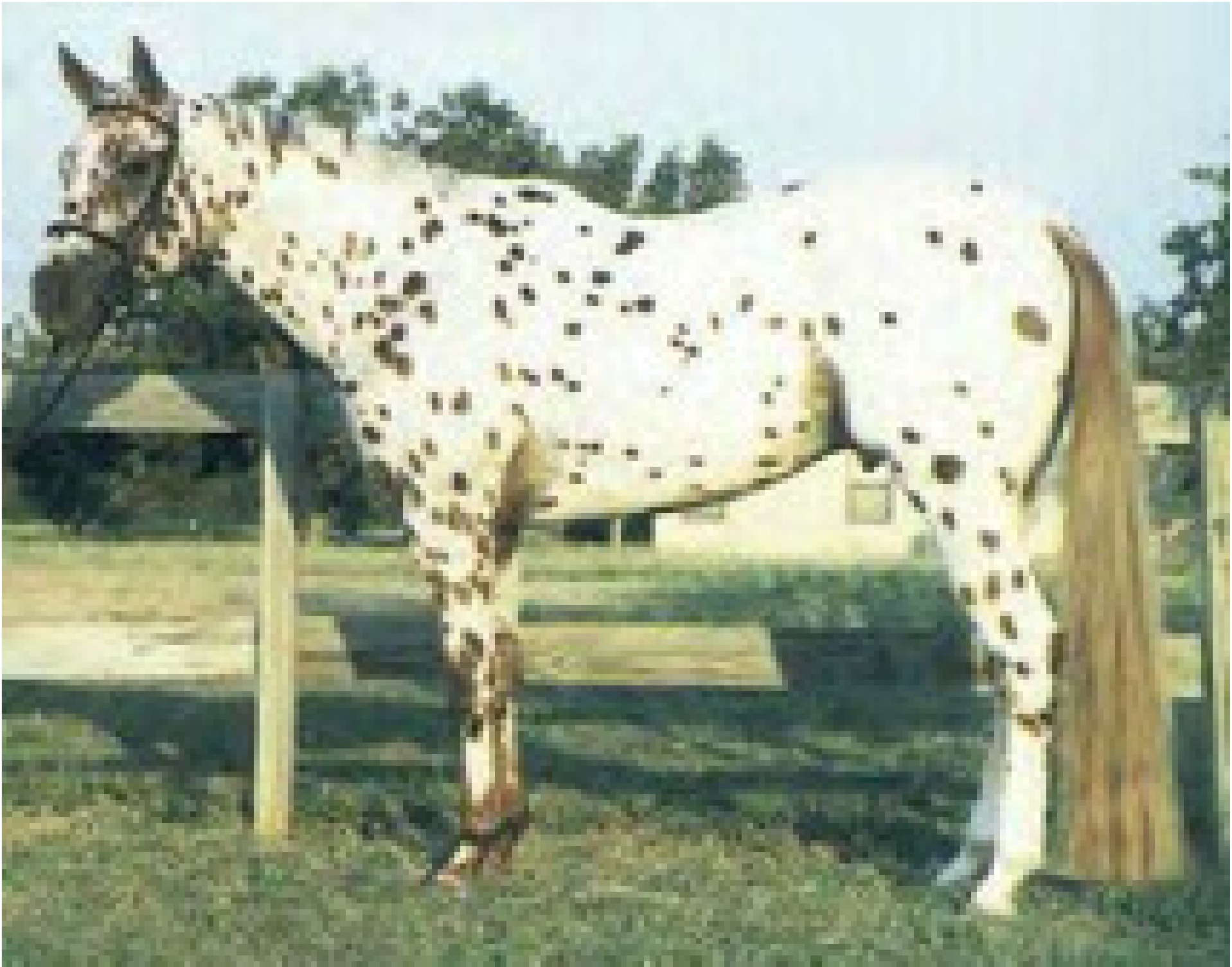}
   \includegraphics[width=0.19\linewidth]{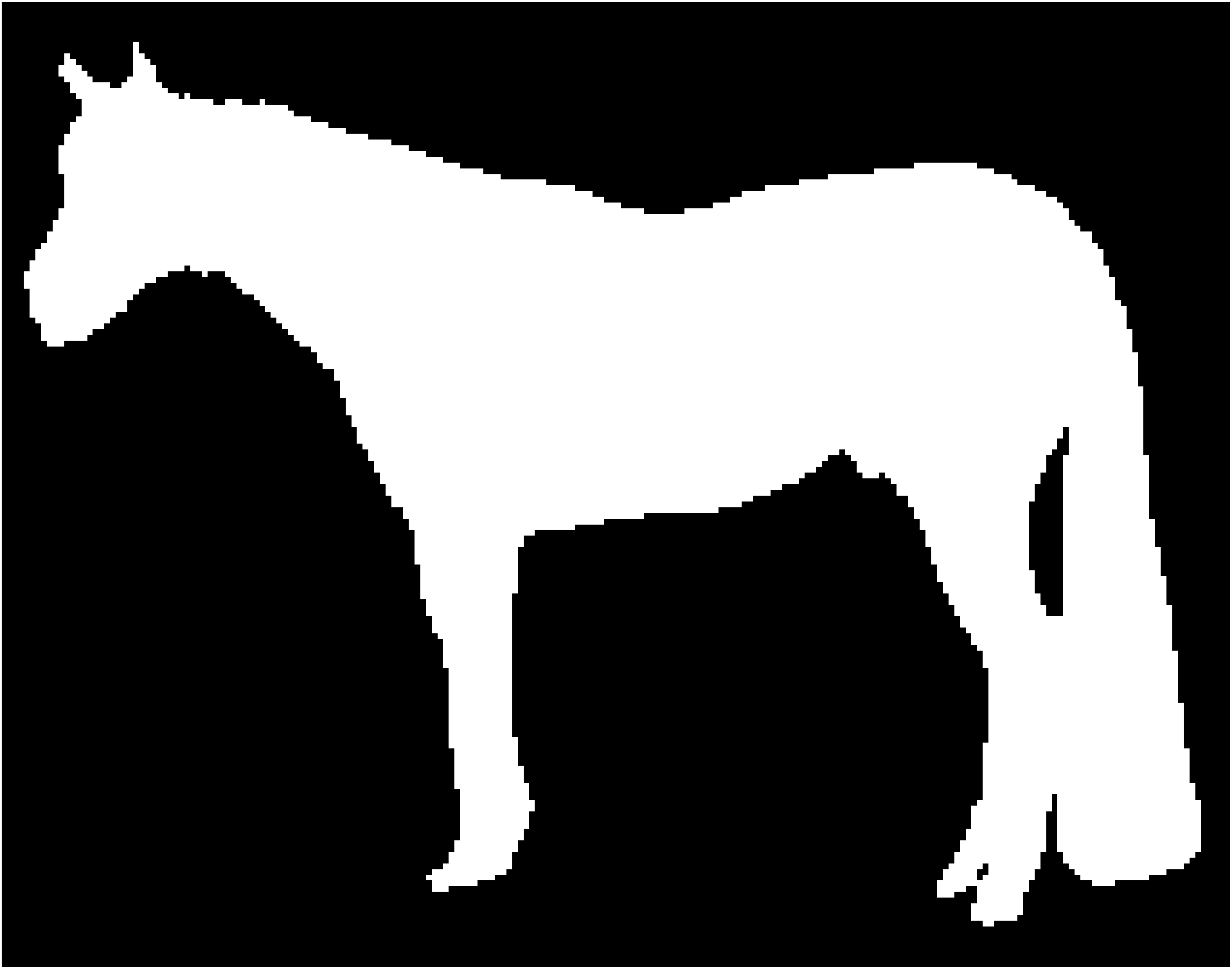}
   \includegraphics[width=0.19\linewidth]{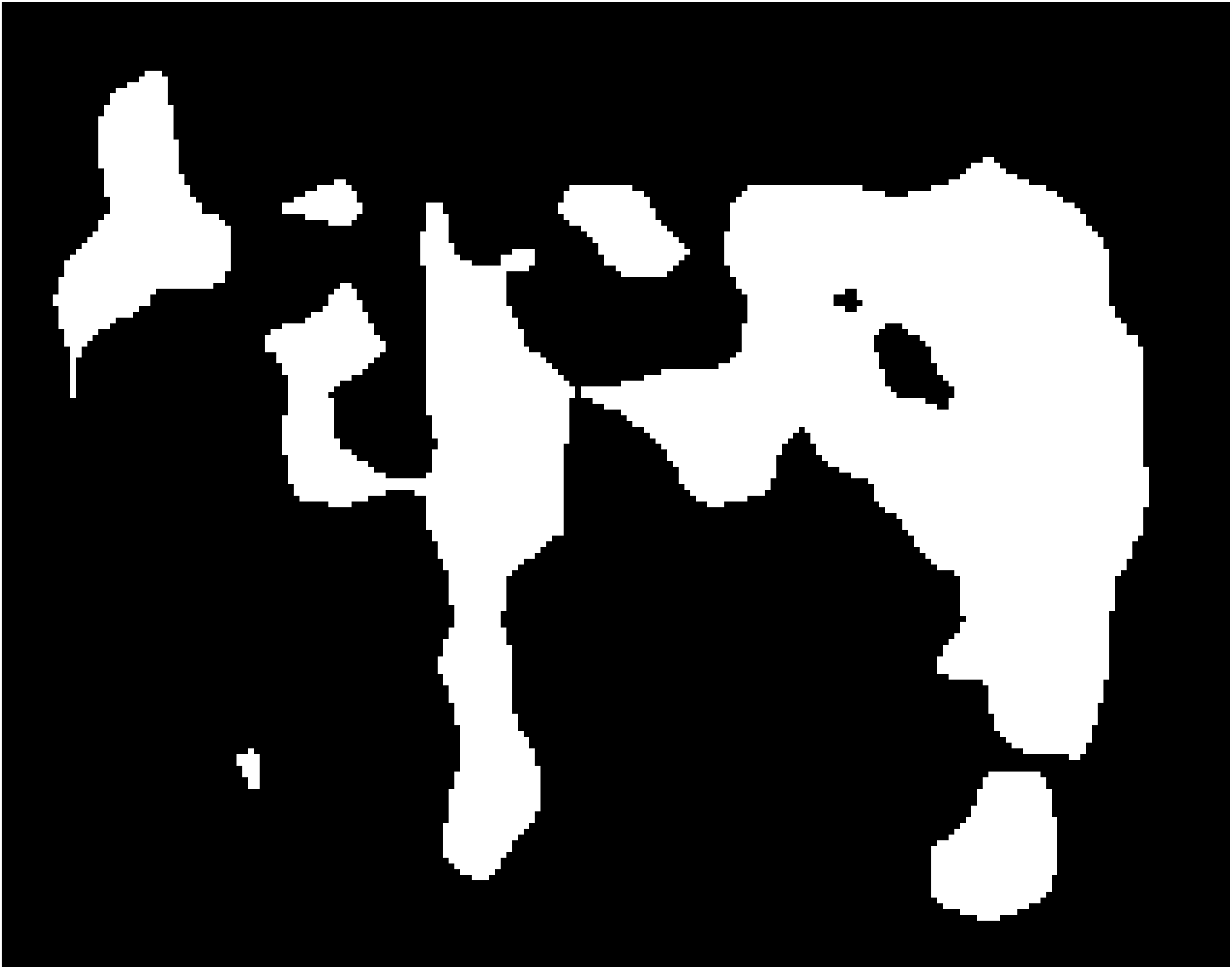}
   \includegraphics[width=0.19\linewidth]{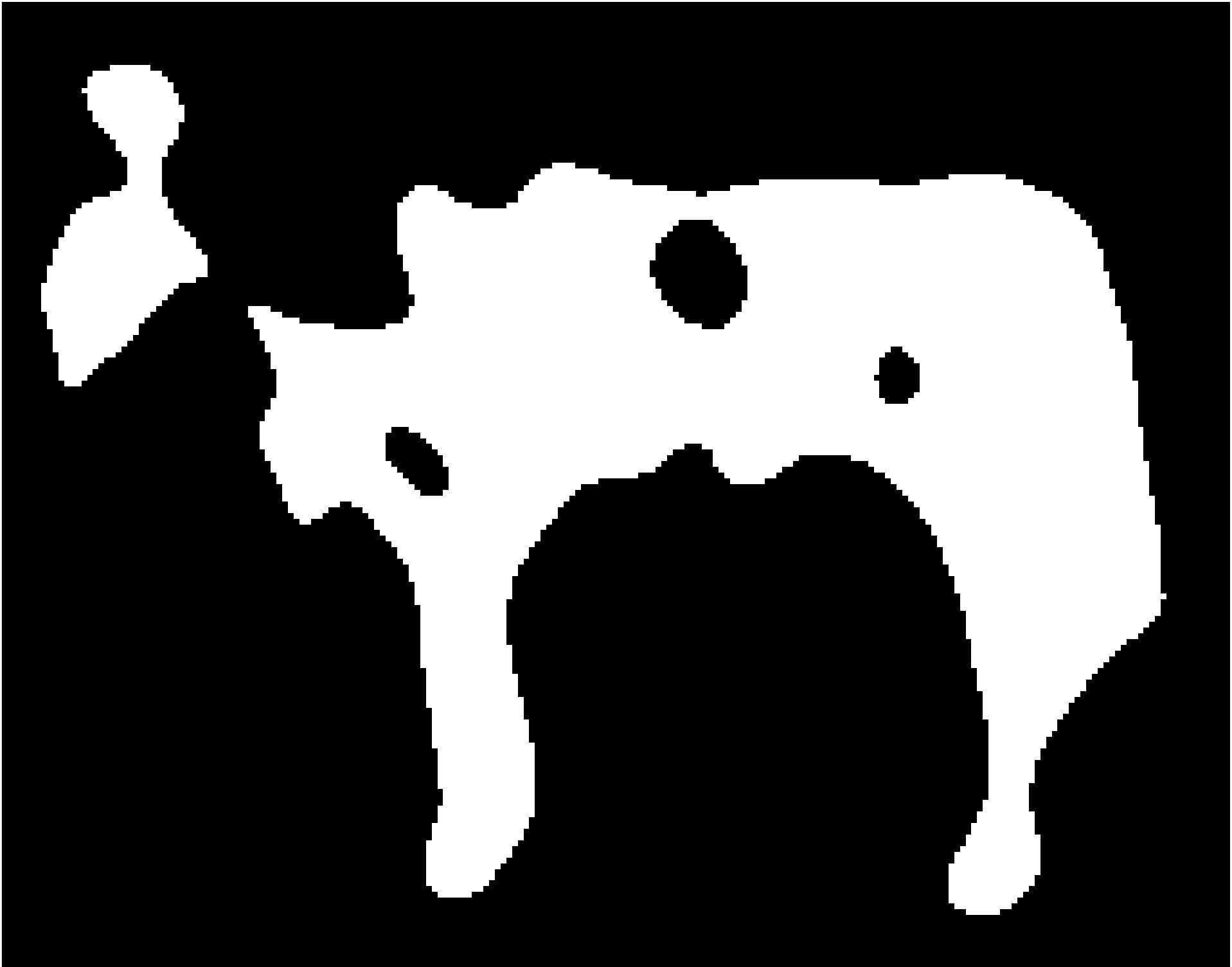}
   \includegraphics[width=0.19\linewidth]{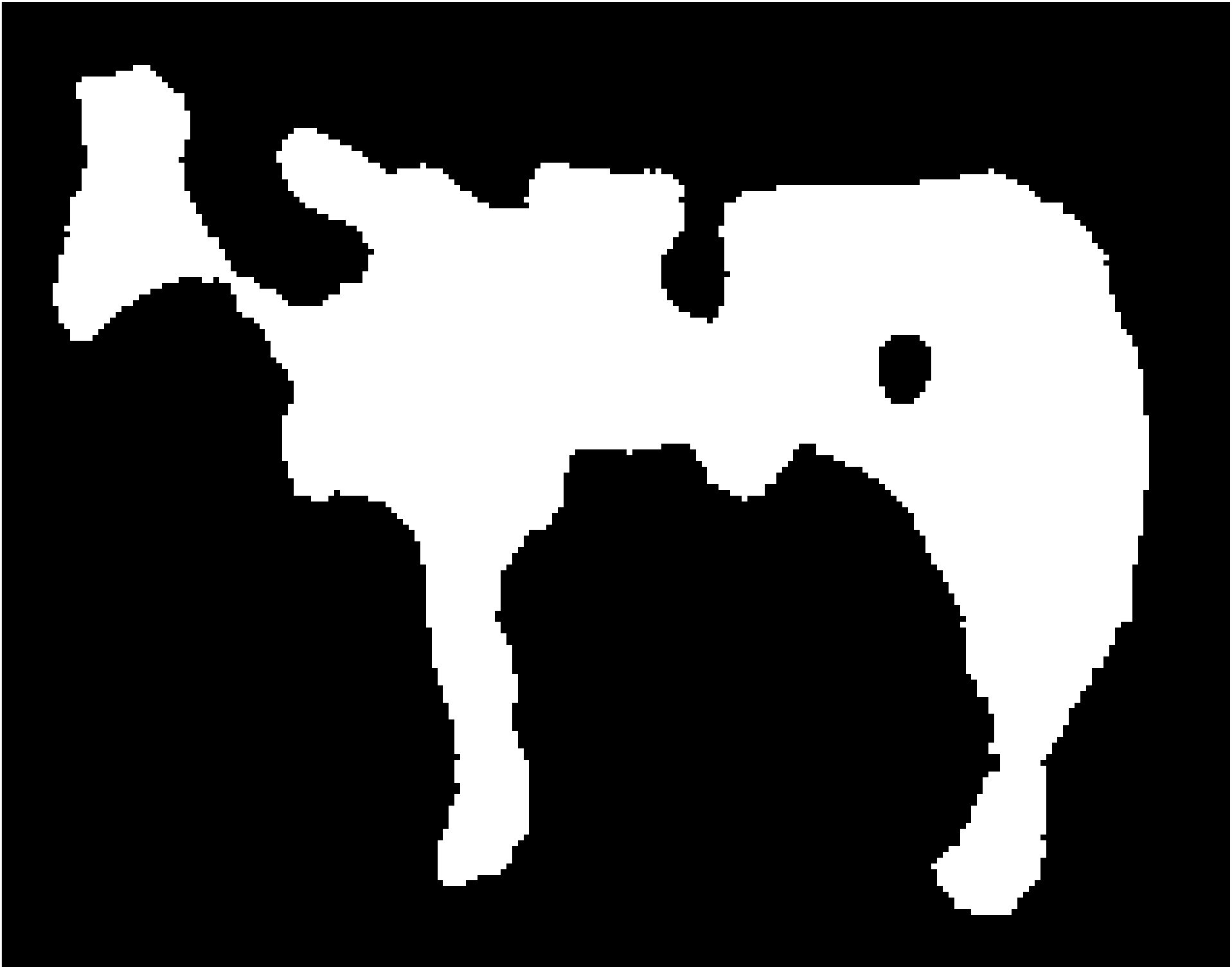}
    
\end{center}
   \caption{Qualitative results on the {\sc Weizmann Horse} dataset. From left: Original image, Ground truth, FCN-8s segmentation results, CRF-Grad  segmentation results (spatial kernel only), CRF-Grad segmentation results (spatial and bilateral kernels). Note how adding the CRF-Grad layer gives more refined segmentations, removing spurious outlier pixels previously classified as horse.}
\label{fig:weizmann_qual_res}
\end{figure*}

\begin{figure*}[t]
\begin{center}
\setlength\tabcolsep{1pt} 
\begin{tabular}{cccc}

    Input & LRR & LRR + CRF-Grad & Ground truth \\
   \includegraphics[width=0.24\linewidth]{figures/cityscapes/vis1_im_small.jpg} &
   \includegraphics[width=0.24\linewidth]{figures/cityscapes/vis_1_gt.pdf} &
   \includegraphics[width=0.24\linewidth]{figures/cityscapes/vis_1_lrr.pdf} &
   \includegraphics[width=0.24\linewidth]{figures/cityscapes/vis_1_ggdr.pdf} \\
  
   \includegraphics[width=0.24\linewidth]{figures/cityscapes/frankfurt_000001_029600_leftImg8bit.jpg} &
   \includegraphics[width=0.24\linewidth]{figures/cityscapes/frankfurt_000001_029600_lrr} &
   \includegraphics[width=0.24\linewidth]{figures/cityscapes/frankfurt_000001_029600_crfgrad} &
   \includegraphics[width=0.24\linewidth]{figures/cityscapes/frankfurt_000001_029600_gtFine_color} \\
   
   \includegraphics[width=0.24\linewidth]{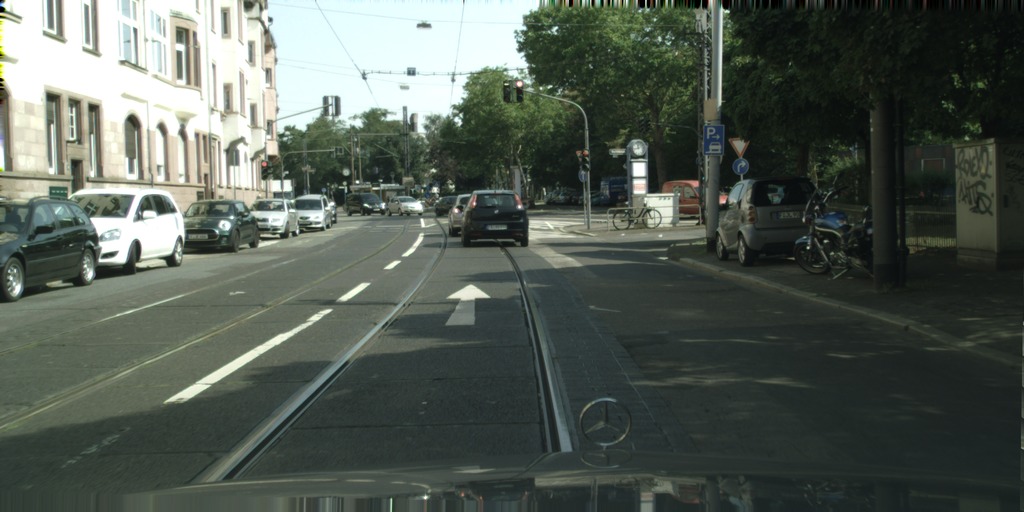} &
   \includegraphics[width=0.24\linewidth]{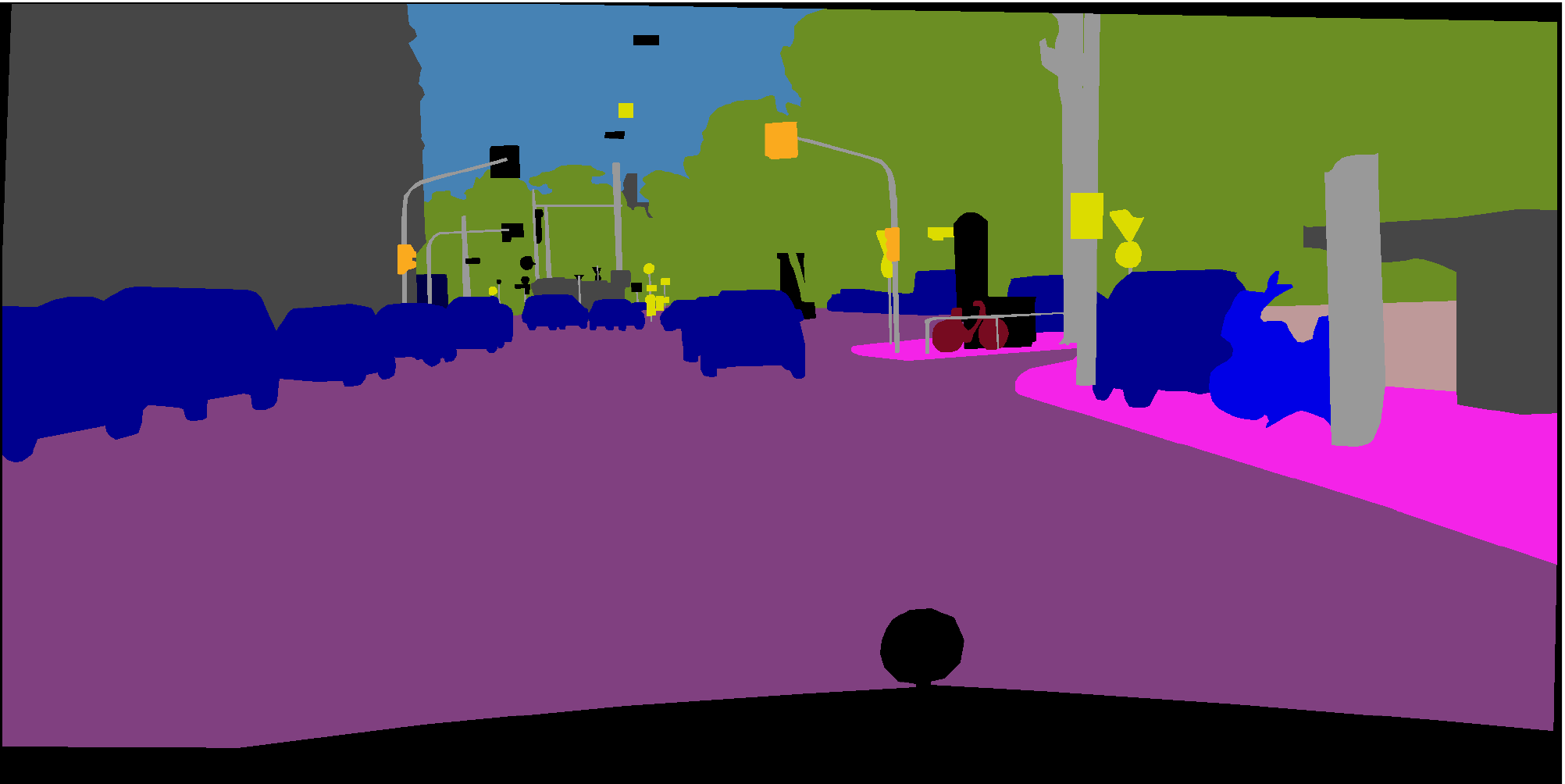} &
   \includegraphics[width=0.24\linewidth]{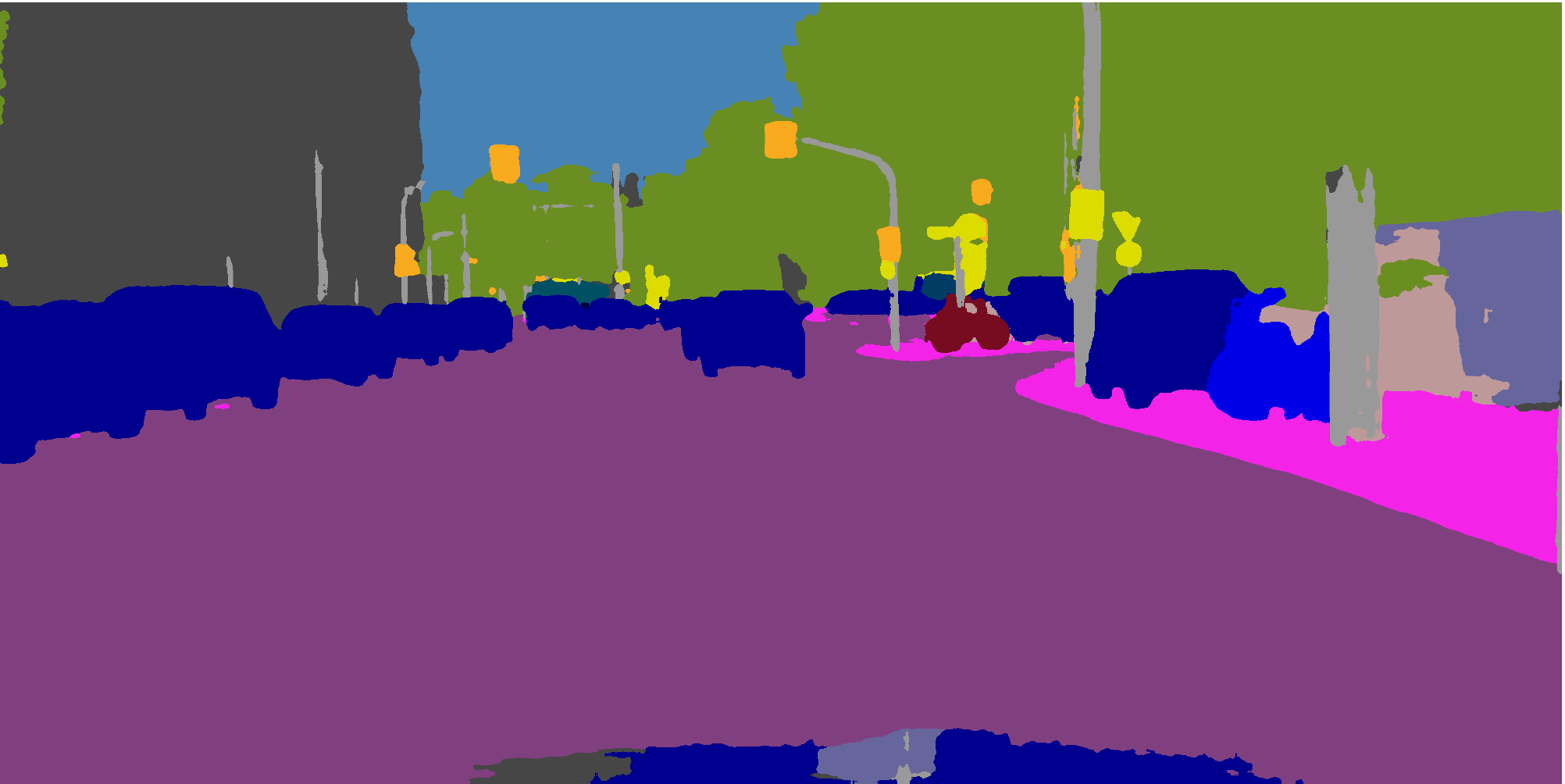} &
   \includegraphics[width=0.24\linewidth]{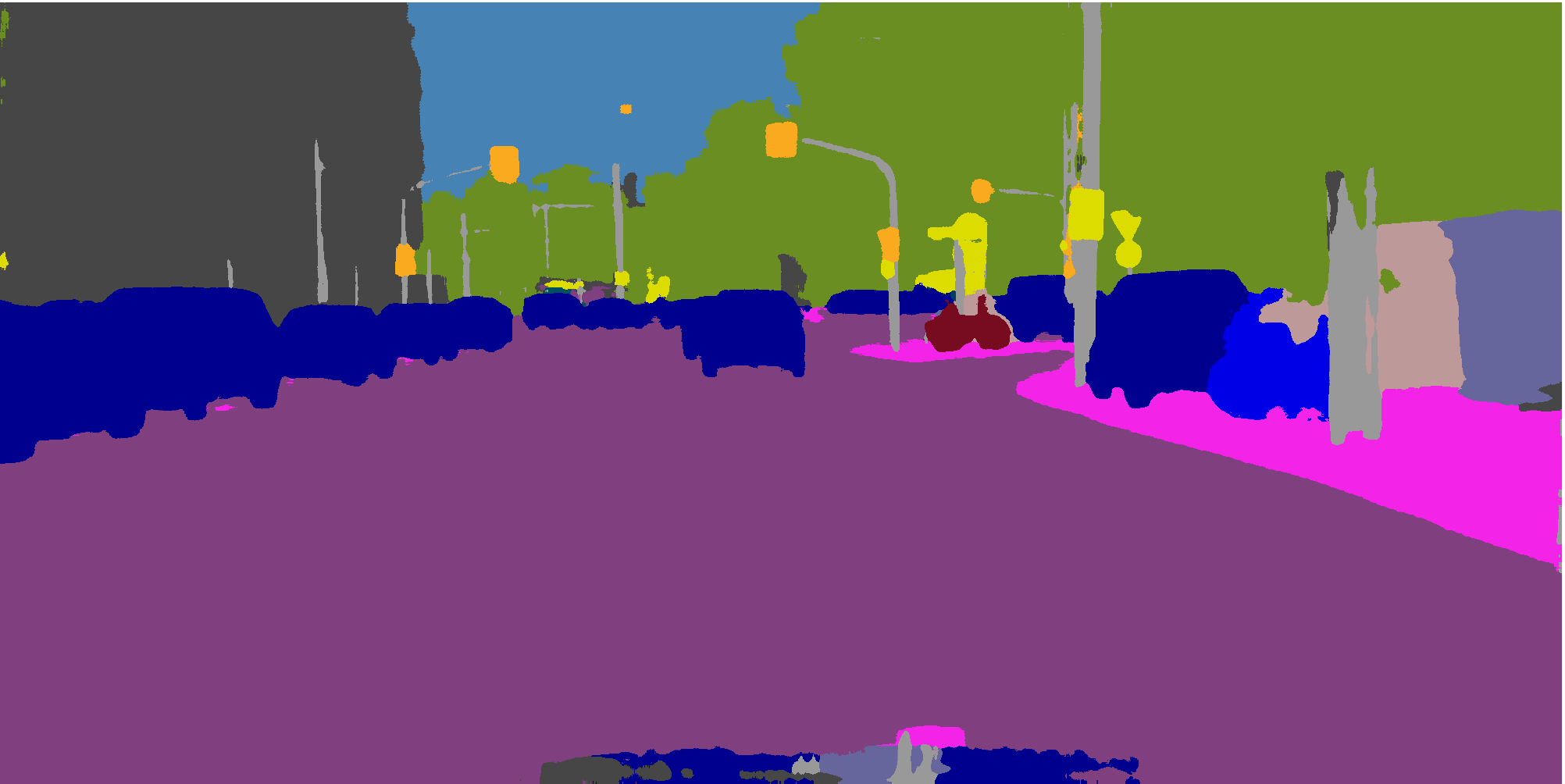} \\
   
   \includegraphics[width=0.24\linewidth]{figures/cityscapes/vis3_im_small.jpg} &
   \includegraphics[width=0.24\linewidth]{figures/cityscapes/vis_3_gt.pdf} &
   \includegraphics[width=0.24\linewidth]{figures/cityscapes/vis_3_lrr.pdf} &
   \includegraphics[width=0.24\linewidth]{figures/cityscapes/vis_3_ggdr.pdf} \\
   
   \includegraphics[width=0.24\linewidth]{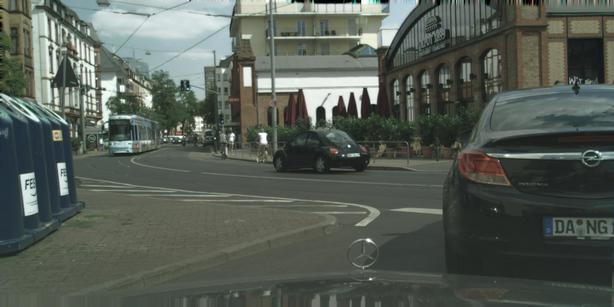} &
   \includegraphics[width=0.24\linewidth]{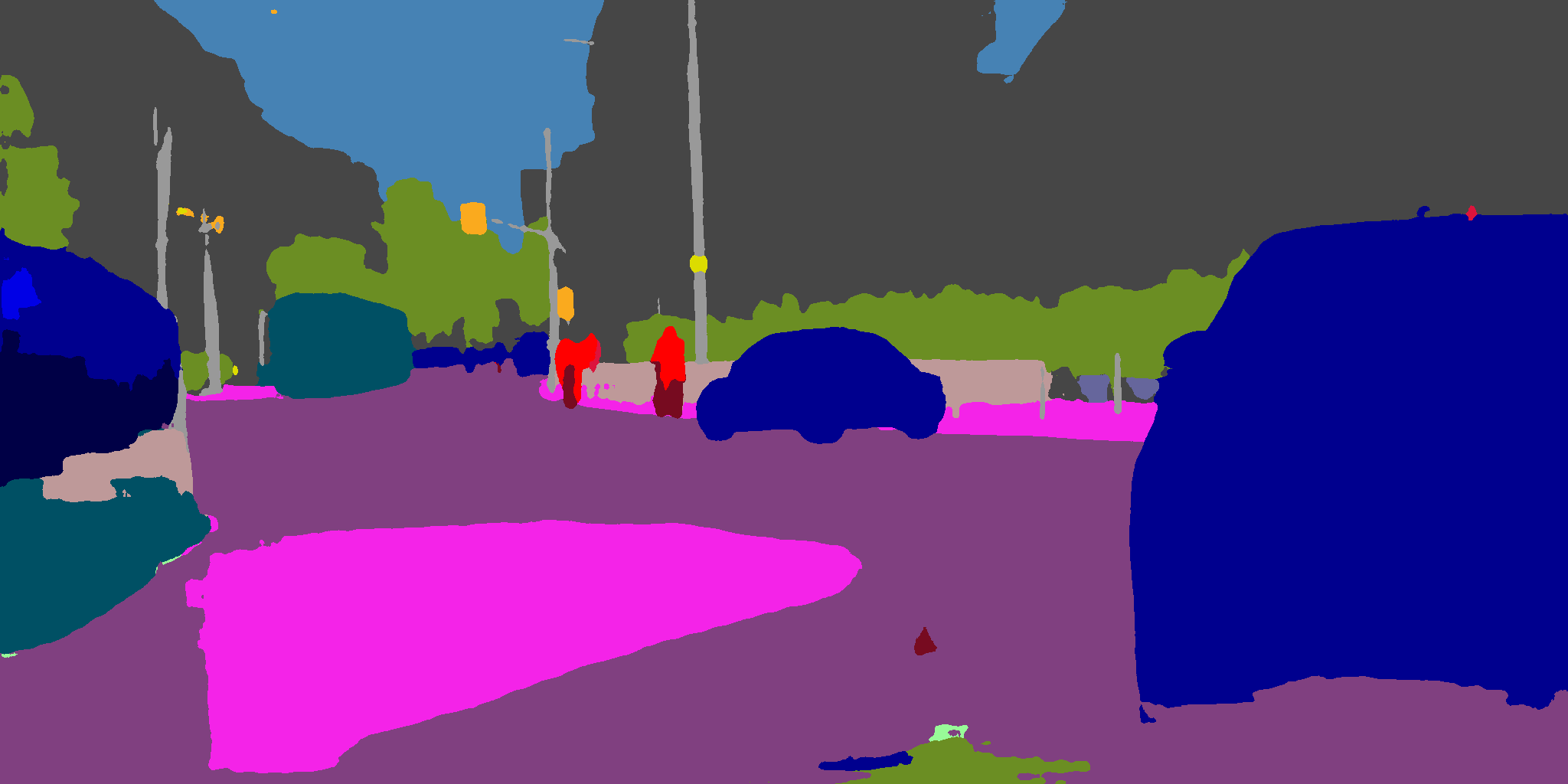} &
   \includegraphics[width=0.24\linewidth]{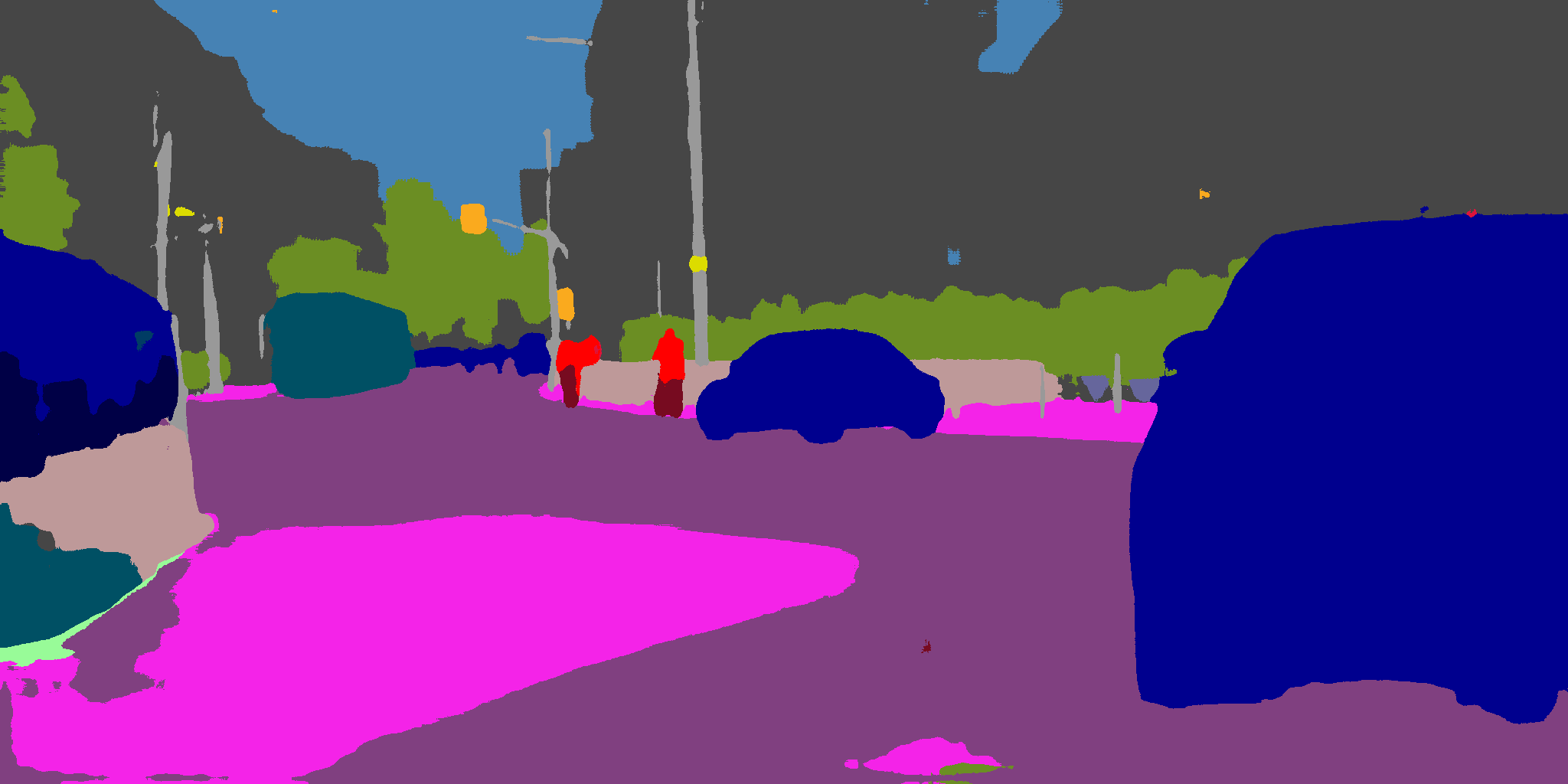} &
   \includegraphics[width=0.24\linewidth]{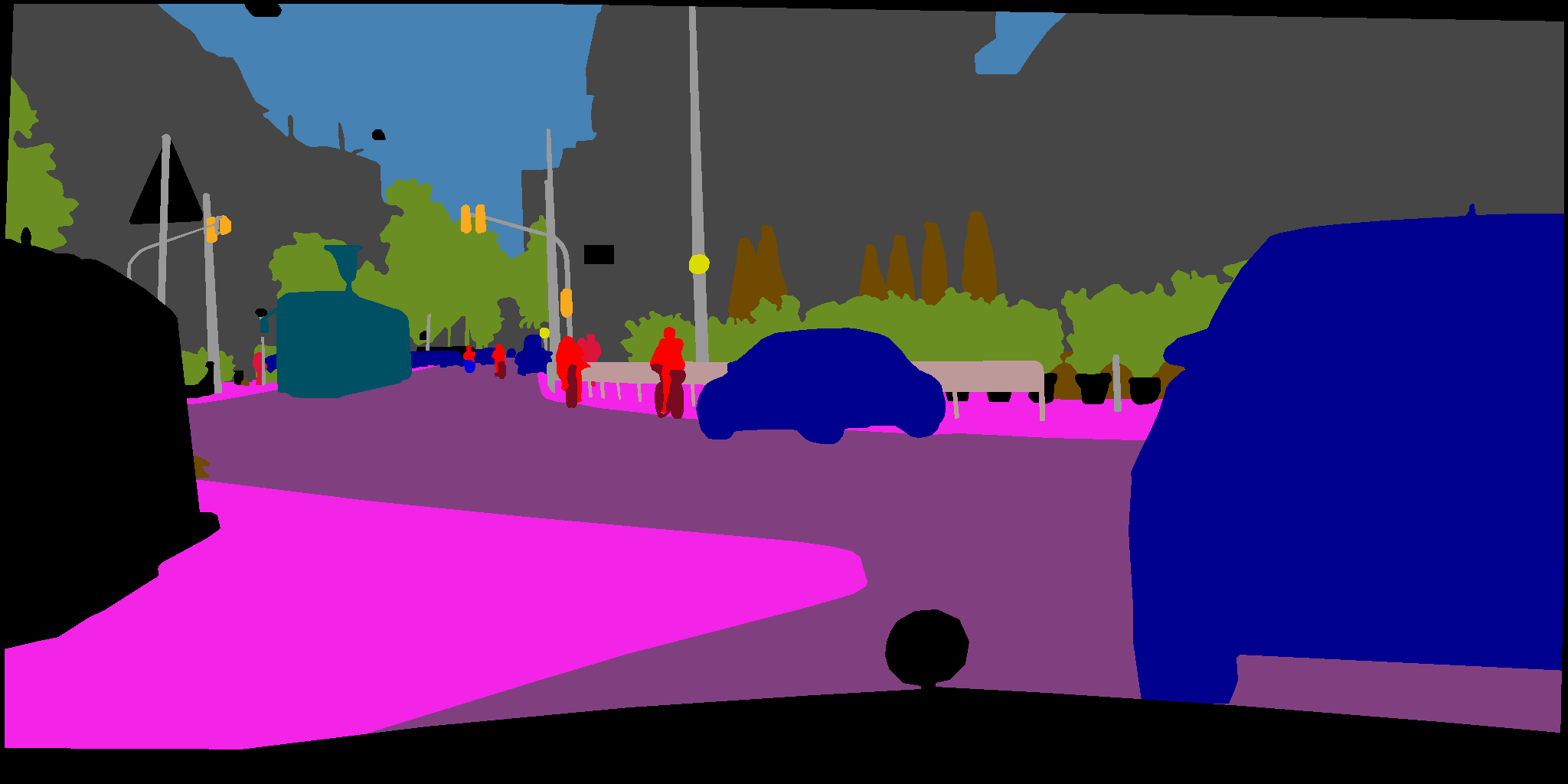} \\
   
   \includegraphics[width=0.24\linewidth]{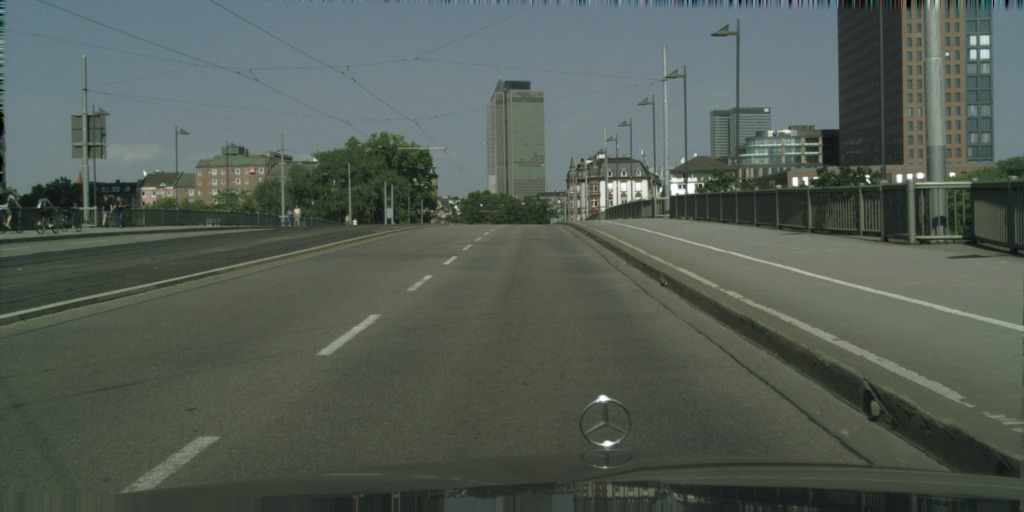} &
   \includegraphics[width=0.24\linewidth]{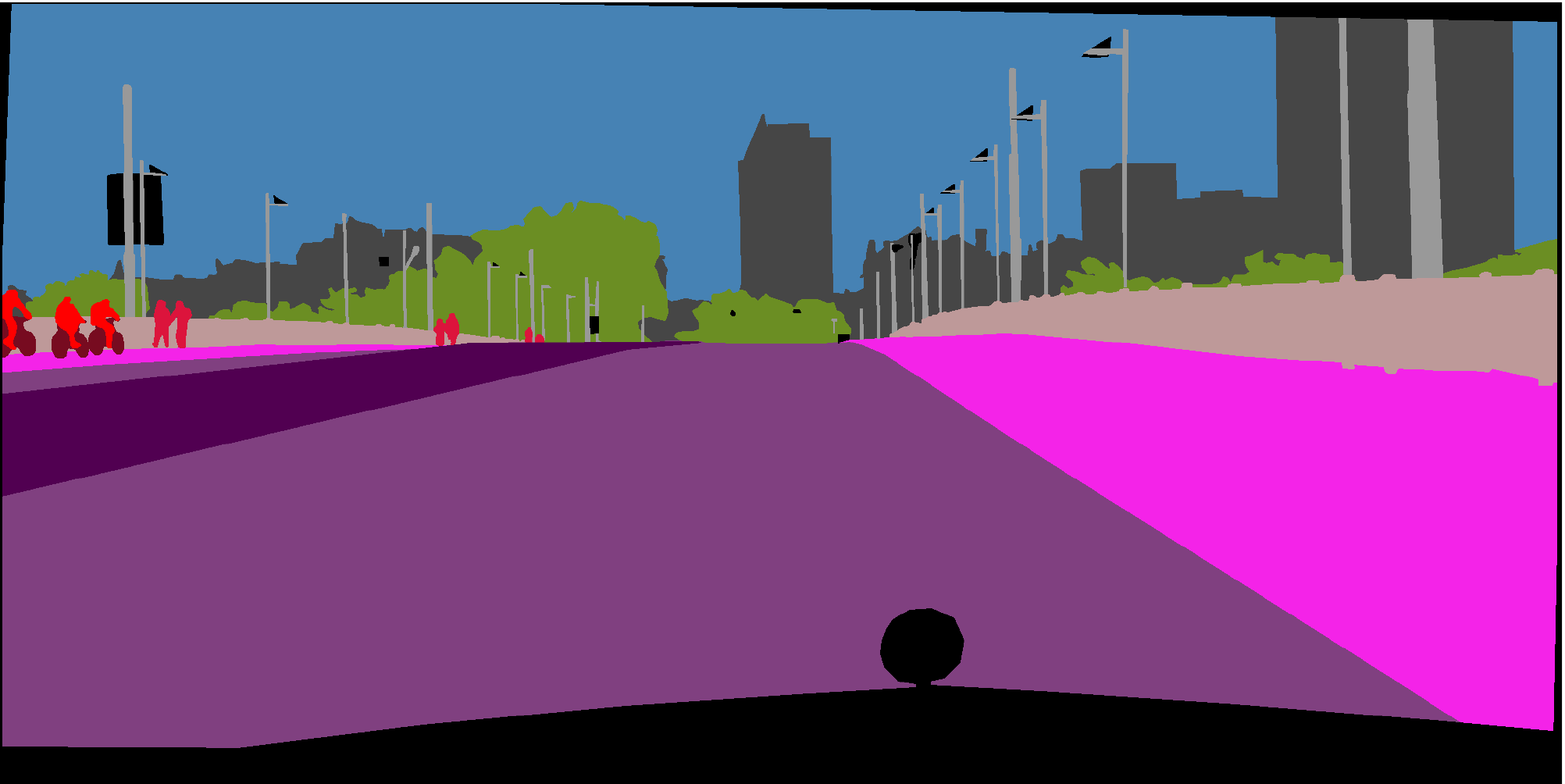} &
   \includegraphics[width=0.24\linewidth]{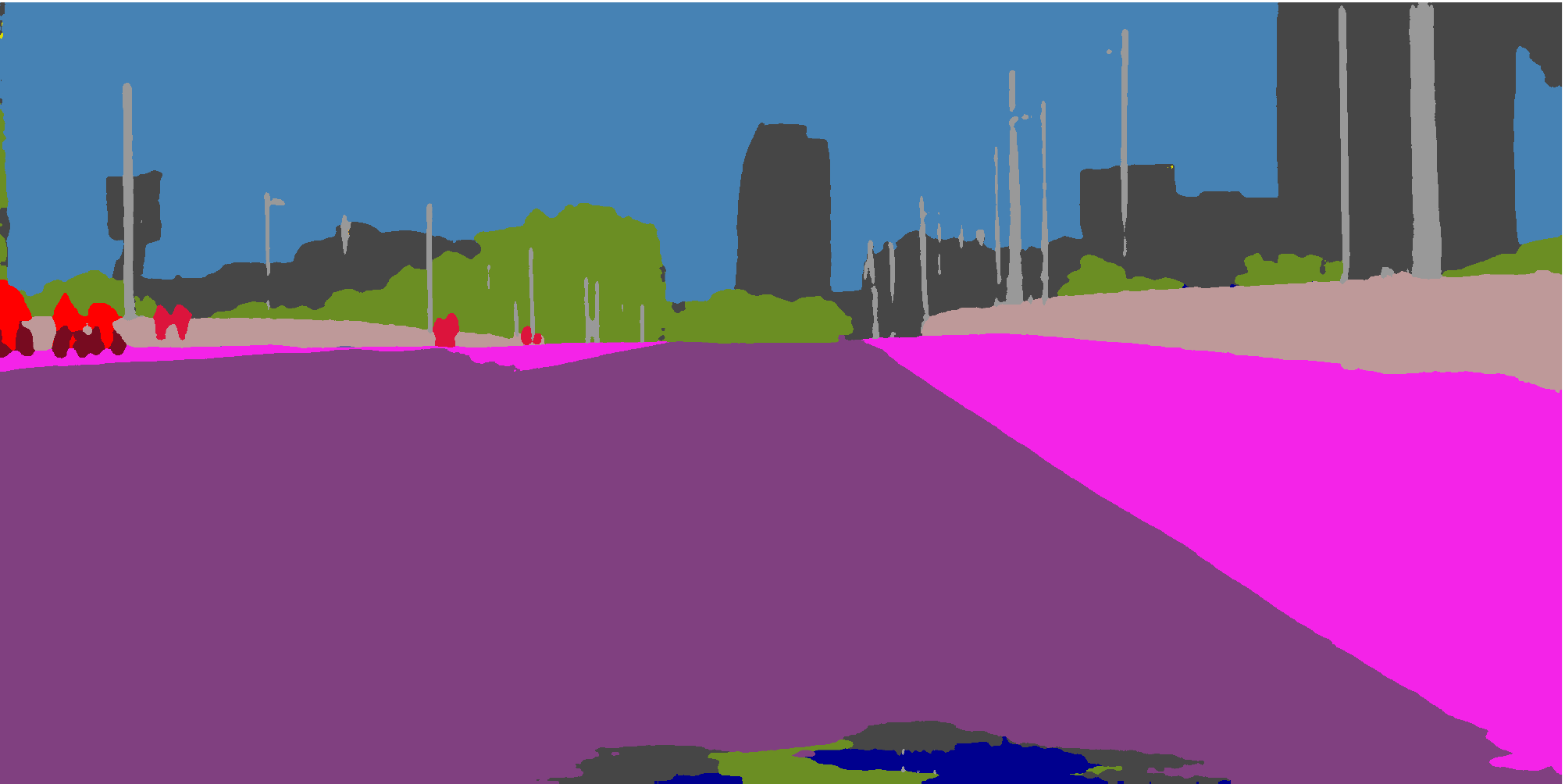} &
   \includegraphics[width=0.24\linewidth]{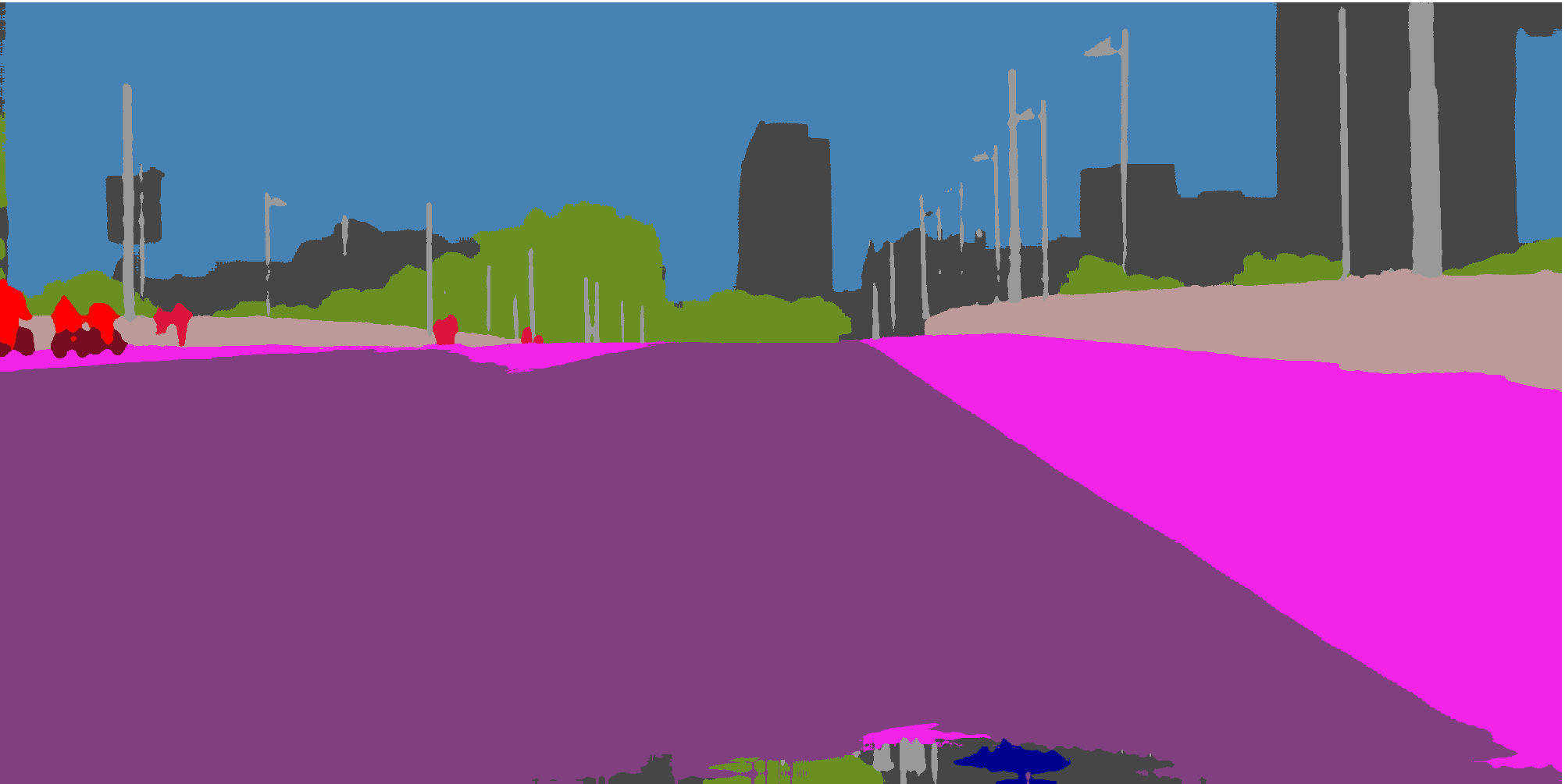} \\
   
   \includegraphics[width=0.24\linewidth]{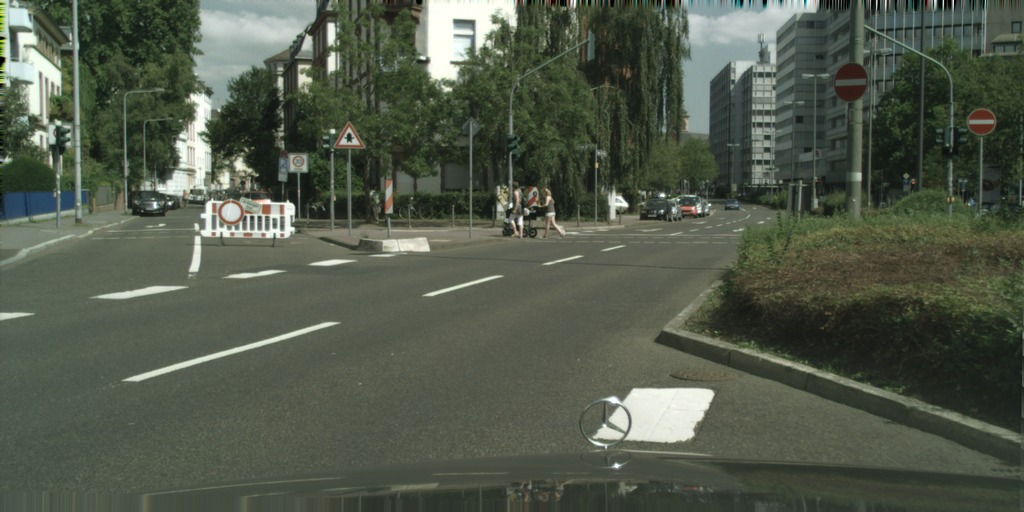} &
   \includegraphics[width=0.24\linewidth]{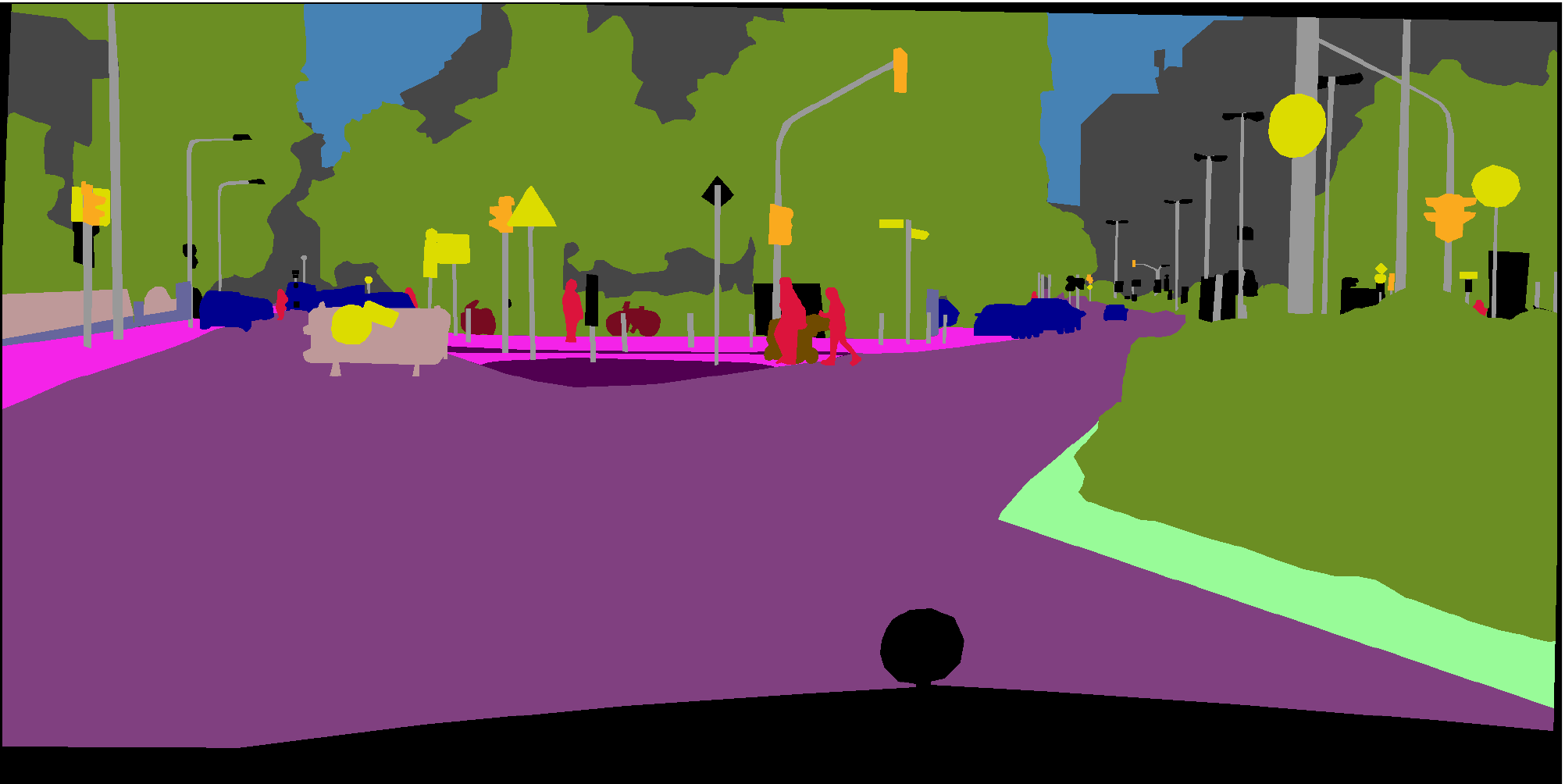} &
   \includegraphics[width=0.24\linewidth]{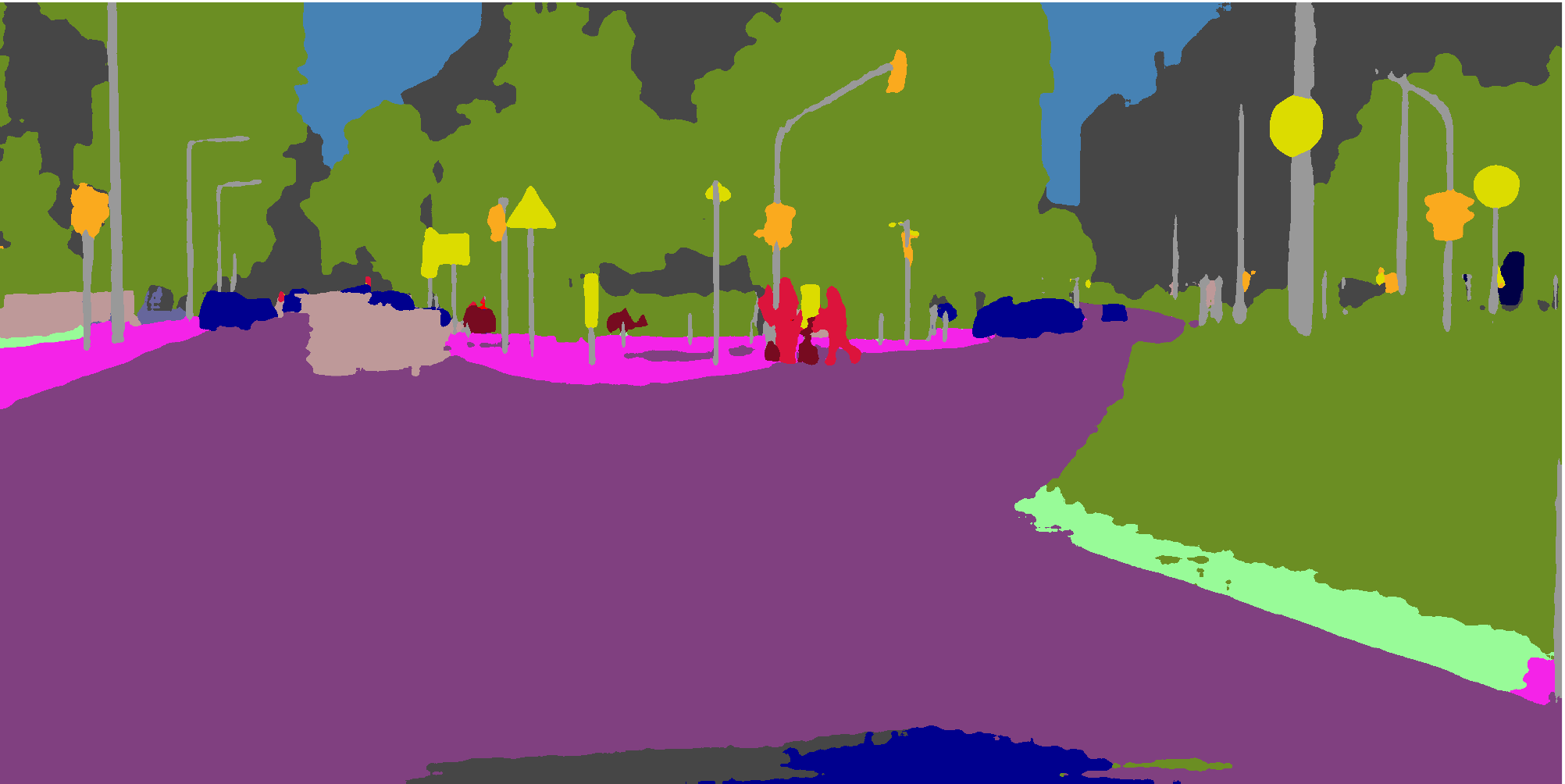} &
   \includegraphics[width=0.24\linewidth]{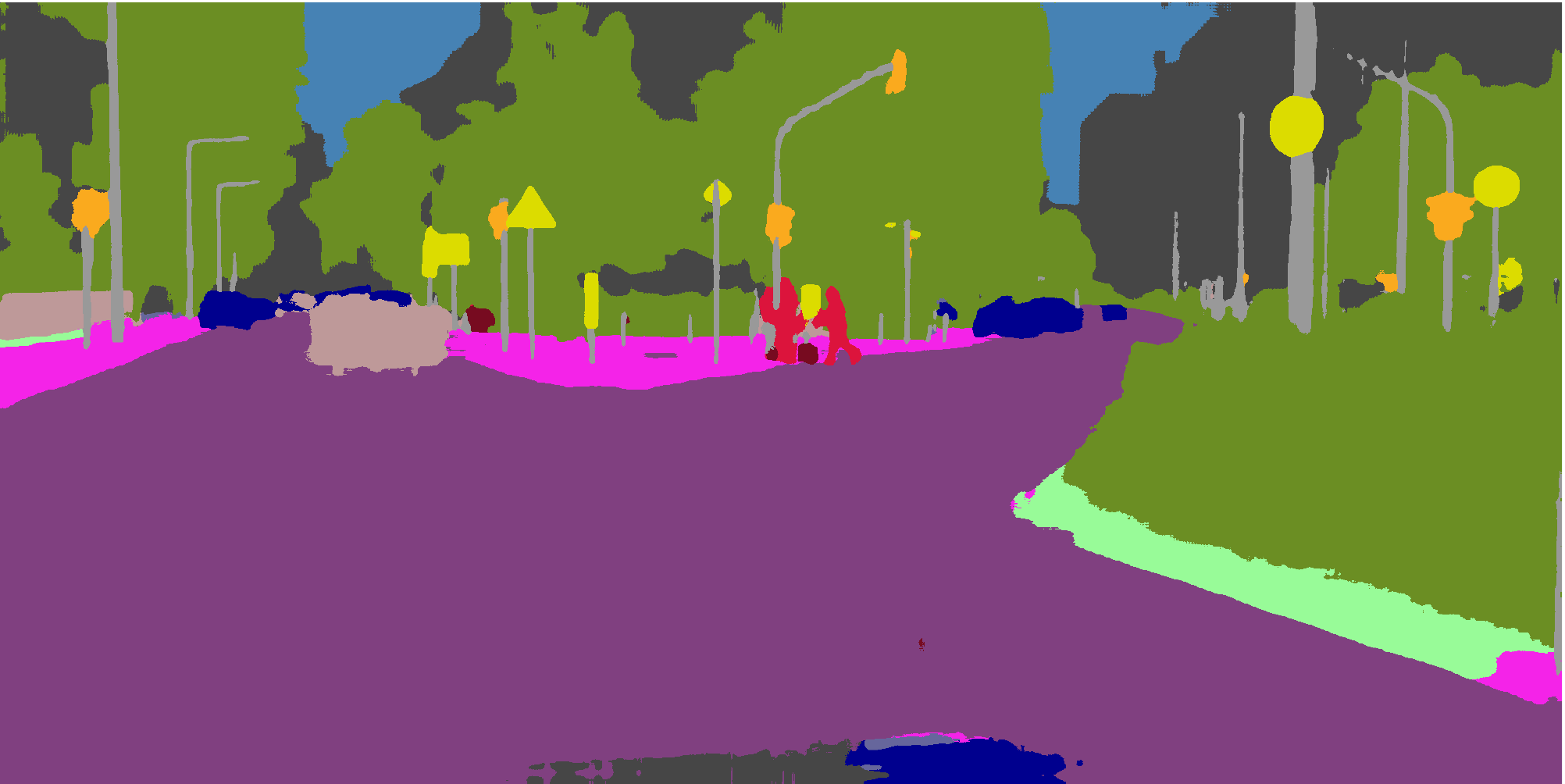} \\
      
   \includegraphics[width=0.24\linewidth]{figures/cityscapes/vis6_im_small.jpg} &
   \includegraphics[width=0.24\linewidth]{figures/cityscapes/vis_6_gt.pdf} &
   \includegraphics[width=0.24\linewidth]{figures/cityscapes/vis_6_lrr.pdf} &
   \includegraphics[width=0.24\linewidth]{figures/cityscapes/vis_6_ggdr.pdf} \\
   
   \includegraphics[width=0.24\linewidth]{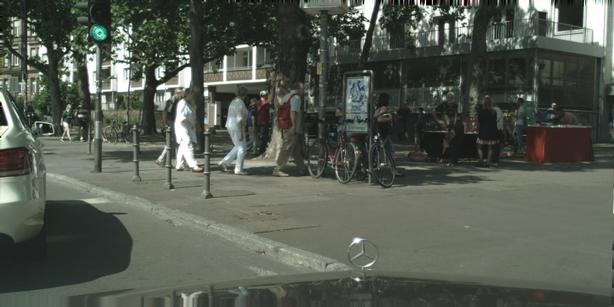} &
   \includegraphics[width=0.24\linewidth]{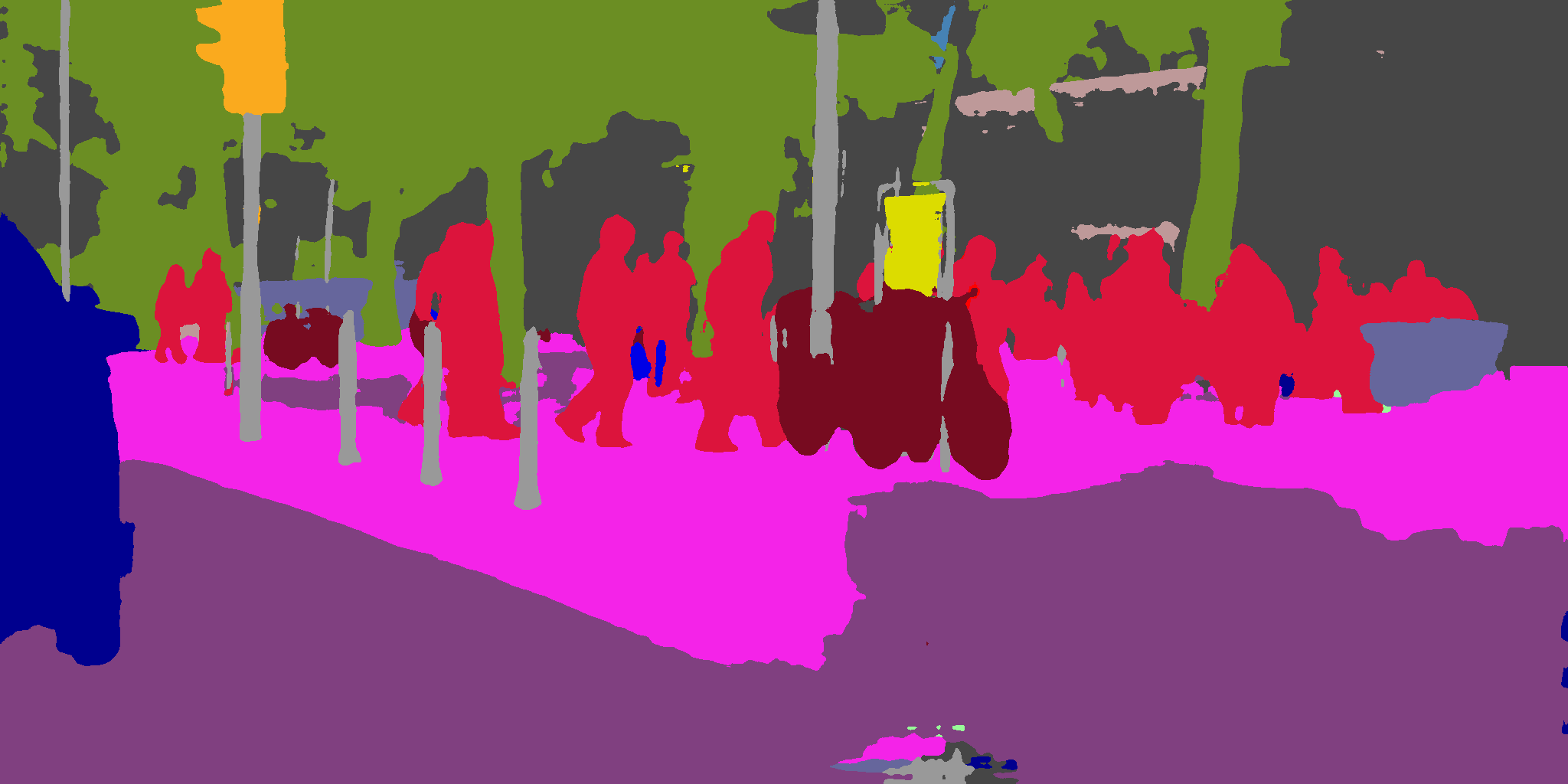} &
   \includegraphics[width=0.24\linewidth]{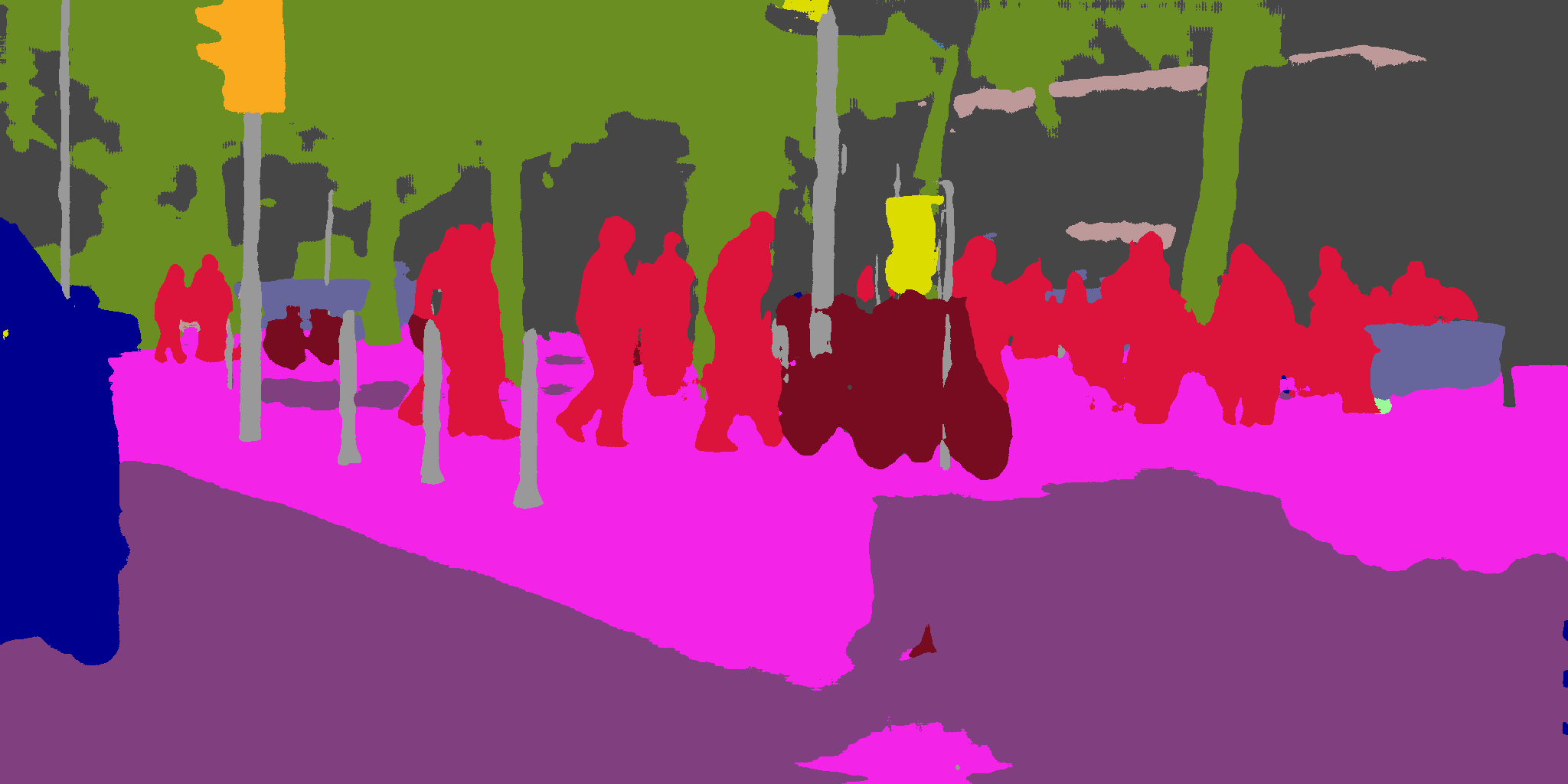} &
   \includegraphics[width=0.24\linewidth]{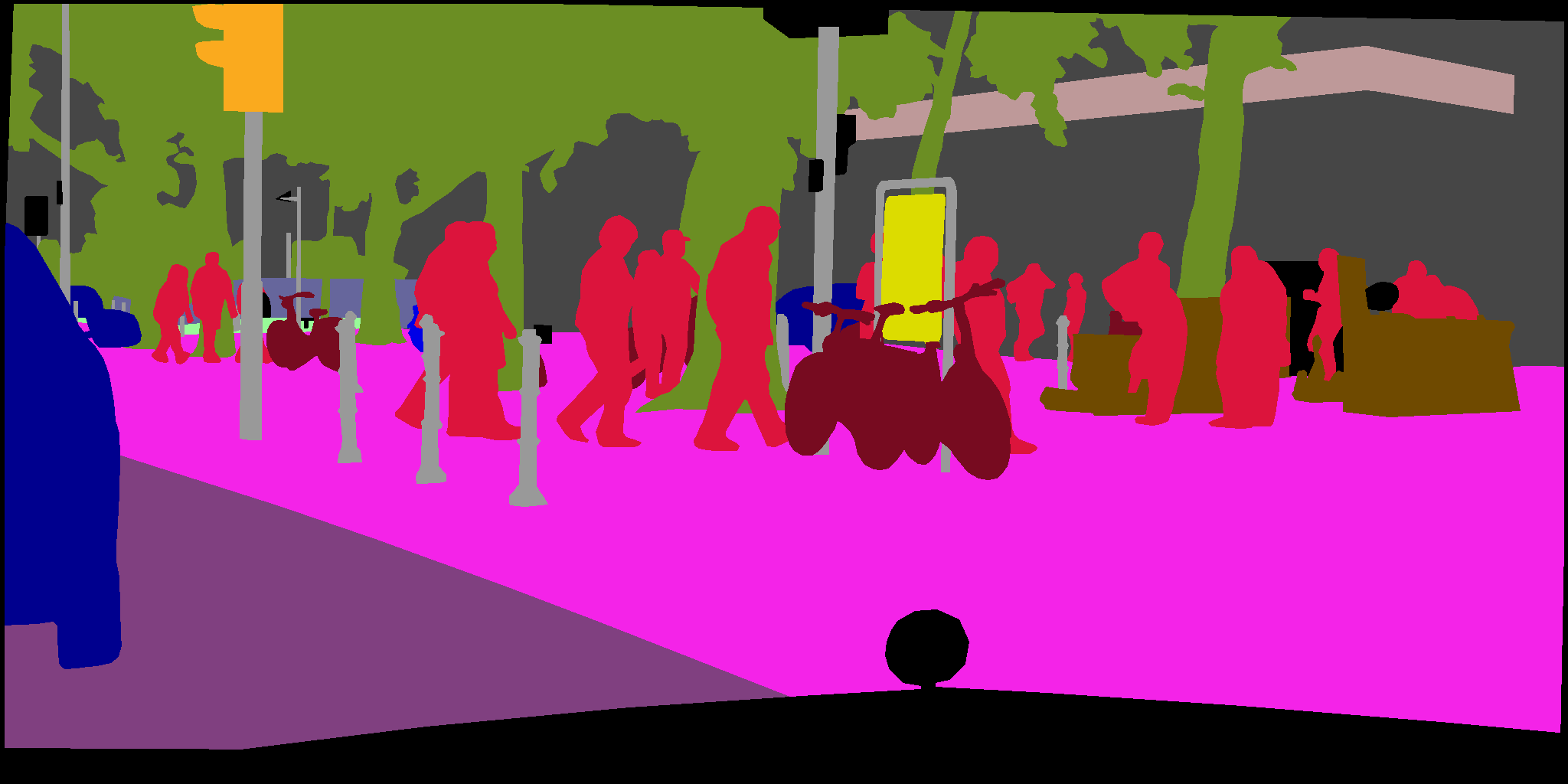} \\
   
   \end{tabular}
   \setlength\tabcolsep{6pt}
   
\end{center}
   \caption{Qualitative results on the {\sc Cityscapes} validation set. Black regions in the ground truth are ignored during evaluation. Our CRF models contextual relationships between classes, hence unlike LRR, it does not label ``road'' as being on top of ``sidewalk'' (Row 2). Note that the traffic lights are better segmented with the additional CRF-Grad layer. Adding the CRF-Grad layer increased the IoU of the class traffic lights from $66.8$ to $68.1$. }
\label{fig:cs_qual_res}
\end{figure*}

\begin{figure*}[t]
\begin{center}
\setlength\tabcolsep{1pt} 
\begin{tabular}{ccccc}

    Input & FCN-8s & CRF-RNN & CRF-Grad & Ground truth  \\
   \includegraphics[width=0.19\linewidth]{figures/nyu/nyu_1395.jpg} &
   \includegraphics[width=0.19\linewidth]{figures/nyu/nyu_1395_gt.png}&
   \includegraphics[width=0.19\linewidth]{figures/nyu/nyu_1395_fcn8.png}&
   \includegraphics[width=0.19\linewidth]{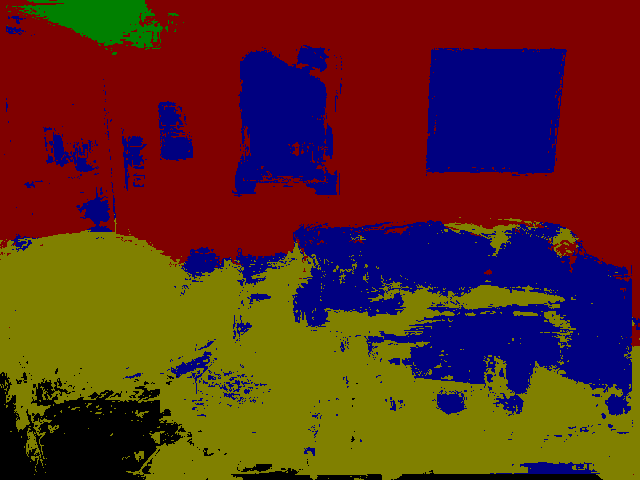}&
   \includegraphics[width=0.19\linewidth]{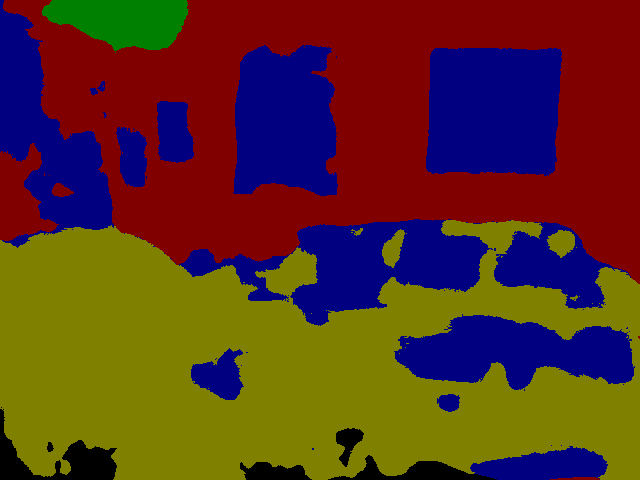}\\
   
   \includegraphics[width=0.19\linewidth]{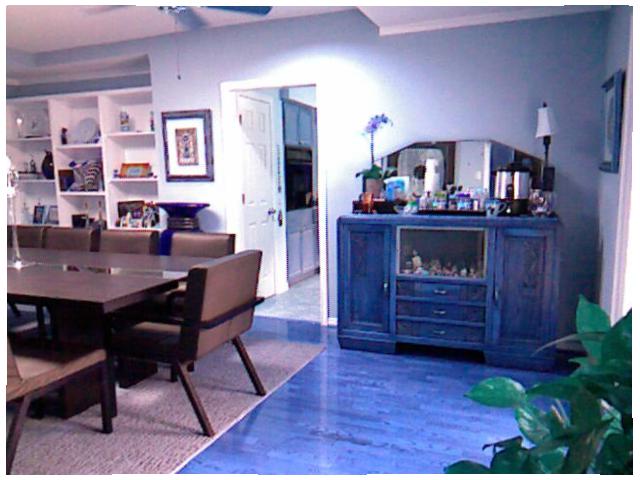}&
   \includegraphics[width=0.19\linewidth]{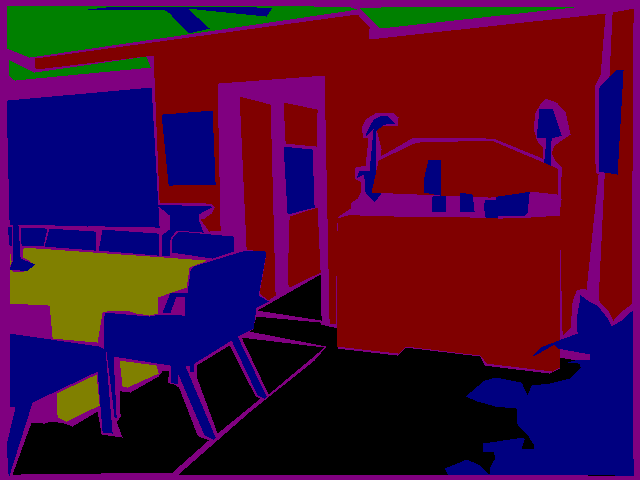}&
   \includegraphics[width=0.19\linewidth]{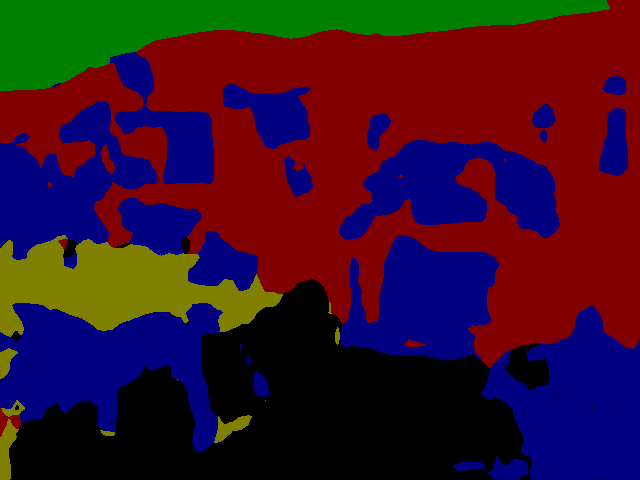}&
   \includegraphics[width=0.19\linewidth]{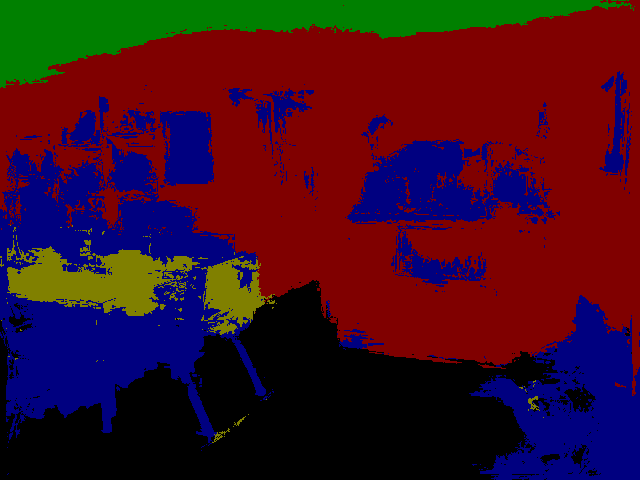}&
   \includegraphics[width=0.19\linewidth]{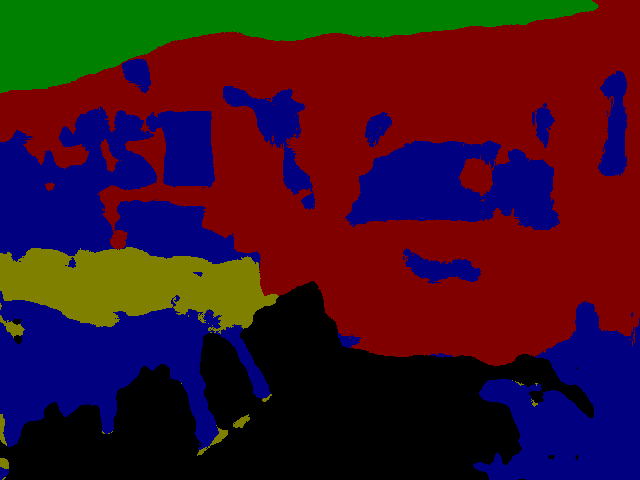}\\
   
   \includegraphics[width=0.19\linewidth]{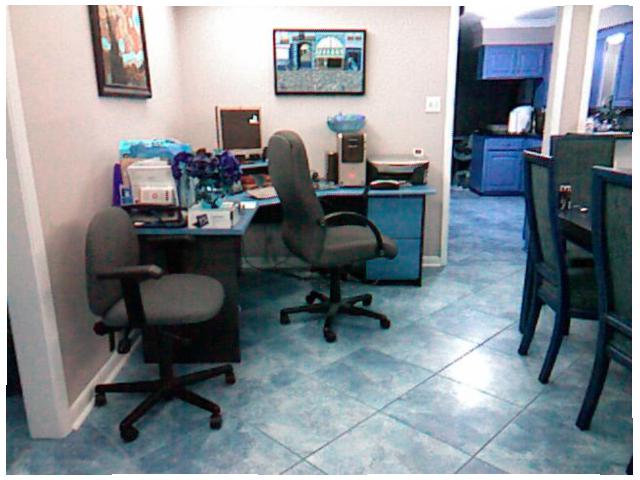}&
   \includegraphics[width=0.19\linewidth]{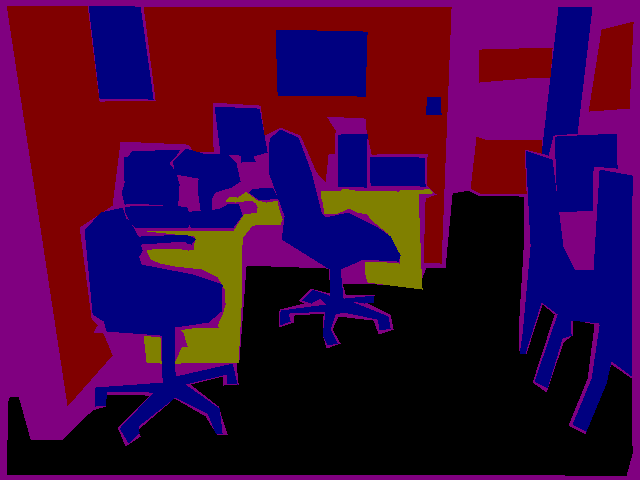}&
   \includegraphics[width=0.19\linewidth]{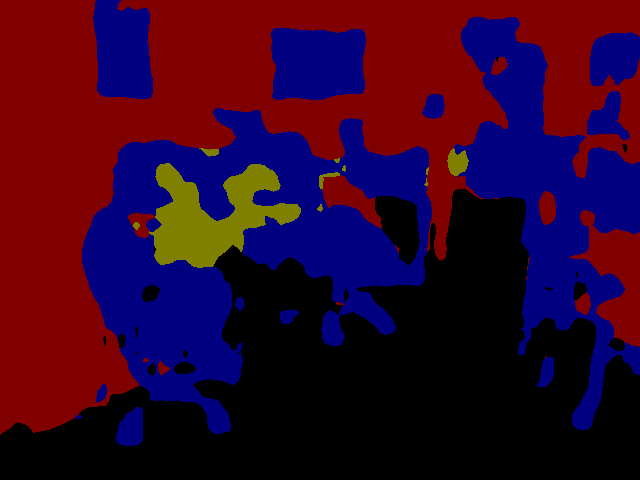}&
   \includegraphics[width=0.19\linewidth]{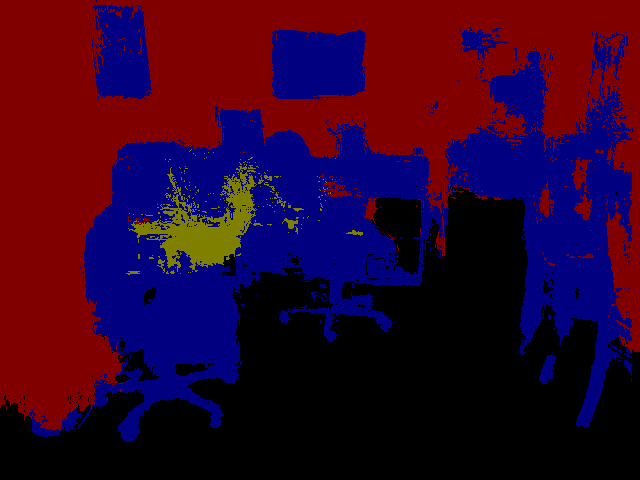}&
   \includegraphics[width=0.19\linewidth]{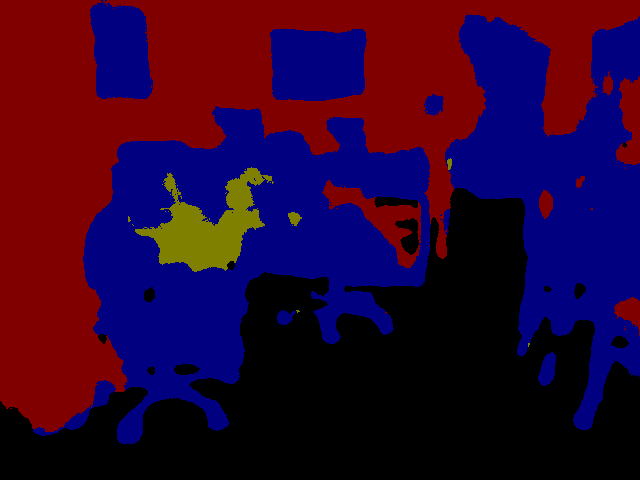}\\
   
   \includegraphics[width=0.19\linewidth]{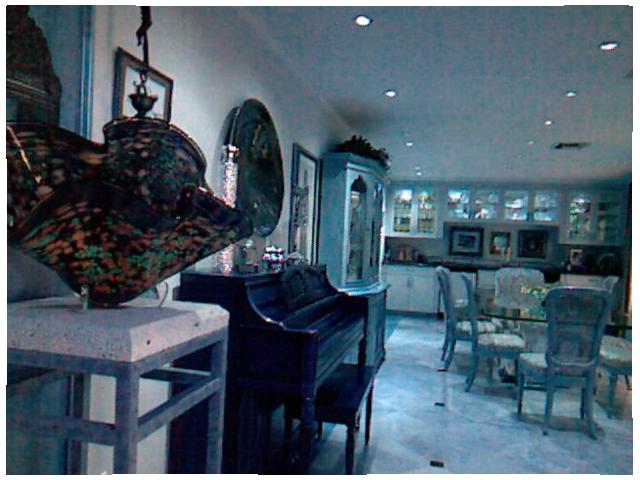}&
   \includegraphics[width=0.19\linewidth]{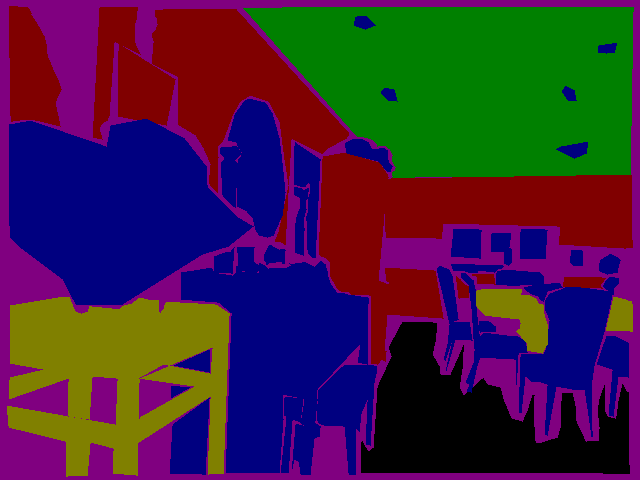}&
   \includegraphics[width=0.19\linewidth]{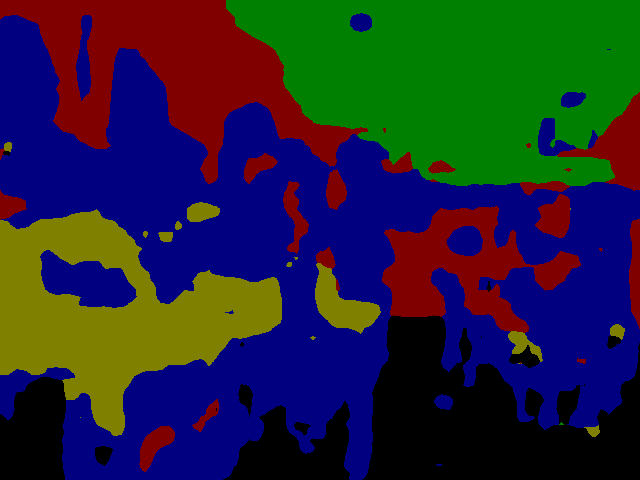}&
   \includegraphics[width=0.19\linewidth]{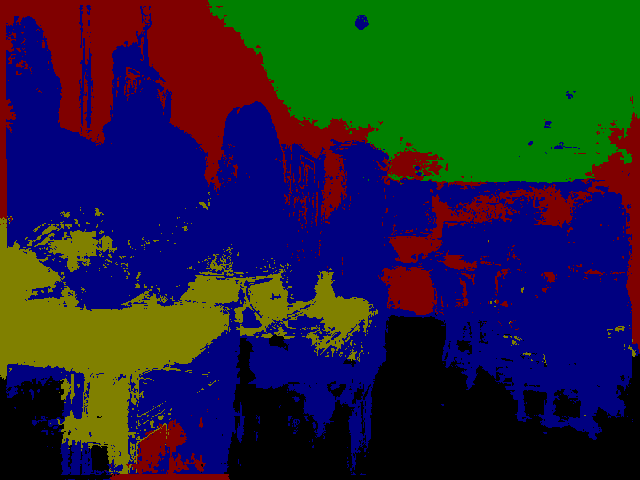}&
   \includegraphics[width=0.19\linewidth]{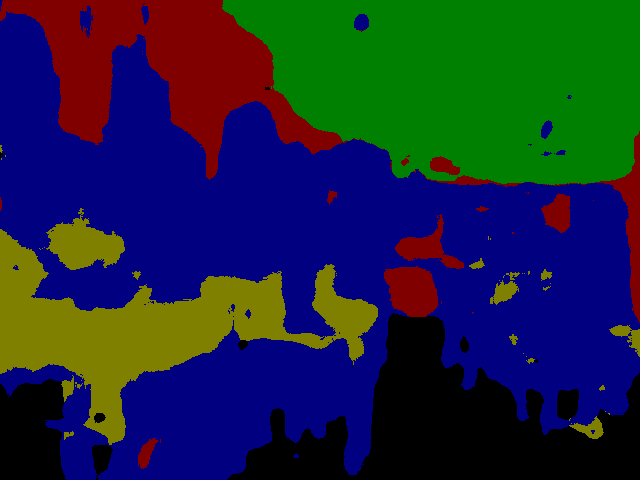}\\

   \includegraphics[width=0.19\linewidth]{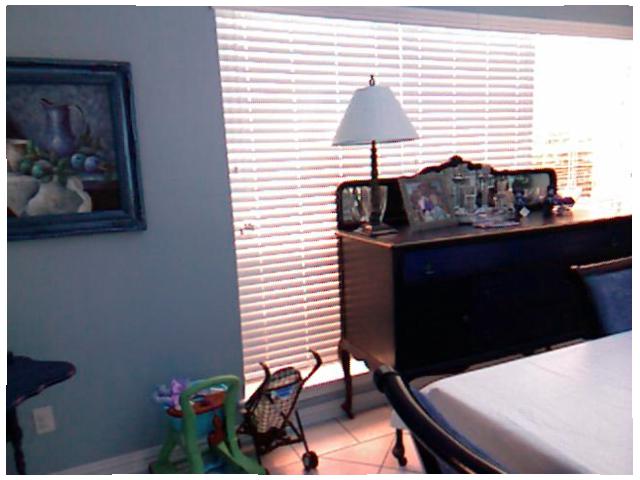}&
   \includegraphics[width=0.19\linewidth]{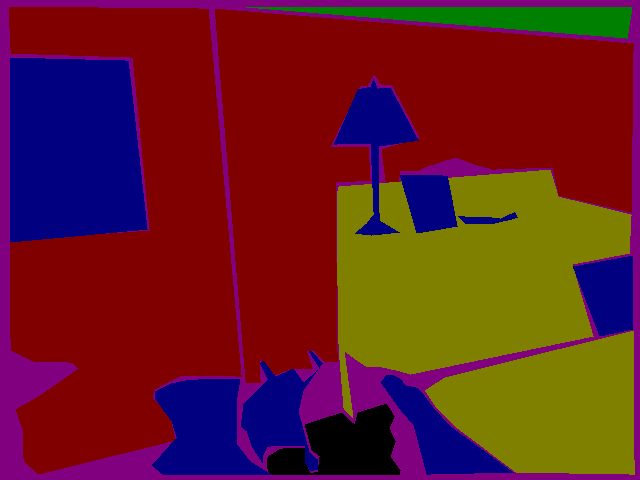}&
   \includegraphics[width=0.19\linewidth]{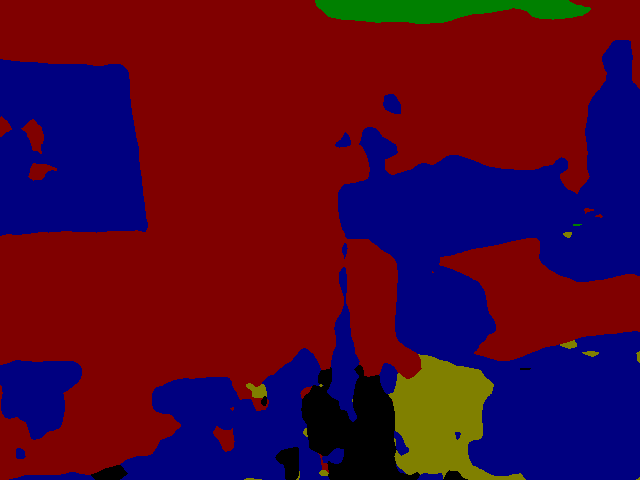}&
   \includegraphics[width=0.19\linewidth]{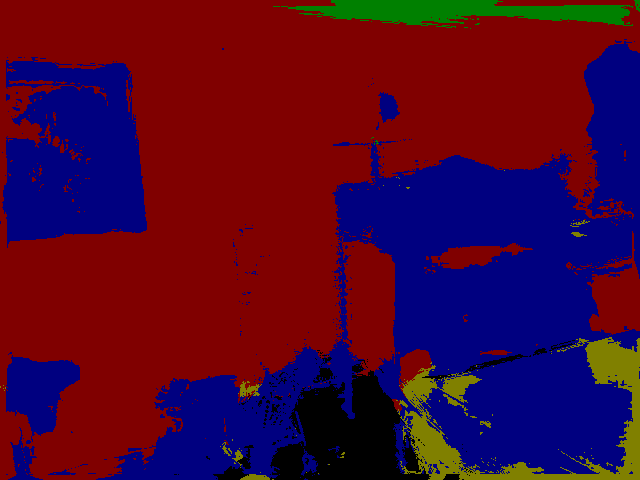}&
   \includegraphics[width=0.19\linewidth]{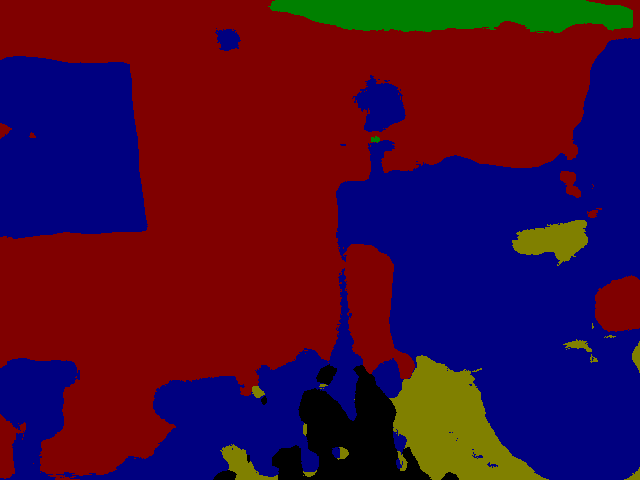}\\
   
   \includegraphics[width=0.19\linewidth]{figures/nyu/nyu_1435.jpg}&
   \includegraphics[width=0.19\linewidth]{figures/nyu/nyu_1435_gt.png}&
   \includegraphics[width=0.19\linewidth]{figures/nyu/nyu_1435_fcn8.png}&
   \includegraphics[width=0.19\linewidth]{figures/nyu/nyu_1435_crfasrnn4arxiv.png}&
   \includegraphics[width=0.19\linewidth]{figures/nyu/nyu_1435_crfgr4arxiv.png}\\
   
   \includegraphics[width=0.19\linewidth]{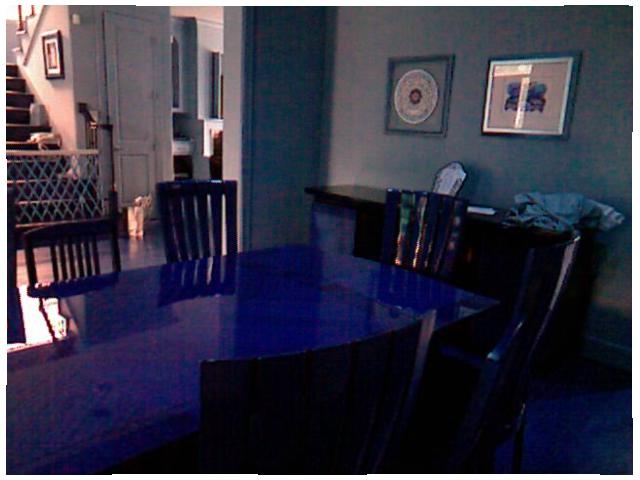}&
   \includegraphics[width=0.19\linewidth]{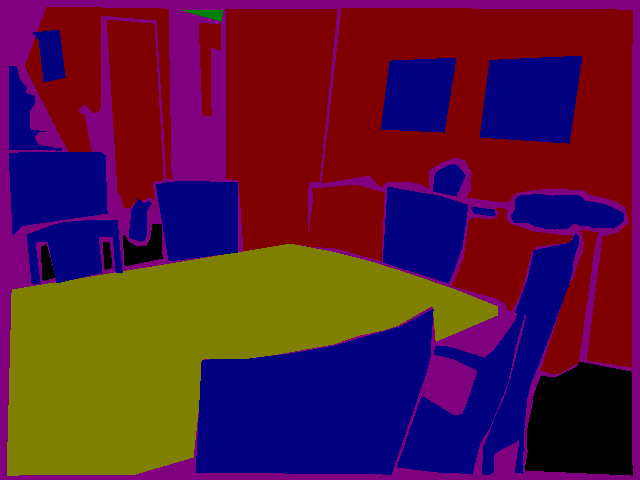}&
   \includegraphics[width=0.19\linewidth]{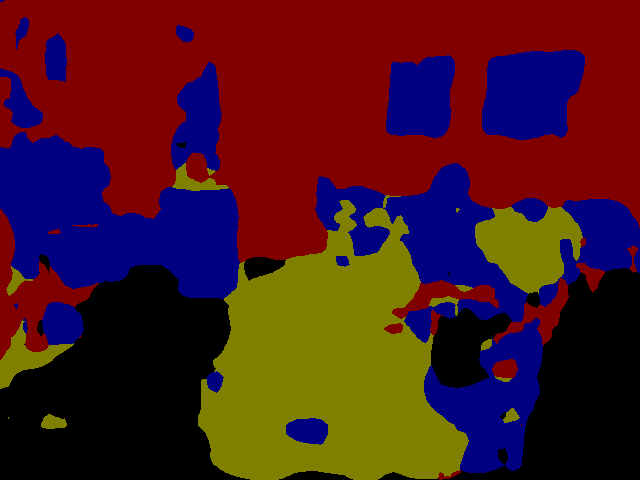}&
   \includegraphics[width=0.19\linewidth]{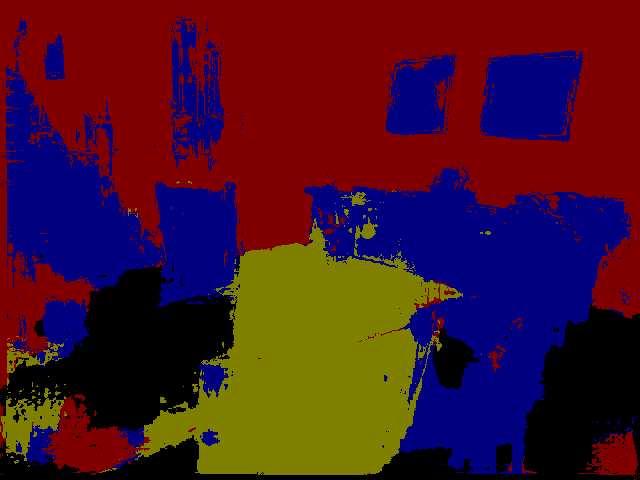}&
   \includegraphics[width=0.19\linewidth]{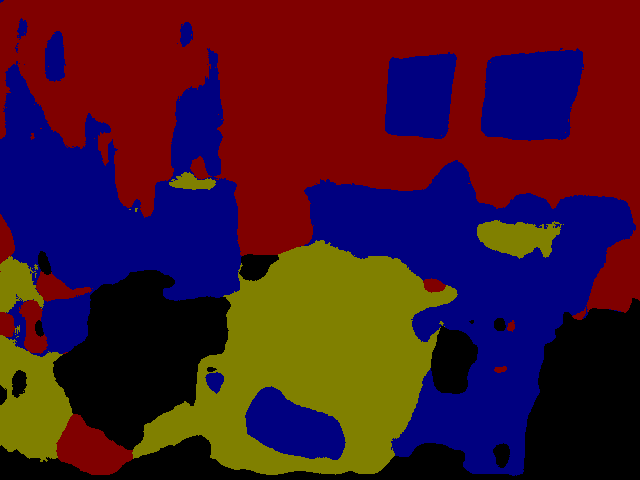}\\
   
   \includegraphics[width=0.19\linewidth]{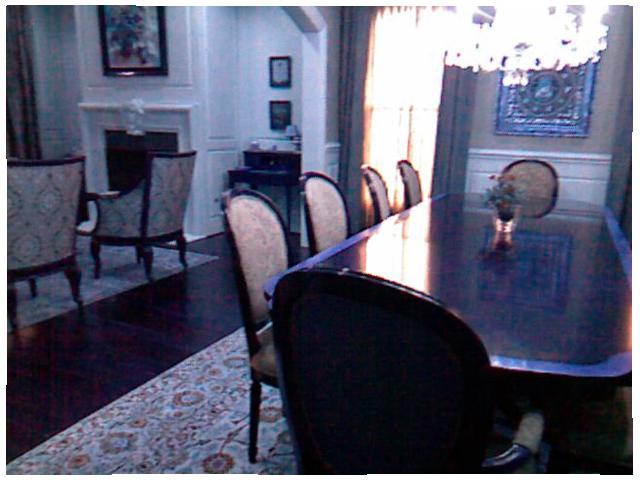}&
   \includegraphics[width=0.19\linewidth]{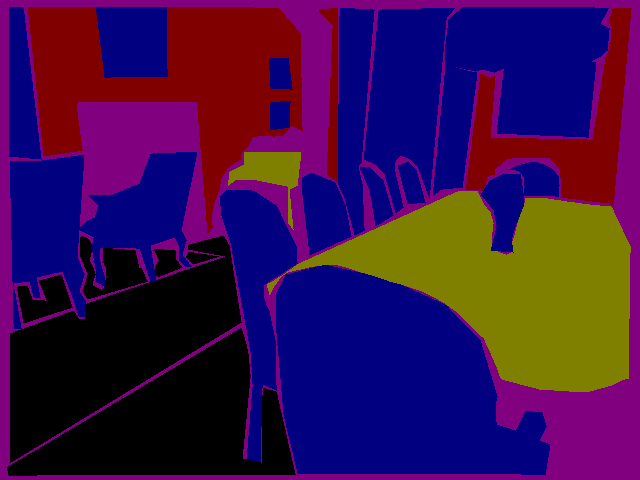}&
   \includegraphics[width=0.19\linewidth]{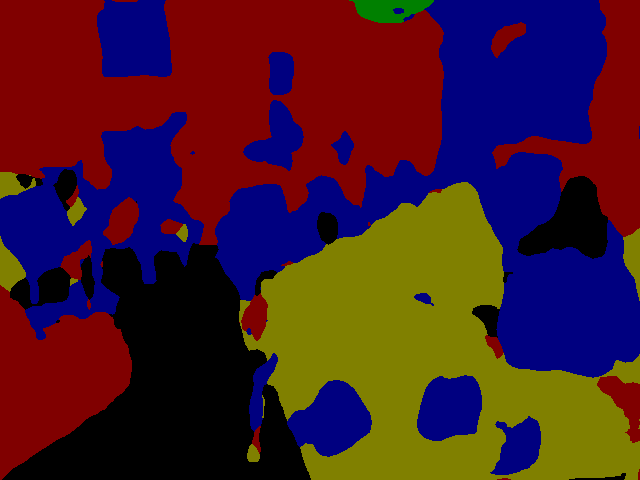}&
   \includegraphics[width=0.19\linewidth]{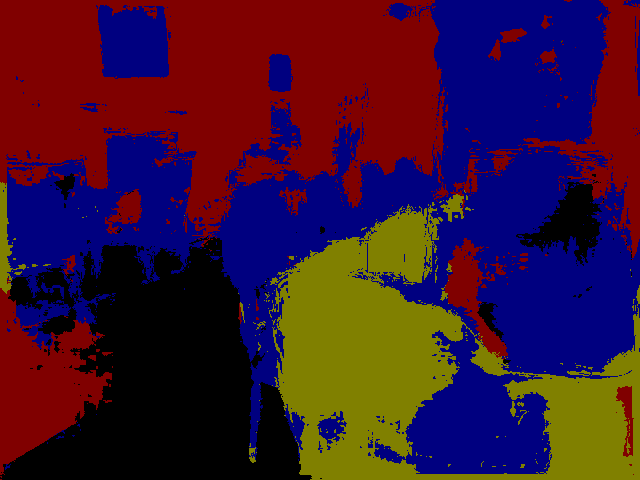}&
   \includegraphics[width=0.19\linewidth]{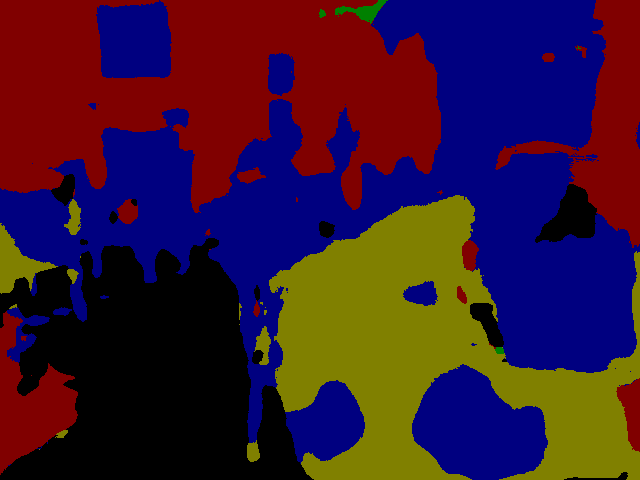}\\

   \end{tabular}
   \setlength\tabcolsep{6pt} 

\end{center}
   \caption{Qualitative results on the {\sc NYU V2} dataset. Note that the CRF-Grad captures the shape of the object instances better compared to the baselines. This effect is perhaps most pronounced for the paintings hanging on the walls.}
\label{fig:nyu_res}
\end{figure*}

In this section we present some additional results from the different datasets. Class-wise results for the {\sc Cityscapes} dataset is presented in Table \ref{tab:cityscapes_classwise}. For the {\sc Weizmann Horse} dataset these results can be seen in Fig.~\ref{fig:weizmann_qual_res}, for the {\sc Cityscapes} dataset in Fig.~\ref{fig:cs_qual_res} and for the {\sc NYU V2} dataset in Fig.~\ref{fig:nyu_res}.

\begin{figure*}[t]
\begin{center}
    \newcommand{\mysize}{0.11}
    
   \setlength\tabcolsep{1pt} 
   \begin{tabular}{cccccccc}
   
   Input & $\bm{z}_{\text{horse}}$ &  $\bm{q}^1_{\text{horse}}$ & $\bm{q}^2_{\text{horse}}$ & $\bm{q}^3_{\text{horse}}$ & $\bm{q}^4_{\text{horse}}$ & $\bm{q}^5_{\text{horse}}$ & Ground truth \\    

   \includegraphics[width=\mysize\linewidth]{figures/weizmann/horse001.jpg} &
   \includegraphics[width=\mysize\linewidth]{figures/weizmann/weizmann_step0_im1.pdf}&
   \includegraphics[width=\mysize\linewidth]{figures/weizmann/weizmann_step1_im1.pdf}&
   \includegraphics[width=\mysize\linewidth]{figures/weizmann/weizmann_step2_im1.pdf}&
   \includegraphics[width=\mysize\linewidth]{figures/weizmann/weizmann_step3_im1.pdf}&
   \includegraphics[width=\mysize\linewidth]{figures/weizmann/weizmann_step4_im1.pdf}&
   \includegraphics[width=\mysize\linewidth]{figures/weizmann/weizmann_step5_im1.pdf}&
   \includegraphics[width=\mysize\linewidth]{figures/weizmann/weizmann_gt_im1.pdf} \\
   
   \includegraphics[width=\mysize\linewidth]{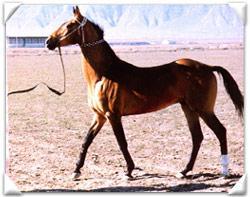} &
   \includegraphics[width=\mysize\linewidth]{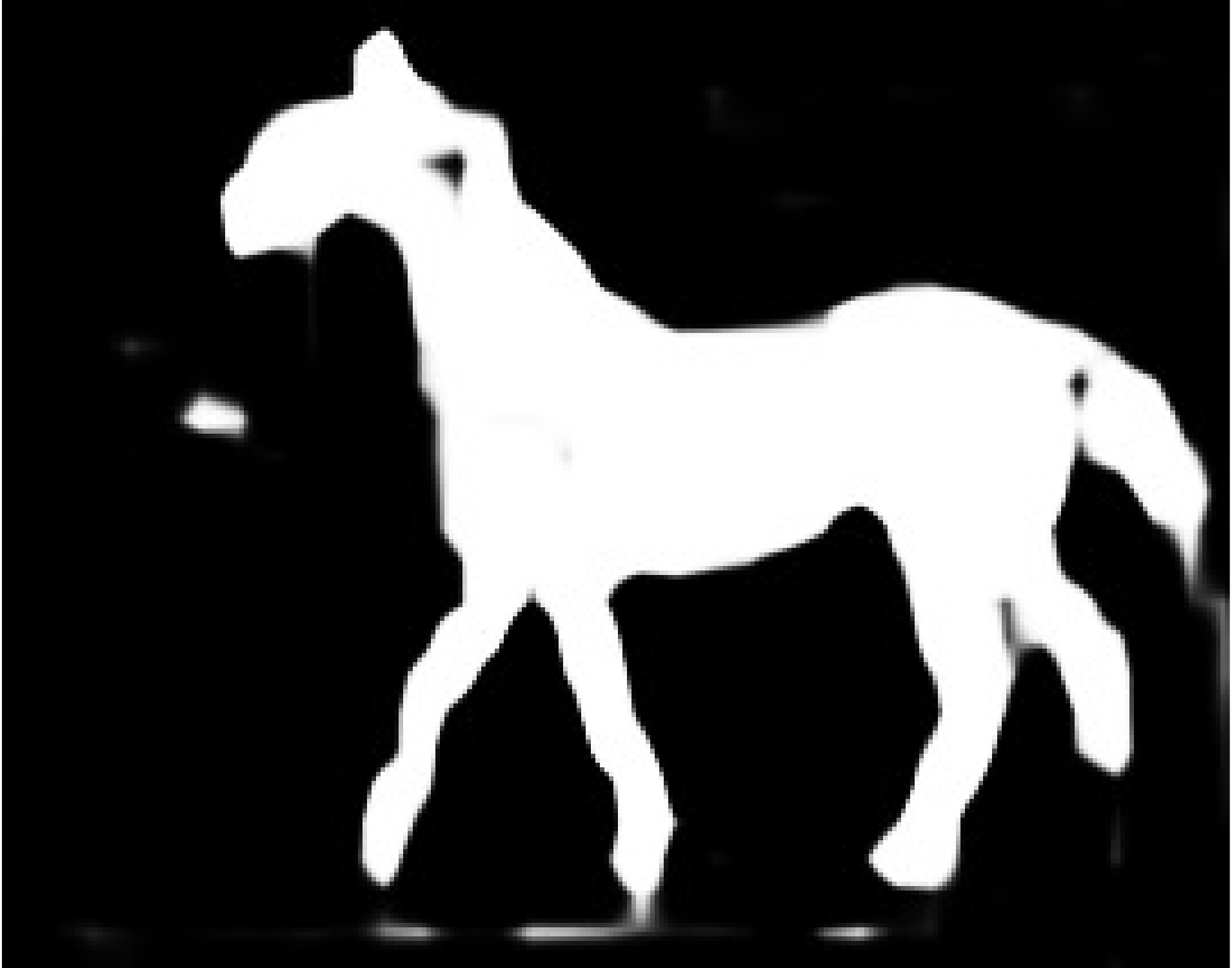}&
   \includegraphics[width=\mysize\linewidth]{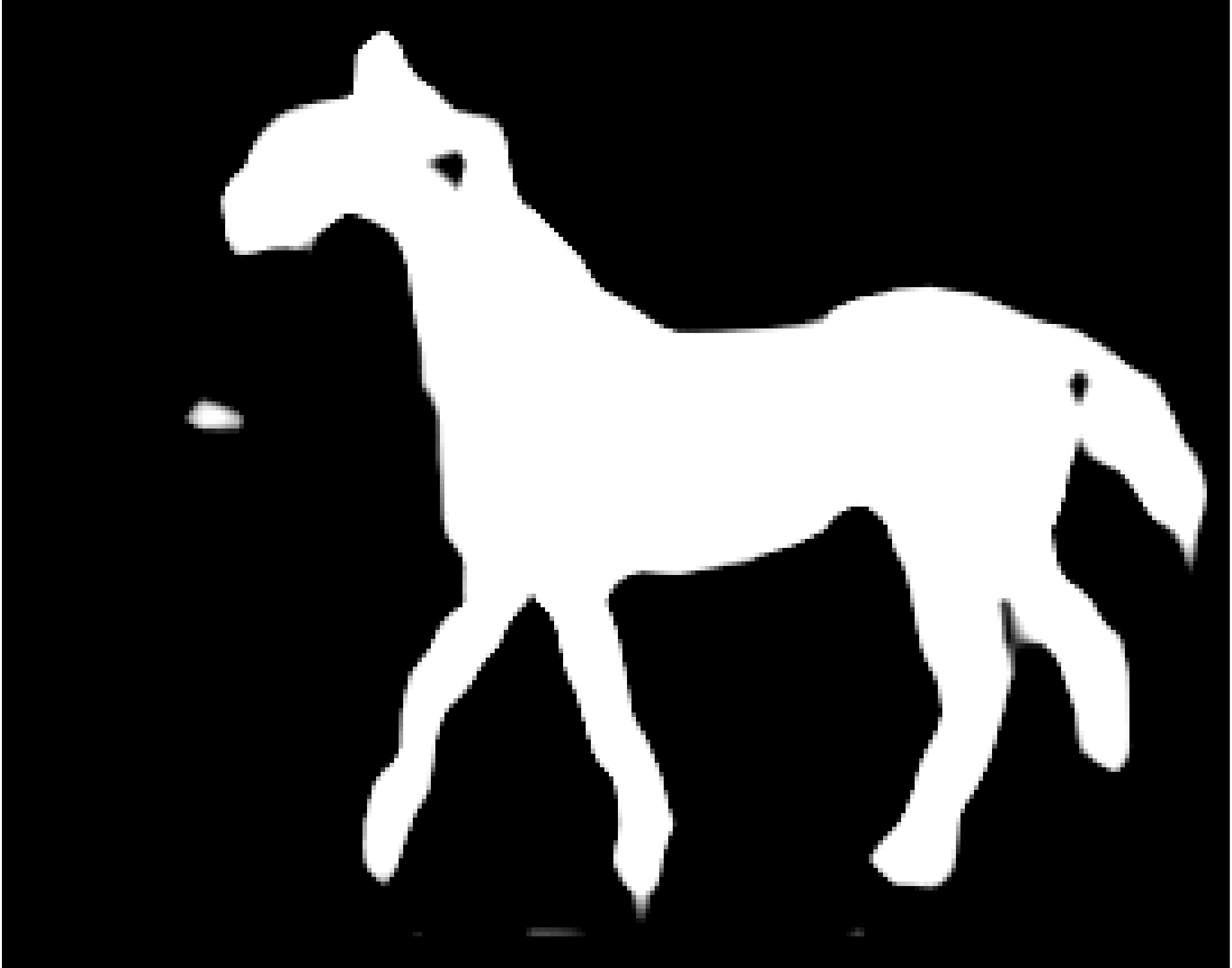}&
   \includegraphics[width=\mysize\linewidth]{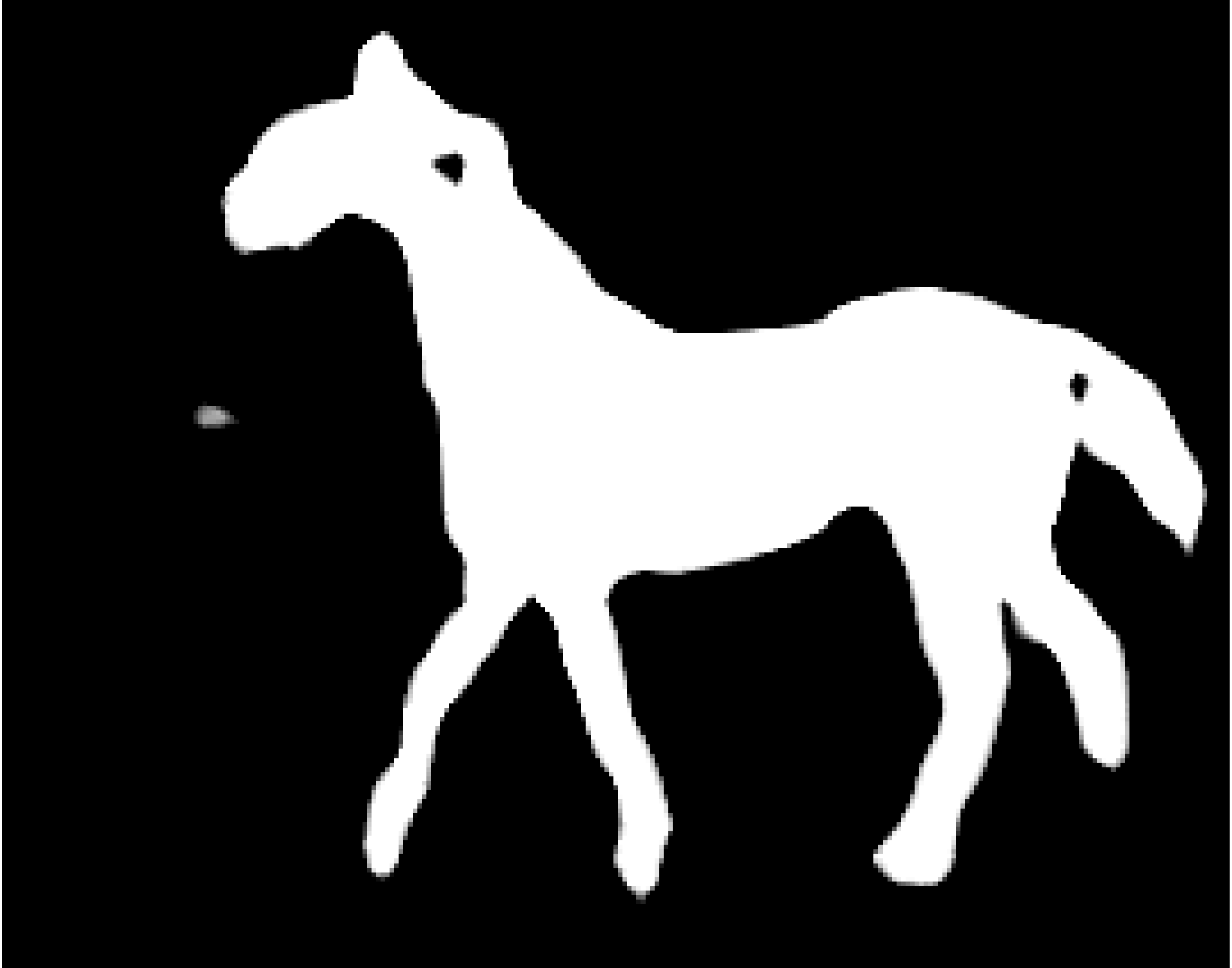}&
   \includegraphics[width=\mysize\linewidth]{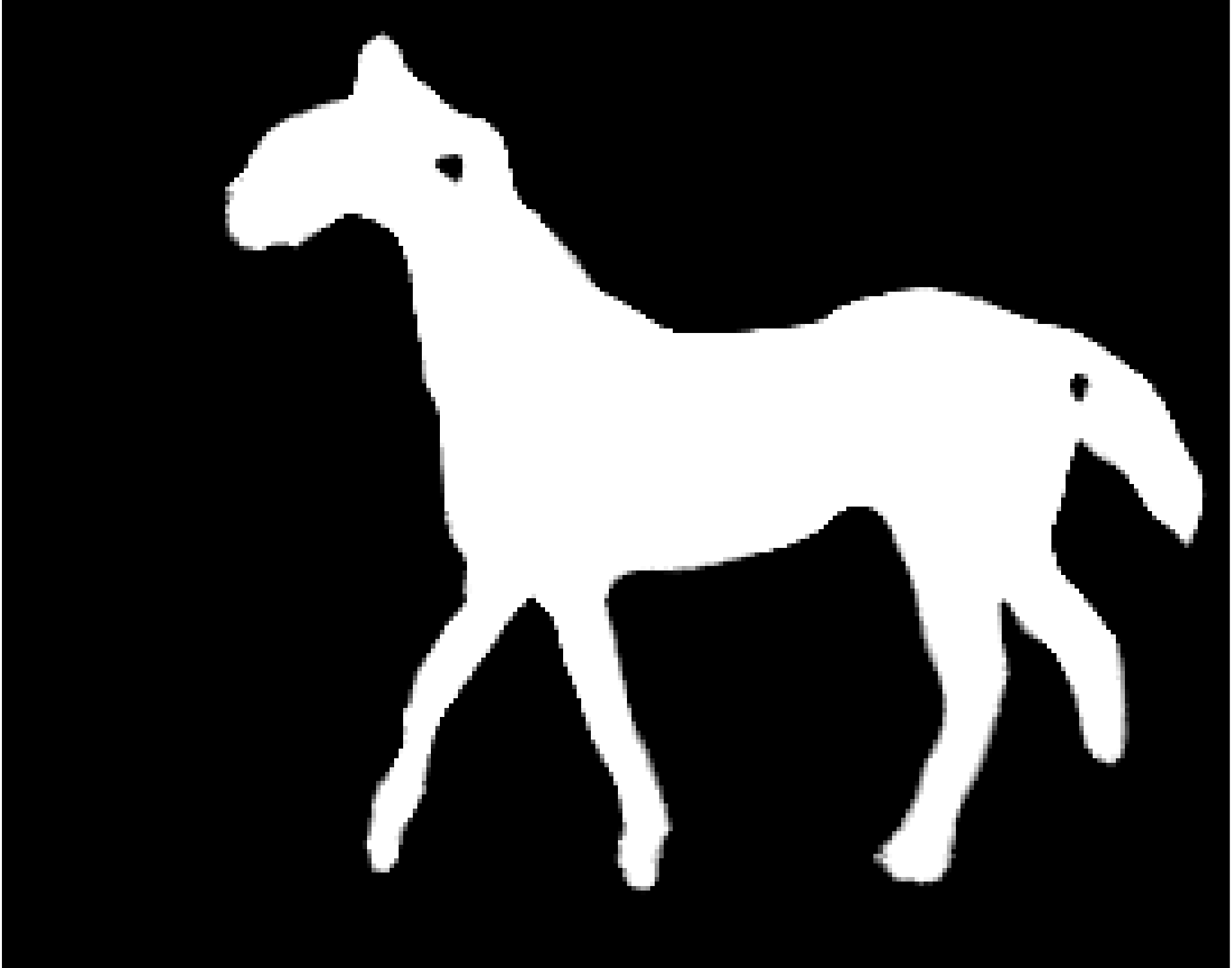}&
   \includegraphics[width=\mysize\linewidth]{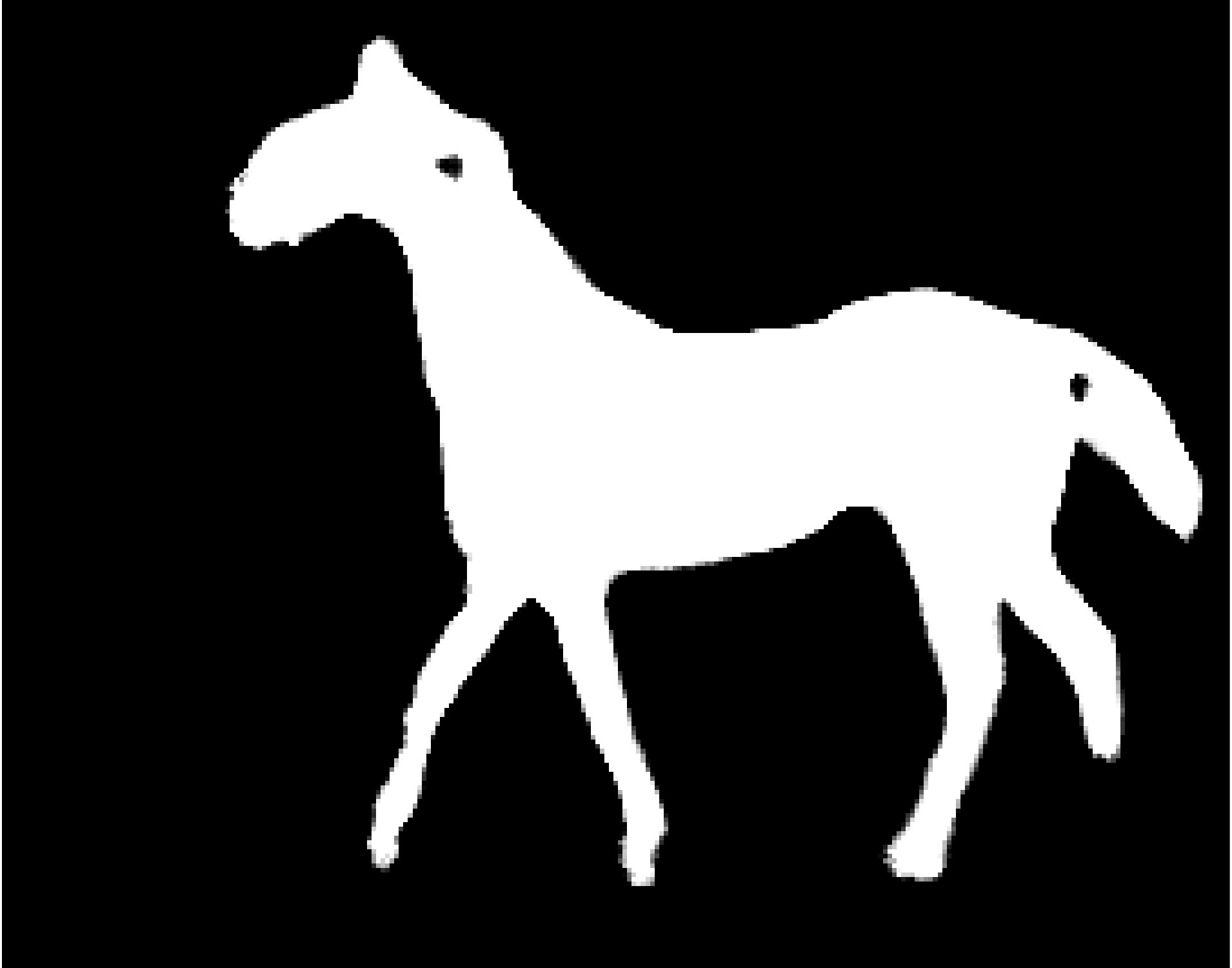}&
   \includegraphics[width=\mysize\linewidth]{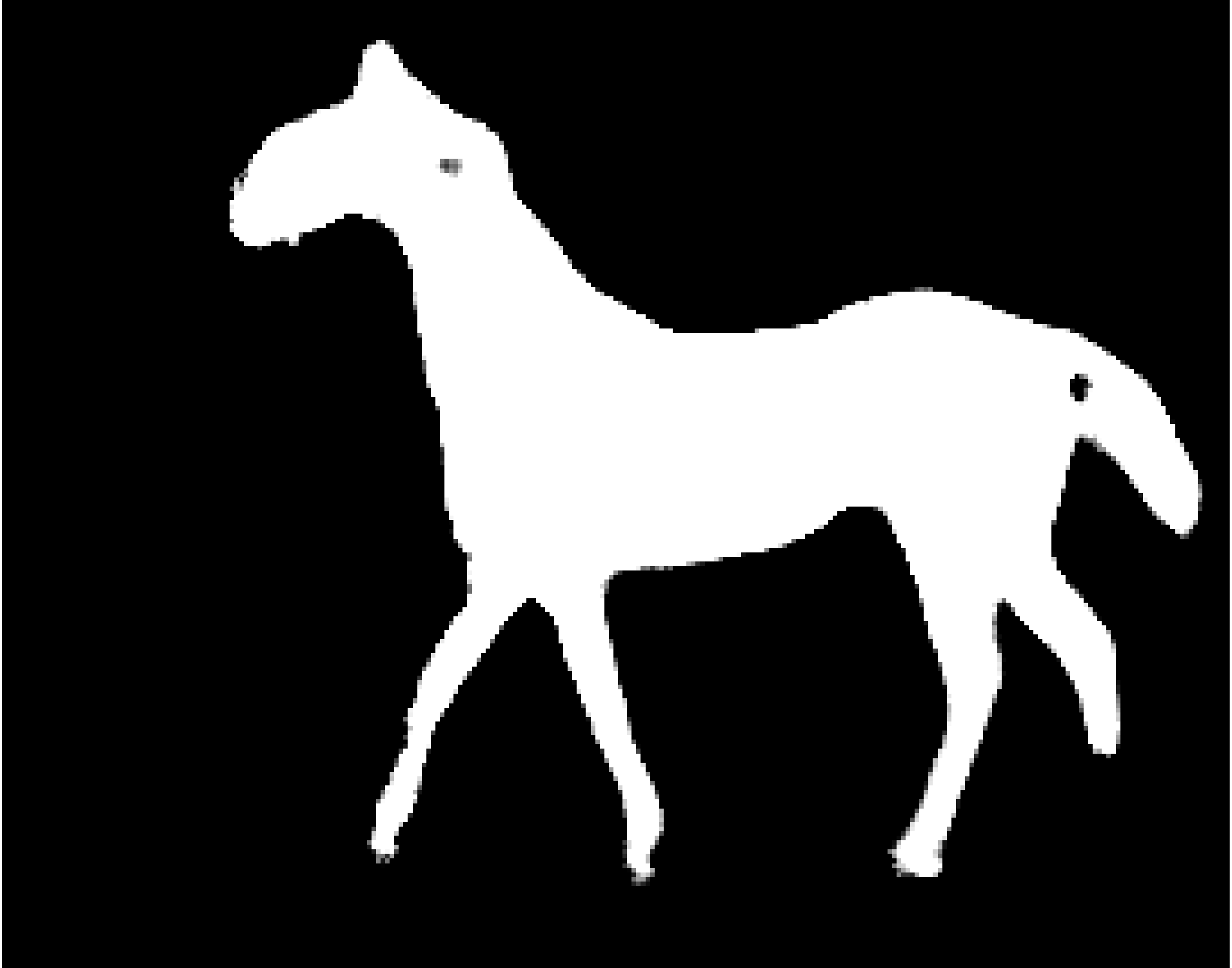}&
   \includegraphics[width=\mysize\linewidth]{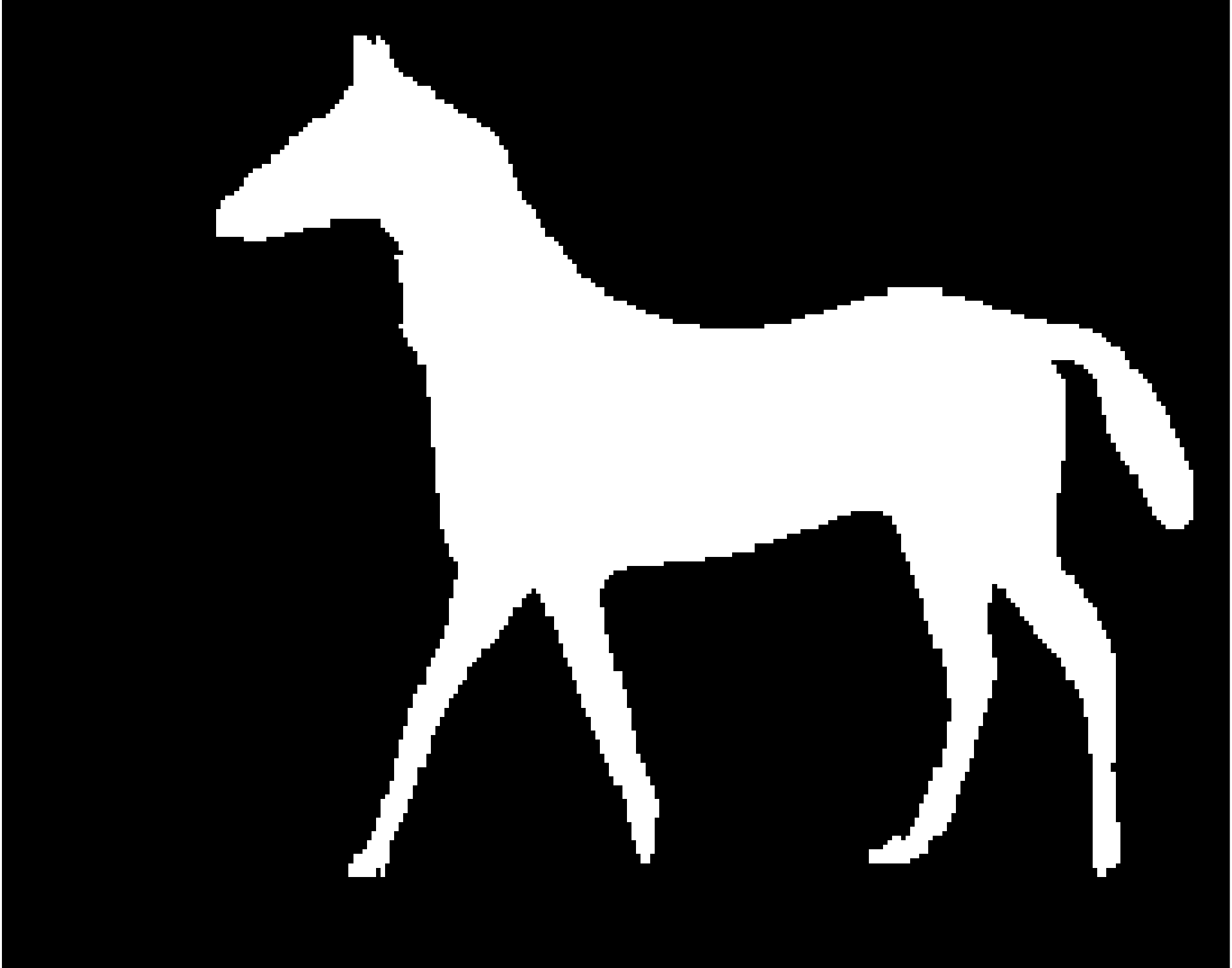}\\
   
   \includegraphics[width=\mysize\linewidth]{figures/weizmann/horse003.jpg}&
   \includegraphics[width=\mysize\linewidth]{figures/weizmann/weizmann_step0_im3.pdf}&
   \includegraphics[width=\mysize\linewidth]{figures/weizmann/weizmann_step1_im3.pdf}&
   \includegraphics[width=\mysize\linewidth]{figures/weizmann/weizmann_step2_im3.pdf}&
   \includegraphics[width=\mysize\linewidth]{figures/weizmann/weizmann_step3_im3.pdf}&
   \includegraphics[width=\mysize\linewidth]{figures/weizmann/weizmann_step4_im3.pdf}&
   \includegraphics[width=\mysize\linewidth]{figures/weizmann/weizmann_step5_im3.pdf}&
   \includegraphics[width=\mysize\linewidth]{figures/weizmann/weizmann_gt_im3.pdf}\\
   
   \includegraphics[width=\mysize\linewidth]{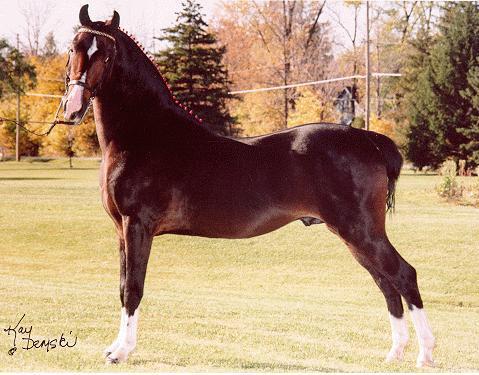}&
   \includegraphics[width=\mysize\linewidth]{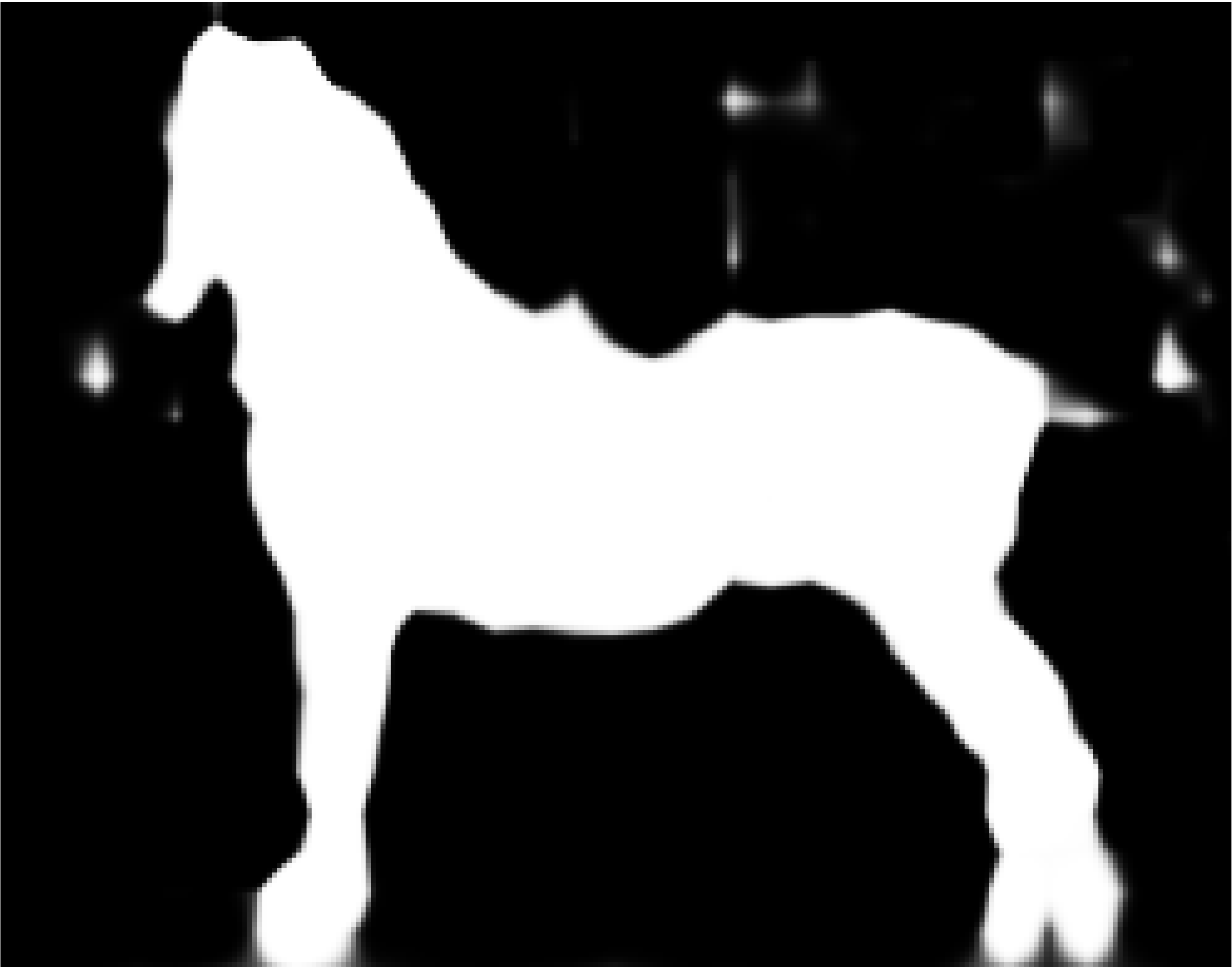}&
   \includegraphics[width=\mysize\linewidth]{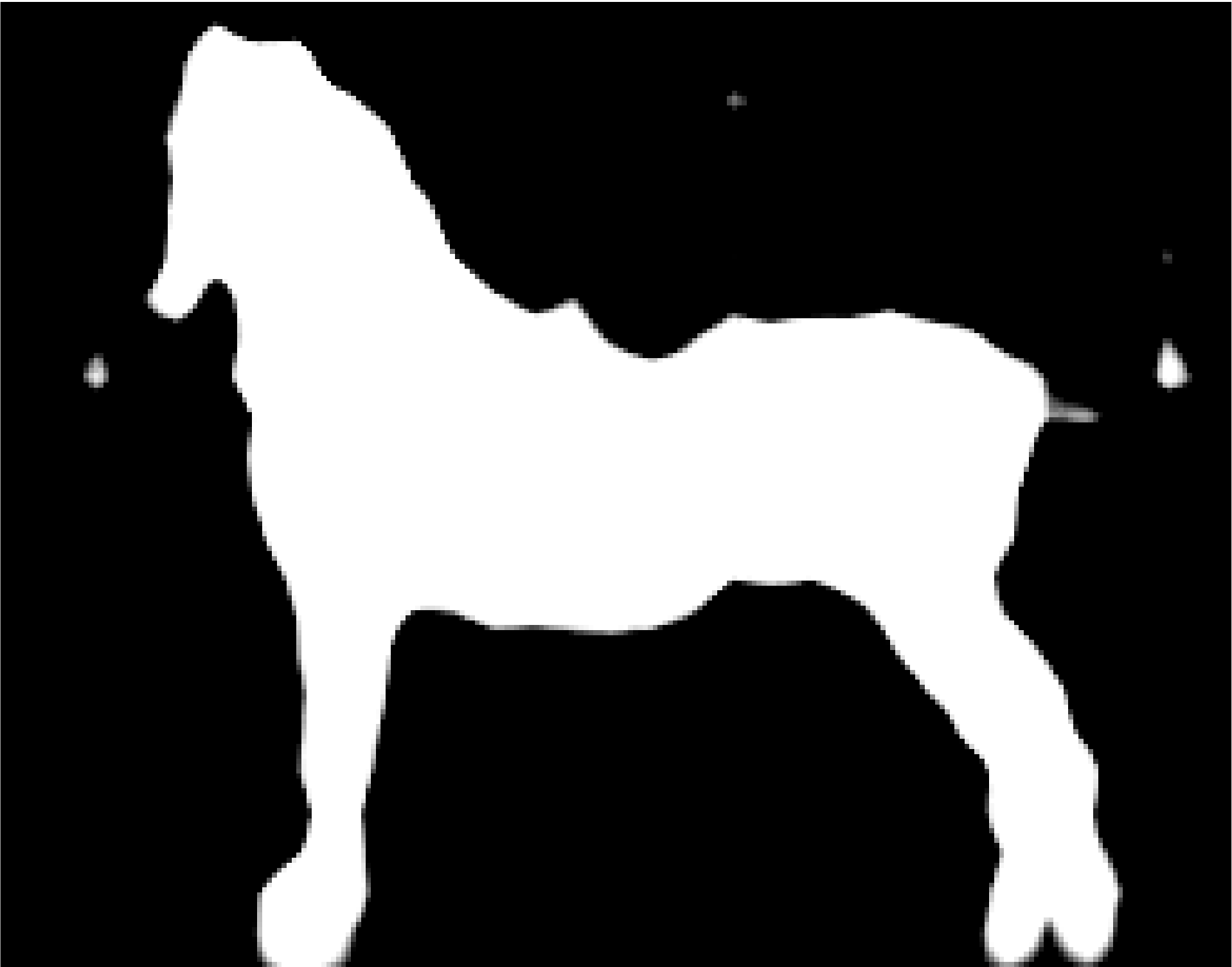}&
   \includegraphics[width=\mysize\linewidth]{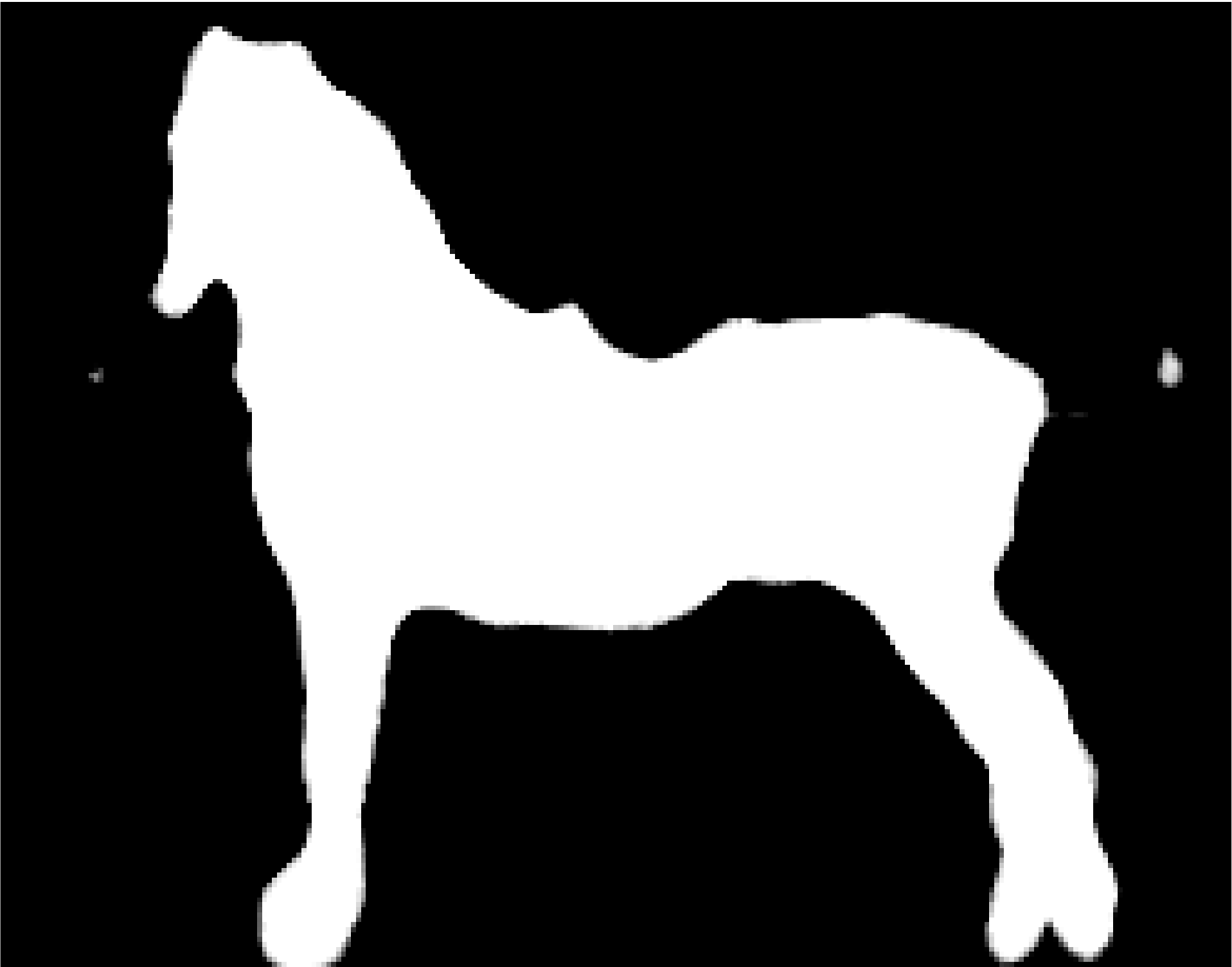}&
   \includegraphics[width=\mysize\linewidth]{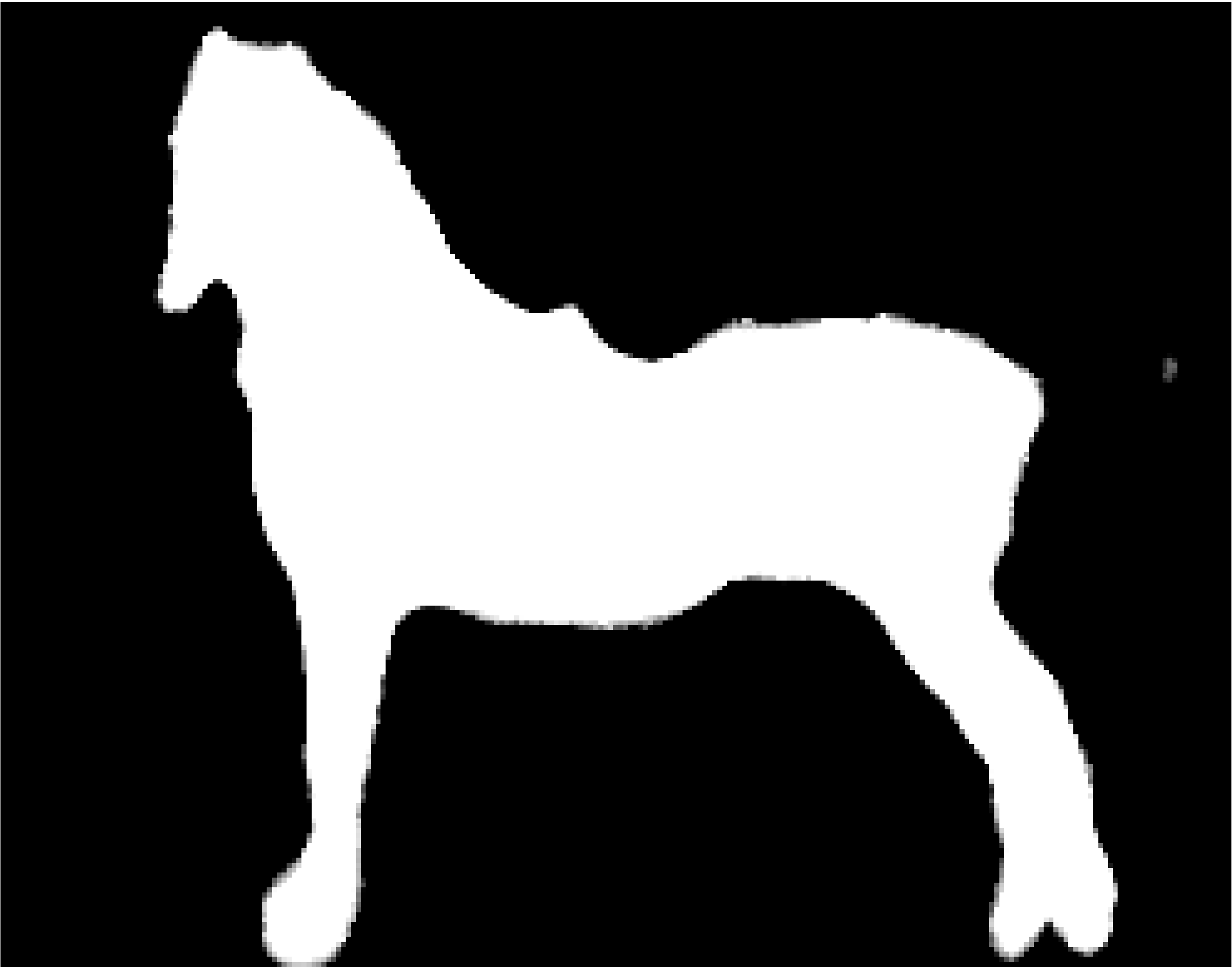}&
   \includegraphics[width=\mysize\linewidth]{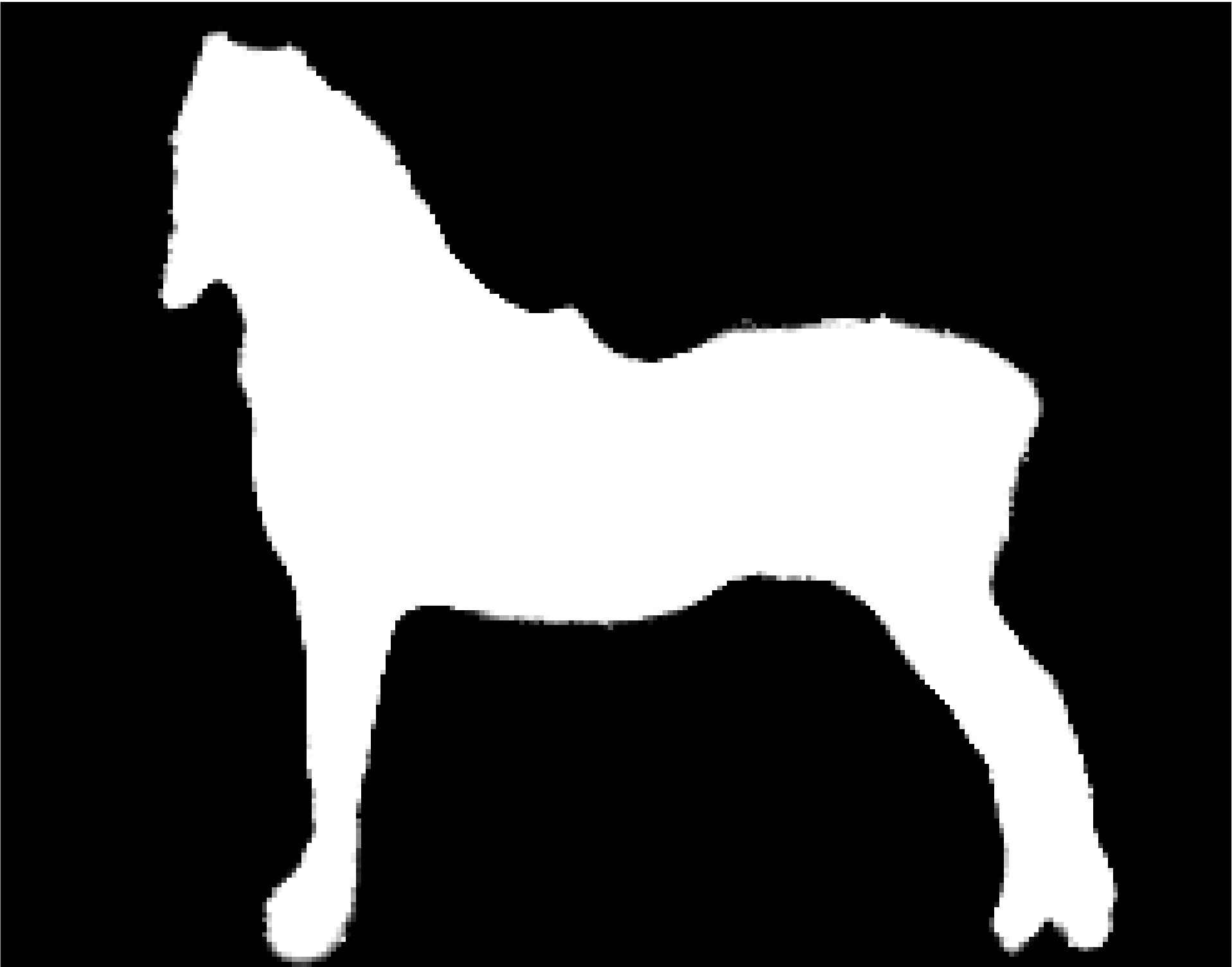}&
   \includegraphics[width=\mysize\linewidth]{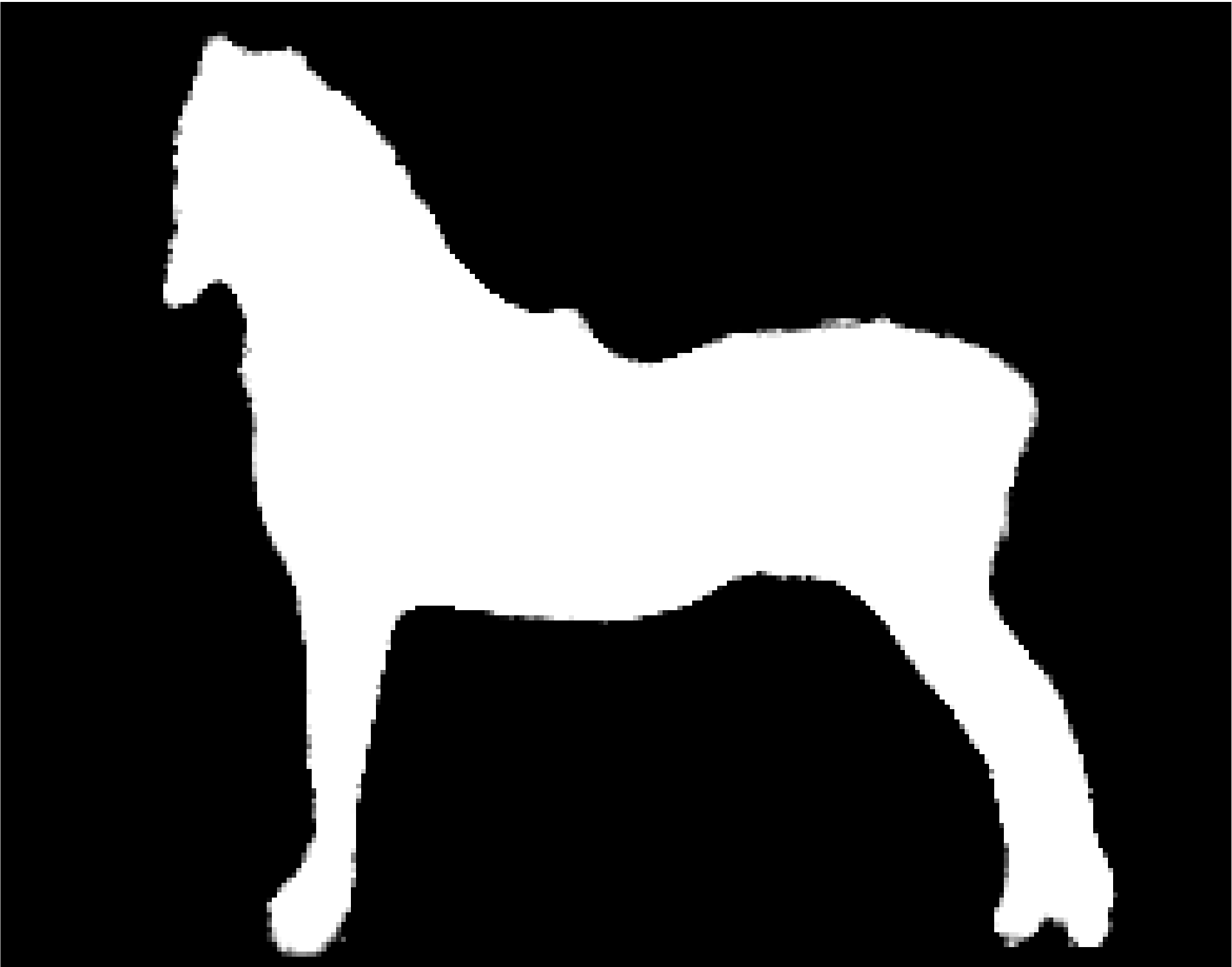}&
   \includegraphics[width=\mysize\linewidth]{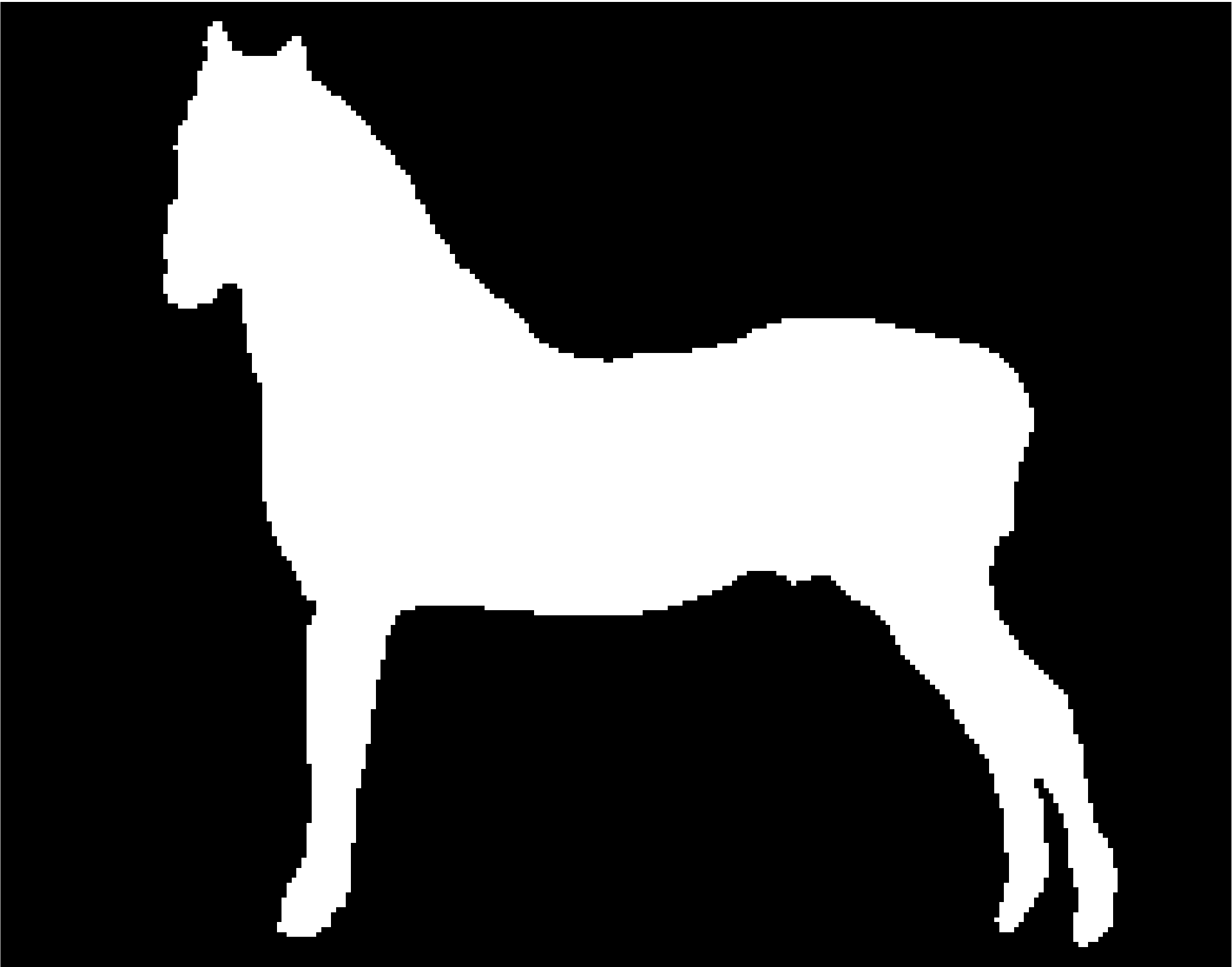}\\
   
   \includegraphics[width=\mysize\linewidth]{figures/weizmann/horse005.jpg}&
   \includegraphics[width=\mysize\linewidth]{figures/weizmann/weizmann_step0_im5.pdf}&
   \includegraphics[width=\mysize\linewidth]{figures/weizmann/weizmann_step1_im5.pdf}&
   \includegraphics[width=\mysize\linewidth]{figures/weizmann/weizmann_step2_im5.pdf}&
   \includegraphics[width=\mysize\linewidth]{figures/weizmann/weizmann_step3_im5.pdf}&
   \includegraphics[width=\mysize\linewidth]{figures/weizmann/weizmann_step4_im5.pdf}&
   \includegraphics[width=\mysize\linewidth]{figures/weizmann/weizmann_step5_im5.pdf}&
   \includegraphics[width=\mysize\linewidth]{figures/weizmann/weizmann_gt_im5.pdf}\\
   
   \end{tabular}
   \setlength\tabcolsep{6pt} 
\end{center}
   \caption{Visualization of intermediate states of the CRF-Grad layer for the {\sc Weizmann Horse} dataset. Note how each step of the projected gradient descent algorithm refines the segmentation slightly, removing spurious outlier pixels classified as horse.}
\label{fig:intmed}
\end{figure*}

\end{document}